\documentclass[english]{article}
\usepackage[T1]{fontenc}
\usepackage[latin9]{inputenc}
\usepackage{geometry}
\usepackage{babel}
\usepackage{verbatim}
\usepackage{float}
\usepackage{bm}
\usepackage{amsmath}
\usepackage{amssymb}
\usepackage{graphicx}
\usepackage[unicode=true,
 bookmarks=false,
 breaklinks=false,pdfborder={0 0 1},colorlinks=false]
 {hyperref}
\hypersetup{
 colorlinks,linkcolor=red,anchorcolor=blue,citecolor=blue}

\makeatletter

\providecommand{\tabularnewline}{\\}
\floatstyle{ruled}
\newfloat{algorithm}{tbp}{loa}
\providecommand{\algorithmname}{Algorithm}
\floatname{algorithm}{\protect\algorithmname}

\usepackage{babel}
\usepackage{babel}
\usepackage{babel}

\usepackage{mathtools}

\usepackage{cite}\usepackage{amsthm}\usepackage{dsfont}\usepackage{array}\usepackage{mathrsfs}\usepackage{comment}\onecolumn

\usepackage{color}\usepackage{babel}

\allowdisplaybreaks

\usepackage{enumitem}
\setlist[itemize]{leftmargin=1.5em}
\setlist[enumerate]{leftmargin=1.5em}

\usepackage{babel}
\usepackage{algorithm}% http://ctan.org/pkg/algorithms
\usepackage{algorithmic}% http://ctan.org/pkg/algorithmicx
\usepackage{arydshln}

\newcommand{\bS}{\bm{S}}

\DeclareMathOperator{\ind}{\mathds{1}}  % Indicator

\numberwithin{equation}{section}

\definecolor{yxc}{RGB}{255,0,0}
\definecolor{yjc}{RGB}{125,0,0}
\definecolor{cm}{RGB}{0,0,200}
\definecolor{yly}{RGB}{0,150,0}

\makeatother

\begin{document}
\theoremstyle{plain} \newtheorem{lemma}{\textbf{Lemma}} \newtheorem{prop}{\textbf{Proposition}}\newtheorem{theorem}{\textbf{Theorem}}\setcounter{theorem}{0}
\newtheorem{corollary}{\textbf{Corollary}} \newtheorem{assumption}{\textbf{Assumption}}
\newtheorem{example}{\textbf{Example}} \newtheorem{definition}{\textbf{Definition}}
\newtheorem{fact}{\textbf{Fact}} \newtheorem{condition}{\textbf{Condition}}\theoremstyle{definition}

\theoremstyle{remark}\newtheorem{remark}{\textbf{Remark}}\newtheorem{claim}{\textbf{Claim}}\newtheorem{conjecture}{\textbf{Conjecture}}
\title{Bridging Convex and Nonconvex Optimization in Robust PCA: Noise, Outliers,
and Missing Data\footnotetext{Author
	names are sorted alphabetically. Corresponding author: Yuxin Chen (Email: \texttt{yuxin.chen@princeton.edu}).}}
\author{Yuxin Chen\thanks{Department of Electrical and Computer Engineering, Princeton University, Princeton,
		NJ 08544, USA; Email: \texttt{yuxin.chen@princeton.edu}.} \and Jianqing Fan\thanks{Department of Operations Research and Financial Engineering, Princeton
		University, Princeton, NJ 08544, USA; Email: \texttt{\{jqfan,
			yulingy\}@princeton.edu}.} \and Cong Ma\thanks{Department of Electrical Engineering and Computer Sciences, UC Berkeley, Berkeley,
		CA 94720, USA; Email: \texttt{congm@berkeley.edu}.} \and Yuling Yan\footnotemark[2]}

%\date{January 2020}

\maketitle
\begin{abstract}
This paper delivers improved theoretical guarantees for the convex
programming approach in low-rank matrix estimation, in the presence
of (1) random noise, (2) gross sparse outliers, and (3) missing data.
This problem, often dubbed as\emph{ robust principal component analysis
(robust PCA)}, finds applications in various domains. Despite the
wide applicability of convex relaxation, the available statistical
support (particularly the stability analysis vis-à-vis random noise)
remains highly suboptimal, which we strengthen in this paper. When
the unknown matrix is well-conditioned, incoherent, and of constant
rank, we demonstrate that a principled convex program achieves near-optimal
statistical accuracy, in terms of both the Euclidean loss and the
$\ell_{\infty}$ loss. All of this happens even when nearly a constant
fraction of observations are corrupted by outliers with arbitrary
magnitudes. The key analysis idea lies in bridging the convex program
in use and an auxiliary nonconvex optimization algorithm, and hence
the title of this paper.
\end{abstract}

\noindent \textbf{Keywords:} robust principal component analysis,
nonconvex optimization, convex relaxation, $\ell_{\infty}$ guarantees,
leave-one-out analysis

\tableofcontents{}

\section{Introduction}

A diverse array of science and engineering applications (e.g.~video
surveillance, joint shape matching, graph clustering, covariance modeling, graphical models) involves estimation
of low-rank matrices \cite{chi2018nonconvex,CanLiMaWri09,chen2014matching,jalali2011clustering,chandrasekaran2012latent, fan2013large,davenport2016overview}.
The imperfectness of data acquisition processes, however, presents
several common yet critical challenges: (1) random noise: which reflects
the uncertainty of the environment and/or the measurement processes;
(2) outliers: which represent a sort of corruption that exhibits abnormal
behavior; and (3) incomplete data, namely, we might only get to observe
a fraction of the matrix entries. Low-rank matrix estimation algorithms
aimed at addressing these challenges have been extensively studied
under the umbrella of \emph{robust principal component analysis (Robust
PCA)} \cite{chandrasekaran2011rank,CanLiMaWri09}, a terminology popularized
by the seminal work \cite{CanLiMaWri09}.

\begin{comment}
Prominent examples include matrix completion --- recovering a low-rank
matrix from incomplete observations of its entries~\cite{ExactMC09,KesMonSew2010} --- which
finds applications in collaborative filtering, sensor localization,
etc., and robust principal component analysis (robust PCA) --- recovering
a low-rank matrix when some of its entries are arbitrarily corrupted
\cite{CanLiMaWri09,chandrasekaran2011rank,chen2013low} --- which
is central to video surveillance, face recognition, etc. The interested
reader is referred to the recent surveys \cite{chi2018nonconvex,chen2018harnessing,davenport2016overview}
for more examples and applications of low-rank matrix recovery.

This paper focuses on the robust PCA problem under a more realistic
setting, namely the low-rank matrix is allowed to be corrupted by
both gross sparse outliers and additive random noise.
\end{comment}

To formulate the above-mentioned problem in a more precise manner,
imagine that we seek to estimate an unknown low-rank matrix $\bm{L}^{\star}\in\mathbb{R}^{n_1\times n_2}$. %\footnote{To avoid cluttered notation, this paper works with square matrices of size $n$ by $n$. Our results and analysis can be extended to accommodate rectangular matrices.} 
What we can obtain is a collection of partially observed and corrupted
entries as follows
\begin{equation}
M_{ij}=L_{ij}^{\star}+S_{ij}^{\star}+E_{ij},\qquad(i,j)\in\Omega_{\mathsf{obs}},\label{eq:M-L-S-E}
\end{equation}
where $\bm{S}^{\star}=[S_{ij}^{\star}]$ is a matrix
consisting of outliers, $\bm{E}=[E_{ij}]$ represents
the random noise, and we only observe entries over an index subset
$\Omega_{\mathsf{obs}}\subseteq[n_1]\times[n_2]$ with $[n]\coloneqq\{1,2,\cdots,n\}$.
The current paper assumes that $\bm{S}^{\star}$ is a relatively sparse
matrix whose non-zero entries might have arbitrary magnitudes. This
assumption has been commonly adopted in prior work to model gross
outliers, while enabling reliable disentanglement of the outlier component
and the low-rank component \cite{chandrasekaran2011rank,CanLiMaWri09,chen2013low,li2011compressed}.
In addition, we suppose that the entries $\{E_{ij}\}$ are independent
zero-mean sub-Gaussian random variables, as commonly assumed in the
statistics literature to model a large family of random noise. The
aim is to reliably estimate $\bm{L}^{\star}$ given the grossly corrupted
and possibly incomplete data (\ref{eq:M-L-S-E}). Ideally, this task
should be accomplished without knowing the locations and magnitudes
of the outliers $\bm{S}^{\star}$.

\subsection{A principled convex programming approach}

Focusing on the noiseless case with $\bm{E}=\bm{0}$, the papers by \cite{chandrasekaran2011rank,CanLiMaWri09}
delivered a positive and somewhat surprising message: both the low-rank
component $\bm{L}^{\star}$ and the sparse component $\bm{S}^{\star}$
can be efficiently recovered with absolutely no error by means of
a principled convex program
\begin{equation}
\underset{\bm{L},\bm{S}\in\mathbb{R}^{n_1\times n_2}}{\text{minimize}}\quad\left\Vert \bm{L}\right\Vert _{\ast}+\tau\left\Vert \bm{S}\right\Vert _{1}\qquad\text{subject to}\quad\mathcal{P}_{\Omega_{\mathsf{obs}}}(\bm{L}+\bm{S}-\bm{M})=\bm{0},\label{eq:PCP}
\end{equation}
provided that certain ``separation'' and ``incoherence'' conditions
on $(\bm{L}^{\star},\bm{S}^{\star},\Omega_{\mathsf{obs}})$ hold\footnote{Clearly, if the low-rank matrix $\bm{L}^{\star}$ is also sparse,
one cannot possibly separate $\bm{S}^{\star}$ from $\bm{L}^{\star}$.
The same holds true if the matrix $\bm{S}^{\star}$ is simultaneously
sparse and low-rank.} and that the regularization parameter $\tau$ is properly chosen.
Here, $\Vert\bm{L}\Vert_{\ast}$ denotes the nuclear norm (i.e.~the
sum of the singular values) of $\bm{L}$, $\Vert\bm{S}\Vert_{1}=\sum_{i,j}|S_{ij}|$
denotes the usual entrywise $\ell_{1}$ norm, and $\mathcal{P}_{\Omega_{\mathsf{obs}}}(\bm{M})$
denotes the Euclidean projection of a matrix $\bm{M}$ onto the subspace
of matrices supported on $\Omega_{\mathsf{obs}}$. Given that the
nuclear norm $\|\cdot\|_{*}$ (resp.~the $\ell_{1}$ norm $\|\cdot\|_{1}$)
is the convex relaxation of the rank function $\mathsf{rank}(\cdot)$
(resp.~the $\ell_{0}$ counting norm $\|\cdot\|_{0}$), the rationale
behind (\ref{eq:PCP}) is rather clear: it seeks a decomposition $(\bm{L},\bm{S})$
of~$\bm{M}$ by promoting the low-rank structure of $\bm{L}$ as
well as the sparsity structure of $\bm{S}$.

Moving on to the more realistic noisy setting, a natural strategy
is to solve the following regularized least-squares problem
\begin{equation}
\underset{\bm{L},\bm{S}\in\mathbb{R}^{n_1\times n_2}}{\text{minimize}}\quad\frac{1}{2}\left\Vert \mathcal{P}_{\Omega_{\mathsf{obs}}}\left(\bm{L}+\bm{S}-\bm{M}\right)\right\Vert _{\mathrm{F}}^{2}+\lambda\left\Vert \bm{L}\right\Vert _{\ast}+\tau\left\Vert \bm{S}\right\Vert _{1}.\label{eq:cvx}
\end{equation}
With the regularization parameters $\lambda,\tau>0$ properly chosen,
one hopes to strike a balance between enhancing the goodness
of fit (by enforcing $\bm{L}+\bm{S}-\bm{M}$ to be small) and promoting
the desired low-complexity structures (by regularizing both the nuclear
norm of $\bm{L}$ and the $\ell_{1}$ norm of $\bm{S}$). A natural
and important question comes into our mind: 

\begin{itemize}[rightmargin=\leftmargin]

\item[] \emph{Where does the algorithm (\ref{eq:cvx}) stand in terms
of its statistical performance vis-\`a-vis random noise, sparse outliers and missing data?}

\end{itemize}

\noindent Unfortunately, however simple this program (\ref{eq:cvx})
might seem, the existing theoretical support remains far from satisfactory, as we shall discuss momentarily.

\subsection{Theory-practice gaps under random noise}
\label{subsec:an-extreme-setting}

To assess the tightness of prior statistical guarantees for (\ref{eq:cvx}),
we find it convenient to first look at a simple setting where (i) $n_1=n_2=n$, (ii)
$\bm{E}$ consists of independent Gaussian components, namely, $E_{ij}\sim\mathcal{N}(0,\sigma^{2})$,
and (iii) there is no missing data. This simple scenario is sufficient
to illustrate the sub-optimality of prior theory.

\paragraph{Prior statistical guarantees} The paper \cite{zhou2010stable}
was the first to derive a sort of statistical performance guarantees
for the above convex program. Under mild conditions, \cite{zhou2010stable}
demonstrated that any minimizer $(\widehat{\bm{L}},\widehat{\bm{S}})$
of (\ref{eq:cvx}) achieves\footnote{Mathematically, \cite{zhou2010stable}
investigated an equivalent constrained form of (\ref{eq:cvx}) and
developed an upper bound on the corresponding estimation error.}
\begin{equation}
\bigl\Vert\widehat{\bm{L}}-\bm{L}^{\star}\bigr\Vert_{\mathrm{F}}=O\left(n\bigl\Vert\bm{E}\bigr\Vert_{\mathrm{F}}\right)=O(\sigma n^{2})\label{eq:L-hat-no-oracle}
\end{equation}
with high probability, where we have substituted in the well-known
high-probability bound $\|\bm{E}\|_{\mathrm{F}}=O\left(\sigma n\right)$
under i.i.d.~Gaussian noise. While this theory corroborates the potential
stability of convex relaxation against both additive noise and sparse
outliers, it remains unclear whether the estimation error bound (\ref{eq:L-hat-no-oracle})
reflects the true performance of the convex program in use. In what
follows, we shall compare it with an oracle error bound and collect
some numerical evidence.

\paragraph{Comparisons with an oracle bound} Consider an idealistic
scenario where an oracle informs us of the outlier matrix $\bm{S}^{\star}$.
With the assistance of this oracle, the task of estimating $\bm{L}^{\star}$
reduces to a low-rank matrix denoising problem \cite{donoho2014minimax}.
By fixing $\bm{S}$ to be $\bm{S}^{\star}$ in (\ref{eq:cvx}), we
arrive at a simplified convex program
\begin{equation}
\underset{\bm{L}\in\mathbb{R}^{n\times n}}{\text{minimize}}\quad\frac{1}{2}\left\Vert \bm{L}-(\bm{L}^{\star}+\bm{E})\right\Vert _{\mathrm{F}}^{2}+\lambda\left\Vert \bm{L}\right\Vert _{\ast}.\label{eq:matrix-denoising}
\end{equation}
It is known that (e.g.~\cite{donoho2014minimax,chen2019noisy}):
under mild conditions and with a properly chosen $\lambda$, the estimation
error of (\ref{eq:matrix-denoising}) satisfies
\begin{equation}
\bigl\Vert\widehat{\bm{L}}-\bm{L}^{\star}\bigr\Vert_{\mathrm{F}}=O\left(\sigma\sqrt{nr}\right),\label{eq:L-hat-oracle}
\end{equation}
where we abuse the notation and denote by $\widehat{\bm{L}}$ the
minimizer of (\ref{eq:matrix-denoising}). The large gap between the
above two bounds (\ref{eq:L-hat-no-oracle}) and (\ref{eq:L-hat-oracle})
is self-evident; in particular, if $r=O(1)$, the gap between these
two bounds can be as large as an order of $n^{1.5}$.

\begin{comment}
In particular, when the noise matrix~$\bm{E}$ is random, say the
entries of $\bm{E}$ are i.i.d.~$\mathcal{N}(0,\sigma^{2})$, one
can choose $\lambda\asymp\|\bm{E}\|\asymp\sigma\sqrt{n}$ and obtain
\begin{equation}
\|\widehat{\bm{L}}-\bm{L}^{\star}\bigr\Vert_{\mathrm{F}}\lesssim\sigma\sqrt{nr}.\label{eq:matrix_denoising_bound}
\end{equation}
\end{comment}

\begin{figure}
	\centering
	
	\begin{tabular}{cc}
		\includegraphics[scale=0.35]{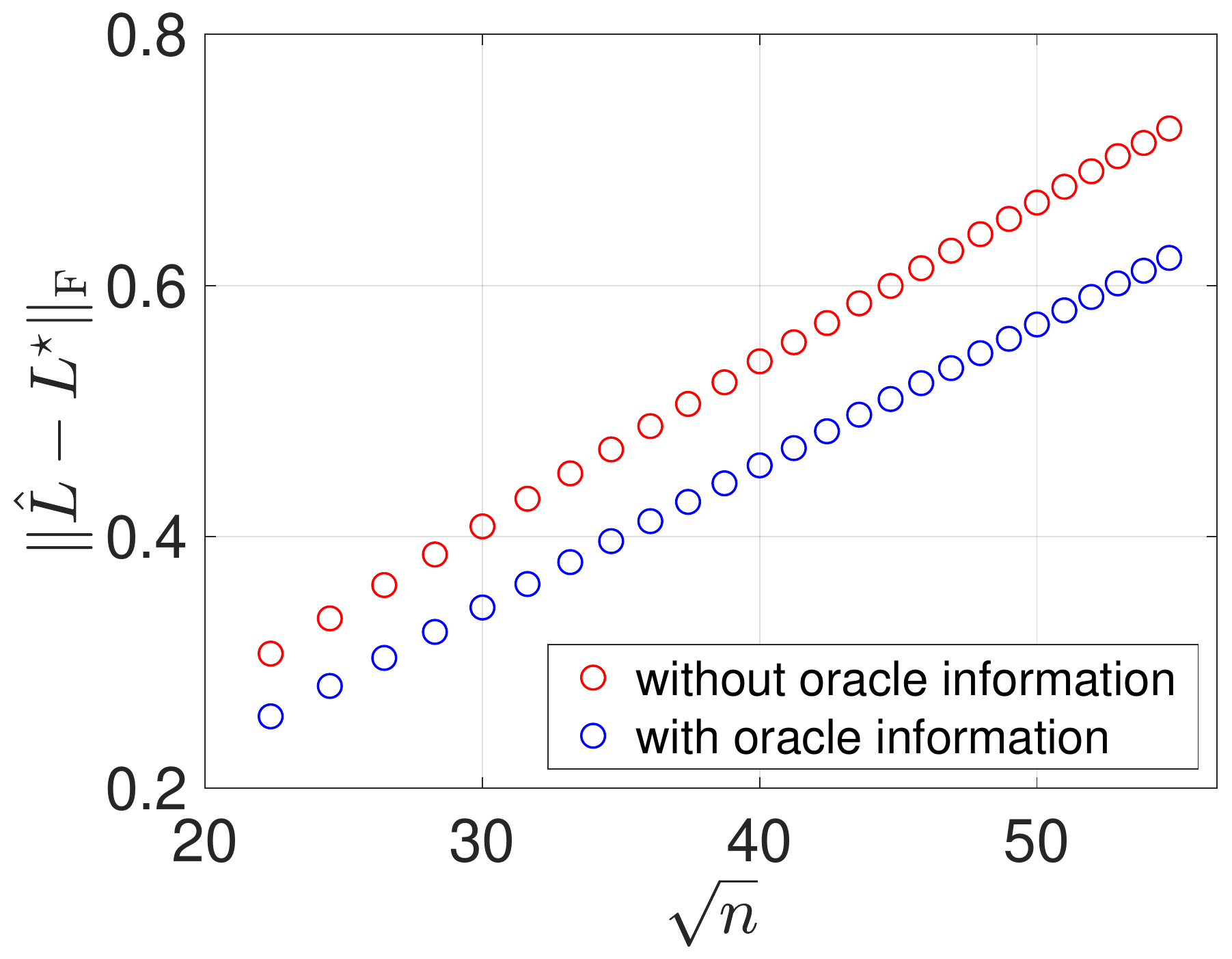} & \includegraphics[scale=0.35]{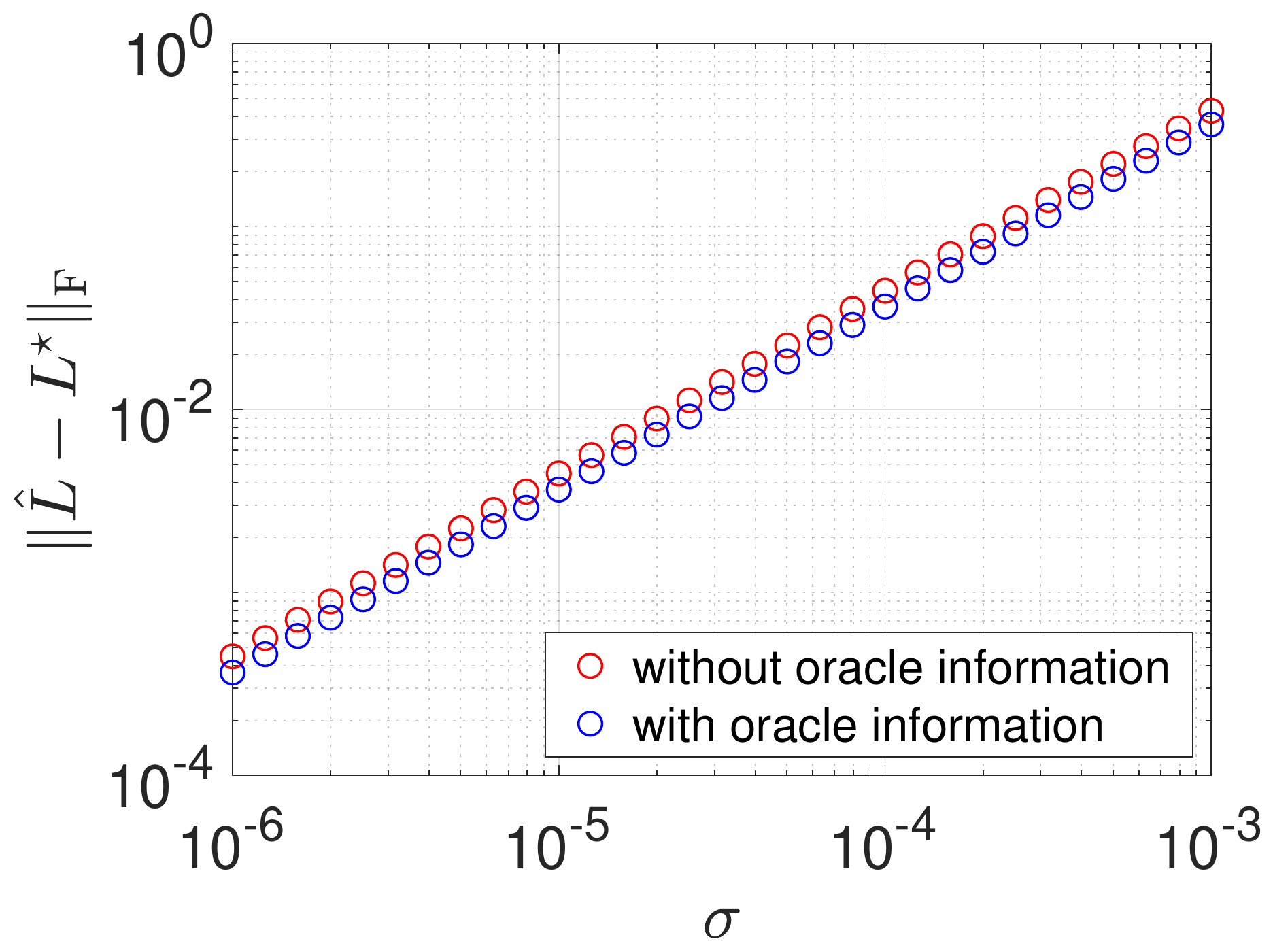}\tabularnewline
		$\quad\quad$(a) & $\quad\quad$(b)\tabularnewline
	\end{tabular}
	
	\caption{(a) Euclidean estimation errors of (\ref{eq:cvx}) and (\ref{eq:matrix-denoising})
		vs.~the problem size~$\sqrt{n}$, where we fix $r=5,\sigma=10^{-3}$; (b) Euclidean
		estimation errors of (\ref{eq:cvx}) and (\ref{eq:matrix-denoising})
		vs.~the noise level $\sigma$ in a log-log plot, where we fix $n=1000,r=5$. For both plots,
		we take $\lambda=5\sigma\sqrt{n}$ and $\tau=2\sigma\sqrt{\log n}$. The results are averaged over 50 independent trials.
		\label{fig:exp1}}
\end{figure}

\paragraph{A numerical example without oracles} One might naturally
wonder whether the discrepancy between the two bounds
(\ref{eq:L-hat-no-oracle}) and (\ref{eq:L-hat-oracle}) stems from
the magical oracle information (i.e.~$\bm{S}^{\star}$) which (\ref{eq:cvx}) does not have
the luxury to know. To demonstrate that this is not the case, we conduct
some numerical experiments to assess the importance of such oracle
information. Generate $\bm{L}^{\star}=\bm{X}^{\star}\bm{Y}^{\star\top}$,
where $\bm{X}^{\star},\bm{Y}^{\star}\in\mathbb{R}^{n\times r}$ are
random orthonormal matrices. Each entry of $\bm{S}^{\star}$ is generated
independently from a mixed distribution: with probability $1/10$,
the entry is drawn from $\mathcal{N}(0,10)$; otherwise, it is set
to be zero. In other words, approximately $10\%$ of the entries in
$\bm{L}^{\star}$ are corrupted by large outliers. Throughout the
experiments, we set $\lambda=5\sigma\sqrt{n}$ and $\tau=2\sigma\sqrt{\log n}$ with $\sigma$ the standard deviation of each noise entry $\{E_{ij}\}$.
Figure~\ref{fig:exp1}(a) fixes $r=5,\sigma=10^{-3}$ and examines
the dependency of the Euclidean error $\|\widehat{\bm{L}}-\bm{L}^{\star}\|_{\mathrm{F}}$
on the size $\sqrt{n}$. Similarly,
Figure~\ref{fig:exp1}(b) fixes $r=5,n=1000$ and displays the estimation
error $\|\widehat{\bm{L}}-\bm{L}^{\star}\|_{\mathrm{F}}$ as the noise
size $\sigma$ varies in a log-log plot. As can be seen from Figure~\ref{fig:exp1},
the performance of the oracle-aided estimator (\ref{eq:matrix-denoising})
matches the theoretical prediction \eqref{eq:L-hat-oracle}, namely,
the numerical estimation error $\|\widehat{\bm{L}}-\bm{L}^{\star}\|_{\mathrm{F}}$
is proportional to both $\sqrt{n}$ and $\sigma$. Perhaps more intriguingly,
even without the help of the oracle, the original estimator~(\ref{eq:cvx})
performs quite well and behaves qualitatively similarly. In comparison
with the bound (\ref{eq:L-hat-no-oracle}) derived in the prior work
\cite{zhou2010stable}, our numerical experiments suggest that the
convex estimator (\ref{eq:cvx}) might perform much better than previously
predicted.

\bigskip\noindent All in all, there seems to be a large gap between
the practical performance of (\ref{eq:cvx}) and the existing theoretical
support. This calls for a new theory that better explains practice,
which we pursue in the current paper. We remark in passing that statistical guarantees have been developed in~\cite{agarwal2012noisy, klopp2017robust} for other convex estimators (i.e.~the ones that are different from the convex estimator~(\ref{eq:cvx}) considered herein). We shall compare our results with theirs later in Section~\ref{subsec:Main-results}.

\subsection{Models, assumptions and notation}

As it turns out, the appealing empirical performance of the convex
program (\ref{eq:cvx}) in the presence of both sparse outliers and
zero-mean random noise can be justified in theory. Towards this end,
we need to introduce several notations and model assumptions that
will be used throughout. Let $\bm{U}^{\star}\bm{\Sigma}^{\star}\bm{V}^{\star\top}$
be the singular value decomposition (SVD) of the unknown rank-$r$
matrix $\bm{L}^{\star}\in\mathbb{R}^{n_1\times n_2}$, where $\bm{U}^{\star}\in\mathbb{R}^{n_1\times r}$ and $\bm{V}^{\star}\in\mathbb{R}^{n_2\times r}$
consist of orthonormal columns and $\bm{\Sigma}^{\star}=\mathsf{diag}\{\sigma_{1}^{\star},\ldots,\sigma_{r}^{\star}\}$
is a diagonal matrix. Here, we let
\[
\sigma_{\max}\coloneqq\sigma_{1}^{\star}\geq\sigma_{2}^{\star}\geq\cdots\geq\sigma_{r}^{\star}\eqqcolon\sigma_{\min}\qquad\text{and}\qquad\kappa\coloneqq\sigma_{\max}/\sigma_{\min}
\]
represent the singular values and the condition number of $\bm{L}^{\star}$,
respectively. We denote by $\Omega^{\star}$ the support set of $\bm{S}^{\star}$,
that is,
\begin{equation}
\Omega^{\star}\coloneqq\{(i,j)\in\Omega_{\mathsf{obs}}:S_{ij}^{\star}\neq0\}.\label{eq:support-Omega-Sstar}
\end{equation}
With this set of notation in place, we list below our key model assumptions.

\begin{assumption}[\textbf{Incoherence}]\label{assumption:incoherence}The
low-rank matrix $\bm{L}^{\star}$ with SVD $\bm{L}^{\star}=\bm{U}^{\star}\bm{\Sigma}^{\star}\bm{V}^{\star\top}$
is assumed to be $\mu$-incoherent in the sense that
\begin{equation}
\left\Vert \bm{U}^{\star}\right\Vert _{2,\infty}\leq\sqrt{\frac{\mu}{n_1}}\left\Vert \bm{U}^{\star}\right\Vert _{\mathrm{F}}=\sqrt{\frac{\mu r}{n_1}}\quad\text{and}\quad\left\Vert \bm{V}^{\star}\right\Vert _{2,\infty}\leq\sqrt{\frac{\mu}{n_2}}\left\Vert \bm{V}^{\star}\right\Vert _{\mathrm{F}}=\sqrt{\frac{\mu r}{n_2}}.\label{eq:incoherence-defn}
\end{equation}
Here, $\left\Vert \bm{U}\right\Vert _{2,\infty}$ denotes the largest
$\ell_{2}$ norm of all rows of a matrix $\bm{U}$. \end{assumption}

\begin{assumption}[\textbf{Random sampling}]\label{assumption:random-sampling}\textbf{
}Each entry is observed independently with probability $p$, namely,
\begin{equation}
\mathbb{P}\left\{ (i,j)\in\Omega_{\mathsf{obs}}\right\} =p.\label{eq:random-sampling-assumption}
\end{equation}
\end{assumption}

\begin{assumption}[\textbf{Random locations of outliers}]\label{assumption:random-location}
Each observed entry is independently corrupted by an outlier with
probability $\rho_{\mathsf{s}}$, namely,
\begin{equation}
\mathbb{P}\left\{ (i,j)\in\Omega^{\star}\mid(i,j)\in\Omega_{\mathsf{obs}}\right\} =\rho_{\mathsf{s}},\label{eq:random-corruption-location-assumption}
\end{equation}
where $\Omega^{\star}\subseteq\Omega_{\mathsf{obs}}$ is the support
of the outlier matrix $\bm{S}^{\star}$. \end{assumption}

\begin{assumption}[\textbf{Random signs of outliers}]\label{assumption:random-sign}\textbf{
}The signs of the nonzero entries of $\bm{S}^{\star}$ are i.i.d.~symmetric Bernoulli
random variables (independent from the locations), namely,
\begin{equation}
\mathsf{sign}(S_{ij}^{\star})\overset{\mathrm{ind.}}{=}\begin{cases}
1, & \text{with probability 1/2},\\
-1, & \text{else},
\end{cases}\qquad\text{for all }(i,j)\in\Omega^{\star}.\label{eq:random-sign-assumption}
\end{equation}
\end{assumption}

\begin{assumption}[\textbf{Random noise}]\label{assumption:random-noise}\textbf{
}The noise matrix $\bm{E}=[E_{ij}]$ is composed
of independent~symmetric\footnote{In fact, we only require $E_{ij}$ to be symmetric for all $(i,j)\in\Omega^{\star}$.} zero-mean sub-Gaussian random variables
with sub-Gaussian norm at most $\sigma>0$, i.e.~$\|E_{ij}\|_{\psi_{2}}\leq\sigma$
(see \cite[Definition 5.7]{Vershynin2012} for precise definitions).\end{assumption}

We take a moment to expand on our model assumptions. Assumption~\ref{assumption:incoherence}
is standard in the low-rank matrix recovery literature~\cite{ExactMC09,CanLiMaWri09,chen2015incoherence,chi2018nonconvex}.
If $\mu$ is small, then this assumption specifies that the singular
spaces of $\bm{L}^{\star}$ is not sparse in the standard basis, thus
ensuring that $\bm{L}^{\star}$ is not simultaneously low-rank and
sparse. Assumption~\ref{assumption:random-location} requires the
sparsity pattern of the outliers $\bm{S}^{\star}$ to be random, which
precludes it from being simultaneously sparse and low-rank. In essence,
Assumptions~\ref{assumption:incoherence}~and~\ref{assumption:random-location} are identifiability conditions,
taken together as a sort of separation condition on $(\bm{L}^{\star},\bm{S}^{\star})$,
which plays a crucial role in guaranteeing exact recovery in the noiseless
case (i.e.~$\bm{E}=\bm{0}$); see~\cite{CanLiMaWri09} for more
discussions on these conditions. Assumption~\ref{assumption:random-sign}
requires the signs of the outliers to be random, which has also been made in \cite{zhou2010stable,wong2017matrix}.\footnote{Note that while the theorems in \cite{zhou2010stable,wong2017matrix} do not make
explicit this random sign assumption, the proofs therein do rely
on this assumption to guarantee the existence of certain approximate dual certificates.} We shall discuss in detail the crucial role of this random sign assumption (as opposed to deterministic sign patterns) in Section~\ref{subsec:discussion-Random-sign}.

\begin{comment}
It turns out that this random sign assumption is crucial for the convex
estimator (\ref{eq:cvx}) to perform optimally in the noise case.
We defer the discussion to Section~\ref{subsec:discussion-Random-sign}.
\end{comment}

Finally, we introduce several notation to be used throughout.
Denote by $f(n)\lesssim g(n)$ or $f(n)=O(g(n))$ the condition $\vert f(n)\vert\leq Cg(n)$
for some constant $C>0$ when $n$ is sufficiently large; we use $f(n)\gtrsim g(n)$
to denote $f(n)\geq C\vert g(n)\vert$ for some constant $C>0$ when
$n$ is sufficiently large; we also use $f(n)\asymp g(n)$ to indicate
that $f(n)\lesssim g(n)$ and $f(n)\gtrsim g(n)$ hold simultaneously.
The notation $f(n)\gg g(n)$ (resp.~$f(n)\ll g(n)$) means that there
exists a sufficiently large (resp.~small) constant $c_{1}>0$ (resp.~$c_{2}>0$)
such that $f(n)\geq c_{1}g(n)$ (resp.~$f(n)\leq c_{2}g(n)$). For
any subspace $T$, we denote by $\mathcal{P}_{T}(\bm{M})$ the Euclidean
projection of a matrix $\bm{M}$ onto the subspace $T$, and let $\mathcal{P}_{T^{\perp}}(\bm{M})\coloneqq\bm{M}-\mathcal{P}_{T}(\bm{M})$.
For any index set $\Omega$, we denote by $\mathcal{P}_{\Omega}(\bm{M})$
the Euclidean projection of a matrix $\bm{M}$ onto the subspace of
matrices supported on $\Omega$, and define $\mathcal{P}_{\Omega^{\mathrm{c}}}(\bm{M})\coloneqq\bm{M}-\mathcal{P}_{\Omega}(\bm{M})$.
For any matrix $\bm{M}$, we let $\|\bm{M}\|$, $\|\bm{M}\|_{\mathrm{F}}$,
$\|\bm{M}\|_{*}$, $\|\bm{M}\|_{1}$ and $\|\bm{M}\|_{\infty}$ denote
its spectral norm, Frobenius norm, nuclear norm, entrywise $\ell_{1}$
norm, and entrywise $\ell_{\infty}$ norm, respectively.

\subsection{Main results\label{subsec:Main-results}}

Armed with the above model assumptions, we are positioned to present
our improved statistical guarantees for convex relaxation (\ref{eq:cvx})
in the random noise setting. Without loss of generality, assume that $$n_1\geq n_2.$$ As we shall elucidate in Section~\ref{subsec:contributions}
and Section~\ref{sec:Architecture-of-the-proof}, our theory is
established by exploiting an intriguing and intimate connection
between convex relaxation and nonconvex optimization, and hence the
title of this paper.

For the sake of simplicity, we shall start by presenting our statistical guarantees when the rank $r$, the condition number $\kappa$ and the incoherence parameter $\mu$ of $\bm{L}^\star$ are all bounded by some constants. Despite its simplicity, this setting subsumes as special cases a wide array of fundamentally important applications, including angular and phase synchronization \cite{singer2011angular} in computational biology, joint shape mapping problem \cite{huang2013consistent,chen2014matching} in computer vision, and so on. All of these problems involve estimating a very well-conditioned matrix $\bm{L}^{\star}$ with a small rank. 
\begin{theorem}\label{thm:main-convex-constant}Suppose that Assumptions~\ref{assumption:incoherence}-\ref{assumption:random-noise}
	hold, and that $r,\kappa,\mu =O(1)$. Take $\lambda=C_{\lambda}\sigma\sqrt{n_1p}$ and $\tau=C_{\tau}\sigma\sqrt{\log n_2}$
	in (\ref{eq:cvx}) for some large enough constants $C_{\lambda},C_\tau>0$.
%	 Define 
%	\begin{align}
%	\label{defn:delta-n}
%		\delta_n \coloneqq \frac{\sigma}{\sigma_{\min}}\sqrt{\frac{n}{p}}.
%	\end{align}
	Assume that
	\begin{equation}
	n_1n_2p\geq C_{\mathsf{sample}}n_1\log^{6}n_1,\quad
	\frac{\sigma}{\sigma_{\min}}\sqrt{\frac{n_1}{p}} \leq \frac{c_{\mathsf{noise}}}{\sqrt{\log n_1}}
	%\sigma\sqrt{\frac{n}{p}}\leq c_{\mathsf{noise}}\frac{\sigma_{\min}}{\sqrt{\log n}},
	\quad\text{and}\quad
	\rho_{s}\leq\frac{c_{\mathsf{outlier}}}{\log n_1}\label{eq:conditions}
	\end{equation}
	for some sufficiently large constant $C_{\mathsf{sample}}>0$ and
	some sufficiently small constants $c_{\mathsf{noise}},c_{\mathsf{outlier}}>0$.
	Then with probability exceeding $1-O(n_2^{-3})$, the following holds:
	\begin{enumerate}
		\item Any minimizer $(\bm{L}_{\mathsf{cvx}},\bm{S}_{\mathsf{cvx}})$ of
		the convex program (\ref{eq:cvx}) obeys \begin{subequations}\label{eq:Zcvx-error}
			\begin{align}
			\left\Vert\bm{L}_{\mathsf{cvx}}-\bm{L}^{\star}\right\Vert_{\mathrm{F}} & \leq C_{\mathsf{err}} \frac{\sigma}{\sigma_{\min}}\sqrt{\frac{n_1}{p}} \left\Vert\bm{L}^{\star}\right\Vert_{\mathrm{F}} \label{eq:main-fro-norm-error}\\
			\left\Vert\bm{L}_{\mathsf{cvx}}-\bm{L}^{\star}\right\Vert_{\infty} & \leq C_{\mathsf{err}} \frac{\sigma}{\sigma_{\min}}\sqrt{\frac{n_1\log n_1}{p}}\left\Vert\bm{L}^{\star}\right\Vert_{\infty} \label{eq:main-inf-norm-error}\\
			\left\Vert\bm{L}_{\mathsf{cvx}}-\bm{L}^{\star}\right\Vert & \leq C_{\mathsf{err}} \frac{\sigma}{\sigma_{\min}}\sqrt{\frac{n_1}{p}} \left\Vert\bm{L}^{\star}\right\Vert \label{eq:main-spectral-norm-error}
			\end{align}
		\end{subequations}
		for some constant $C_{\mathsf{err}}>0$.  
		\item Letting $\bm{L}_{\mathsf{cvx},r}\coloneqq\mathrm{arg}\min_{\bm{L}:\mathsf{rank}(\bm{L})\leq r}\|\bm{L}-\bm{L}_{\mathsf{cvx}}\|_{\mathrm{F}}$
		be the best rank-$r$ approximation of $\bm{L}_{\mathsf{cvx}}$, we
		have
		\begin{equation}
		\|\bm{L}_{\mathsf{cvx},r}-\bm{L}_{\mathsf{cvx}}\|_{\mathrm{F}}\leq
		\frac{1}{n_2^{5}}\cdot\frac{\sigma}{\sigma_{\min}}\sqrt{\frac{n_1}{p}}\, \left\Vert\bm{L}^{\star}\right\Vert_{\mathrm{F}},\label{eq:Zcvx-r-bound}
		\end{equation}
		and the statistical guarantees (\ref{eq:Zcvx-error}) hold unchanged
		if $\bm{L}_{\mathsf{cvx}}$ is replaced by $\bm{L}_{\mathsf{cvx},r}$.
	\end{enumerate}
\end{theorem}

Before we embark on interpreting our statistical guarantees, let us
first parse the required conditions~(\ref{eq:conditions}) in Theorem~\ref{thm:main-convex-constant}. For simplicity we assume that $n_1=n_2=n$. %See Table \ref{table:comparison} for a summary of our results vs.~prior statistical guarantees.

\begin{itemize}
	\item \emph{Missing data}. Theorem \ref{thm:main-convex-constant} accommodates the case where a dominant fraction of entries are unobserved (more precisely, the sample size can be as low as an order of $n\,\mathrm{poly}\log n$). This is an appealing result since, even when there is no noise and no outlier (i.e.~$\bm{E}=\bm{0}$ and $\rho_{s}=0$), the minimal sample size
required for exact matrix completion is at least on the order of $n\log n$
		\cite{CanTao10}. In comparison, prior theory on robust PCA with both sparse outliers and dense additive noise is either based on full observations \cite{zhou2010stable,agarwal2012noisy}, or assumes the sampling rate $p$ exceeds some universal constant \cite{wong2017matrix}. In other words, these prior results  require the number of observed entries to exceed the order of $n^2$. The only exception is~\cite{klopp2017robust}, which also allows a significant amount of missing data (i.e.~$p\gtrsim(\mathrm{poly}\log n)/ n$).
\item \emph{Noise levels}. The noise condition, namely $\sigma\sqrt{n\log n/p}\lesssim\sigma_{\min}$,
accommodates a wide range of noise levels. To see this, it is straightforward
to check that this noise condition is equivalent to
\[
\sigma\lesssim\sqrt{\frac{np}{\log n}}\,\|\bm{L}^{\star}\|_{\infty}
\]
as long as $r,\mu,\kappa\asymp1$. In other words, the entrywise noise
level $\sigma$ is allowed to be significantly larger than the maximum
magnitude of the entries in the low-rank matrix $\bm{L}^{\star}$, as long as
$p\gg(\log n)/n$.
\item \emph{Tolerable fraction of outliers}. The above theorem assumes that
no more than a fraction $\rho_{s}\lesssim1/\log n$ of observations
are corrupted by outliers. In words, our theory  allows \emph{nearly}
a constant proportion (up to a logarithmic order) of the entries of $\bm{L}^{\star}$ to be corrupted
with arbitrary magnitudes.
\end{itemize}

Next, we move on to the interpretation of our statistical guarantees. Note that we still assume that $n_1=n_2=n$ for ease of presentation.

\begin{itemize}
\item \emph{Near-optimal statistical guarantees}. Our first result (\ref{eq:main-fro-norm-error})
gives an Euclidean estimation error bound of~(\ref{eq:cvx})
\begin{equation}
\left\Vert \bm{L}_{\mathsf{cvx}}-\bm{L}^{\star}\right\Vert _{\mathrm{F}}\lesssim\sigma\sqrt{\frac{n}{p}}.\label{eq:L2-error-simplified}
\end{equation}
This cannot be improved even when an oracle has informed us of the
outliers $\bm{S}^{\star}$ and the tangent space of $\bm{L}^{\star}$;
see~\cite[Section III.B]{CanPla10}. We remark that under similar
model assumptions, the paper \cite{wong2017matrix} derived an estimation
error bound for a constrained version of the convex program (\ref{eq:cvx}), which asserts that this convex estimator $\widetilde{\bm{L}}_{\mathsf{cvx}}$ satisfies 
\footnote{More specifically, \cite[Theorem 4]{wong2017matrix} studies the following convex program  $\text{minimize}_{\bm{L},\bm{S}\in\mathbb{R}^{n\times n}}\Vert \bm{L}\Vert _{\ast}+\lambda\Vert \bm{S}\Vert_{1}$ s.t.~$\Vert\mathcal{P}_{\Omega_{\mathsf{obs}}}(\bm{L}+\bm{S}-\bm{M})\Vert_{\mathrm{F}}\leq\delta$. Here, the quantity $\delta$ needs to be larger than $\Vert\mathcal{P}_{\Omega_{\mathsf{obs}}}(\bm{L}+\bm{S}-\bm{M})\Vert_{\mathrm{F}}$. Under our setting, the minimum level of $\delta$ should be a high-probability upper bound on $\Vert\mathcal{P}_{\Omega_{\mathsf{obs}}}(\bm{E})\Vert_{\mathrm{F}}$, which is on the order of $\sigma n\sqrt{p}$. With this choice of $\delta$, \cite[Theorem 4]{wong2017matrix} yields
	$\Vert\widetilde{\bm{L}}_{\mathsf{cvx}}-\bm{L}^\star\Vert_{\mathrm{F}}\leq[2+8\sqrt{n}(1+\sqrt{8/p})]\delta\lesssim\sigma n^{1.5}$.}
\begin{equation}
\big\Vert \widetilde{\bm{L}}_{\mathsf{cvx}}-\bm{L}^{\star}\big\Vert _{\mathrm{F}}\lesssim\sigma n^{1.5},\label{eq:other-bound}
\end{equation}
with the proviso that $p$ is at least on the constant order. The restriction on $p$ arises from the dual certificate constructed in \cite{CanLiMaWri09}, which is also used in the Proof of Theorem 4 in \cite{wong2017matrix}.
While this is sub-optimal compared to our results in the setting considered herein, 
it is worth pointing out that the bound therein accommodates arbitrary
noise matrix $\bm{E}$ (e.g.~deterministic, adversary), and here
in~(\ref{eq:other-bound}) we specialize their result to the random
noise setting, namely the noise $\bm{E}$ obeys Assumption~\ref{assumption:random-noise}.
In addition, under the full observation (i.e.~$p=1$) setting, the
paper \cite{agarwal2012noisy} derived an estimation error bound for
a convex program similar to (\ref{eq:cvx}), but with an additional
constraint regularizing the spikiness of the low-rank component. Note that instead of imposing the incoherence condition as in Assumption~\ref{assumption:incoherence}, the prior work \cite{agarwal2012noisy} assumes a milder spikiness condition on $\bm{L}^\star$, which only constrains the maximum entry in the matrix $\bm{L}^\star$ is not too large.
When $\{E_{ij}\}$ are i.i.d.~drawn from $\mathcal{N}(0,\sigma^{2})$
and when there is no missing data (i.e.~$p=1$), the Euclidean estimation
error bound achievable by their estimator $\bm{L}_{\mathsf{cvx}}^{\mathsf{ANW}}$
reads
\begin{equation}
\left\Vert \bm{L}_{\mathsf{cvx}}^{\mathsf{ANW}}-\bm{L}^{\star}\right\Vert _{\mathrm{F}}\lesssim\sigma\sqrt{n}\max\left\{ 1,\sqrt{n\rho_{\mathsf{s}}\log n}\right\} +\Vert\bm{L}^{\star}\Vert_{\infty}n\sqrt{\rho_{\mathsf{s}}},\label{eq:martin-L2bound}
\end{equation}
		which is sub-optimal compared to our results. In particular,  (i) the bound (\ref{eq:martin-L2bound}) does not vanish
		even as the noise level decreases to zero, and (ii) it becomes looser as $\rho_{\mathsf{s}}$ grows (e.g.~if $\rho_{\mathsf{s}}\asymp 1 / \log n$, the bound  (\ref{eq:martin-L2bound}) is $O(\sqrt{n})$ larger than our bound). Moreover, the work \cite{agarwal2012noisy}
		did not account for missing data. Similar to~\cite{agarwal2012noisy} (but with an additional spikiness condition on $\bm{S}^\star$), the paper \cite{klopp2017robust} derived an estimation error bound for a constrained convex program, with a new constraint regularizing the spikiness of the sparse outliers. Their Euclidean estimation error bound reads 
\begin{equation}
\left\Vert\bm{L}_{\mathsf{cvx}}^{\mathsf{KLT}}-\bm{L}^\star\right\Vert_{\mathrm{F}}\lesssim\max\left\{\sigma,\left\Vert\bm{L}^\star\right\Vert_\infty,\left\Vert\bm{S}^\star\right\Vert_\infty\right\}\sqrt{\frac{n\log n}{p}}\max\left\{1,\sqrt{np\rho_{\mathrm{s}}}\right\},\label{eq:KLT-L2bound}
\end{equation}
		which is also sub-optimal compared to our results. In particular, (1) their error bound degrades as the magnitude $\|\bm{S}^{\star}\|_{\infty}$ of the outlier increases; (2) when there is no missing data (i.e.~$p=1$), their bound might be off by a factor as large as $O(\sqrt{n})$. It is worth emphasizing that the theory developed in these prior works is developed to accommodate a broader range of matrices. For example, both \cite{agarwal2012noisy} and \cite{klopp2017robust} study the set of entrywise bounded low-rank matrices (without assuming the incoherence condition); \cite{agarwal2012noisy} even allows $\bm{L}^\star$ to be approximately low rank. To ease comparison, Table~\ref{table:comparison} displays a summary of our results vs.~prior statistical guarantees when specialized to the settings considered herein.
		\begin{table}[t]
			\caption{Comparison of our statistical guarantee and prior theory.}\label{table:comparison}
			\vspace{0.8em}
			\centering
			%	\begin{tabular}{c|c|c|c|c}
			%		\hline
			%		$\vphantom{2_{2_{2_{2}}}^{2^{2^{2}}}}$ & \cite{zhou2010stable} & \cite{wong2017matrix} & \cite{agarwal2012noisy} & This paper\tabularnewline
			%		\hline
			%		Euclidean error & $\sigma n^{2}\vphantom{2_{2_{2_{2}}}^{2^{2^{2}}}}$ & $\sigma n^{1.5}\vphantom{2_{2_{2_{2}}}^{2^{2^{2}}}}$ & $\sigma\sqrt{n}\max\{1,\sqrt{n\rho_{\mathsf{s}}\log n}\}+\Vert\bm{L}^{\star}\Vert_{\infty}n\sqrt{\rho_{\mathsf{s}}}\vphantom{2_{2_{2_{2}}}^{2^{2^{2}}}}$ & $\sigma\sqrt{n/p}\vphantom{2_{2_{2_{2}}}^{2^{2^{2}}}}$\tabularnewline
			%		\hline
			%		\begin{tabular}[c]{@{}c@{}}Accounting for\\ missing data\end{tabular} & no & yes $(p\gtrsim 1)$ & no & yes $(p\gtrsim \frac{\mathrm{poly}\log(n)}{n})$ \tabularnewline
			%		\hline
			%	\end{tabular}
			\begin{tabular}{c|c|c}
				\hline
				$\vphantom{2_{2_{2_{2}}}^{2^{2^{2}}}}$ & Euclidean estimation error & Accounting for missing data \tabularnewline
				\hline
				\cite{zhou2010stable} & $\sigma n^{2}\vphantom{2_{2_{2_{2}}}^{2^{2^{2}}}}$ & no  \tabularnewline
				\hline
				\cite{agarwal2012noisy} & $\sigma\sqrt{n}\max\{\sqrt{r},\sqrt{n\rho_{\mathsf{s}}\log n}\}+\Vert\bm{L}^{\star}\Vert_{\infty}n\sqrt{\rho_{\mathsf{s}}}\vphantom{2_{2_{2_{2}}}^{2^{2^{2}}}}$ & no \tabularnewline
				\hline
				\cite{wong2017matrix} & $\sigma n^{1.5}\vphantom{2_{2_{2_{2}}}^{2^{2^{2}}}}$ & yes ($p\gtrsim1$)  \tabularnewline
				\hline
				\cite{klopp2017robust} & $\max\{\sigma,\Vert\bm{L}^\star\Vert_\infty,\Vert\bm{S}^\star\Vert_\infty\}\sqrt{(n\log n)/p}\max\{1,\sqrt{np\rho_{\mathrm{s}}}\}\vphantom{2_{2_{2_{2}}}^{2^{2^{2}}}}$ & yes ($p\gtrsim(\mathrm{poly}\log n)/n$) \tabularnewline
				\hline
				This paper & $\sigma\sqrt{nr/p}\vphantom{2_{2_{2_{2}}}^{2^{2^{2}}}}$ & yes ($p\gtrsim\kappa^4\mu^2r^2(\mathrm{poly}\log n)/ n$) \tabularnewline
				\hline
			\end{tabular}
		\end{table}

\item \emph{Entrywise and spectral norm error control}. Moving beyond Euclidean
estimation errors, our theory also provides statistical guarantees
measured by two other important metrics: the entrywise $\ell_{\infty}$
norm (cf.~(\ref{eq:main-inf-norm-error})) and the spectral norm
(cf.~(\ref{eq:main-spectral-norm-error})). In particular, our entrywise
error bound (\ref{eq:main-inf-norm-error}) in reads
\begin{equation}
\left\Vert \bm{L}_{\mathsf{cvx}}-\bm{L}^{\star}\right\Vert _{\mathrm{\infty}}\lesssim\sigma\sqrt{\frac{\log n}{np}}\label{eq:Linf-error-simplified}
\end{equation}
as long as $r,\kappa,\mu\asymp1$, which is about $O(n)$ times small
than the Euclidean loss (\ref{eq:L2-error-simplified}) modulo some
logarithmic factor. This uncovers an appealing ``delocalization''
behavior of the estimation errors, namely, the estimation errors of
$\bm{L}^{\star}$ are fairly spread out across all entries. This can
also be viewed as an \textquotedblleft implicit regularization\textquotedblright{}
phenomenon: the convex program automatically controls the spikiness
of the low-rank solution, without the need of explicitly regularizing
it (e.g.~adding a constraint $\|\bm{L}\|_{\infty}\leq\alpha$ as
adopted in the prior work \cite{agarwal2012noisy,klopp2017robust}). See Figure~\ref{fig:infty_op} for the numerical evidence for the relative entrywise and spectral norm error of $\bm{L}_{\mathsf{cvx}}$.
\begin{figure}[t!]
	\center
	
	\includegraphics[scale=0.35]{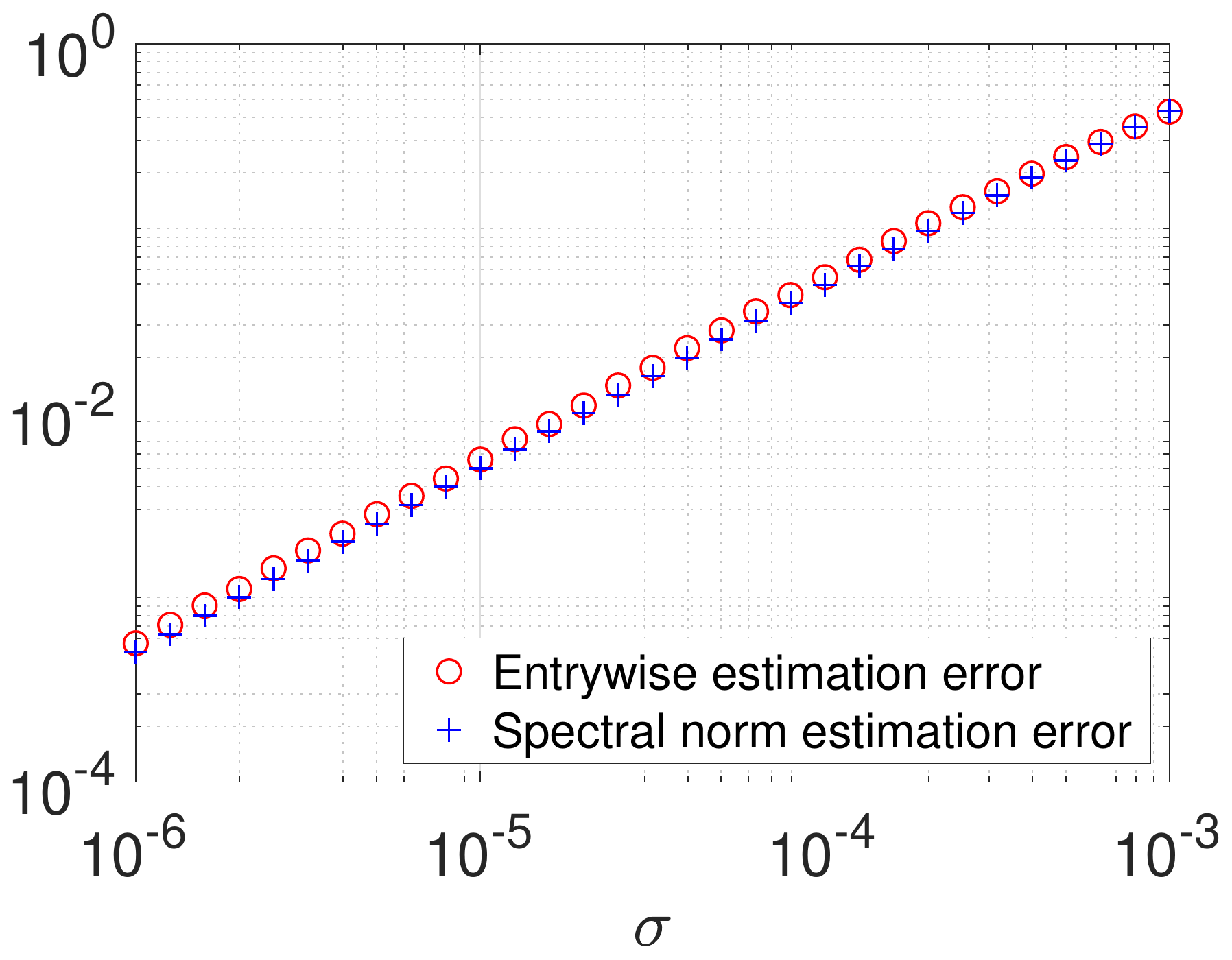}
	
	\caption{The relative estimation error of $\bm{L}_{\mathsf{cvx}}$ measured
		by both $\|\cdot\|_{\infty}$ (i.e.~$\|\bm{L}_{\mathsf{cvx}}-\bm{L}^{\star}\|_{\infty}/\|\bm{L}^{\star}\|_{\infty}$)
		and $\|\cdot\|$ (i.e.~$\|\bm{L}_{\mathsf{cvx}}-\bm{L}^{\star}\|/\|\bm{L}^{\star}\|$)
		vs.~the standard deviation $\sigma$ of the noise in a log-log plot. The results are
		reported for $n=1000$, $r=5$, $p=0.2$, $\rho_{s}=0.1$, $\lambda=5\sigma\sqrt{np}$, $\tau=2\sigma\sqrt{\log n}$,
		and are averaged over 50 independent trials. In addition, the data generating process is similar to that in Figure~\ref{fig:exp1}. \label{fig:infty_op}}
\end{figure}

\item \emph{Approximate low-rank structure of the convex estimator $\bm{L}_{\mathsf{cvx}}$}.
Last but not least, Theorem~\ref{thm:main-convex-constant} ensures that the convex
estimate $\bm{L}_{\mathsf{cvx}}$ is nearly rank-$r$, so that a rank-$r$
approximation of $\bm{L}_{\mathsf{cvx}}$ is extremely accurate. In
other words, the convex program automatically adapts to the true rank of
$\bm{L}^{\star}$ without having any prior knowledge about $r$. As we
shall see shortly, this is a crucial observation underlying the intimate
connection between convex relaxation and a certain nonconvex approach.
\end{itemize}
%
%\yly{I move it to the next subsubsection.}In addition, we make note of several aspects of our theory that call
%for further improvement. For instance, when there is no missing data,
%the rank $r$ of the unknown matrix $\bm{L}^{\star}$ needs to satisfy
%$r\lesssim\sqrt{n}$. On the positive side, our result allows $r$
%to grow with the problem dimension $n$. However, prior results in
%the noiseless case \cite{CanLiMaWri09,li2011compressed} allow $r$
%to grow almost linearly with $n$. This unsatisfactory aspect arises from the suboptimal analysis (in terms of the dependency on $r$) of a tightly related nonconvex estimation algorithm (to be elaborated on later), which, to the best
%of our knowledge, has not been resolved in the nonconvex low-rank
%matrix recovery literature~\cite{ma2017implicit,chen2019nonconvex}.
%See Section~\ref{sec:Prior-art} for more discussions about this point.
%Moreover, when $\bm{E}=\bm{0}$, it is known that $\rho_{s}$ can
%be as large as a constant even when $r$ grows with~$n$~\cite{li2011compressed,chen2013low}
%--- a case not covered by our current theory for noisy case.

%\subsubsection{Case 2: general settings}
Moving beyond the setting with $r,\kappa,\mu \asymp 1$, we have developed theoretical guarantees that allow $r,\kappa,\mu$ to grow with the problem dimension~$n_1,n_2$. The result is this.
\begin{theorem}\label{thm:main-convex}Suppose that Assumptions~\ref{assumption:incoherence}-\ref{assumption:random-noise}
	hold and that $n_1\geq n_2$. Take $\lambda=C_{\lambda}\sigma\sqrt{n_1 p}$ and $\tau=C_{\tau}\sigma\sqrt{\log n_2}$
	in (\ref{eq:cvx}) for some large enough constants $C_{\lambda},C_\tau>0$. Assume that
	\begin{equation}\label{eq:conditions-general}
	n_1 n_2 p\geq C_{\mathsf{sample}}\kappa^{4}\mu^{2}r^{2}n_1\log^{6}n_1,\quad
	\frac{\sigma}{\sigma_{\min}}\sqrt{\frac{n_1}{p}} \leq \frac{c_{\mathsf{noise}}}{\sqrt{\kappa^{4}\mu r\log n_1}},
	\ \text{and}\quad\rho_{s}\leq\frac{c_{\mathsf{outlier}}}{\kappa^{3}\mu r\log n_1}
	\end{equation}
	for some sufficiently large constant $C_{\mathsf{sample}}>0$ and
	some sufficiently small constants $c_{\mathsf{noise}},c_{\mathsf{outlier}}>0$.
	Then with probability exceeding $1-O(n_2^{-3})$, the following holds:
	\begin{enumerate}
		\item Any minimizer $(\bm{L}_{\mathsf{cvx}},\bm{S}_{\mathsf{cvx}})$ of
		the convex program (\ref{eq:cvx}) obeys \begin{subequations}\label{eq:Zcvx-error-general}
			\begin{align}
				\left\Vert\bm{L}_{\mathsf{cvx}}-\bm{L}^{\star}\right\Vert_{\mathrm{F}} & \leq C_{\mathsf{err}}  \kappa\frac{\sigma}{\sigma_{\min}}\sqrt{\frac{n_1}{p}}\,\left\Vert\bm{L}^{\star}\right\Vert_{\mathrm{F}} \label{eq:main-fro-norm-error-general}\\
			\left\Vert\bm{L}_{\mathsf{cvx}}-\bm{L}^{\star}\right\Vert_{\infty} & \leq C_{\mathsf{err}} \sqrt{\kappa^{3}\mu r}\cdot\frac{\sigma}{\sigma_{\min}}\sqrt{\frac{n_1\log n_1}{p}}\left\Vert\bm{L}^{\star}\right\Vert_{\infty} \label{eq:main-inf-norm-error-general}\\
			\left\Vert\bm{L}_{\mathsf{cvx}}-\bm{L}^{\star}\right\Vert & \leq C_{\mathsf{err}} \frac{\sigma}{\sigma_{\min}}\sqrt{\frac{n_1}{p}} \left\Vert\bm{L}^{\star}\right\Vert\label{eq:main-spectral-norm-error-general}
			\end{align}
		\end{subequations}
			for some constant $C_{\mathsf{err}}>0$. 
		%Here, we recall the definition of  from (\ref{defn:delta-n}).
		\item Letting $\bm{L}_{\mathsf{cvx},r}\coloneqq\mathrm{arg}\min_{\bm{L}:\mathsf{rank}(\bm{L})\leq r}\|\bm{L}-\bm{L}_{\mathsf{cvx}}\|_{\mathrm{F}}$
		be the best rank-$r$ approximation of $\bm{L}_{\mathsf{cvx}}$, we
		have
		\begin{equation}
		\|\bm{L}_{\mathsf{cvx},r}-\bm{L}_{\mathsf{cvx}}\|_{\mathrm{F}}\leq
		\frac{1}{n_2^{5}}\cdot\frac{\sigma}{\sigma_{\min}}\sqrt{\frac{n_1}{p}} \left\Vert\bm{L}^{\star}\right\Vert_{\mathrm{F}},\label{eq:Zcvx-r-bound-general}
		\end{equation}
		and the statistical guarantees (\ref{eq:Zcvx-error-general}) hold unchanged
		if $\bm{L}_{\mathsf{cvx}}$ is replaced by $\bm{L}_{\mathsf{cvx},r}$.
	\end{enumerate}
\end{theorem}

Similar to Theorem~\ref{thm:main-convex-constant}, our general theory (i.e.~Theorem~\ref{thm:main-convex}) provides the estimation error of the convex estimator $\bm{L}_{\mathsf{cvx}}$ in three different norms (i.e.~the Euclidean, entrywise and operator norms), and reveals the near low-rankness of the convex estimator (cf.~(\ref{eq:Zcvx-r-bound-general})) as well as the implicit regularization phenomenon (cf.~(\ref{eq:main-inf-norm-error-general})).  

\begin{figure}
	\center
	
	\includegraphics[scale=0.35]{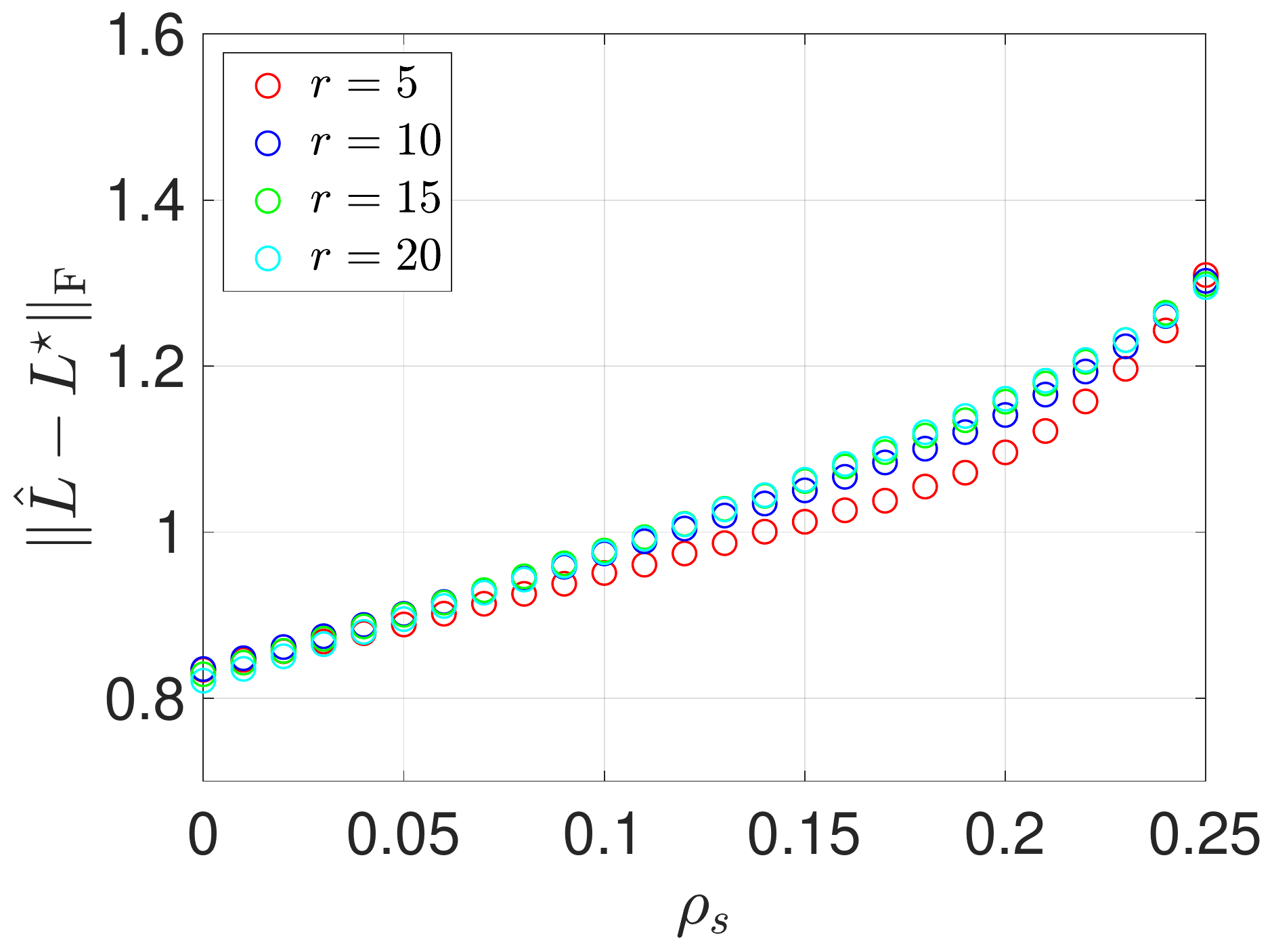}
	
	\caption{Euclidean estimation errors of \eqref{eq:cvx}
		vs.~$\rho_{\mathsf{s}}$ under four different ranks $r=5,10,15,20$. The results are
		reported for $n=1000$, $p=0.04r$, $\sigma=10^{-3}$, $\lambda=5\sigma\sqrt{np}$, $\tau=2\sigma\sqrt{\log n}$,
		and are averaged over 50 independent trials. In addition, the data generating process is similar to that in Figure~\ref{fig:exp1}. \label{fig:rho-r}}
\end{figure}

Finally, we make note of several aspects of our general theory that call
for further improvement. For instance, when there is no missing data and $n_1=n_2=n$,
the rank $r$ of the unknown matrix $\bm{L}^{\star}$ needs to satisfy
$r\lesssim\sqrt{n}$. On the positive side, our result allows $r$
to grow with the problem dimension $n$. However, prior results in
the noiseless case \cite{CanLiMaWri09,li2011compressed} allow $r$
to grow almost linearly with $n$. This unsatisfactory aspect arises from the suboptimal analysis (in terms of the dependency on $r$) of a tightly related nonconvex estimation algorithm (to be elaborated on later), which, to the best
of our knowledge, has not been resolved in the nonconvex low-rank
matrix recovery literature~\cite{ma2017implicit,chen2019nonconvex}.
See Section~\ref{sec:Prior-art} for more discussions about this point.
Moreover, when $\bm{E}=\bm{0}$, it is known that $\rho_{s}$ can
	be as large as a constant even when the rank $r$ is allowed to grow with the dimension $n$~\cite{li2011compressed,chen2013low}. Our current theory, however, fails to cover the case with $\rho_{s}\asymp 1$ in the presence of noise. 
	We demonstrate through numerical experiments that the dependence of $\rho_{\mathsf{s}}$ on $r$ might indeed by suboptimal in our current theory. More specifically, Figure \ref{fig:rho-r} depicts the numerical Euclidean estimation errors w.r.t.~the corruption probability $\rho_{\mathsf{s}}$ as we vary the rank while fixing the sampling ratio. It can be seen that the estimation error curves corresponding to different ranks align very well with each other, thus suggesting the capability of convex relaxation in tolerating a constant fraction $\rho_{\mathsf{s}}$ of outliers.

\subsection{A peek at our technical approach\label{subsec:contributions}}

Before delving into the proof details, we immediately highlight our
key technical ideas and novelties. For simplicity we assume $n_1=n_2=n$ throughout this section.

\begin{figure}
	\centering
	
	\begin{tabular}{cc}
		\includegraphics[scale=0.35]{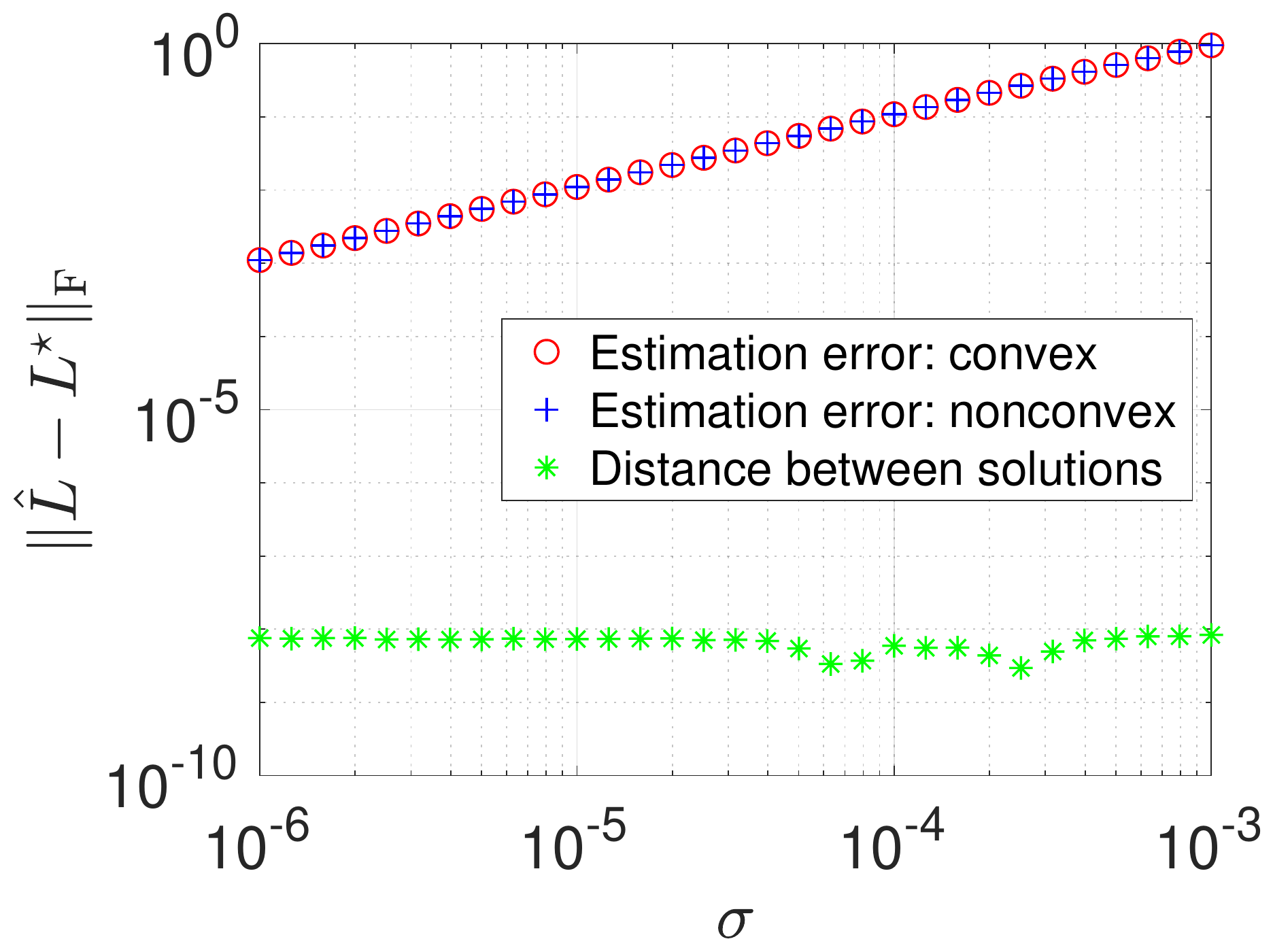} &\quad \includegraphics[scale=0.35]{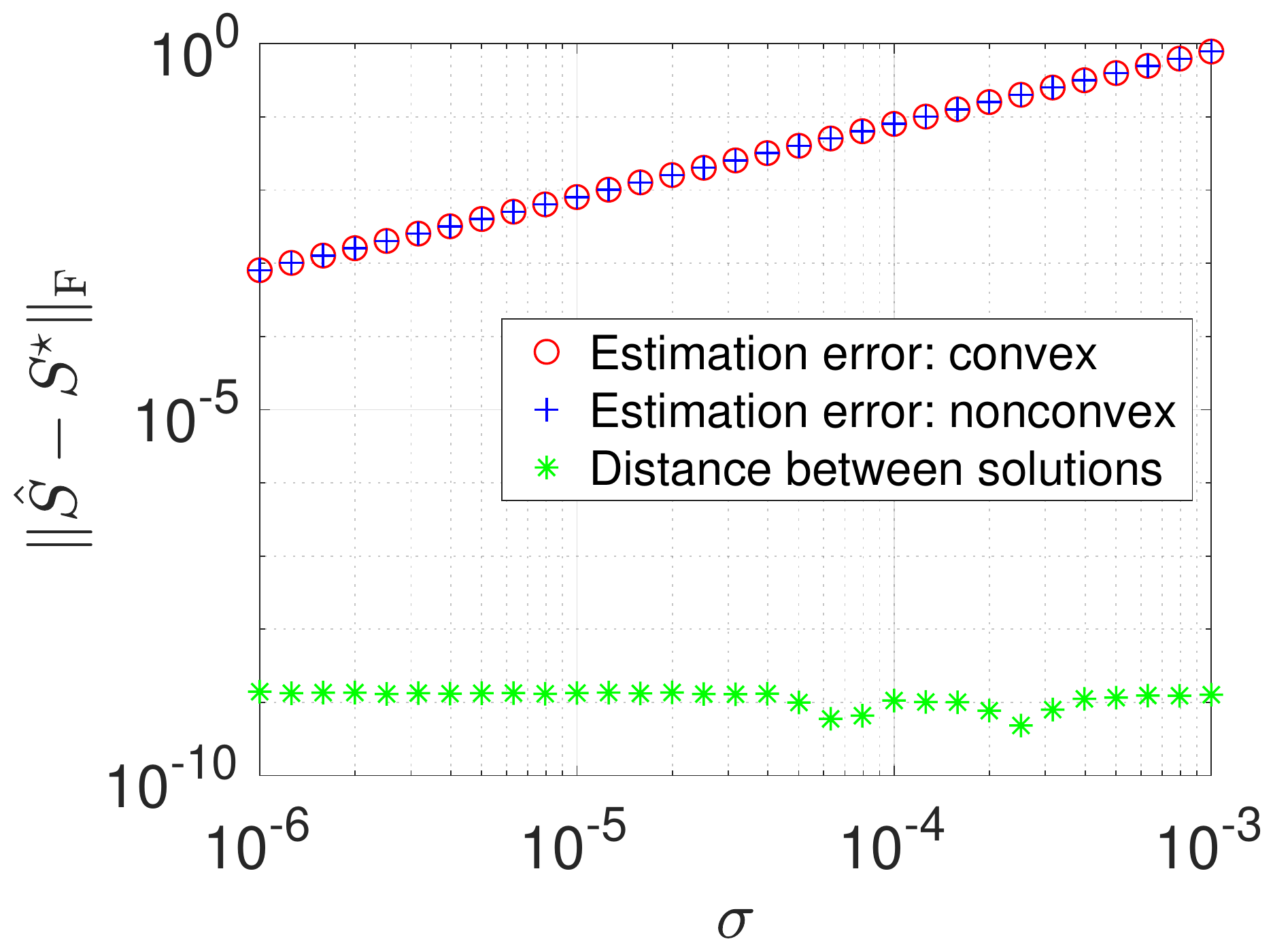}\tabularnewline
		$\quad\quad$(a) & $\quad\quad$(b)\tabularnewline
	\end{tabular}
	
	\caption{(a) The relative estimation errors of both $\bm{L}_{\mathsf{cvx}}$
		(the convex estimator (\ref{eq:cvx})) and $\bm{L}_{\mathsf{ncvx}}$
		(the estimate returned by the nonconvex approach tailored to (\ref{eq:peek-ncvx}))
		and the relative distance between them vs.~the standard deviation
		$\sigma$ of the noise. (b) The relative estimation errors of both
		$\bm{S}_{\mathsf{cvx}}$ (the convex estimator in (\ref{eq:cvx}))
		and $\bm{S}_{\mathsf{ncvx}}$ (the estimate returned by the nonconvex
		approach tailored to (\ref{eq:peek-ncvx})) and the relative distance
		between them vs.~the standard deviation $\sigma$ of the noise. The
		results are reported for $n=1000$, $r=5$, $p=0.2$, $\rho_{s}=0.1$,
		$\lambda=5\sigma\sqrt{np}$, $\tau=2\sigma\sqrt{\log n}$ and are averaged over 50 independent trials.
		\label{fig:exp2}}
\end{figure}

\paragraph{Connections between convex and nonconvex optimization.} Instead of directly analyzing the convex program~(\ref{eq:cvx}),
we turn attention to a seemingly different, but in fact closely related,
nonconvex program
\begin{equation}
\underset{\bm{X},\bm{Y}\in\mathbb{R}^{n\times r},\bm{S}\in\mathbb{R}^{n\times n}}{\text{minimize}}\quad\frac{1}{2}\left\Vert \mathcal{P}_{\Omega_{\mathsf{obs}}}\left(\bm{X}\bm{Y}^{\top}+\bm{S}-\bm{M}\right)\right\Vert _{\mathrm{F}}^{2}+\frac{\lambda}{2}\big(\left\Vert \bm{X}\right\Vert _{\mathrm{F}}^{2}+\left\Vert \bm{Y}\right\Vert _{\mathrm{F}}^{2}\big)+\tau\left\Vert \bm{S}\right\Vert _{1}.\label{eq:peek-ncvx}
\end{equation}
This idea is inspired by an interesting numerical finding (cf.~Figure~\ref{fig:exp2}) that the
solution to the convex program~(\ref{eq:cvx}), and an estimate obtained
by attempting to solve the nonconvex formulation (\ref{eq:peek-ncvx}),
are exceedingly close in our experiments. If such an intimate connection
can be formalized, then it suffices to analyze the statistical performance
of the nonconvex approach instead.\footnote{On the surface, the convex program (\ref{eq:cvx}) and the nonconvex
	one (\ref{eq:peek-ncvx}) are closely related: the convex solution
	$(\bm{L}_{\mathsf{cvx}},\bm{S}_{\mathsf{cvx}})$ coincides with that
	of the nonconvex program (\ref{eq:peek-ncvx}) if $\bm{L}_{\mathsf{cvx}}$
	is rank-$r$. This is an immediate consequence of the algebraic identity
	$\|\bm{Z}\|_{*}=\inf_{\bm{X},\bm{Y}\in\mathbb{R}^{n\times r}:\bm{X}\bm{Y}^{\top}=\bm{Z}}(\|\bm{X}\|_{\mathrm{F}}^{2}+\|\bm{Y}\|_{\mathrm{F}}^{2})$
	\cite{srebro2005rank,mazumder2010spectral}. However, it is difficult
	to know \emph{a priori }the rank of the convex solution. Hence such
	a connection does not prove useful in establishing the statistical
	properties of the convex estimator.} Fortunately, recent advances in nonconvex low-rank factorization
(see \cite{chi2018nonconvex} for an overview) provide powerful tools
for analyzing nonconvex low-rank estimation, allowing us to derive
the desired statistical guarantees that can then be transferred to
the convex approach. Of course, this is merely a high-level picture
of our proof strategy, and we defer the details to Section~\ref{sec:Architecture-of-the-proof}.

\begin{comment}
rationale is two-fold: (1) the convex solution $(\bm{L}_{\mathsf{cvx}},\bm{S}_{\mathsf{cvx}})$
is well approximated

by the nonconvex solution $(\bm{X}_{\mathsf{ncvx}}\bm{Y}_{\mathsf{ncvx}}^{\top},\bm{S}_{\mathsf{ncvx}})$
resulting from solving (\ref{eq:peek-ncvx}), and (2) the nonconvex
solution obeys the desired stability guarantees listed in (\ref{eq:Zcvx-error}).
It the hopes were true, the stability guarantees of the nonconvex
solution would be seamlessly transferred to that of the convex solution
$(\bm{L}_{\mathsf{cvx}},\bm{S}_{\mathsf{cvx}})$, hence finishing
the proof of Theorem~\ref{thm:main-convex}.
\end{comment}

It is worth emphasizing that our key idea --- that is, bridging convex
and nonconvex optimization  ---  is drastically different from previous
technical approaches for analyzing convex estimators (e.g.~(\ref{eq:cvx})).
As it turns out, these prior approaches, which include constructing
dual certificates and/or exploiting restricted strong convexity, have
their own deficiencies in analyzing (\ref{eq:cvx}) and fall short
of explaining the effectiveness of (\ref{eq:cvx}) in the random noise
setting. For instance, constructing dual certificates in the noisy
case is notoriously challenging given that we do not have closed-form
expressions for the primal solutions (so that it is difficult to invoke
the powerful dual construction strategies like the golfing scheme
\cite{Gross2011recovering} developed for the noiseless case). If
we directly utilize the dual certificates constructed for the noiseless
case, we would end up with an overly conservative bound like (\ref{eq:L-hat-no-oracle}),
which is exactly why the results in \cite{zhou2010stable,wong2017matrix} are sub-optimal.
On the other hand, while it is viable to show certain strong convexity
of (\ref{eq:cvx}) when restricted to some highly local sets and directions,
it is unclear how (\ref{eq:cvx}) forces its solution to stay within
the desired set and follow the desired directions, without adding
further (and often unnecessary) constraints to (\ref{eq:cvx}).

\paragraph{Nonconvex low-rank estimation with nonsmooth loss functions.} It is worth noting that a similar connection between convex and nonconvex optimization has
been pointed out by~\cite{chen2019noisy} towards understanding the power
of convex relaxation for noisy matrix completion. Due to the absence
of sparse outliers in the noisy matrix completion problem, the nonconvex
loss function considered therein is smooth in nature, which greatly simplifies
both the algorithmic and theoretical development. By contrast, the
nonsmoothness inherent in (\ref{eq:peek-ncvx}) makes it particularly
challenging to achieve the two desiderata mentioned above, namely,
connecting the convex and nonconvex solutions and establishing the
optimality of the nonconvex solution. In fact, to establish the connection between convex and nonconvex solutions, we put forward a novel two-step analysis strategy. Specifically, we first develop a crude upper bound on the Euclidean estimation error leveraging the idea of approximate dual certificates; see Theorem~\ref{thm:crude-bound}. While this crude upper bound is far from optimal, it serves as an important starting point towards formalizing the intimate relation between the convex solution $(\bm{L}_{\mathsf{cvx}},\bm{S}_{\mathsf{cvx}})$ and the nonconvex solution $(\bm{X},\bm{Y},\bm{S})$, since it is challenging to establish $\bm{X}\bm{Y}\approx\bm{L}_{\mathsf{cvx}}$ and $\bm{S}\approx\bm{S}_{\mathsf{cvx}}$ simultaneously without the aid of a crude bound. Second, in establishing the optimality of the nonconvex solution, the nonsmoothness nature of the nonconvex loss prevents us from applying the vanilla gradient descent scheme (as has been done in \cite{chen2019noisy}). To address this issue, we develop
an alternating minimization scheme --- which alternates between gradient
updates on $(\bm{X},\bm{Y})$ and minimization of $\bm{S}$ --- aimed
at minimizing the nonsmooth nonconvex loss function (\ref{eq:peek-ncvx});
see Algorithm~\ref{alg:gd-mc-ncvx} for details. As it turns out,
such a simple algorithm allows us to track the proximity of the convex
and nonconvex solutions and establish the optimality of the nonconvex
solution all at once.

	\begin{figure}
	\centering
	
	\includegraphics[scale=0.35]{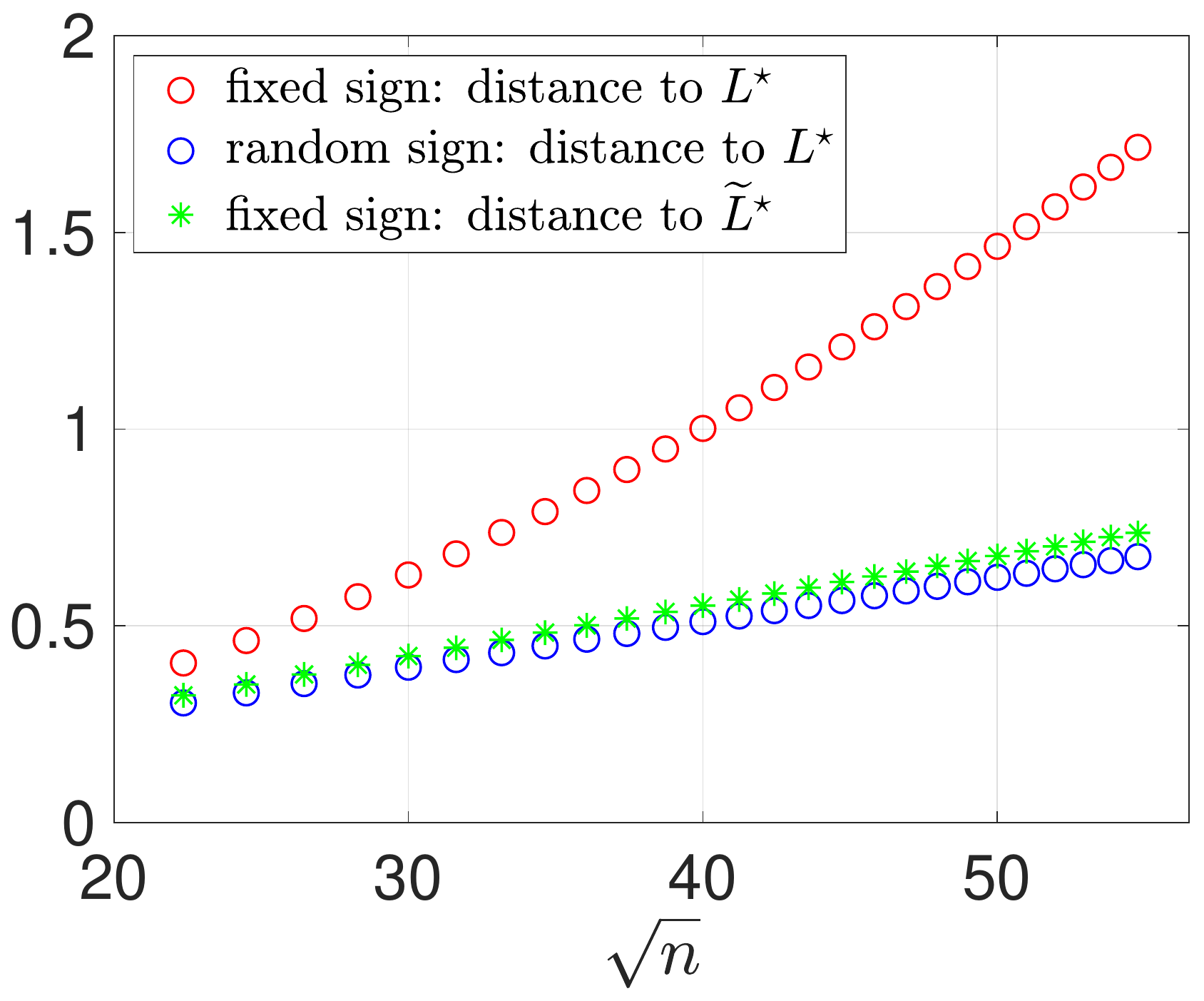}\caption{The red (resp.~blue) line displays the Euclidean estimation error of (\ref{eq:cvx}) vs.~$\sqrt{n}$ under fixed (resp.~random) sign patterns of $\bm{S}^{\star}$. The green line displays the Euclidean distance between $\bm{L}_{\mathsf{cvx}}$ and $\widetilde{\bm{L}}^\star$ under fixed sign patterns of $\bm{S}^{\star}$. The results
		are reported for $r=5$, $p=1$, $\sigma=10^{-3}$, and $\rho_{\mathsf{s}}=1/\log n$, with $\lambda=5\sigma\sqrt{np}$
		and $\tau=2\sigma\sqrt{\log n}$ and are averaged over 50 independent trials. For the random sign setting, the
		nonzero entries of $\bm{S}^{\star}$ are independently generated as $z\cdot5\sigma$, where $z$ follows a Rademacher distribution. For the fixed sign setting, each nonzero
		entry of $\bm{S}^{\star}$ equals to $5\sigma$.
		\label{fig:random-sign-1}}
\end{figure}

%\begin{figure}
%	\centering
%	
%	\includegraphics[scale=0.35]{figure/Exp2_new}\caption{The Euclidean estimation error of (\ref{eq:cvx}) vs.~$\sqrt{n}$
%		under two different sign patterns of $\bm{S}^{\star}$. The results
%		are reported for $r=5$, $p=1$, $\sigma=10^{-3}$, and $\rho_{\mathsf{s}}=0.1$, with $\lambda=5\sigma\sqrt{np}$
%		and $\tau=2\sigma\sqrt{\log n}$ and are averaged over 50 independent trials. For the random sign setting, the
%		nonzero entries of $\bm{S}^{\star}$ are independently generated from
%		$\mathcal{N}(0,10)$. For the fixed sign setting, each nonzero
%		entry of $\bm{S}^{\star}$ is independently generated following the
%		same distribution as $|z|$, where $z\sim\mathcal{N}(0,10)$.
%		\label{fig:random-sign-2}}
%\end{figure}

\subsection{Random signs of outliers\label{subsec:discussion-Random-sign}}

The careful reader might wonder whether it is possible to remove the
random sign assumption on $\bm{S}^{\star}$ (namely, Assumption~\ref{assumption:random-sign})
without compromising our statistical guarantees. After all, the results
of \cite{chandrasekaran2011rank,CanLiMaWri09,li2011compressed} derived
for the noise-free case do not rely on such a random sign assumption
at all.\footnote{Notably, in the noisy setting, prior theory \cite{zhou2010stable,wong2017matrix} also implicitly assumes this random sign condition, while \cite{agarwal2012noisy,klopp2017robust} do not require this condition.}  Unfortunately, removal of such a condition might be problematic
in general, as illustrated by the following example.

\paragraph{An example with non-random signs} Suppose that (i) $n_1=n_2=n$, (ii) each
non-zero entry of $\bm{S}^{\star}$ obeys $S_{ij}^{\star}=c_0\sigma$,
(iii) $\rho_{\mathsf{s}}=c_{1}/\log n$ for some sufficiently small
constant $c_{1}>0$, and (iv) there is no missing data (i.e.~$p=1$).
In such a scenario, the data matrix can be decomposed as
\[
\bm{M}=\bm{L}^{\star}+\bm{S}^{\star}+\bm{E}=\underset{\eqqcolon\widetilde{\bm{L}}^{\star}}{\underbrace{\bm{L}^{\star}+\mathbb{E}[\bm{S}^{\star}]}}+\underset{\eqqcolon\widetilde{\bm{E}}}{\underbrace{\bm{S}^{\star}-\mathbb{E}[\bm{S}^{\star}]+\bm{E}}}.
\]
Two observations are worth noting: (1) given that $\mathbb{E}[\bm{S}^{\star}]=c_0\rho_{\mathsf{s}}\sigma\bm{1}\bm{1}^{\top}$
with $\bm{1}$ the all-one vector, the rank of the matrix $\widetilde{\bm{L}}^{\star}=\bm{L}^{\star}+\mathbb{E}[\bm{S}^{\star}]$
is at most $r+1$; (2) $\widetilde{\bm{E}}$ is a zero-mean random
matrix consisting of independent entries with sub-Gaussian norm $O(\sigma)$.
In other words, the decomposition $\bm{M}=\widetilde{\bm{L}}^{\star}+\widetilde{\bm{E}}$
corresponds to a case with random noise but no outliers. Consequently,
we can invoke Theorem \ref{thm:main-convex-constant} to conclude that (assuming $r=O(1)$
and $\widetilde{\bm{L}}^{\star}$ is incoherent with condition number
$O(1)$): any minimizer $(\bm{L}_{\mathsf{cvx}},\bm{S}_{\mathsf{cvx}})$
of (\ref{eq:cvx}) obeys
\begin{align*}
\|\bm{L}_{\mathsf{cvx}}-\bm{L}^{\star}-\rho_{\mathsf{s}}\sigma\bm{1}\bm{1}^{\top}\|_{\mathrm{F}}&=\|\bm{L}_{\mathsf{cvx}}-\widetilde{\bm{L}}^{\star}\|_{\mathrm{F}}\lesssim\frac{\sigma}{\sigma_{\min}(\widetilde{\bm{L}}^\star)}\sqrt{n}\|\widetilde{\bm{L}}^\star\|_{\mathrm{F}}\\
&\lesssim\sigma\sqrt{nr}\frac{\sigma_{\max}(\widetilde{\bm{L}}^\star)}{\sigma_{\min}(\widetilde{\bm{L}}^\star)}\lesssim\sigma\sqrt{n}
\end{align*}
with high probability. Here the last step follows since $\widetilde{\bm{L}}^\star$ is of constant rank and condition number. This, however, leads to a lower bound on the
estimation error
\begin{align*}
\|\bm{L}_{\mathsf{cvx}}-\bm{L}^{\star}\|_{\mathrm{F}} & \geq\|c_0\rho_{\mathsf{s}}\sigma\bm{1}\bm{1}^{\top}\|_{\mathrm{F}}-\|\bm{L}_{\mathsf{cvx}}-\bm{L}^{\star}-\rho_{\mathsf{s}}\sigma\bm{1}\bm{1}^{\top}\|_{\mathrm{F}}=\sigma\big(c_0\rho_{\mathsf{s}}n-O(\sqrt{n})\big)\\
 & =(1-o(1))\frac{c_0c_{1}\sigma n}{\log n},
\end{align*}
which can be $O(\sqrt{n}/\log n)$ times larger than the desired estimation
error $O(\sigma\sqrt{n})$. Numerical experiments under the above setting (with $c_0=5$ and $c_1=1$) also suggest that (i) the estimation error under the fixed sign setting might be orderwise larger than that under the random sign setting; and (ii) under the fixed sign setting, the estimator \eqref{eq:cvx} approximately recovers $\widetilde{\bm{L}}^\star$ instead of $\bm{L}^\star$; see Figure~\ref{fig:random-sign-1}.

\bigskip\noindent The take-away message is this: when the entries
of $\bm{S}^{\star}$ are of non-random signs, it might sometimes be
possible to decompose $\bm{S}^{\star}$ into (1) a low-rank bias component
with a large Euclidean norm, and (2) a random fluctuation component
whose typical size does not exceed that of $\bm{E}$. If this is the
case, then the convex program (\ref{eq:cvx}) might mistakenly treat
the bias component as a part of the low-rank matrix $\bm{L}^{\star}$,
thus dramatically hampering its estimation accuracy. 
%\yly{Numerical experiments also indicate that non-random signs of $\bm{S}^{\star}$ might also harm the estimation quality of (\ref{eq:cvx}) beyond the setting of the above example; see Figure~\ref{fig:random-sign-2}.}

\section{Prior art\label{sec:Prior-art}}

Principal component analysis (PCA) \cite{pearson1901liii, jolliffe2011principal,fan2018principal}
is one of the most widely used statistical methods for dimension reduction
in data analysis. However, PCA is known to be quite sensitive to
adversarial outliers --- even a single corrupted data point can make
PCA completely off. This motivated the investigation of robust PCA,
which aims at making PCA robust to gross adversarial outliers. As
formulated in \cite{CanLiMaWri09,chandrasekaran2011rank}, this is closely related to the problem of disentangling a low-rank matrix $\bm{L}^{\star}$ and
a sparse outlier matrix $\bm{S}^{\star}$ (with unknown locations
and magnitudes) from a superposition of them. Consequently, robust
PCA can be viewed as an outlier-robust extension of the low-rank
matrix estimation/completion tasks \cite{ExactMC09,KesMonSew2010,chi2018nonconvex}.
In a similar vein, robust PCA has also been extensively studied in the context of  structured covariance
estimation under approximate factor models \cite{fan2008high,fan2013large,JMLR:v18:16-140,fan2019robust},
where the population covariance of certain random sample vectors is a mixture of a low-rank matrix and
a sparse matrix,  corresponding to the factor component and the idiosyncratic
component, respectively. 

Focusing on the convex relaxation approach, \cite{chandrasekaran2011rank,CanLiMaWri09}
started by considering the noiseless case with no missing data (i.e.~$\bm{E}=\bm{0}$
and $p= 1$) and demonstrated that, under
mild conditions, convex relaxation succeeds in exactly decomposing
both $\bm{L}^{\star}$ and $\bm{S}^{\star}$ from the data matrix $\bm{L}^{\star}+\bm{S}^{\star}$.
More specifically, \cite{chandrasekaran2011rank} adopted a deterministic
model without assuming any probabilistic structure on the outlier
matrix $\bm{S}^{\star}$. As shown in \cite{chandrasekaran2011rank} and several subsequent work \cite{chen2013low,hsu2011robust},
convex relaxation is guaranteed to work as long as the fraction of
outliers in each row/column does not exceed $O(1/r)$.
In contrast, \cite{CanLiMaWri09} proposed a random model by assuming that
$\bm{S}^{\star}$ has random support (cf.~Assumption~\ref{assumption:random-location});
under this model, exact recovery is guaranteed even if a constant
fraction of the entries of $\bm{S}^{\star}$ are nonzero with arbitrary
magnitudes. Following the random location model proposed in \cite{CanLiMaWri09},
the paper \cite{ganesh2010dense} showed that, in the absence of noise,
convex programming can provably tolerate a dominant fraction of outliers,
provided that the signs of the nonzero entries of $\bm{S}^{\star}$
are randomly generated (cf.~Assumption~\ref{assumption:random-sign}).
Later, the papers \cite{chen2013low,li2011compressed} extended these
results to the case when most entries of the matrix are unseen; even
in the presence of highly incomplete data, convex relaxation still
succeeds when a constant proportion of the observed entries are arbitrarily
corrupted. It is worth noting that the results of \cite{chen2013low}
accommodated both models proposed in \cite{chandrasekaran2011rank}
and \cite{CanLiMaWri09}, while the results of \cite{li2011compressed}
focused on the latter model.

The literature on robust PCA with not only sparse outliers but also
dense noise --- namely, when the measurements take the form
$\bm{M}=\mathcal{P}_{\Omega_{\mathsf{obs}}}(\bm{L}^{\star}+\bm{S}^{\star}+\bm{E})$
--- is relatively scarce. \cite{zhou2010stable, agarwal2012noisy} were among the first to present a general theory for robust PCA with dense noise, which was further extended in \cite{wong2017matrix, klopp2017robust}. {As we mentioned before, the first three \cite{zhou2010stable, agarwal2012noisy, wong2017matrix} accommodated arbitrary noise with the last one \cite{klopp2017robust} focusing on the random noise.} As we have discussed in Section~\ref{subsec:Main-results},
the statistical guarantees provided in these papers are highly suboptimal
when it comes to the random noise setting considered herein. The paper \cite{chen2014robust} extended the robust PCA results to the case where the truth is not only low-rank but also of Hankel structure. The results therein, however, suffered from the same sub-optimality issue.  

Moving beyond convex relaxation methods, another line of work proposed
nonconvex approaches for robust PCA \cite{netrapalli2014non,gu2016low,yi2016fast,cherapanamjeri2017nearly,zhang2018unified,charisopoulos2019low,li2018nonconvex,cai2019accelerated}, largely motivated by the recent
success of nonconvex methods in low-rank matrix factorization \cite{chi2018nonconvex,KesMonSew2010,candes2014wirtinger,sun2016guaranteed,ChenCandes15solving,chen2015fast,zhang2016provable,chen2016projected,jain2013low,netrapalli2013phase,ma2017implicit,chen2018gradient,wang2017solving,wei2016guarantees,cai2018solving,zheng2016convergence,charisopoulos2019composite}.
Following the deterministic model of \cite{chandrasekaran2011rank}, the paper
\cite{netrapalli2014non} proposed an alternating projection\,/\,minimization
scheme to seek a low-rank and sparse decomposition of the observed data matrix. In the noiseless setting,
i.e.~$\bm{E}=\bm{0}$, this alternating minimization scheme provably
disentangles the low-rank and sparse matrix from their superposition under mild conditions.
In addition, \cite{netrapalli2014non} extended their result to the arbitrary noise case where the size of the noise is extremely small, namely, $\|\bm{E}\|_{\infty}\ll\sigma_{\min}/n$.
When the noise $\{E_{ij}\}\sim\mathcal{N}(0,\sigma^{2})$, this is
equivalent to the condition $\sigma\ll\sigma_{\min}/(n\sqrt{\log n})$.
Comparing this with our noise condition $\sigma\ll\sigma_{\min}/(\sqrt{n\log n})$
(cf.~(\ref{eq:conditions})) when $r,\mu,\kappa \asymp 1$, one sees that our theoretical guarantees
cover a wider range of noise levels. Similarly, \cite{yi2016fast}
applied regularized gradient descent on a smooth nonconvex loss function which enjoys provable convergence
guarantees to $(\bm{L}^{\star},\bm{S}^{\star})$ under the noiseless
and partial observation setting. A recent paper \cite{charisopoulos2019low}
considered the nonsmooth nonconvex formulation for robust PCA and established
rigorously the convergence of subgradient-type methods in the rank-1 setting, i.e.~$r=1$. However, the extension to more general rank remains out of reach.

It is worth noting that noisy matrix completion problem \cite{CanPla10,chen2019noisy}
is subsumed as a special case by the model studied in this paper (namely, it is a special case with $\bm{S}^{\star}=\bm{0}$).
Statistical optimality under the random noise setting (cf.~Assumption~\ref{assumption:random-noise})
--- including the convex relaxation approach \cite{chen2019noisy,Negahban2012restricted,MR2906869,klopp2014noisy} and the
nonconvex approach \cite{ma2017implicit,chen2019nonconvex} --- has been extensively
studied. Focusing on arbitrary deterministic noise, \cite{CanPla10} established the stability
of the convex approach, whose resulting estimation error bound is similar
to the one established for robust PCA with noise in \cite{zhou2010stable}) (see~(\ref{eq:L-hat-no-oracle})). The paper \cite{krahmer2019convex} later confirmed
that the estimation error bound established in \cite{CanPla10} is
the best one can hope for in the arbitrary noise setting for matrix
completion, although it might be highly suboptimal if we restrict attention to random noise.

%
%Setting aside the algorithmic issue, a recent line of work \cite{ge2016matrix,ge2017no}
%studies the geometric properties of the nonconvex loss function associated
%with low-rank matrix recovery. In particular, for matrix completion
%(i.e.~when $\bm{S}^{\star}=0$) and robust PCA, 

Finally, there is also a large literature considering robust PCA under different settings and/or from different perspectives. For instance, the computational efficiency in solving the convex optimization
problem~(\ref{eq:cvx}) and its variants has been studied in the optimization literature (e.g.~\cite{tao2011recovering,goldfarb2013fast,shen2014augmented,ma2018efficient}). The problem has also been investigated under a streaming\,/\,online setting \cite{guo2014online, qiu2010real, feng2013online, zhan2016online, qiu2014recursive,vaswani2018static}. These are beyond the scope of the current paper. 

%\yly{where shall we put \cite{zhang2014novel,ha2015robust}?}

\section{Architecture of the proof \label{sec:Architecture-of-the-proof}}
In this section, we give an outline for proving Theorem~\ref{thm:main-convex}. The proof of Theorem~\ref{thm:main-convex-constant} follows immediately as it is a special case of Theorem~\ref{thm:main-convex}. For simplicity of presentation, our proof sets $n_1=n_2=n$. It is straightforward to obtain the proof for the general rectangular case via minor modification.

The main ingredient of the proof lies in establishing an intimate
link between convex and nonconvex optimization. Unless otherwise noted,
we shall set the regularization parameters as 
\begin{equation}
\lambda=C_{\lambda}\sigma\sqrt{np}\qquad\text{and}\qquad\tau=C_{\tau}\sigma\sqrt{\log n}
\end{equation}
throughout. In addition, the soft thresholding operator at level $\tau$
is defined such that
\begin{equation}
\mathcal{S}_{\tau}\left(x\right)\coloneqq\mathsf{sign}\left(x\right)\max\left(\left|x\right|-\tau,0\right)
%=\begin{cases}
%x-\tau, & \text{if}\ x>\tau,\\
%x+\tau, & \text{if}\ x<-\tau,\\
%0, & \text{otherwise.}
%\end{cases}
\label{eq:defn-soft-thresholding}
\end{equation}
For any matrix $\bm{X}$, the matrix $\mathcal{S}_{\tau}(\bm{X})$
is obtained by applying the soft thresholding operator $\mathcal{S}_{\tau}(\cdot)$
to each entry of $\bm{X}$ separately. Additionally, we define the
true low-rank factors as follows

\begin{equation}
\bm{X}^{\star}\coloneqq\bm{U}^{\star}\left(\bm{\Sigma}^{\star}\right)^{1/2}\qquad\text{and}\qquad\bm{Y}^{\star}\coloneqq\bm{V}^{\star}\left(\bm{\Sigma}^{\star}\right)^{1/2},\label{eq:Xstar-Ystar-defn}
\end{equation}
where $\bm{U}^{\star}\bm{\Sigma}^{\star}\bm{V}^{\star\top}$ is the
SVD of the true low-rank matrix $\bm{L}^{\star}$.

\subsection{Crude estimation error bounds for convex relaxation \label{subsec:Crude-bound-cvx}}

We start by delivering a crude upper bound on the Euclidean estimation
error, built upon the (approximate) duality certificate previously
constructed in \cite{chen2013low}. The proof is postponed
to Appendix~\ref{subsec:Proof-of-Lemma-crude-bound}.

\begin{theorem}\label{thm:crude-bound}Consider any given $\lambda>0$
and set $\tau\asymp\lambda\sqrt{({\log n})/{np}}$. Suppose that Assumptions
\ref{assumption:incoherence}-\ref{assumption:random-sign} hold,
and that
\[
n^{2}p\geq C\mu^{2}r^{2}n\log^{6}n\qquad\text{and}\qquad\rho_{\mathsf{s}}\leq c
\]
hold for some sufficiently large (resp.~small) constant $C>0$ (resp.~$c>0$).
Then with probability at least $1-O(n^{-10})$, any minimizer $(\bm{L}_{\mathsf{cvx}},\bm{S}_{\mathsf{cvx}})$
of the convex program \eqref{eq:cvx} satisfies
\begin{equation}
\left\Vert \bm{L}_{\mathsf{cvx}}-\bm{L}^{\star}\right\Vert _{\mathrm{F}}^{2}+\left\Vert \bm{S}_{\mathsf{cvx}}-\bm{S}^{\star}\right\Vert _{\mathrm{F}}^{2}\lesssim\lambda^{2}n^{5}\log^3 n+\frac{n}{\lambda^{2}}\big\|\mathcal{P}_{\Omega_{\mathsf{obs}}}\left(\bm{E}\right)\big\|_{\mathrm{F}}^{4}.
\end{equation}
\end{theorem}

It is worth noting that the above theorem holds true for an arbitrary
noise matrix $\bm{E}$. When specialized to the case with independent
sub-Gaussian noise, this crude bound admits a simpler expression as
follows.

\begin{corollary}\label{corollary:crude-bound}Take $\lambda=C_{\lambda}\sigma\sqrt{np}$ and $\tau=C_\tau\sigma\sqrt{\log n}$
for some universal constant $C_{\lambda},C_\tau>0$. Under the assumptions
of Theorem \ref{thm:crude-bound} and Assumption \ref{assumption:random-noise},
we have --- with probability exceeding $1-O(n^{-10})$ --- that
\begin{equation}
\left\Vert \bm{L}_{\mathsf{cvx}}-\bm{L}^{\star}\right\Vert _{\mathrm{F}}\lesssim\sigma n^{3}\log^{3/2} n\quad\text{and}\quad\left\Vert \bm{S}_{\mathsf{cvx}}-\bm{S}^{\star}\right\Vert _{\mathrm{F}}\lesssim\sigma n^{3}\log^{3/2} n.\label{eq:Lcvx-Scvx-crude-bound}
\end{equation}
\end{corollary}

\begin{proof}This corollary follows immediately by combining Theorem
\ref{thm:crude-bound} and Lemma~\ref{lemma:noise-bound} below.
\end{proof}

\begin{lemma}\label{lemma:noise-bound}Suppose that Assumption~\ref{assumption:random-noise}
holds and that $n^{2}p>C_{1}n\log^{2}n$ for some sufficiently large
constant $C_{1}>0$. Then with probability exceeding $1-O(n^{-10})$,
one has
\[
\big\|\mathcal{P}_{\Omega_{\mathsf{obs}}}\left(\bm{E}\right)\big\|\lesssim\sigma\sqrt{np}\qquad\text{and}\qquad\big\|\mathcal{P}_{\Omega_{\mathsf{obs}}}\left(\bm{E}\right)\big\|_{\mathrm{F}}\lesssim\sigma n\sqrt{p}.
\]
\end{lemma}

While the above results often lose a polynomial factor in $n$ vis-\`a-vis
the optimal error bound, it serves as an important starting point
that paves the way for subsequent analytical refinement.

\subsection{Approximate stationary points of the nonconvex formulation}

Instead of analyzing the convex estimator directly, we take a detour
by considering the following nonconvex optimization problem
\begin{equation}
\underset{\bm{X},\bm{Y}\in\mathbb{R}^{n\times r},\,\bm{S}\in\mathbb{R}^{n\times n}}{\text{minimize}}\quad F\left(\bm{X},\bm{Y},\bm{S}\right)\coloneqq\underbrace{\frac{1}{2p}\left\Vert \mathcal{P}_{\Omega_{\mathsf{obs}}}\left(\bm{X}\bm{Y}^{\top}+\bm{S}-\bm{M}\right)\right\Vert _{\mathrm{F}}^{2}+\frac{\lambda}{2p}\left\Vert \bm{X}\right\Vert _{\mathrm{F}}^{2}+\frac{\lambda}{2p}\left\Vert \bm{Y}\right\Vert _{\mathrm{F}}^{2}}_{\eqqcolon\,f\left(\bm{X},\bm{Y};\bm{S}\right)}+\frac{\tau}{p}\left\Vert \bm{S}\right\Vert _{1}.\label{eq:ncvx}
\end{equation}
Here, $f\left(\bm{X},\bm{Y};\bm{S}\right)$ is a function of $\bm{X}$
and $\bm{Y}$ with $\bm{S}$ frozen, which contains the smooth component
of the loss function $F(\bm{X},\bm{Y},\bm{S})$. As it turns out,
the solution to convex relaxation \eqref{eq:cvx} is exceedingly close
to an estimate  $(\bm{X},\bm{Y},\bm{S})$ obtained by a nonconvex algorithm aimed at solving
(\ref{eq:ncvx}) --- to be detailed in Section~\ref{sec:approximate_stationary_point}. This fundamental connection between the two algorithmic
paradigms provides a powerful framework that allows us to understand
convex relaxation by studying nonconvex optimization.

In what follows, we set out to develop the afforementioned intimate
connection. Before proceeding, we first state the following conditions
concerned with the interplay between the noise size, the estimation
accuracy of the nonconvex estimate $(\bm{X},\bm{Y},\bm{S})$, and the regularization parameters.

\begin{condition}\label{assumption:link-cvx-ncvx-noise} The regularization
parameters $\lambda$ and $\tau\asymp\lambda\sqrt{({\log n})/{np}}$
satisfy 
	%{\bf Q: notation of $\bX$ and $\bS$}
\begin{itemize}
\item \textbf{$\|\mathcal{P}_{\Omega_{\mathsf{obs}}}(\bm{E})\|<\lambda/16$   
}and $\Vert\mathcal{P}_{\Omega_{\mathsf{obs}}}(\bm{E})\Vert_{\infty}\leq\tau/4$;
\item \textbf{$\|\bm{S}-\bm{S}^{\star}\|<\lambda/16$} and $\Vert\bm{X}\bm{Y}^{\top}-\bm{L}^{\star}\Vert_{\infty}\leq\tau/4$;
\item $\|\mathcal{P}_{\Omega_{\mathsf{obs}}}(\bm{X}\bm{Y}^{\top}-\bm{L}^{\star})-p(\bm{X}\bm{Y}^{\top}-\bm{L}^{\star})\|<\lambda/8$.
\end{itemize}
\end{condition}

As an interpretation, the above condition says that: (1) the regularization
parameters are not too small compared to the size of the noise, so
as to ensure that we enforce a sufficiently large degree of regularization;
(2) the estimate represented by the point $(\bm{X}\bm{Y}^{\top},\bm{S})$
is sufficiently close to the truth. At this point, whether this condition
is meaningful or not remains far from clear; we shall return to justify
its feasibility shortly.

In addition, we need another condition concerning the injectivity
of $\mathcal{P}_{\Omega^{\star}}$ w.r.t.~a certain tangent space. For a rank-$r$ matrix $\bm{L}$ with singular value decomposition $\bm{U}\bm{\Sigma}\bm{V}^\top$ where $\bm{U},\bm{V}\in\mathbb{R}^{n\times r}$, the tangent space of the set of  rank-$r$ matrices at the point $\bm{L}$ is given by
\[
\left\{\bm{U}\bm{A}^\top+\bm{B}\bm{V}^\top\,|\,\bm{A},\bm{B}\in\mathbb{R}^{n\times r}\right\}.
\]
Again, the validity of this condition will be discussed momentarily.

\begin{condition}[\textbf{Injectivity}]\label{assumption:injectivity}
	Let $T$ be the tangent space of the set of rank-$r$ matrices at
	the point $\bm{X}\bm{Y}^{\top}$. Assume that there exist a constants
	$c_{\mathrm{inj}}>0$ such that for all $\bm{H}\in T$, one has
	\[
	p^{-1}\left\Vert \mathcal{P}_{\Omega_{\mathsf{obs}}}\left(\bm{H}\right)\right\Vert _{\mathrm{F}}^{2}\geq\frac{c_{\mathrm{inj}}}{\kappa}\left\Vert \bm{H}\right\Vert _{\mathrm{F}}^{2}\qquad\text{and}\qquad p^{-1}\left\Vert \mathcal{P}_{\Omega^{\star}}\left(\bm{H}\right)\right\Vert _{\mathrm{F}}^{2}\leq\frac{c_{\mathrm{inj}}}{4\kappa}\left\Vert \bm{H}\right\Vert _{\mathrm{F}}^{2}.
	\]
\end{condition}

With the above conditions in place, we are ready to make precise the
intimate link between convex relaxation and a candidate nonconvex
solution. The proof is deferred to Appendix~\ref{sec:proof-Equivalence-cvx-ncvx}.

\begin{theorem}\label{thm:cvx-ncvx-equivalence}Suppose that $n\geq\kappa$
and $\rho_{\mathsf{s}}\leq c/\kappa$ for some sufficiently small
constant $c>0$. Assume that there exists a triple $(\bm{X},\bm{Y},\bm{S})$
such that
\begin{equation}
\left\Vert \nabla f\left(\bm{X},\bm{Y};\bm{S}\right)\right\Vert _{\mathrm{F}}\leq\frac{1}{n^{20}}\frac{\lambda}{p}\sqrt{\sigma_{\min}},\quad\text{and}\quad\bm{S}=\mathcal{P}_{\Omega_{\mathsf{obs}}}\left(\mathcal{S}_{\tau}\left(\bm{M}-\bm{X}\bm{Y}^{\top}\right)\right).\label{eq:thm-cvx-ncvx-equivalence-cond-1}
\end{equation}
Further, assume that any singular value of $\bm{X}$ and $\bm{Y}$
lies in $[\sqrt{\sigma_{\min}/2},\sqrt{2\sigma_{\max}}]$. If the
solution $(\bm{L}_{\mathsf{cvx}},\bm{S}_{\mathsf{cvx}})$ to the convex
program (\ref{eq:cvx}) admits the following crude error bound
\begin{equation}
\left\Vert \bm{L}_{\mathsf{cvx}}-\bm{L}^{\star}\right\Vert _{\mathrm{F}}\lesssim\sigma n^{4},\label{eq:thm-cvx-ncvx-equivalence-cond-2}
\end{equation}
then under Conditions \ref{assumption:link-cvx-ncvx-noise}-\ref{assumption:injectivity}
we have
\[
\left\Vert \bm{X}\bm{Y}^{\top}-\bm{L}_{\mathsf{cvx}}\right\Vert _{\mathrm{F}}\lesssim\frac{\sigma}{n^{5}}\qquad\text{and}\qquad\left\Vert \bm{S}-\bm{S}_{\mathsf{cvx}}\right\Vert _{\mathrm{F}}\lesssim\frac{\sigma}{n^{5}}.
\]
\end{theorem}

This theorem is a deterministic result, focusing on some sort of ``approximate
stationary points'' of $F(\bm{X},\bm{Y},\bm{S})$. To interpret this,
observe that in view of \eqref{eq:thm-cvx-ncvx-equivalence-cond-1},
one has $\nabla f\left(\bm{X},\bm{Y};\bm{S}\right)\approx\bm{0}$,
and $\bm{S}$ minimizes $F(\bm{X},\bm{Y},\cdot)$ for any fixed $\bm{X}$
and $\bm{Y}$. If one can identify such an approximate stationary
point that is sufficiently close to the truth (so that it satisfies
Condition \ref{assumption:link-cvx-ncvx-noise}), then under mild
conditions our theory asserts that
\[
\bm{X}\bm{Y}^{\top}\approx\bm{L}_{\mathsf{cvx}}\qquad\text{and}\qquad\bm{S}\approx\bm{S}_{\mathsf{cvx}}.
\]
This would in turn formalize the intimate relation between the solution
to convex relaxation and an approximate stationary point of the nonconvex
formulation. The existence of such approximate stationary points will be verified shortly in Section \ref{sec:approximate_stationary_point}.

The careful reader might immediately remark that this theorem does
not say anything explicit about the minimizer of the nonconvex optimization
problem (\ref{eq:ncvx}); rather, it only pays attention to a special
class of approximate stationary points of the nonconvex formulation.
This arises mainly due to a technical consideration: it seems more
difficult to analyze the nonconvex optimizer directly than to study
certain approximate stationary points. Fortunately, our theorem indicates
that any approximate stationary point obeying the above conditions
serves as an extremely tight approximation of the convex estimate,
and, therefore, it suffices to identify and analyze any such points. 

\subsection{Constructing an approximate stationary point via nonconvex algorithms} \label{sec:approximate_stationary_point}

By virtue of Theorem \ref{thm:cvx-ncvx-equivalence}, the key to understanding
convex relaxation is to construct an approximate stationary point
of the nonconvex problem (\ref{eq:ncvx}) that enjoys desired statistical
properties. For this purpose, we resort to the following iterative
algorithm (Algorithm \ref{alg:gd-mc-ncvx}) to solve the nonconvex
program (\ref{eq:ncvx}).

\begin{algorithm}[h]
\caption{Alternating minimization method for solving the nonconvex problem
(\ref{eq:ncvx}).}

\label{alg:gd-mc-ncvx}\begin{algorithmic}

\STATE \textbf{{Suitable initialization}}: $\bm{X}^{0}=\bm{X}^{\star}$,
$\bm{Y}^{0}=\bm{Y}^{\star}$, $\bm{S}^{0}=\bm{S}^{\star}$.

\STATE \textbf{{Gradient updates}}: \textbf{for }$t=0,1,\ldots,t_{0}-1$
\textbf{do}

\STATE \vspace{-1em}
 \begin{subequations}\label{subeq:gradient_update_ncvx}
\begin{align}
\bm{X}^{t+1}= & \bm{X}^{t}-\eta\nabla_{\bm{X}}f\left(\bm{X}^{t},\bm{Y}^{t};\bm{S}^{t}\right)=\bm{X}^{t}-\frac{\eta}{p}\left[\mathcal{P}_{\Omega_{\mathsf{obs}}}\left(\bm{X}^{t}\bm{Y}^{t\top}+\bm{S}^{t}-\bm{M}\right)\bm{Y}^{t}+\lambda\bm{X}^{t}\right];\label{eq:gradient_update_ncvx_X}\\
\bm{Y}^{t+1}= & \bm{Y}^{t}-\eta\nabla_{\bm{Y}}f\left(\bm{X}^{t},\bm{Y}^{t};\bm{S}^{t}\right)=\bm{Y}^{t}-\frac{\eta}{p}\left\{ \left[\mathcal{P}_{\Omega_{\mathsf{obs}}}\left(\bm{X}^{t}\bm{Y}^{t\top}+\bm{S}^{t}-\bm{M}\right)\right]^{\top}\bm{X}^{t}+\lambda\bm{Y}^{t}\right\} ;\label{eq:gradient_update_ncvx_Y}\\
\bm{S}^{t+1}= & \mathcal{S}_{\tau}\left[\mathcal{P}_{\Omega_{\mathsf{obs}}}\left(\bm{M}-\bm{X}^{t+1}\bm{Y}^{t+1\top}\right)\right].\label{eq:shrink_update_S}
\end{align}
\end{subequations}

\end{algorithmic}
\end{algorithm}

In a nutshell, Algorithm \ref{alg:gd-mc-ncvx} alternates between
one iteration of gradient updates (w.r.t.~the decision matrices $\bm{X}$
and $\bm{Y}$) and optimization of the non-smooth problem w.r.t.~$\bm{S}$
(with $\bm{X}$ and $\bm{Y}$ frozen).\footnote{Note that for any given $\bm{X}$ and $\bm{Y}$, the solution to $\text{minimize}_{\bm{S}}\ F(\bm{X},\bm{Y},\bm{S})$
is given precisely by $\mathcal{S}_{\tau}(\mathcal{P}_{\Omega_{\mathsf{obs}}}(\bm{M}-\bm{X}\bm{Y}^{\top}))$.} For the sake of simplicity, we initialize this algorithm from the
ground truth $(\bm{X}^{\star},\bm{Y}^{\star},\bm{S}^{\star})$, but
our analysis framework might be extended to accommodate other more
practical initialization (e.g.~the one obtained by a spectral method \cite{chen2020spectral}).

The following theorem makes precise the statistical guarantees of
the above nonconvex optimization algorithm; the proof is deferred
to Appendix~\ref{sec:Analysis-nonconvex}. Here and throughout, we
define
\begin{equation}
\bm{H}^{t}\coloneqq\underset{\bm{R}\in\mathcal{O}^{r\times r}}{\arg\min}\ \ \big(\|\bm{X}^{t}\bm{R}-\bm{X}^{\star}\|_{\mathrm{F}}^{2}+\|\bm{Y}^{t}\bm{R}-\bm{Y}^{\star}\|_{\mathrm{F}}^{2}\big)^{1/2},\label{eq:defn-Ht}
\end{equation}
where $\mathcal{O}^{r\times r}$ denotes the set of $r\times r$ orthonormal
matrices.

\begin{theorem}\label{thm:ncvx-property}Instate the assumptions
of Theorem \ref{thm:main-convex} and define $$\delta_n \coloneqq \frac{\sigma}{\sigma_{\min}}\sqrt{\frac{n}{p}}.$$ Take $t_{0}=n^{47}$ and $\eta\asymp1/(n\kappa^{3}\sigma_{\max})$
in Algorithm~\ref{alg:gd-mc-ncvx}. With probability at least $1-O(n^{-3})$,
the iterates $\{(\bm{X}^{t},\bm{Y}^{t},\bm{S}^{t})\}_{0\leq t\leq t_{0}}$
of Algorithm \ref{alg:gd-mc-ncvx} satisfy
\begin{subequations}
	\begin{align}
	\max\left\{ \left\Vert \bm{X}^{t}\bm{H}^{t}-\bm{X}^{\star}\right\Vert _{\mathrm{F}},\left\Vert \bm{Y}^{t}\bm{H}^{t}-\bm{Y}^{\star}\right\Vert _{\mathrm{F}}\right\}  & \lesssim \delta_n \left\Vert \bm{X}^{\star}\right\Vert _{\mathrm{F}},\label{eq:thm-ncvx-fro}\\
	\max\left\{ \left\Vert \bm{X}^{t}\bm{H}^{t}-\bm{X}^{\star}\right\Vert ,\left\Vert \bm{Y}^{t}\bm{H}^{t}-\bm{Y}^{\star}\right\Vert \right\}  & \lesssim \delta_n \left\Vert \bm{X}^{\star}\right\Vert ,\label{eq:thm-ncvx-op}\\
	\max\left\{ \left\Vert \bm{X}^{t}\bm{H}^{t}-\bm{X}^{\star}\right\Vert _{2,\infty},\left\Vert \bm{Y}^{t}\bm{H}^{t}-\bm{Y}^{\star}\right\Vert _{2,\infty}\right\}  & \lesssim\kappa \sqrt{\log n}\, \delta_n \max\left\{ \left\Vert \bm{X}^{\star}\right\Vert _{2,\infty},\left\Vert \bm{Y}^{\star}\right\Vert _{2,\infty}\right\} ,\label{eq:thm-ncvx-two-infty}\\
	\left\Vert \bm{S}^{t}-\bm{S}^{\star}\right\Vert  & \lesssim\sigma\sqrt{np}.\label{eq:thm-ncvx-S}
	\end{align}
\end{subequations} In addition, with probability at least $1-O(n^{-3})$,
one has
\begin{equation}
\min_{0\leq t<t_{0}}\left\Vert \nabla f\left(\bm{X}^{t},\bm{Y}^{t};\bm{S}^{t}\right)\right\Vert _{\mathrm{F}}\leq\frac{1}{n^{20}}\frac{\lambda}{p}\sqrt{\sigma_{\min}}.\label{eq:small-gradient}
\end{equation}
\end{theorem}

In short, the bounds (\ref{eq:thm-ncvx-fro})-(\ref{eq:thm-ncvx-two-infty})
reveal that the entire sequence $\{\bm{X}^{t},\bm{Y}^{t}\}_{t=0}^{t_{0}}$
stays sufficiently close to the truth (measured by $\|\cdot\|_{\mathrm{F}}$,
$\|\cdot\|$, and more importantly, $\|\cdot\|_{2,\infty}$), the
inequality (\ref{eq:thm-ncvx-S}) demonstrates the goodness of fit
of $\{\bm{S}^{t}\}_{0\leq t\leq t_{0}}$ in terms of the spectral
norm accuracy, whereas the last bound (\ref{eq:small-gradient}) indicates
that there is at least one point in the sequence $\{\bm{X}^{t},\bm{Y}^{t},\bm{S}^{t}\}_{0\leq t\leq t_{0}}$
that can serve as an approximate stationary point of the nonconvex
formulation.

We shall also gather a few immediate consequences of Theorem \ref{thm:ncvx-property}
as follows, which contain basic properties that will be useful throughout.

\begin{corollary}\label{cor:ncvx-property}Instate the assumptions
of Theorem \ref{thm:ncvx-property}. Suppose that the sample size
obeys $n^{2}p\gg\kappa^{4}\mu^{2}r^{2}n\log^{4}n$, the noise satisfies
$\delta_n \ll1/\sqrt{\kappa^{4}\mu r\log n}$,
the outlier fraction satisfies $\rho_{\mathsf{s}}\ll1/(\kappa^{3}\mu r\log n)$.
With probability at least $1-O(n^{-3})$, the iterates of Algorithm
\ref{alg:gd-mc-ncvx} satisfy \begin{subequations}
\begin{align}
\left\Vert \bm{X}^{t}\bm{Y}^{t\top}-\bm{L}^{\star}\right\Vert _{\mathrm{F}} & \lesssim\kappa\delta_n \left\Vert \bm{L}^{\star}\right\Vert _{\mathrm{F}},\label{eq:cor-fro-error}\\
\left\Vert \bm{X}^{t}\bm{Y}^{t\top}-\bm{L}^{\star}\right\Vert _{\infty} & \lesssim\sqrt{\kappa^{3}\mu r\log n}\,\delta_n\left\Vert \bm{L}^{\star}\right\Vert _{\infty},\label{eq:cor-infty-error}\\
\left\Vert \bm{X}^{t}\bm{Y}^{t\top}-\bm{L}^{\star}\right\Vert  & \lesssim\delta_n\left\Vert \bm{L}^{\star}\right\Vert \label{eq:cor-op-error}
\end{align}
\end{subequations}simultaneously for all $t\leq t_{0}$.\end{corollary}\begin{proof}See
\cite[Appendix D.12]{chen2019noisy}.\end{proof}

\subsection{Proof of Theorem \ref{thm:main-convex}} \label{sec:proof_main_theorem}

Define
\begin{align}
t_{\ast} & \coloneqq\arg\min_{0\leq t<t_{0}}\Vert\nabla f(\bm{X}^{t},\bm{Y}^{t};\bm{S}^{t})\Vert_{\mathrm{F}};\label{eq:defn-tstar}\\
\big(\bm{X}_{\mathsf{ncvx}},\bm{Y}_{\mathsf{ncvx}},\bm{S}_{\mathsf{ncvx}}\big) & \coloneqq\big(\bm{X}^{t_{\ast}}\bm{H}^{t_{\ast}},\bm{Y}^{t_{\ast}}\bm{H}^{t_{\ast}},\bm{S}^{t_{\ast}}\big).\label{eq:defn-xystar}
\end{align}
Theorem~\ref{thm:ncvx-property} and Corollary~\ref{cor:ncvx-property}
have established appealing statistical performance of the nonconvex
solution $(\bm{X}_{\mathsf{ncvx}},\bm{Y}_{\mathsf{ncvx}},\bm{S}_{\mathsf{ncvx}})$.
To transfer this desired statistical property to that of $(\bm{L}_{\mathsf{cvx}},\bm{S}_{\mathsf{cvx}})$,
it remains to show that the nonconvex estimator $\big(\bm{X}_{\mathsf{ncvx}}\bm{Y}_{\mathsf{ncvx}}^{\top},\bm{S}_{\mathsf{ncvx}}\big)$
is extremely close to the convex estimator $(\bm{L}_{\mathsf{cvx}},\bm{S}_{\mathsf{cvx}})$.
Towards this end, we intend to invoke Theorem~\ref{thm:cvx-ncvx-equivalence};
therefore, it boils down to verifying the conditions therein.
\begin{enumerate}
\item The small gradient condition (cf.~(\ref{eq:thm-cvx-ncvx-equivalence-cond-1}))
holds automatically under~(\ref{eq:small-gradient}).
\item By virtue of the spectral norm bound~(\ref{eq:thm-ncvx-op}), one
has
\[
\left\Vert \bm{X}_{\mathsf{ncvx}}-\bm{X}^{\star}\right\Vert =\left\Vert \bm{X}^{t_{\ast}}\bm{H}^{t_{\ast}}-\bm{X}^{\star}\right\Vert \lesssim\frac{\sigma}{\sigma_{\min}}\sqrt{\frac{n}{p}}\left\Vert \bm{L}^{\star}\right\Vert \leq\frac{\sqrt{\sigma_{\min}}}{10},
\]
as long as $\sigma\sqrt{\kappa n/p}\ll\sigma_{\min}$. This together
with the Weyl inequality verifies the constraints on the singular
values of $(\bm{X}_{\mathsf{ncvx}},\bm{Y}_{\mathsf{ncvx}})$.
\item The crude error bounds are valid in view of Theorem~\ref{thm:crude-bound}.
\item Regarding Condition~\ref{assumption:link-cvx-ncvx-noise} and Condition~\ref{assumption:injectivity},
Lemma~\ref{lemma:noise-bound} and standard inequalities about sub-Gaussian
random variables imply that \textbf{$\|\mathcal{P}_{\Omega_{\mathsf{obs}}}(\bm{E})\|<\lambda/16$
}and $\Vert\mathcal{P}_{\Omega_{\mathsf{obs}}}(\bm{E})\Vert_{\infty}\leq\tau/4$.
In addition, the bounds~(\ref{eq:thm-ncvx-S}) and~(\ref{eq:cor-infty-error})
ensure the second assumption $\|\bm{S}_{\mathsf{ncvx}}-\bm{S}^{\star}\|\leq\lambda/16$
and $\Vert\bm{X}\bm{Y}^{\top}-\bm{L}^{\star}\Vert_{\infty}\leq\tau/4$
in Condition~\ref{assumption:link-cvx-ncvx-noise}. We are left with
the last assumption in Condition~\ref{assumption:link-cvx-ncvx-noise}
and Condition~\ref{assumption:injectivity}, which are guaranteed
to hold in view of the following lemma (see Appendix~\ref{sec:Proof-of-Lemma-inj-main}
for the proof).

\begin{lemma}\label{lemma:injectivity-main}Instate the notations
and assumptions of Theorem \ref{thm:main-convex}. Then with probability
exceeding $1-O(n^{-10})$, we have \begin{subequations}
	\begin{align}
	\big\Vert \mathcal{P}_{\Omega}\big(\bm{X}\bm{Y}^{\top}-\bm{M}^{\star}\big)&-p\big(\bm{X}\bm{Y}^{\top}-\bm{M}^{\star}\big)\big\Vert   <\lambda/8,\label{eq:P-debias-condition}\\
	\frac{1}{p}\left\Vert \mathcal{P}_{\Omega_{\mathsf{obs}}}\left(\bm{H}\right)\right\Vert _{\mathrm{F}}^{2} & \geq\frac{1}{32\kappa}\left\Vert \bm{H}\right\Vert _{\mathrm{F}}^{2},\qquad\forall\bm{H}\in T,\label{eq:inj-lower}\\
	p^{-1}\left\Vert \mathcal{P}_{\Omega^{\star}}\left(\bm{H}\right)\right\Vert _{\mathrm{F}}^{2} & \leq\frac{1}{128\kappa}\left\Vert \bm{H}\right\Vert _{\mathrm{F}}^{2},\qquad\forall\bm{H}\in T\label{eq:inj-upper}
	\end{align}
\end{subequations}simultaneously for all $(\bm{X},\bm{Y})$ obeying\begin{subequations}\label{subeq:condition-inf}
\begin{align}
\left\Vert \bm{X}-\bm{X}^{\star}\right\Vert _{2,\infty} & \leq C_{\infty}\kappa\frac{\sigma}{\sigma_{\min}}\sqrt{\frac{n\log n}{p}}\max\left\{ \left\Vert \bm{X}^{\star}\right\Vert _{2,\infty},\left\Vert \bm{Y}^{\star}\right\Vert _{2,\infty}\right\} ;\label{condition-inf-X}\\
\left\Vert \bm{Y}-\bm{Y}^{\star}\right\Vert _{2,\infty} & \leq C_{\infty}\kappa\frac{\sigma}{\sigma_{\min}}\sqrt{\frac{n\log n}{p}}\max\left\{ \left\Vert \bm{X}^{\star}\right\Vert _{2,\infty},\left\Vert \bm{Y}^{\star}\right\Vert _{2,\infty}\right\} .\label{condition-inf-2}
\end{align}
\end{subequations}Here, $T$ denotes the tangent space of the set
of rank-$r$ matrices at the point $\bm{X}\bm{Y}^{\top}$, and $C_{\infty}>0$
is an absolute constant. \end{lemma}
\end{enumerate}
Armed with the above conditions, we can readily invoke Theorem \ref{thm:cvx-ncvx-equivalence}
to reach
\[
\left\Vert \bm{X}_{\mathsf{ncvx}}\bm{Y}_{\mathsf{ncvx}}^{\top}-\bm{L}_{\mathsf{cvx}}\right\Vert _{\mathrm{F}}\lesssim\frac{\sigma}{n^{5}}\qquad\text{and}\qquad\left\Vert \bm{S}_{\mathsf{ncvx}}-\bm{S}_{\mathsf{cvx}}\right\Vert _{\mathrm{F}}\lesssim\frac{\sigma}{n^{5}}
\]
with high probability. This taken collectively with Corollary \ref{cor:ncvx-property}
gives
\begin{align*}
\left\Vert \bm{L}_{\mathsf{cvx}}-\bm{L}^{\star}\right\Vert _{\mathrm{F}} & \leq\left\Vert \bm{X}_{\mathsf{ncvx}}\bm{Y}_{\mathsf{ncvx}}^{\top}-\bm{L}_{\mathsf{cvx}}\right\Vert _{\mathrm{F}}+\left\Vert \bm{X}_{\mathsf{ncvx}}\bm{Y}_{\mathsf{ncvx}}^{\top}-\bm{L}^{\star}\right\Vert _{\mathrm{F}}\\
 & \lesssim\frac{\sigma}{n^{5}}+\kappa\frac{\sigma}{\sigma_{\min}}\sqrt{\frac{n}{p}}\left\Vert \bm{L}^{\star}\right\Vert _{\mathrm{F}}\\
 & \asymp\kappa\frac{\sigma}{\sigma_{\min}}\sqrt{\frac{n}{p}}\left\Vert \bm{L}^{\star}\right\Vert _{\mathrm{F}}.
\end{align*}
Similar arguments lead to the advertised high-probability bounds 
\begin{align*}
\left\Vert \bm{L}_{\mathsf{cvx}}-\bm{L}^{\star}\right\Vert _{\infty} & \lesssim\sqrt{\kappa^{3}\mu r}\frac{\sigma}{\sigma_{\min}}\sqrt{\frac{n\log n}{p}}\left\Vert \bm{L}^{\star}\right\Vert _{\infty},\\
\left\Vert \bm{L}_{\mathsf{cvx}}-\bm{L}^{\star}\right\Vert  & \lesssim\frac{\sigma}{\sigma_{\min}}\sqrt{\frac{n}{p}}\left\Vert \bm{L}^{\star}\right\Vert .
\end{align*}

Finally, given that $\bm{X}_{\mathsf{ncvx}}\bm{Y}_{\mathsf{ncvx}}^{\top}$
is a rank-$r$ matrix, the rank-$r$ approximation $\bm{L}_{\mathsf{cvx},r}\coloneqq\arg\min_{\bm{Z}:\text{rank}(\bm{Z})\leq r}\|\bm{Z}-\bm{L}_{\mathsf{cvx}}\|_{\mathrm{F}}$
of $\bm{L}_{\mathsf{cvx}}$ necessarily satisfies
\[
\left\Vert \bm{L}_{\mathsf{cvx},r}-\bm{L}_{\mathsf{cvx}}\right\Vert _{\mathrm{F}}\leq\left\Vert \bm{X}_{\mathsf{ncvx}}\bm{Y}_{\mathsf{ncvx}}^{\top}-\bm{L}_{\mathsf{cvx}}\right\Vert _{\mathrm{F}}\lesssim\frac{\sigma}{n^{5}}\leq\frac{1}{n^{5}}
\cdot\frac{\sigma}{\sigma_{\min}}\sqrt{\frac{n}{p}} \left\Vert \bm{L}^{\star}\right\Vert ,
\]
which establishes (\ref{eq:Zcvx-r-bound-general}). In view of the triangle
inequality, the properties (\ref{eq:Zcvx-error-general}) hold unchanged if
$\bm{L}_{\mathsf{cvx}}$ is replaced by $\bm{L}_{\mathsf{cvx},r}$.

\section{Discussion\label{sec:Discussion}}

This paper investigates the unreasonable effectiveness of convex programming
in estimating an unknown low-rank matrix from grossly corrupted data.
We develop an improved theory that confirms the optimality of convex
relaxation in the presence of random noise, gross sparse outliers,
and missing data. In particular, our results significantly improve
upon the prior statistical guarantees \cite{zhou2010stable} under
random noise, while further allowing for missing data. Our theoretical
analysis is built upon an appealing connection between convex and
nonconvex optimization, which has not been established previously.

Having said this, our current work leaves open several important issues
that call for further investigation. To begin with, the conditions
(\ref{eq:conditions-general}) stated in the main theorem are likely suboptimal
in terms of the dependency on both the rank $r$ and the condition
number $\kappa$. For example, we shall keep in mind that in the noise-free
setting, the sample size can be as low as $O(nr\mathrm{poly}\log n)$
and the tolerable outlier fraction can be as large as a constant \cite{li2011compressed,chen2013low},
both of which exhibit more favorable scalings w.r.t.~$r$ and $\kappa$
compared to our current condition (\ref{eq:conditions-general}). Moving forward,
our analysis ideas suggest a possible route for analyzing convex relaxation
for other structured estimation problems under both random noise and
outliers, including but not limited to sparse PCA (the case with a
simultaneously low-rank and sparse matrix) \cite{cai2013sparse},
low-rank Hankel matrix estimation (the case involving a low-rank Hankel
matrix) \cite{chen2014robust}, and blind deconvolution\footnote{Our ongoing work \cite{chen2020convex} is pursuing this direction.} (the case
that aims to recover a low-rank matrix from structured Fourier measurements)
\cite{ahmed2014blind}. Last but not least, we would like to point out that it is possible to design a similar debiasing procedure as in \cite{Chen22931} for correcting the bias in the  convex estimator, which further allows uncertainty quantification and statistical inference on the unknown low-rank matrix of interest.

\section*{Acknowledgements}

Y.~Chen is supported in part by the AFOSR YIP award FA9550-19-1-0030,
by the ONR grant N00014-19-1-2120, by the ARO grants W911NF-20-1-0097 and W911NF-18-1-0303,
by the NSF grants CCF-1907661, IIS-1900140 and DMS-2014279, and by the Princeton SEAS innovation award. J.~Fan is supported
in part by the NSF grants DMS-1662139 and DMS-1712591, the ONR grant
N00014-19-1-2120, and the NIH grant 2R01-GM072611-14.

\appendix

\section{An equivalent probabilistic model of $\Omega^{\star}$ used throughout
the proof \label{sec:An-equivalent-probabilistic}}

Recall that $\Omega^\star$ is the support of the sparse component $\bS^*$.
In this section, we introduce an equivalent probabilistic model of
$\Omega^{\star}$, which is more amenable to analysis and shall
be assumed throughout the proof.
\begin{itemize}
\item \textbf{The original model.} Recall from Assumption~\ref{assumption:random-location}
the way we generate $\Omega^{\star}$ : (1) sample $\Omega_{\mathsf{obs}}$
from the i.i.d.~Bernoulli model with parameter $p$; (2) for each
$(i,j)\in\Omega_{\mathsf{obs}}$, let $(i,j)\in\Omega^{\star}$ independently
with probability $\rho_{\mathsf{s}}$.
\item \textbf{An equivalent model. } The model involves over-sampling and rejection method: (1) sample $\Omega_{\mathsf{obs}}$
from the i.i.d.~Bernoulli model with parameter $p$; (2) generate
an augmented index set $\Omega_{\mathsf{aug}}\subseteq\Omega_{\mathsf{obs}}$
such that: for each $(i,j)\in\Omega_{\mathsf{obs}}$, we generate $(i,j)\in\Omega_{\mathsf{aug}}$
independently with probability $\rho_{\mathsf{aug}}$; (3) for any
$(i,j)\in\Omega_{\mathsf{aug}}$, include $(i,j)$ in $\Omega^{\star}$
independently with probability $\rho_{\mathsf{s}}/\rho_{\mathsf{aug}}$.
\end{itemize}
It is straightforward to verify that the two models for $\Omega^{\star}$
are equivalent as long as $\rho_{\mathsf{s}}\leq\rho_{\mathsf{aug}}\leq1$.
Two important remarks are in order. First, by construction, we have
$\Omega^{\star}\subseteq\Omega_{\mathsf{aug}}$. Second, the choice
of $\rho_{\mathsf{aug}}$ can vary as needed as long as $\rho_{\mathsf{s}}\leq\rho_{\mathsf{aug}}\leq1$. 

The introduction of this augmented index set $\Omega_{\mathsf{aug}}$ comes in handy when we would like to control the size $\|\mathcal{P}_{\Omega^{\star}}(\bm{A})\|_{\mathrm{F}}$ for some matrix $\bm{A} \in \mathbb{R}^{n \times n}$. The first inclusion property $\Omega^{\star}\subseteq\Omega_{\mathsf{aug}}$ allows us to upper bound $\|\mathcal{P}_{\Omega^{\star}}(\bm{A})\|_{\mathrm{F}}$ by $\|\mathcal{P}_{\Omega_{\mathsf{aug}}}(\bm{A})\|_{\mathrm{F}}$, and the freedom to choose $\rho_{\mathsf{aug}}$ allows us to leverage stronger concentration results, which might not hold for the smaller $\rho_{\mathsf{s}}$. See Corollary~\ref{lemma:P_omega_star_P_T} in the next section for an example.

\section{Preliminaries\label{sec:Preliminary-facts}}

\subsection{A few preliminary facts}

This subsection collects several results that are useful throughout
the proof. To begin with, the incoherence assumption (cf.~Assumption~\ref{assumption:incoherence}) asserts that
\begin{equation}\label{eq:incoherence-X}
\left\Vert\bm{X}^\star\right\Vert_{2,\infty}\leq\sqrt{\mu r/n}\left\Vert\bm{X}^\star\right\Vert\qquad\text{and}\qquad\left\Vert\bm{Y}^\star\right\Vert_{2,\infty}\leq\sqrt{\mu r/n}\left\Vert\bm{Y}^\star\right\Vert.
\end{equation}
This is because
\begin{equation*}
\left\Vert\bm{X}^\star\right\Vert_{2,\infty}=\big\Vert\bm{U}^\star(\bm{\Sigma}^\star)^{1/2}\big\Vert_{2,\infty}\leq\left\Vert\bm{U}^\star\right\Vert_{2,\infty}\big\Vert(\bm{\Sigma}^\star)^{1/2}\big\Vert\leq\sqrt{\mu r/n}\left\Vert\bm{X}^\star\right\Vert,
\end{equation*}where the first inequality comes from the elementary inequality $\Vert \bm{A}\bm{B}\Vert_{2,\infty}\leq\Vert\bm{A}\Vert_{2,\infty}\Vert\bm{B}\Vert$, and the last inequality is a consequence of the incoherence assumption as well as the fact that $\Vert(\bm{\Sigma}^\star)^{1/2}\Vert=\Vert\bm{X}^\star\Vert$. 

The next lemma is extensively used in the low-rank matrix
completion literature.

\begin{lemma}\label{lemma:inj-general}Suppose that each $(i,j)$
is included in $\Omega_{0}\subseteq[n]\times[n]$ independently with
probability $\rho_{0}$. Then with probability exceeding $1-O(n^{-10})$,
one has 
\begin{equation}
\left\Vert \mathcal{P}_{T^{\star}}-\rho_{0}^{-1}\mathcal{P}_{T^{\star}}\mathcal{P}_{\Omega_{0}}\mathcal{P}_{T^{\star}}\right\Vert \leq\frac{1}{2},\label{eq:tangent_space_property}
\end{equation}
provided that $n^{2}\rho_{0}\gg\mu rn\log n$. Here, $T^{\star}$
denotes the tangent space of the set of rank-$r$ matrices at the
point $\bm{L}^{\star}=\bm{X}^{\star}\bm{Y}^{\star\top}$. \end{lemma}\begin{proof}See
\cite[Theorem 4.1]{ExactMC09}\end{proof}

In fact, the bound~(\ref{eq:tangent_space_property}) uncovers certain
near-isometry of the operator $\rho_{0}^{-1}\mathcal{P}_{\Omega_{0}}(\cdot)$
when restricted to the tangent space $T^{\star}$. This property is
formalized in the following fact.

\begin{fact}\label{fact:near-isometry}Suppose that $\|\mathcal{P}_{T^{\star}}-\rho_{0}^{-1}\mathcal{P}_{T^{\star}}\mathcal{P}_{\Omega_{0}}\mathcal{P}_{T^{\star}}\|\leq1/2$.
Then one has 
\[
\frac{1}{2}\left\Vert \bm{H}\right\Vert _{\mathrm{F}}^{2}\leq\frac{1}{\rho_{0}}\left\Vert \mathcal{P}_{\Omega_{0}}\left(\bm{H}\right)\right\Vert _{\mathrm{F}}^{2}\leq\frac{3}{2}\left\Vert \bm{H}\right\Vert _{\mathrm{F}}^{2},\qquad\text{for all }\bm{H}\in T^{\star}.
\]
\end{fact}\begin{proof}The proof has actually been documented in
the literature. For completeness, we present the proof for the lower
bound here; the upper bound follows from a very similar argument.
For any $\bm{H}\in\mathbb{R}^{n\times n}$, one has 
\begin{align*}
\left\Vert \mathcal{P}_{\Omega_{0}}\mathcal{P}_{T^{\star}}\left(\bm{H}\right)\right\Vert _{\mathrm{F}}^{2} & =\left\langle \mathcal{P}_{\Omega_{0}}\mathcal{P}_{T^{\star}}\left(\bm{H}\right),\mathcal{P}_{\Omega_{0}}\mathcal{P}_{T^{\star}}\left(\bm{H}\right)\right\rangle =\left\langle \mathcal{P}_{T^{\star}}\left(\bm{H}\right),\mathcal{P}_{T^{\star}}\mathcal{P}_{\Omega_{0}}\mathcal{P}_{T^{\star}}\left(\bm{H}\right)\right\rangle \\
 & =\rho_{0}\left\Vert \mathcal{P}_{T^{\star}}\left(\bm{H}\right)\right\Vert _{\mathrm{F}}^{2}-\rho_{0}\left\langle \mathcal{P}_{T^{\star}}\left(\bm{H}\right),\left(\mathcal{P}_{T^{\star}}-\rho_{0}^{-1}\mathcal{P}_{T^{\star}}\mathcal{P}_{\Omega_{0}}\mathcal{P}_{T^{\star}}\right)\left(\mathcal{P}_{T^{\star}}\left(\bm{H}\right)\right)\right\rangle \\
 & \geq\rho_{0}\left\Vert \mathcal{P}_{T^{\star}}\left(\bm{H}\right)\right\Vert _{\mathrm{F}}^{2}-\rho_{0}\left\Vert \mathcal{P}_{T^{\star}}-\rho_{0}^{-1}\mathcal{P}_{T^{\star}}\mathcal{P}_{\Omega_{0}}\mathcal{P}_{T^{\star}}\right\Vert \left\Vert \mathcal{P}_{T^{\star}}\left(\bm{H}\right)\right\Vert _{\mathrm{F}}^{2}\\
 & \geq\frac{\rho_{0}}{2}\left\Vert \mathcal{P}_{T^{\star}}\left(\bm{H}\right)\right\Vert _{\mathrm{F}}^{2}.
\end{align*}
Here, the penultimate inequality relies on the elementary fact that
$\langle\bm{A},\bm{B}\rangle\leq\|\bm{A}\|_{\mathrm{F}}\|\bm{B}\|_{\mathrm{F}}$,
and the last step follows from the assumption $\|\mathcal{P}_{T^{\star}}-\rho_{0}^{-1}\mathcal{P}_{T^{\star}}\mathcal{P}_{\Omega_{0}}\mathcal{P}_{T^{\star}}\|\leq1/2$.
\end{proof}

The following corollary is an immediate consequence of Lemma~\ref{lemma:inj-general}
and Fact~\ref{fact:near-isometry}.

\begin{corollary}\label{lemma:P_omega_star_P_T}Suppose that $\rho_{\mathsf{s}}\leq\rho_{\mathsf{aug}}\leq1/12$
and that $n^{2}p\rho_{\mathsf{aug}}\gg\mu rn\log n$. Then with probability
at least $1-O(n^{-10})$, we have 
\[
\|\mathcal{P}_{\Omega^{\star}}\mathcal{P}_{T^{\star}}\|^{2}\leq p/8.
\]
\end{corollary}\begin{proof}Recall the auxiliary index set $\Omega_{\mathsf{aug}}$
introduced in Appendix~\ref{sec:An-equivalent-probabilistic}. Since
$\Omega^{\star}\subseteq\Omega_{\mathsf{aug}}$, we have for any $\bm{H}\in\mathbb{R}^{n\times n}$
\[
\left\Vert \mathcal{P}_{\Omega^{\star}}\mathcal{P}_{T^{\star}}\left(\bm{H}\right)\right\Vert _{\mathrm{F}}^{2}\leq\left\Vert \mathcal{P}_{\Omega_{\mathsf{aug}}}\mathcal{P}_{T^{\star}}\left(\bm{H}\right)\right\Vert _{\mathrm{F}}^{2}\leq\frac{3p\rho_{\mathsf{aug}}}{2}\left\Vert \mathcal{P}_{T^{\star}}\left(\bm{H}\right)\right\Vert _{\mathrm{F}}^{2}\leq\frac{3p\rho_{\mathsf{aug}}}{2}\left\Vert \bm{H}\right\Vert _{\mathrm{F}}^{2}.
\]
Here, the second inequality arises from Lemma~\ref{lemma:inj-general}
and Fact~\ref{fact:near-isometry} (by taking $\Omega_{0}=\Omega_{\mathsf{aug}}$
and $\rho_{0}=p\rho_{\mathsf{aug}}$). The proof is complete by recognizing
the assumption $\rho_{\mathsf{aug}}\leq1/12$. \end{proof}

As it turns out, the near-isometry property of $\rho_{0}^{-1}\mathcal{P}_{\Omega_{0}}(\cdot)$
can be strengthened to a uniform version (uniform over a large collection
of tangent spaces), as shown in the lemma below.

\begin{lemma}\label{lemma:sampling-operator-norm-in-T}Suppose that
each $(i,j)$ is included in $\Omega_{0}\subseteq[n]\times[n]$ independently
with probability $\rho_{0}$, and that $n^{2}\rho_{0}\gg\mu rn\log n$.
Then with probability at least $1-O(n^{-10})$, 
\[
\frac{1}{32\kappa}\left\Vert \bm{H}\right\Vert _{\mathrm{F}}^{2}\leq\frac{1}{\rho_{0}}\left\Vert \mathcal{P}_{\Omega_{0}}\left(\bm{H}\right)\right\Vert _{\mathrm{F}}^{2}\leq40\kappa\left\Vert \bm{H}\right\Vert _{\mathrm{F}}^{2},\qquad\text{for all }\bm{H}\in T
\]
holds simultaneously for all $(\bm{X},\bm{Y})$ obeying 
\[
\max\left\{ \left\Vert \bm{X}-\bm{X}^{\star}\right\Vert _{2,\infty},\left\Vert \bm{Y}-\bm{Y}^{\star}\right\Vert _{2,\infty}\right\} \leq\frac{c}{\kappa\sqrt{n}}\left\Vert \bm{X}^{\star}\right\Vert .
\]
Here, $c>0$ is some sufficiently small constant, and $T$ denotes
the tangent space of the set of rank-$r$ matrices at the point $\bm{X}\bm{Y}^{\top}$.\end{lemma}\begin{proof}See
Appendix~\ref{sec:Proof-of-Lemma-sampling-operator-norm-in-T}.\end{proof}

In the end, we recall a useful lemma which relates the operator norm
to the $\ell_{2,\infty}$ norm of a matrix.

\begin{lemma}\label{lemma:spectral-projection}Suppose that each
$(i,j)$ is included in $\Omega_{0}\subseteq[n]\times[n]$ independently
with probability $\rho_{0}$, and that $n^{2}\rho_{0}\gg n\log n$.
Then there exists some absolute constant $C>0$ such that with probability
at least $1-O(n^{-10})$, 
\[
\left\Vert \mathcal{P}_{\Omega_{0}}\left(\bm{A}\bm{B}^{\top}\right)-\rho_{0}\bm{A}\bm{B}^{\top}\right\Vert \leq C\sqrt{n\rho_{0}}\left\Vert \bm{A}\right\Vert _{2,\infty}\left\Vert \bm{B}\right\Vert _{2,\infty}
\]
holds simultaneously for all $\bm{A}$ and $\bm{B}$. \end{lemma}\begin{proof}See
\cite[Lemmas 4.2 and 4.3]{chen2019nonconvex}.\end{proof}

\subsection{Proof of Lemma~\ref{lemma:sampling-operator-norm-in-T}\label{sec:Proof-of-Lemma-sampling-operator-norm-in-T}}

The lower bound has been established in \cite[Lemma 7]{chen2019noisy},
and hence we focus on the upper bound. We start by expressing $\bm{H}\in T$
as $\bm{H}=\bm{X}\bm{A}^{\top}+\bm{B}\bm{Y}^{\top}$, where $\bm{A},\bm{B}\in\mathbb{R}^{n\times r}$
are chosen to be
\begin{align*}
\left(\bm{A},\bm{B}\right) & \coloneqq\underset{(\tilde{\bm{A}},\tilde{\bm{B}}):\,\bm{H}=\bm{X}\tilde{\bm{A}}^{\top}+\tilde{\bm{B}}\bm{Y}^{\top}}{\arg\min}\Big\{\Vert\tilde{\bm{A}}\Vert_{\mathrm{F}}^{2}/2+\Vert\tilde{\bm{B}}\Vert_{\mathrm{F}}^{2}/2\Big\}.
\end{align*}
The optimality condition of $(\bm{A},\bm{B})$ requires 
\begin{equation}
\bm{X}^{\top}\bm{B}=\bm{A}^{\top}\bm{Y};
\end{equation}
see \cite[Section C.3.1]{chen2019noisy} for the justification of
this identity. The proof then consists of two steps:
\begin{enumerate}
\item Showing that $\|\bm{H}\|_{\mathrm{F}}^{2}$ is bounded from below,
namely,
\[
\left\Vert \bm{H}\right\Vert _{\mathrm{F}}^{2}\geq\frac{49}{100}\sigma_{\min}\left(\left\Vert \bm{A}\right\Vert _{\mathrm{F}}^{2}+\left\Vert \bm{B}\right\Vert _{\mathrm{F}}^{2}\right).
\]
To see this, we can invoke the bound on $\alpha_{2}$ stated in \cite[Appendix C.3.1]{chen2019noisy}
to yield 
\begin{align*}
\left\Vert \bm{H}\right\Vert _{\mathrm{F}}^{2} & =\left\Vert \bm{X}\bm{A}^{\top}+\bm{B}\bm{Y}^{\top}\right\Vert _{\mathrm{F}}^{2}\geq\frac{1}{2}\left(\left\Vert \bm{X}^{\star}\bm{A}^{\top}\right\Vert _{\mathrm{F}}^{2}+\left\Vert \bm{B}\bm{Y}^{\star}\right\Vert _{\mathrm{F}}^{2}\right)-\frac{1}{100}\sigma_{\min}\left(\left\Vert \bm{A}\right\Vert _{\mathrm{F}}^{2}+\left\Vert \bm{B}\right\Vert _{\mathrm{F}}^{2}\right)\\
 & \geq\frac{1}{2}\sigma_{\min}\left(\left\Vert \bm{A}\right\Vert _{\mathrm{F}}^{2}+\left\Vert \bm{B}\right\Vert _{\mathrm{F}}^{2}\right)-\frac{1}{100}\sigma_{\min}\left(\left\Vert \bm{A}\right\Vert _{\mathrm{F}}^{2}+\left\Vert \bm{B}\right\Vert _{\mathrm{F}}^{2}\right)\geq\frac{49}{100}\sigma_{\min}\left(\left\Vert \bm{A}\right\Vert _{\mathrm{F}}^{2}+\left\Vert \bm{B}\right\Vert _{\mathrm{F}}^{2}\right).
\end{align*}
\item Showing that $\|\mathcal{P}_{\Omega^{\star}}(\bm{H})\|_{\mathrm{F}}^{2}$
is bounded from above, namely,
\[
\frac{1}{2\rho_{0}}\left\Vert \mathcal{P}_{\Omega^{\star}}\left(\bm{H}\right)\right\Vert _{\mathrm{F}}^{2}\leq9\sigma_{\max}\left(\left\Vert \bm{A}\right\Vert _{\mathrm{F}}^{2}+\left\Vert \bm{B}\right\Vert _{\mathrm{F}}^{2}\right).
\]
To this end, one starts with the following decomposition 
\begin{equation}
\frac{1}{2\rho_{0}}\left\Vert \mathcal{P}_{\Omega_{0}}\left(\bm{H}\right)\right\Vert _{\mathrm{F}}^{2}=\frac{1}{2}\left\Vert \bm{H}\right\Vert _{\mathrm{F}}^{2}+\frac{1}{2\rho_{0}}\left\Vert \mathcal{P}_{\Omega_{0}}\left(\bm{H}\right)\right\Vert _{\mathrm{F}}^{2}-\frac{1}{2}\left\Vert \bm{H}\right\Vert _{\mathrm{F}}^{2}.\label{eq:decomposition-Pomegao-H-rho0}
\end{equation}
Apply \cite[Equation (83)]{chen2019noisy} to obtain 
\[
\frac{1}{2}\left\Vert \bm{X}\bm{A}^{\top}+\bm{B}\bm{Y}^{\top}\right\Vert _{\mathrm{F}}^{2}\leq8\sigma_{\max}\left(\left\Vert \bm{A}\right\Vert _{\mathrm{F}}^{2}+\left\Vert \bm{B}\right\Vert _{\mathrm{F}}^{2}\right).
\]
In addition, the bound on $\alpha_{1}$ stated in \cite[Appendix C.3.1]{chen2019noisy}
tells us that 
\begin{align*}
\frac{1}{2\rho_{0}}\left\Vert \mathcal{P}_{\Omega_{0}}\left(\bm{H}\right)\right\Vert _{\mathrm{F}}^{2}-\frac{1}{2}\left\Vert \bm{H}\right\Vert _{\mathrm{F}}^{2} & =\frac{1}{2\rho_{0}}\left\Vert \mathcal{P}_{\Omega_{0}}\left(\bm{X}\bm{A}^{\top}+\bm{B}\bm{Y}^{\top}\right)\right\Vert _{\mathrm{F}}^{2}-\frac{1}{2}\left\Vert \bm{X}\bm{A}^{\top}+\bm{B}\bm{Y}^{\top}\right\Vert _{\mathrm{F}}^{2}\\
 & \quad\leq\frac{1}{32}\left(\left\Vert \bm{X}^{\star}\bm{A}^{\top}\right\Vert _{\mathrm{F}}^{2}+\left\Vert \bm{B}\bm{Y}^{\star}\right\Vert _{\mathrm{F}}^{2}\right)+\frac{1}{25}\sigma_{\min}\left(\left\Vert \bm{A}\right\Vert _{\mathrm{F}}^{2}+\left\Vert \bm{B}\right\Vert _{\mathrm{F}}^{2}\right)\\
 & \quad\leq\left(\frac{1}{32}\sigma_{\max}+\frac{1}{25}\sigma_{\min}\right)\left(\left\Vert \bm{A}\right\Vert _{\mathrm{F}}^{2}+\left\Vert \bm{B}\right\Vert _{\mathrm{F}}^{2}\right).
\end{align*}
Substitution into (\ref{eq:decomposition-Pomegao-H-rho0}) gives
\begin{align*}
\frac{1}{2\rho_{0}}\left\Vert \mathcal{P}_{\Omega_{0}}\left(\bm{H}\right)\right\Vert _{\mathrm{F}}^{2} & \leq8\sigma_{\max}\left(\left\Vert \bm{A}\right\Vert _{\mathrm{F}}^{2}+\left\Vert \bm{B}\right\Vert _{\mathrm{F}}^{2}\right)+\left(\frac{1}{32}\sigma_{\max}+\frac{1}{25}\sigma_{\min}\right)\left(\left\Vert \bm{A}\right\Vert _{\mathrm{F}}^{2}+\left\Vert \bm{B}\right\Vert _{\mathrm{F}}^{2}\right)\\
 & \leq9\sigma_{\max}\left(\left\Vert \bm{A}\right\Vert _{\mathrm{F}}^{2}+\left\Vert \bm{B}\right\Vert _{\mathrm{F}}^{2}\right).
\end{align*}
\end{enumerate}
Putting the above two bounds together, we conclude that 
\[
\frac{1}{2\rho_{0}}\left\Vert \mathcal{P}_{\Omega_{0}}\left(\bm{H}\right)\right\Vert _{\mathrm{F}}^{2}\leq9\sigma_{\max}\left(\left\Vert \bm{A}\right\Vert _{\mathrm{F}}^{2}+\left\Vert \bm{B}\right\Vert _{\mathrm{F}}^{2}\right)\leq\frac{900}{49}\kappa\cdot\frac{49}{100}\sigma_{\min}\left(\left\Vert \bm{A}\right\Vert _{\mathrm{F}}^{2}+\left\Vert \bm{B}\right\Vert _{\mathrm{F}}^{2}\right)\leq20\kappa\left\Vert \bm{H}\right\Vert _{\mathrm{F}}^{2}
\]
as claimed.

\section{Proof of Lemma~\ref{lemma:injectivity-main} \label{sec:Proof-of-Lemma-inj-main}}

With Lemma~\ref{lemma:sampling-operator-norm-in-T} in place, we
can immediately justify Lemma~\ref{lemma:injectivity-main}.

To begin with, the first two parts (\ref{eq:P-debias-condition})
and (\ref{eq:inj-lower}) are the same as \cite[Lemma 4]{chen2019noisy}.
Hence, it suffices to verify the last one (\ref{eq:inj-upper}). Recall
from Appendix~\ref{sec:An-equivalent-probabilistic} that $\Omega^{\star}\subseteq\Omega_{\mathsf{aug}}$,
where $\Omega_{\mathsf{aug}}$ is randomly sampled such that each
$(i,j)$ is included in $\Omega_{\mathsf{aug}}$ independently with
probability $p\rho_{\mathsf{aug}}$. Applying Lemma~\ref{lemma:sampling-operator-norm-in-T}
on $\Omega_{\mathsf{aug}}$ finishes the proof, with the proviso that
$\rho_{\mathsf{aug}}\asymp1/\kappa^{2}$ and $\rho_{\mathsf{s}}\leq\rho_{\mathsf{aug}}$.

\section{Crude error bounds (Proof of Theorem~\ref{thm:crude-bound})\label{subsec:Proof-of-Lemma-crude-bound}}

This section is devoted to establishing our crude statistical error
bounds on $\|\bm{L}_{\mathsf{cvx}}-\bm{L}^{\star}\|_{\mathrm{F}}$
and $\|\bm{S}_{\mathsf{cvx}}-\bm{S}^{\star}\|_{\mathrm{F}}$. Without loss of generality, we only consider the case when $\tau=\lambda\sqrt{\frac{\log n}{np}}$. The proof works for general choices $\tau\asymp\lambda\sqrt{\frac{\log n}{np}}$ with slight modification. To simplify
the notation hereafter, we denote 
\begin{align*}
\bm{\Lambda}_{\bm{L}}\coloneqq\bm{L}_{\mathsf{cvx}}-\bm{L}^{\star} & ,\qquad\text{and}\qquad\bm{\Lambda}_{\bm{S}}\coloneqq\bm{S}_{\mathsf{cvx}}-\bm{S}^{\star},\\
\bm{\Lambda}^{+}\coloneqq(\mathcal{P}_{\Omega_{\mathsf{obs}}}\left(\bm{\Lambda}_{\bm{L}}\right)+\bm{\Lambda}_{\bm{S}})/2, & \qquad\text{and}\qquad\bm{\Lambda}^{-}\coloneqq(\mathcal{P}_{\Omega_{\mathsf{obs}}}\left(\bm{\Lambda}_{\bm{L}}\right)-\bm{\Lambda}_{\bm{S}})/2,
\end{align*}
which immediately imply
\[
\bm{\Lambda}_{\bm{L}}=\bm{\Lambda}^{+}+\bm{\Lambda}^{-}+\mathcal{P}_{\Omega_{\mathsf{obs}}^{\mathrm{c}}}\left(\bm{\Lambda}_{\bm{L}}\right),\qquad\text{and}\qquad\bm{\Lambda}_{\bm{S}}=\bm{\Lambda}^{+}-\bm{\Lambda}^{-}.
\]
These in turn allow us to decompose $\|\bm{\Lambda}_{\bm{L}}\|_{\mathrm{F}}^{2}+\|\bm{\Lambda}_{\bm{S}}\|_{\mathrm{F}}^{2}$
as follows
\begin{align}
 & \left\Vert \bm{\Lambda}_{\bm{L}}\right\Vert _{\mathrm{F}}^{2}+\left\Vert \bm{\Lambda}_{\bm{S}}\right\Vert _{\mathrm{F}}^{2}=\left\Vert \bm{\Lambda}^{+}+\bm{\Lambda}^{-}+\mathcal{P}_{\Omega_{\mathsf{obs}}^{\mathrm{c}}}\left(\bm{\Lambda}_{\bm{L}}\right)\right\Vert _{\mathrm{F}}^{2}+\left\Vert \bm{\Lambda}^{+}-\bm{\Lambda}^{-}\right\Vert _{\mathrm{F}}^{2}\nonumber \\
 & \quad=\left\Vert \bm{\Lambda}^{+}\right\Vert _{\mathrm{F}}^{2}+\left\Vert \bm{\Lambda}^{-}+\mathcal{P}_{\Omega_{\mathsf{obs}}^{\mathrm{c}}}\left(\bm{\Lambda}_{\bm{L}}\right)\right\Vert _{\mathrm{F}}^{2}+2\left\langle \bm{\Lambda}^{+},\bm{\Lambda}^{-}+\mathcal{P}_{\Omega_{\mathsf{obs}}^{\mathrm{c}}}\left(\bm{\Lambda}_{\bm{L}}\right)\right\rangle +\left\Vert \bm{\Lambda}^{+}\right\Vert _{\mathrm{F}}^{2}+\left\Vert \bm{\Lambda}^{-}\right\Vert _{\mathrm{F}}^{2}-2\left\langle \bm{\Lambda}^{+},\bm{\Lambda}^{-}\right\rangle \nonumber \\
 & \quad=2\left\Vert \bm{\Lambda}^{+}\right\Vert _{\mathrm{F}}^{2}+\left\Vert \bm{\Lambda}^{-}+\mathcal{P}_{\Omega_{\mathsf{obs}}^{\mathrm{c}}}\left(\bm{\Lambda}_{\bm{L}}\right)\right\Vert _{\mathrm{F}}^{2}+\left\Vert \bm{\Lambda}^{-}\right\Vert _{\mathrm{F}}^{2}+2\left\langle \bm{\Lambda}^{+},\mathcal{P}_{\Omega_{\mathsf{obs}}^{\mathrm{c}}}\left(\bm{\Lambda}_{\bm{L}}\right)\right\rangle .\label{eq:decomposition-Lambda-LS-1}
\end{align}
Since $(\bm{L}_{\mathsf{cvx}},\bm{S}_{\mathsf{cvx}})$ is the minimizer
of (\ref{eq:cvx}), it is self-evident that $\bm{S}_{\mathsf{cvx}}$
must be supported on $\Omega_{\mathsf{obs}}$. Then by construction,
$\bm{\Lambda}_{\bm{S}},\bm{\Lambda}^{+}$ and $\bm{\Lambda}^{-}$
are all necessarily supported on $\Omega_{\mathsf{obs}}$, thus indicating
that
\[
\langle\bm{\Lambda}^{+},\mathcal{P}_{\Omega_{\mathsf{obs}}^{\mathrm{c}}}(\bm{\Lambda}_{\bm{L}})\rangle=0.
\]
Making use of this relation, we can continue the derivation (\ref{eq:decomposition-Lambda-LS-1})
above to obtain
\begin{align*}
 & \left\Vert \bm{\Lambda}_{\bm{L}}\right\Vert _{\mathrm{F}}^{2}+\left\Vert \bm{\Lambda}_{\bm{S}}\right\Vert _{\mathrm{F}}^{2}=2\left\Vert \bm{\Lambda}^{+}\right\Vert _{\mathrm{F}}^{2}+\left\Vert \bm{\Lambda}^{-}+\mathcal{P}_{\Omega_{\mathsf{obs}}^{\mathrm{c}}}\left(\bm{\Lambda}_{\bm{L}}\right)\right\Vert _{\mathrm{F}}^{2}+\left\Vert \bm{\Lambda}^{-}\right\Vert _{\mathrm{F}}^{2}\\
 & \quad=\underbrace{2\left\Vert \bm{\Lambda}^{+}\right\Vert _{\mathrm{F}}^{2}}_{\eqqcolon\alpha_{1}}+\underbrace{\left\Vert \mathcal{P}_{T^{\star}}\left(\bm{\Lambda}^{-}+\mathcal{P}_{\Omega_{\mathsf{obs}}^{\mathrm{c}}}\left(\bm{\Lambda}_{\bm{L}}\right)\right)\right\Vert _{\mathrm{F}}^{2}+\left\Vert \mathcal{P}_{\Omega^{\star}}\left(\bm{\Lambda}^{-}\right)\right\Vert _{\mathrm{F}}^{2}}_{\eqqcolon\alpha_{2}}+\underbrace{\left\Vert \mathcal{P}_{T^{\star\perp}}\left(\bm{\Lambda}^{-}+\mathcal{P}_{\Omega_{\mathsf{obs}}^{\mathrm{c}}}\left(\bm{\Lambda}_{\bm{L}}\right)\right)\right\Vert _{\mathrm{F}}^{2}+\left\Vert \mathcal{P}_{\Omega^{\star\mathrm{c}}}\left(\bm{\Lambda}^{-}\right)\right\Vert _{\mathrm{F}}^{2}}_{\eqqcolon\alpha_{3}}.
\end{align*}
In the sequel, we shall control the three terms $\alpha_{1},\alpha_{2}$
and $\alpha_{3}$ separately.

\paragraph{Step 1: bounding $\alpha_{1}$.}

By definition, we have 
\begin{align}
\alpha_{1} & =2\left\Vert \bm{\Lambda}^{+}\right\Vert _{\mathrm{F}}^{2}=\frac{1}{2}\left\Vert \mathcal{P}_{\Omega_{\mathsf{obs}}}\left(\bm{\Lambda}_{\bm{L}}\right)+\bm{\Lambda}_{\bm{S}}\right\Vert _{\mathrm{F}}^{2}=\frac{1}{2}\left\Vert \mathcal{P}_{\Omega_{\mathsf{obs}}}\left(\bm{\Lambda}_{\bm{L}}+\bm{\Lambda}_{\bm{S}}\right)\right\Vert _{\mathrm{F}}^{2}\nonumber \\
 & =\frac{1}{2}\left\Vert \mathcal{P}_{\Omega_{\mathsf{obs}}}\left(\bm{L}_{\mathsf{cvx}}+\bm{S}_{\mathsf{cvx}}-\bm{M}+\bm{M}-\bm{L}^{\star}-\bm{S}^{\star}\right)\right\Vert _{\mathrm{F}}^{2}\nonumber \\
 & \leq\left\Vert \mathcal{P}_{\Omega_{\mathsf{obs}}}\left(\bm{L}_{\mathsf{cvx}}+\bm{S}_{\mathsf{cvx}}-\bm{M}\right)\right\Vert _{\mathrm{F}}^{2}+\left\Vert \mathcal{P}_{\Omega_{\mathsf{obs}}}\left(\bm{L}^{\star}+\bm{S}^{\star}-\bm{M}\right)\right\Vert _{\mathrm{F}}^{2}\nonumber \\
 & =\left\Vert \mathcal{P}_{\Omega_{\mathsf{obs}}}\left(\bm{L}_{\mathsf{cvx}}+\bm{S}_{\mathsf{cvx}}-\bm{M}\right)\right\Vert _{\mathrm{F}}^{2}+\left\Vert \mathcal{P}_{\Omega_{\mathsf{obs}}}\left(\bm{E}\right)\right\Vert _{\mathrm{F}}^{2},\label{eq:crude_alpha_1}
\end{align}
where the third identity holds true since $\bm{\Lambda}_{\bm{S}}=\mathcal{P}_{\Omega_{\mathsf{obs}}}(\bm{\Lambda}_{\bm{S}})$,
the penultimate relation is due to the elementary inequality $\|\bm{A}+\bm{B}\|_{\mathrm{F}}^{2}\leq2\|\bm{A}\|_{\mathrm{F}}^{2}+2\|\bm{B}\|_{\mathrm{F}}^{2}$,
and the last line follows since $\mathcal{P}_{\Omega_{\mathsf{obs}}}(\bm{L}^{\star}+\bm{S}^{\star}-\bm{M})=\mathcal{P}_{\Omega_{\mathsf{obs}}}(\bm{E})$.
To upper bound $\|\mathcal{P}_{\Omega_{\mathsf{obs}}}(\bm{L}_{\mathsf{cvx}}+\bm{S}_{\mathsf{cvx}}-\bm{M})\|_{\mathrm{F}}^{2}$,
we leverage the optimality of $(\bm{L}_{\mathsf{cvx}},\bm{S}_{\mathsf{cvx}})$
w.r.t.~the convex program~(\ref{eq:cvx}) to obtain
\begin{align}
 & \frac{1}{2}\left\Vert \mathcal{P}_{\Omega_{\mathsf{obs}}}\left(\bm{L}_{\mathsf{cvx}}+\bm{S}_{\mathsf{cvx}}-\bm{M}\right)\right\Vert _{\mathrm{F}}^{2}+\lambda\left\Vert \bm{L}_{\mathsf{cvx}}\right\Vert _{\ast}+\tau\left\Vert \bm{S}_{\mathsf{cvx}}\right\Vert _{1}\nonumber \\
 & \quad\qquad\leq\frac{1}{2}\left\Vert \mathcal{P}_{\Omega_{\mathsf{obs}}}\left(\bm{L}^{\star}+\bm{S}^{\star}-\bm{M}\right)\right\Vert _{\mathrm{F}}^{2}+\lambda\left\Vert \bm{L}^{\star}\right\Vert _{\ast}+\tau\left\Vert \bm{S}^{\star}\right\Vert _{1}.\label{eq:crude-bound-proof-optimality}
\end{align}
Recognizing again that $\mathcal{P}_{\Omega_{\mathsf{obs}}}(\bm{L}^{\star}+\bm{S}^{\star}-\bm{M})=\mathcal{P}_{\Omega_{\mathsf{obs}}}(\bm{E})$,
we can rearrange terms in~(\ref{eq:crude-bound-proof-optimality})
to derive 
\begin{align}
\left\Vert \mathcal{P}_{\Omega_{\mathsf{obs}}}\left(\bm{L}_{\mathsf{cvx}}+\bm{S}_{\mathsf{cvx}}-\bm{M}\right)\right\Vert _{\mathrm{F}}^{2} & \leq\|\mathcal{P}_{\Omega_{\mathsf{obs}}}\left(\bm{E}\right)\|_{\mathrm{F}}^{2}+2\lambda\left\Vert \bm{L}^{\star}\right\Vert _{\ast}+2\tau\left\Vert \bm{S}^{\star}\right\Vert _{1}-2\lambda\big\|\bm{L}_{\mathsf{cvx}}\big\|_{\ast}-2\tau\big\|\bm{S}_{\mathsf{cvx}}\big\|_{1}\nonumber \\
 & \overset{(\text{i})}{\leq}\|\mathcal{P}_{\Omega_{\mathsf{obs}}}\left(\bm{E}\right)\|_{\mathrm{F}}^{2}+2\lambda\left\Vert \bm{\Lambda}_{\bm{L}}\right\Vert _{*}+2\tau\left\Vert \bm{\Lambda}_{\bm{S}}\right\Vert _{1}\nonumber \\
 & \overset{(\text{ii})}{\leq}\|\mathcal{P}_{\Omega_{\mathsf{obs}}}\left(\bm{E}\right)\|_{\mathrm{F}}^{2}+2\lambda\sqrt{n}\left\Vert \bm{\Lambda}_{\bm{L}}\right\Vert _{\mathrm{F}}+2\tau\sqrt{|\Omega_{\mathsf{obs}}|}\left\Vert \bm{\Lambda}_{\bm{S}}\right\Vert _{\mathrm{F}}\nonumber \\
 & \overset{(\text{iii})}{\leq}\|\mathcal{P}_{\Omega_{\mathsf{obs}}}\left(\bm{E}\right)\|_{\mathrm{F}}^{2}+2\sqrt{2}\lambda\sqrt{n\log n}\left(\left\Vert \bm{\Lambda}_{\bm{L}}\right\Vert _{\mathrm{F}}+\left\Vert \bm{\Lambda}_{\bm{S}}\right\Vert _{\mathrm{F}}\right),\label{eq:crude_alpha_1_2}
\end{align}
where $|\Omega_{\mathsf{obs}}|$ denotes the cardinality of $\Omega_{\mathsf{obs}}$.
Here, the relation (i) results from the triangle inequality, the inequality
(ii) holds true since $\|\bm{A}\|_{*}\leq\sqrt{n}\|\bm{A}\|_{\mathrm{F}}$
for any $\bm{A}\in\mathbb{R}^{n\times n}$ and $\|\bm{\Lambda}_{\bm{S}}\|_{1}=\|\mathcal{P}_{\Omega_{\mathsf{obs}}}(\bm{\Lambda}_{\bm{S}})\|_{1}\leq\sqrt{|\Omega_{\mathsf{obs}}|}\|\bm{\Lambda}_{\bm{S}}\|_{\mathrm{F}}$,
and the last line (iii) arises from the fact that $|\Omega_{\mathsf{obs}}|\leq2n^{2}p$
with high probability as well as the choice $\tau=\lambda\sqrt{\frac{\log n}{np}}$.
Combine (\ref{eq:crude_alpha_1}) and (\ref{eq:crude_alpha_1_2})
to reach
\begin{align}
\alpha_{1} & \leq2\|\mathcal{P}_{\Omega_{\mathsf{obs}}}\left(\bm{E}\right)\|_{\mathrm{F}}^{2}+2\sqrt{2}\lambda\sqrt{n\log n}\left(\left\Vert \bm{\Lambda}_{\bm{L}}\right\Vert _{\mathrm{F}}+\left\Vert \bm{\Lambda}_{\bm{S}}\right\Vert _{\mathrm{F}}\right)\nonumber \\
 & \leq2\left\Vert \mathcal{P}_{\Omega_{\mathsf{obs}}}\left(\bm{E}\right)\right\Vert _{\mathrm{F}}^{2}+4\lambda\sqrt{n\log n}\sqrt{\left\Vert \bm{\Lambda}_{\bm{L}}\right\Vert _{\mathrm{F}}^{2}+\left\Vert \bm{\Lambda}_{\bm{S}}\right\Vert _{\mathrm{F}}^{2}},\label{eq:UB-alpha1}
\end{align}
where we use the elementary inequality $a+b\leq\sqrt{2}\cdot\sqrt{a^{2}+b^{2}}$.

\paragraph{Step 2: bounding $\alpha_{2}$ via $\alpha_{3}$.}

To relate $\alpha_{2}$ to $\alpha_{3}$, the following lemma plays
a crucial role, whose proof is deferred to Appendix~\ref{subsec:Proof-of-Lemma-crude-bound-tool-2}.

\begin{lemma}\label{lemma:crude-bound-tool-2}Suppose that $\|\mathcal{P}_{\Omega^{\star}}\mathcal{P}_{T^{\star}}\|^{2}\leq p/8$
and that $\Vert\mathcal{P}_{T^{\star}}-p^{-1}\mathcal{P}_{T^{\star}}\mathcal{P}_{\Omega_{\mathsf{obs}}}\mathcal{P}_{T^{\star}}\Vert\leq1/2$.
Then for any pair $(\bm{A},\bm{B})$ of matrices, we have 
\begin{equation}
\left\Vert \mathcal{P}_{T^{\star}}\left(\bm{A}\right)\right\Vert _{\mathrm{F}}^{2}+\left\Vert \mathcal{P}_{\Omega^{\star}}\left(\bm{B}\right)\right\Vert _{\mathrm{F}}^{2}\leq\frac{4}{p}\left\Vert \mathcal{P}_{\Omega_{\mathsf{obs}}}\left[\mathcal{P}_{T^{\star}}\left(\bm{A}\right)+\mathcal{P}_{\Omega^{\star}}\left(\bm{B}\right)\right]\right\Vert _{\mathrm{F}}^{2}.
\end{equation}

\end{lemma}

Suppose for the moment that the assumptions of Lemma~\ref{lemma:crude-bound-tool-2}
hold. Taking $(\bm{A},\bm{B})$ as $(\bm{\Lambda}^{-}+\mathcal{P}_{\Omega_{\mathsf{obs}}^{\mathrm{c}}}(\bm{\Lambda}_{\bm{L}}),-\bm{\Lambda}^{-})$
in Lemma~\ref{lemma:crude-bound-tool-2} yields 
\[
\alpha_{2}=\left\Vert \mathcal{P}_{T^{\star}}\left(\bm{\Lambda}^{-}+\mathcal{P}_{\Omega_{\mathsf{obs}}^{\mathrm{c}}}\left(\bm{\Lambda}_{\bm{L}}\right)\right)\right\Vert _{\mathrm{F}}^{2}+\left\Vert \mathcal{P}_{\Omega^{\star}}\left(\bm{\Lambda}^{-}\right)\right\Vert _{\mathrm{F}}^{2}\leq\frac{4}{p}\left\Vert \mathcal{P}_{\Omega_{\mathsf{obs}}}\left[\mathcal{P}_{T^{\star}}\left(\bm{\Lambda}^{-}+\mathcal{P}_{\Omega_{\mathsf{obs}}^{\mathrm{c}}}\left(\bm{\Lambda}_{\bm{L}}\right)\right)-\mathcal{P}_{\Omega^{\star}}\left(\bm{\Lambda}^{-}\right)\right]\right\Vert _{\mathrm{F}}^{2}.
\]
By virtue of the identity 
\begin{align*}
\mathcal{P}_{T^{\star}}\left(\bm{\Lambda}^{-}+\mathcal{P}_{\Omega_{\mathsf{obs}}^{\mathrm{c}}}\left(\bm{\Lambda}_{\bm{L}}\right)\right)-\mathcal{P}_{\Omega^{\star}}\left(\bm{\Lambda}^{-}\right) & =\bm{\Lambda}^{-}+\mathcal{P}_{\Omega_{\mathsf{obs}}^{\mathrm{c}}}\left(\bm{\Lambda}_{\bm{L}}\right)-\mathcal{P}_{T^{\star\perp}}\left(\bm{\Lambda}^{-}+\mathcal{P}_{\Omega_{\mathsf{obs}}^{\mathrm{c}}}\left(\bm{\Lambda}_{\bm{L}}\right)\right)-\bm{\Lambda}^{-}+\mathcal{P}_{(\Omega^{\star})^{\mathrm{c}}}(\bm{\Lambda}^{-})\\
 & =\mathcal{P}_{\Omega_{\mathsf{obs}}^{\mathrm{c}}}\left(\bm{\Lambda}_{\bm{L}}\right)-\mathcal{P}_{T^{\star\perp}}\left(\bm{\Lambda}^{-}+\mathcal{P}_{\Omega_{\mathsf{obs}}^{\mathrm{c}}}\left(\bm{\Lambda}_{\bm{L}}\right)\right)+\mathcal{P}_{(\Omega^{\star})^{\mathrm{c}}}(\bm{\Lambda}^{-}),
\end{align*}
we further obtain 
\begin{align}
\alpha_{2} & \leq\frac{4}{p}\left\Vert \mathcal{P}_{\Omega_{\mathsf{obs}}}\left[\mathcal{P}_{\Omega_{\mathsf{obs}}^{\mathrm{c}}}\left(\bm{\Lambda}_{\bm{L}}\right)-\mathcal{P}_{T^{\star\perp}}\left(\bm{\Lambda}^{-}+\mathcal{P}_{\Omega_{\mathsf{obs}}^{\mathrm{c}}}\left(\bm{\Lambda}_{\bm{L}}\right)\right)+\mathcal{P}_{(\Omega^{\star})^{\mathrm{c}}}(\bm{\Lambda}^{-})\right]\right\Vert _{\mathrm{F}}^{2}\nonumber \\
 & =\frac{4}{p}\left\Vert \mathcal{P}_{\Omega_{\mathsf{obs}}}\left[\mathcal{P}_{T^{\star\perp}}\left(\bm{\Lambda}^{-}+\mathcal{P}_{\Omega_{\mathsf{obs}}^{\mathrm{c}}}\left(\bm{\Lambda}_{\bm{L}}\right)\right)-\mathcal{P}_{(\Omega^{\star})^{\mathrm{c}}}(\bm{\Lambda}^{-})\right]\right\Vert _{\mathrm{F}}^{2}\nonumber \\
 & \leq\frac{4}{p}\left\Vert \mathcal{P}_{T^{\star\perp}}\left(\bm{\Lambda}^{-}+\mathcal{P}_{\Omega_{\mathsf{obs}}^{\mathrm{c}}}\left(\bm{\Lambda}_{\bm{L}}\right)\right)-\mathcal{P}_{(\Omega^{\star})^{\mathrm{c}}}\left(\bm{\Lambda}^{-}\right)\right\Vert _{\mathrm{F}}^{2}\nonumber \\
 & \leq\frac{8}{p}\left\Vert \mathcal{P}_{T^{\star\perp}}\left(\bm{\Lambda}^{-}+\mathcal{P}_{\Omega_{\mathsf{obs}}^{\mathrm{c}}}\left(\bm{\Lambda}_{\bm{L}}\right)\right)\right\Vert _{\mathrm{F}}^{2}+\frac{8}{p}\left\Vert \mathcal{P}_{(\Omega^{\star})^{\mathrm{c}}}\left(\bm{\Lambda}^{-}\right)\right\Vert _{\mathrm{F}}^{2}=\frac{8}{p}\alpha_{3}.\label{eq:alpha2-alpha3}
\end{align}
Once again, the derivation has made use of the elementary inequality
$\|\bm{A}+\bm{B}\|_{\mathrm{F}}^{2}\leq2\|\bm{A}\|_{\mathrm{F}}^{2}+2\|\bm{B}\|_{\mathrm{F}}^{2}$.

\paragraph{Step 3: bounding $\alpha_{3}$ via $\alpha_{1}$ and $\|\mathcal{P}_{\Omega_{\mathsf{obs}}}(\bm{E})\|_{\mathrm{F}}$.}

The following lemma proves useful in linking $\alpha_{3}$ with $\alpha_{1}$,
and we postpone the proof to Appendix~\ref{subsec:Proof-of-Lemma-crude-bound-tool-1}.

\begin{lemma}\label{lemma:crude-bound-tool-1}Assume that $n^{2}p\gg n\log n$,
$\rho_{\mathsf{s}}\ll1$ and $\Vert\mathcal{P}_{T^{\star}}-p^{-1}(1-\rho_{\mathsf{s}})^{-1}\mathcal{P}_{T^{\star}}\mathcal{P}_{\Omega_{\mathsf{obs}}\setminus\Omega^{\star}}\mathcal{P}_{T^{\star}}\Vert\leq1/2$.
Further assume that there exists a dual certificate $\bm{W}\in\mathbb{R}^{n\times n}$
such that \begin{subequations}\label{eq:dc} 
\begin{align}
\left\Vert \mathcal{P}_{T^{\star}}\left[\lambda\bm{W}+\tau\mathsf{sgn}\left(\bm{S}^{\star}\right)-\lambda\bm{U}^{\star}\bm{V}^{\star\top}\right]\right\Vert _{\mathrm{F}} & \leq\tau/\sqrt{n},\label{eq:dc-1}\\
\left\Vert \mathcal{P}_{T^{\star\perp}}\left[\lambda\bm{W}+\tau\mathsf{sign}\left(\bm{S}^{\star}\right)\right]\right\Vert  & <\lambda/2,\label{eq:dc-2}\\
\mathcal{P}_{(\Omega_{\mathsf{obs}}\setminus\Omega^{\star})^{\mathrm{c}}}\left(\bm{W}\right) & =\bm{0},\label{eq:dc-3}\\
\left\Vert \lambda\bm{W}\right\Vert _{\infty} & <\tau/2,\label{eq:dc-4}
\end{align}
\end{subequations}where $\mathsf{sign}\left(\bm{S}^{\star}\right)\coloneqq[\mathsf{sign}(S_{ij}^{\star})]_{1\leq i,j\leq n}$.
Then for any $\bm{H}_{\bm{L}},\bm{H}_{\bm{S}}\in\mathbb{R}^{n\times n}$
satisfying $\mathcal{P}_{\Omega_{\mathsf{obs}}}(\bm{H}_{\bm{L}})+\bm{H}_{\bm{S}}=\bm{0}$,
one has 
\begin{align*}
\lambda\left\Vert \bm{L}^{\star}+\bm{H}_{\bm{L}}\right\Vert _{*}+\tau\left\Vert \bm{S}^{\star}+\bm{H}_{\bm{S}}\right\Vert _{1} & \geq\lambda\left\Vert \bm{L}^{\star}\right\Vert _{*}+\tau\left\Vert \bm{S}^{\star}\right\Vert _{1}+\frac{\lambda}{4}\left\Vert \mathcal{P}_{T^{\star\perp}}\left(\bm{H}_{\bm{L}}\right)\right\Vert _{*}+\frac{\tau}{4}\left\Vert \mathcal{P}_{\Omega_{\mathsf{obs}}\setminus\Omega^{\star}}\left(\bm{H}_{\bm{S}}\right)\right\Vert _{1}.
\end{align*}
\end{lemma}

Again, we assume for the moment that the assumptions in Lemma~\ref{lemma:crude-bound-tool-1}
hold. Setting $\bm{H}_{L}=\bm{\Lambda}^{-}+\mathcal{P}_{\Omega_{\mathsf{obs}}^{\mathrm{c}}}(\bm{\Lambda}_{\bm{L}})$
and $\bm{H}_{\bm{S}}=-\bm{\Lambda}^{-}$ in Lemma~\ref{lemma:crude-bound-tool-1}
gives 
\begin{align*}
 & \lambda\left\Vert \bm{L}^{\star}+\bm{\Lambda}^{-}+\mathcal{P}_{\Omega_{\mathsf{obs}}^{\mathrm{c}}}\left(\bm{\Lambda}_{\bm{L}}\right)\right\Vert _{\ast}+\tau\left\Vert \bm{S}^{\star}-\bm{\Lambda}^{-}\right\Vert _{1}\\
 & \quad\geq\lambda\left\Vert \bm{L}^{\star}\right\Vert _{\ast}+\tau\left\Vert \bm{S}^{\star}\right\Vert _{1}+\frac{\lambda}{4}\left\Vert \mathcal{P}_{T^{\star\perp}}\left(\bm{\Lambda}^{-}+\mathcal{P}_{\Omega_{\mathsf{obs}}^{\mathrm{c}}}\left(\bm{\Lambda}_{\bm{L}}\right)\right)\right\Vert _{*}+\frac{\tau}{4}\left\Vert \mathcal{P}_{\Omega_{\mathsf{obs}}\setminus\Omega^{\star}}\left(\bm{\Lambda}^{-}\right)\right\Vert _{1}.
\end{align*}
In addition, recalling the identities $\bm{L}_{\mathsf{cvx}}=\bm{L}^{\star}+\bm{\Lambda}^{+}+\bm{\Lambda}^{-}+\mathcal{P}_{\Omega_{\mathsf{obs}}^{\mathrm{c}}}(\bm{\Lambda}_{\bm{L}})$
and $\bm{S}_{\mathsf{cvx}}=\bm{S}^{\star}+\bm{\Lambda}^{+}-\bm{\Lambda}^{-}$,
we can invoke the triangle inequality to obtain 
\begin{align*}
\lambda\left\Vert \bm{L}_{\mathsf{cvx}}\right\Vert _{\ast}+\tau\left\Vert \bm{S}_{\mathsf{cvx}}\right\Vert _{1} & =\lambda\left\Vert \bm{L}^{\star}+\bm{\Lambda}^{-}+\bm{\Lambda}^{+}+\mathcal{P}_{\Omega_{\mathsf{obs}}^{\mathrm{c}}}\left(\bm{\Lambda}_{\bm{L}}\right)\right\Vert _{\ast}+\tau\left\Vert \bm{S}^{\star}-\bm{\Lambda}^{-}+\bm{\Lambda}^{+}\right\Vert _{1}\\
 & \geq\lambda\left\Vert \bm{L}^{\star}+\bm{\Lambda}^{-}+\mathcal{P}_{\Omega_{\mathsf{obs}}^{\mathrm{c}}}\left(\bm{\Lambda}_{\bm{L}}\right)\right\Vert _{\ast}+\tau\left\Vert \bm{S}^{\star}-\bm{\Lambda}^{-}\right\Vert _{1}-\lambda\left\Vert \bm{\Lambda}^{+}\right\Vert _{\ast}-\tau\left\Vert \bm{\Lambda}^{+}\right\Vert _{1}.
\end{align*}
Adding the above two inequalities and using the fact $\mathsf{support}(\bm{\Lambda}^{-})\subseteq\Omega_{\mathrm{obs}}$
lead to
\begin{align}
 & \frac{\lambda}{4}\left\Vert \mathcal{P}_{T^{\star\perp}}\left(\bm{\Lambda}^{-}+\mathcal{P}_{\Omega_{\mathsf{obs}}^{\mathrm{c}}}\left(\bm{\Lambda}_{\bm{L}}\right)\right)\right\Vert _{*}+\frac{\tau}{4}\left\Vert \mathcal{P}_{(\Omega^{\star})^{\mathrm{c}}}\left(\bm{\Lambda}^{-}\right)\right\Vert _{1}=\frac{\lambda}{4}\left\Vert \mathcal{P}_{T^{\star\perp}}\left(\bm{\Lambda}^{-}+\mathcal{P}_{\Omega_{\mathsf{obs}}^{\mathrm{c}}}\left(\bm{\Lambda}_{\bm{L}}\right)\right)\right\Vert _{*}+\frac{\tau}{4}\left\Vert \mathcal{P}_{\Omega_{\mathsf{obs}}\setminus\Omega^{\star}}\left(\bm{\Lambda}^{-}\right)\right\Vert _{1}\nonumber \\
 & \quad\leq\lambda\left\Vert \bm{L}_{\mathsf{cvx}}\right\Vert _{\ast}+\tau\left\Vert \bm{S}_{\mathsf{cvx}}\right\Vert _{1}+\lambda\left\Vert \bm{\Lambda}^{+}\right\Vert _{\ast}+\tau\left\Vert \bm{\Lambda}^{+}\right\Vert _{1}-\lambda\left\Vert \bm{L}^{\star}\right\Vert _{\ast}-\tau\left\Vert \bm{S}^{\star}\right\Vert _{1}\nonumber \\
 & \quad\leq\frac{1}{2}\left\Vert \mathcal{P}_{\Omega_{\mathsf{obs}}}\left(\bm{L}^{\star}+\bm{S}^{\star}-\bm{M}\right)\right\Vert _{\mathrm{F}}^{2}-\frac{1}{2}\left\Vert \mathcal{P}_{\Omega_{\mathsf{obs}}}\left(\bm{L}_{\mathsf{cvx}}+\bm{S}_{\mathsf{cvx}}-\bm{M}\right)\right\Vert _{\mathrm{F}}^{2}+\lambda\left\Vert \bm{\Lambda}^{+}\right\Vert _{\ast}+\tau\left\Vert \bm{\Lambda}^{+}\right\Vert _{1}\nonumber \\
 & \quad\leq\frac{1}{2}\left\Vert \mathcal{P}_{\Omega_{\mathsf{obs}}}\left(\bm{E}\right)\right\Vert _{\mathrm{F}}^{2}+4\lambda\sqrt{n\log n}\left\Vert \bm{\Lambda}^{+}\right\Vert _{\mathrm{F}}.\label{eq:alpha3-UB-2}
\end{align}
Here, the penultimate line results from the inequality (\ref{eq:crude-bound-proof-optimality})
and last line follows from the same argument in obtaining~(\ref{eq:crude_alpha_1_2}).

We are now ready to establish the upper bound on $\alpha_{3}$. Invoke
the elementary inequalities $\|\bm{A}\|_{\mathrm{F}}\leq\|\bm{A}\|_{*}$
and $\|\bm{A}\|_{\mathrm{F}}\leq\|\bm{A}\|_{1}$ for any $\bm{A}\in\mathbb{R}^{n\times n}$
to show that 
\begin{align*}
\alpha_{3} & =\left\Vert \mathcal{P}_{T^{\star\perp}}\left(\bm{\Lambda}^{-}+\mathcal{P}_{\Omega_{\mathsf{obs}}^{\mathrm{c}}}\left(\bm{\Lambda}_{\bm{L}}\right)\right)\right\Vert _{\mathrm{F}}^{2}+\left\Vert \mathcal{P}_{(\Omega^{\star})^{\mathrm{c}}}\left(\bm{\Lambda}^{-}\right)\right\Vert _{\mathrm{F}}^{2}\leq\left\Vert \mathcal{P}_{T^{\star\perp}}\left(\bm{\Lambda}^{-}+\mathcal{P}_{\Omega_{\mathsf{obs}}^{\mathrm{c}}}\left(\bm{\Lambda}_{\bm{L}}\right)\right)\right\Vert _{*}^{2}+\left\Vert \mathcal{P}_{(\Omega^{\star})^{\mathrm{c}}}\left(\bm{\Lambda}^{-}\right)\right\Vert _{1}^{2}\\
 & \leq\left(\frac{16}{\lambda^{2}}+\frac{16}{\tau^{2}}\right)\left(\frac{\lambda}{4}\left\Vert \mathcal{P}_{T^{\star\perp}}\left(\bm{\Lambda}^{-}+\mathcal{P}_{\Omega_{\mathsf{obs}}^{\mathrm{c}}}\left(\bm{\Lambda}_{\bm{L}}\right)\right)\right\Vert _{*}+\frac{\tau}{4}\left\Vert \mathcal{P}_{(\Omega^{\star})^{\mathrm{c}}}\left(\bm{\Lambda}^{-}\right)\right\Vert _{1}\right)^{2}.
\end{align*}
This combined with (\ref{eq:alpha3-UB-2}) allows us to obtain 
\[
\alpha_{3}\leq\left(\frac{16}{\lambda^{2}}+\frac{16}{\tau^{2}}\right)\left(\frac{1}{2}\left\Vert \mathcal{P}_{\Omega_{\mathsf{obs}}}\left(\bm{E}\right)\right\Vert _{\mathrm{F}}^{2}+4\lambda\sqrt{n}\left\Vert \bm{\Lambda}^{+}\right\Vert _{\mathrm{F}}\right)^{2}\leq\left(\frac{32}{\lambda^{2}}+\frac{32}{\tau^{2}}\right)\left(\frac{1}{4}\left\Vert \mathcal{P}_{\Omega_{\mathsf{obs}}}\left(\bm{E}\right)\right\Vert _{\mathrm{F}}^{4}+16\lambda^{2}n\log n\left\Vert \bm{\Lambda}^{+}\right\Vert _{\mathrm{F}}^{2}\right),
\]
where we have used the elementary inequality $(a+b)^{2}\leq2a^{2}+2b^{2}$.
Recalling that $\tau=\lambda/\sqrt{np/\log n}$ and that $np\geq1$,
we arrive at 
\begin{equation}
\alpha_{3}\leq\frac{64np}{\lambda^{2}}\left(\frac{1}{4}\left\Vert \mathcal{P}_{\Omega_{\mathsf{obs}}}\left(\bm{E}\right)\right\Vert _{\mathrm{F}}^{4}+16\lambda^{2}n\log n\left\Vert \bm{\Lambda}^{+}\right\Vert _{\mathrm{F}}^{2}\right)=\frac{16np}{\lambda^{2}}\left\Vert \mathcal{P}_{\Omega_{\mathsf{obs}}}\left(\bm{E}\right)\right\Vert _{\mathrm{F}}^{4}+2^{9}n^{2}p\alpha_{1}\log n,\label{eq:crude_alpha_3}
\end{equation}
where we have identified $2\|\bm{\Lambda}^{+}\|_{\mathrm{F}}^{2}$
with $\alpha_{1}$.

\paragraph{Step 4: putting the above bounds on $\alpha_{1},\alpha_{2},\alpha_{3}$
together.}

Taking the preceding bounds on $\alpha_{1}$, $\alpha_{2}$ and $\alpha_{3}$
collectively yields 
\begin{align*}
\left\Vert \bm{\Lambda}_{\bm{L}}\right\Vert _{\mathrm{F}}^{2}+\left\Vert \bm{\Lambda}_{\bm{S}}\right\Vert _{\mathrm{F}}^{2} & =\alpha_{1}+\alpha_{2}+\alpha_{3}\overset{(\mathrm{i})}{\leq}\alpha_{1}+\left(1+\frac{8}{p}\right)\alpha_{3}\overset{(\text{ii})}{\leq}\alpha_{1}+\frac{16}{p}\alpha_{3}\\
 & \overset{(\mathrm{iii})}{\leq}\left(2^{13}n^{2}\log n+1\right)\alpha_{1}+\frac{2^{8}n}{\lambda^{2}}\left\Vert \mathcal{P}_{\Omega_{\mathsf{obs}}}\left(\bm{E}\right)\right\Vert _{\mathrm{F}}^{4}\\
 & \overset{(\mathrm{iv})}{\leq}\left(2^{13}n^{2}\log n+1\right)\left[2\left\Vert \mathcal{P}_{\Omega_{\mathsf{obs}}}\left(\bm{E}\right)\right\Vert _{\mathrm{F}}^{2}+4\lambda\sqrt{n\log n}\sqrt{\left\Vert \bm{\Lambda}_{\bm{L}}\right\Vert _{\mathrm{F}}^{2}+\left\Vert \bm{\Lambda}_{\bm{S}}\right\Vert _{\mathrm{F}}^{2}}\right]+\frac{2^{8}n}{\lambda^{2}}\left\Vert \mathcal{P}_{\Omega_{\mathsf{obs}}}\left(\bm{E}\right)\right\Vert _{\mathrm{F}}^{4}.
\end{align*}
Here, the first inequality (i) comes from (\ref{eq:alpha2-alpha3}),
the second inequality (ii) follows from the fact $1\leq8/p$, the third relation
(iii) is a consequence of (\ref{eq:crude_alpha_3}), and the last
line (iv) results from~(\ref{eq:UB-alpha1}). Note that this forms
a quadratic inequality in~$\sqrt{\|\bm{\Lambda}_{\bm{L}}\|_{\mathrm{F}}^{2}+\|\bm{\Lambda}_{\bm{S}}\|_{\mathrm{F}}^{2}}$.
Solving the inequality yields the claimed bound
\begin{align*}
\left\Vert \bm{\Lambda}_{\bm{L}}\right\Vert _{\mathrm{F}}^{2}+\left\Vert \bm{\Lambda}_{\bm{S}}\right\Vert _{\mathrm{F}}^{2} & \lesssim\lambda^{2}n^{5}\log^3 n+n^{2}\log n\left\Vert \mathcal{P}_{\Omega_{\mathsf{obs}}}\left(\bm{E}\right)\right\Vert _{\mathrm{F}}^{2}+\frac{n}{\lambda^{2}}\left\Vert \mathcal{P}_{\Omega_{\mathsf{obs}}}\left(\bm{E}\right)\right\Vert _{\mathrm{F}}^{4}.
\end{align*}
Further, the elementary inequality $a^{2}+b^{2}\geq2ab$ yields
\begin{align*}
\lambda^{2}n^{5}\log^3 n+\frac{n}{\lambda^{2}}\left\Vert \mathcal{P}_{\Omega_{\mathsf{obs}}}\left(\bm{E}\right)\right\Vert _{\mathrm{F}}^{4} & \geq2n^{3}\log^{3/2}n\left\Vert \mathcal{P}_{\Omega_{\mathsf{obs}}}\left(\bm{E}\right)\right\Vert _{\mathrm{F}}^{2}\geq n^{2}\log n\left\Vert \mathcal{P}_{\Omega_{\mathsf{obs}}}\left(\bm{E}\right)\right\Vert _{\mathrm{F}}^{2},
\end{align*}
leading to the simplified bound
\begin{align*}
\left\Vert \bm{\Lambda}_{\bm{L}}\right\Vert _{\mathrm{F}}^{2}+\left\Vert \bm{\Lambda}_{\bm{S}}\right\Vert _{\mathrm{F}}^{2} & \lesssim\lambda^{2}n^{5}\log^3 n+\frac{n}{\lambda^{2}}\left\Vert \mathcal{P}_{\Omega_{\mathsf{obs}}}\left(\bm{E}\right)\right\Vert _{\mathrm{F}}^{4}.
\end{align*}

\paragraph{Step 5: checking the conditions in Lemmas~\ref{lemma:crude-bound-tool-2}
and~\ref{lemma:crude-bound-tool-1}.}

We are left with proving that the conditions in Lemmas~\ref{lemma:crude-bound-tool-2}
and~\ref{lemma:crude-bound-tool-1} hold with high probability. In
view of Lemma~\ref{lemma:inj-general} and Corollary~\ref{lemma:P_omega_star_P_T},
the conditions $\Vert\mathcal{P}_{T^{\star}}-p^{-1}\mathcal{P}_{T^{\star}}\mathcal{P}_{\Omega_{\mathsf{obs}}}\mathcal{P}_{T^{\star}}\Vert\leq1/2$
and $\Vert\mathcal{P}_{\Omega^{\star}}\mathcal{P}_{T^{\star}}\Vert^{2}\leq p/8$
hold with high probability, provided that $n^{2}p\gg\mu rn\log n$
and $\rho_{\mathsf{s}}\leq1/12$. In addition, Lemma \ref{lemma:inj-general}
ensures that $\Vert\mathcal{P}_{T^{\star}}-p^{-1}(1-\rho_{\mathsf{s}})^{-1}\mathcal{P}_{T^{\star}}\mathcal{P}_{\Omega_{\mathsf{obs}}\setminus\Omega^{\star}}\mathcal{P}_{T^{\star}}\Vert\leq1/2$
holds with high probability, with the proviso that $n^{2}p(1-\rho_{\mathsf{s}})\gg\mu rn\log n$,
which holds true under the assumptions $\rho_{\mathsf{s}}\leq1/12$
and $n^{2}p\gg\mu rn\log n$. Last but not least, the existence of
the dual certificate $\bm{W}$ obeying~(\ref{eq:dc}) is guaranteed
with high probability according to \cite[Section III.D]{chen2013low},
under the conditions $\rho_{\mathsf{s}}\ll1$ and $n^{2}p\gg\mu^{2}r^{2}n\log^{6}n$.\footnote{Note that \cite[Section III.D]{chen2013low} requires $n^{2}p\gg\max\{\mu,\mu_{2}\}rn\log^{6}n$
under an additional incoherence condition $\|\bm{U}^{\star}\bm{V}^{\star\top}\|_{\infty}\leq\sqrt{\mu_{2}r/n^{2}}$.
While we do not impose this extra condition, it is easily seen that
$\|\bm{U}^{\star}\bm{V}^{\star\top}\|_{\infty}\leq\|\bm{U}^{\star}\|_{2,\infty}\|\bm{V}^{\star}\|_{2,\infty}\leq\mu r/n$
and hence $\mu_{2}\leq\mu^{2}r$.}

\subsection{Proof of Lemma~\ref{lemma:crude-bound-tool-2} \label{subsec:Proof-of-Lemma-crude-bound-tool-2}}

Expand $\|\mathcal{P}_{\Omega_{\mathsf{obs}}}(\mathcal{P}_{T^{\star}}(\bm{A})+\mathcal{P}_{\Omega^{\star}}(\bm{B}))\|_{\mathrm{F}}^{2}$
to obtain 
\begin{align*}
\left\Vert \mathcal{P}_{\Omega_{\mathsf{obs}}}\left[\mathcal{P}_{T^{\star}}\left(\bm{A}\right)+\mathcal{P}_{\Omega^{\star}}\left(\bm{B}\right)\right]\right\Vert _{\mathrm{F}}^{2} & =\left\Vert \mathcal{P}_{\Omega_{\mathsf{obs}}}\mathcal{P}_{T^{\star}}\left(\bm{A}\right)\right\Vert _{\mathrm{F}}^{2}+\left\Vert \mathcal{P}_{\Omega^{\star}}\left(\bm{B}\right)\right\Vert _{\mathrm{F}}^{2}+2\left\langle \mathcal{P}_{\Omega_{\mathsf{obs}}}\mathcal{P}_{T^{\star}}\left(\bm{A}\right),\mathcal{P}_{\Omega^{\star}}\left(\bm{B}\right)\right\rangle \\
 & \geq\frac{p}{2}\left\Vert \mathcal{P}_{T^{\star}}\left(\bm{A}\right)\right\Vert _{\mathrm{F}}^{2}+\left\Vert \mathcal{P}_{\Omega^{\star}}\left(\bm{B}\right)\right\Vert _{\mathrm{F}}^{2}+2\left\langle \mathcal{P}_{\Omega_{\mathsf{obs}}}\mathcal{P}_{T^{\star}}\left(\bm{A}\right),\mathcal{P}_{\Omega^{\star}}\left(\bm{B}\right)\right\rangle .
\end{align*}
Here, the equality uses the fact $\Omega^{\star}\subseteq\Omega_{\mathsf{obs}}$,
and the inequality holds because of the assumption $\Vert\mathcal{P}_{T^{\star}}-p^{-1}\mathcal{P}_{T^{\star}}\mathcal{P}_{\Omega_{\mathsf{obs}}}\mathcal{P}_{T^{\star}}\Vert\leq1/2$
and Fact~\ref{fact:near-isometry}. Use $\Omega^{\star}\subseteq\Omega_{\mathsf{obs}}$
once again to obtain 
\begin{align*}
2\left\langle \mathcal{P}_{\Omega_{\mathsf{obs}}}\mathcal{P}_{T^{\star}}\left(\bm{A}\right),\mathcal{P}_{\Omega^{\star}}\left(\bm{B}\right)\right\rangle  & =2\left\langle \mathcal{P}_{\Omega^{\star}}\mathcal{P}_{T^{\star}}\mathcal{P}_{T^{\star}}\left(\bm{A}\right),\mathcal{P}_{\Omega^{\star}}\left(\bm{B}\right)\right\rangle \\
 & \geq-2\left\Vert \mathcal{P}_{\Omega^{\star}}\mathcal{P}_{T^{\star}}\right\Vert \left\Vert \mathcal{P}_{T^{\star}}\left(\bm{A}\right)\right\Vert _{\mathrm{F}}\left\Vert \mathcal{P}_{\Omega^{\star}}\left(\bm{B}\right)\right\Vert _{\mathrm{F}}\\
 & \geq-2\left\Vert \mathcal{P}_{\Omega^{\star}}\mathcal{P}_{T^{\star}}\right\Vert ^{2}\left\Vert \mathcal{P}_{T^{\star}}\left(\bm{A}\right)\right\Vert _{\mathrm{F}}^{2}-\frac{1}{2}\left\Vert \mathcal{P}_{\Omega^{\star}}\left(\bm{B}\right)\right\Vert _{\mathrm{F}}^{2}.
\end{align*}
Here, the last relation arises from the elementary inequality $ab\leq(a^{2}+b^{2})/2$
and the fact that $\left\Vert \mathcal{P}_{\Omega^{\star}}\mathcal{P}_{T^{\star}}\right\Vert \leq1$.
Combine the above two inequalities to obtain 
\begin{align*}
\left\Vert \mathcal{P}_{\Omega_{\mathsf{obs}}}\left[\mathcal{P}_{T^{\star}}\left(\bm{A}\right)+\mathcal{P}_{\Omega^{\star}}\left(\bm{B}\right)\right]\right\Vert _{\mathrm{F}}^{2} & \geq\left(\frac{p}{2}-2\left\Vert \mathcal{P}_{\Omega^{\star}}\mathcal{P}_{T^{\star}}\right\Vert ^{2}\right)\left\Vert \mathcal{P}_{T^{\star}}\left(\bm{A}\right)\right\Vert _{\mathrm{F}}^{2}+\frac{1}{2}\left\Vert \mathcal{P}_{\Omega^{\star}}\left(\bm{B}\right)\right\Vert _{\mathrm{F}}^{2}\\
 & \geq\frac{p}{4}\left\Vert \mathcal{P}_{T^{\star}}\left(\bm{A}\right)\right\Vert _{\mathrm{F}}^{2}+\frac{1}{2}\left\Vert \mathcal{P}_{\Omega^{\star}}\left(\bm{B}\right)\right\Vert _{\mathrm{F}}^{2}\\
 & \geq\frac{p}{4}\left(\left\Vert \mathcal{P}_{T^{\star}}\left(\bm{A}\right)\right\Vert _{\mathrm{F}}^{2}+\left\Vert \mathcal{P}_{\Omega^{\star}}\left(\bm{B}\right)\right\Vert _{\mathrm{F}}^{2}\right)
\end{align*}
as claimed, where we have used the assumption $\Vert\mathcal{P}_{\Omega^{\star}}\mathcal{P}_{T^{\star}}\Vert^{2}\leq p/8$
in the middle line and the fact $1/2\geq p/4$ in the last inequality.

\subsection{Proof of Lemma~\ref{lemma:crude-bound-tool-1} \label{subsec:Proof-of-Lemma-crude-bound-tool-1}}

In view of the convexity of the nuclear norm $\Vert\cdot\Vert_{\ast}$,
one has 
\begin{align*}
\left\Vert \bm{L}^{\star}+\bm{H}_{\bm{L}}\right\Vert _{\ast} & \geq\left\Vert \bm{L}^{\star}\right\Vert _{\ast}+\left\langle \bm{U}^{\star}\bm{V}^{\star\top}+\bm{G}_{1},\bm{H}_{\bm{L}}\right\rangle =\left\Vert \bm{L}^{\star}\right\Vert _{\ast}+\left\langle \bm{U}^{\star}\bm{V}^{\star\top},\bm{H}_{\bm{L}}\right\rangle +\left\Vert \mathcal{P}_{T^{\star\perp}}\left(\bm{H}_{\bm{L}}\right)\right\Vert _{\ast}.
\end{align*}
Here, $\bm{U}^{\star}\bm{V}^{\star\top}+\bm{G}_{1}$ is a sub-gradient
of $\|\cdot\|_{*}$ at $\bm{L}^{\star}$. The last identity holds
by choosing $\bm{G}_{1}$ such that $\langle\bm{G}_{1},\bm{H}_{\bm{L}}\rangle=\Vert\mathcal{P}_{T^{\star\perp}}(\bm{H}_{\bm{L}})\Vert_{\ast}$.
Similarly, using the assumption $\mathcal{P}_{\Omega_{\mathsf{obs}}}(\bm{H}_{\bm{L}})+\bm{H}_{\bm{S}}=\bm{0}$
and the convexity of the $\ell_{1}$ norm $\|\cdot\|_{1}$, we can
obtain 
\begin{align*}
\left\Vert \bm{S}^{\star}+\bm{H}_{\bm{S}}\right\Vert _{1} & =\left\Vert \bm{S}^{\star}-\mathcal{P}_{\Omega_{\mathsf{obs}}}\left(\bm{H}_{\bm{L}}\right)\right\Vert _{1}\\
 & \geq\left\Vert \bm{S}^{\star}\right\Vert _{1}-\left\langle \mathsf{sign}\left(\bm{S}^{\star}\right)+\bm{G}_{2},\mathcal{P}_{\Omega_{\mathsf{obs}}}\left(\bm{H}_{\bm{L}}\right)\right\rangle \\
 & \overset{\text{(i)}}{=}\left\Vert \bm{S}^{\star}\right\Vert _{1}-\left\langle \mathsf{sign}\left(\bm{S}^{\star}\right),\mathcal{P}_{\Omega_{\mathsf{obs}}}\left(\bm{H}_{\bm{L}}\right)\right\rangle +\left\Vert \mathcal{P}_{\Omega_{\mathsf{obs}}\setminus\Omega^{\star}}\left(\bm{H}_{\bm{L}}\right)\right\Vert _{1}\\
 & \overset{\text{(ii)}}{=}\left\Vert \bm{S}^{\star}\right\Vert _{1}-\left\langle \mathsf{sign}\left(\bm{S}^{\star}\right),\bm{H}_{\bm{L}}\right\rangle +\left\Vert \mathcal{P}_{\Omega_{\mathsf{obs}}\setminus\Omega^{\star}}\left(\bm{H}_{\bm{L}}\right)\right\Vert _{1},
\end{align*}
where $\mathsf{sign}(\bm{S}^{\star})\coloneqq[\mathsf{sign}(S_{ij}^{\star})]_{1\leq i,j\leq n}$,
and $\mathsf{sign}(\bm{S}^{\star})+\bm{G}_{2}$ is a sub-gradient
of $\|\cdot\|_{1}$ at $\bm{S}^{\star}$. The first equality (i) holds
by choosing $\bm{G}_{2}$ such that $-\langle\bm{G}_{2},\mathcal{P}_{\Omega_{\mathsf{obs}}}(\bm{H}_{\bm{L}})\rangle=\Vert\mathcal{P}_{\Omega_{\mathsf{obs}}\setminus\Omega^{\star}}(\bm{H}_{\bm{L}})\Vert_{1}$,
and the last relation (ii) arises since $\mathsf{sign}(\bm{S}^{\star})$
is supported on $\Omega^{\star}\subseteq\Omega_{\mathsf{obs}}$. Combine
the above two bounds to deduce that
\begin{align}
\Delta & \coloneqq\lambda\left\Vert \bm{L}^{\star}+\bm{H}_{\bm{L}}\right\Vert _{\ast}+\tau\left\Vert \bm{S}^{\star}+\bm{H}_{\bm{S}}\right\Vert _{1}-\lambda\left\Vert \bm{L}^{\star}\right\Vert _{*}-\tau\left\Vert \bm{S}^{\star}\right\Vert _{1}\nonumber \\
 & \geq\lambda\left\langle \bm{U}^{\star}\bm{V}^{\star\top},\bm{H}_{\bm{L}}\right\rangle +\lambda\left\Vert \mathcal{P}_{T^{\star\perp}}\left(\bm{H}_{\bm{L}}\right)\right\Vert _{\ast}-\tau\left\langle \mathsf{sgn}\left(\bm{S}^{\star}\right),\bm{H}_{\bm{L}}\right\rangle +\tau\left\Vert \mathcal{P}_{\Omega_{\mathsf{obs}}\setminus\Omega^{\star}}\left(\bm{H}_{\bm{L}}\right)\right\Vert _{1}\nonumber \\
 & =\left\langle \lambda\bm{U}^{\star}\bm{V}^{\star\top}-\tau\mathsf{sgn}\left(\bm{S}^{\star}\right),\bm{H}_{\bm{L}}\right\rangle +\lambda\left\Vert \mathcal{P}_{T^{\star\perp}}\left(\bm{H}_{\bm{L}}\right)\right\Vert _{\ast}+\tau\left\Vert \mathcal{P}_{\Omega_{\mathsf{obs}}\setminus\Omega^{\star}}\left(\bm{H}_{\bm{L}}\right)\right\Vert _{1}\nonumber \\
 & =\underbrace{\left\langle \lambda\bm{U}^{\star}\bm{V}^{\star\top}-\tau\mathsf{sgn}\left(\bm{S}^{\star}\right)-\lambda\bm{W},\bm{H}_{\bm{L}}\right\rangle }_{\eqqcolon\theta_{1}}+\underbrace{\left\langle \lambda\bm{W},\bm{H}_{\bm{L}}\right\rangle }_{\eqqcolon\theta_{2}}+\lambda\left\Vert \mathcal{P}_{T^{\star\perp}}\left(\bm{H}_{\bm{L}}\right)\right\Vert _{\ast}+\tau\left\Vert \mathcal{P}_{\Omega_{\mathsf{obs}}\setminus\Omega^{\star}}\left(\bm{H}_{\bm{L}}\right)\right\Vert _{1},\label{eq:crude-proof-1}
\end{align}
where $\bm{W}\in\mathbb{R}^{n\times n}$ is the dual certificate stated
in Lemma~\ref{lemma:crude-bound-tool-1}.

In what follows, we shall lower bound the right-hand side of~(\ref{eq:crude-proof-1}).
To begin with, for $\theta_{1}$ we have 
\begin{align*}
\theta_{1} & =\left\langle \mathcal{P}_{T^{\star}}\left[\lambda\bm{U}^{\star}\bm{V}^{\star\top}-\tau\mathsf{sign}\left(\bm{S}^{\star}\right)-\lambda\bm{W}\right],\mathcal{P}_{T^{\star}}\left(\bm{H}_{\bm{L}}\right)\right\rangle +\left\langle \mathcal{P}_{T^{\star\perp}}\left[\lambda\bm{U}^{\star}\bm{V}^{\star\top}-\tau\mathsf{sign}\left(\bm{S}^{\star}\right)-\lambda\bm{W}\right],\mathcal{P}_{T^{\star\perp}}\left(\bm{H}_{\bm{L}}\right)\right\rangle \\
 & =\left\langle \mathcal{P}_{T^{\star}}\left[\lambda\bm{U}^{\star}\bm{V}^{\star\top}-\tau\mathsf{sign}\left(\bm{S}^{\star}\right)-\lambda\bm{W}\right],\mathcal{P}_{T^{\star}}\left(\bm{H}_{\bm{L}}\right)\right\rangle -\left\langle \mathcal{P}_{T^{\star\perp}}\left[\tau\mathsf{sign}\left(\bm{S}^{\star}\right)+\lambda\bm{W}\right],\mathcal{P}_{T^{\star\perp}}\left(\bm{H}_{\bm{L}}\right)\right\rangle \\
 & \geq-\left\Vert \mathcal{P}_{T^{\star}}\left[\lambda\bm{U}^{\star}\bm{V}^{\star\top}-\tau\mathsf{sign}\left(\bm{S}^{\star}\right)-\lambda\bm{W}\right]\right\Vert _{\mathrm{F}}\left\Vert \mathcal{P}_{T^{\star}}\left(\bm{H}_{\bm{L}}\right)\right\Vert _{\mathrm{F}}-\left\Vert \mathcal{P}_{T^{\star\perp}}\left[\tau\mathsf{sign}\left(\bm{S}^{\star}\right)+\lambda\bm{W}\right]\right\Vert \left\Vert \mathcal{P}_{T^{\star\perp}}\left(\bm{H}_{\bm{L}}\right)\right\Vert _{\ast}\\
 & \geq-\frac{\tau}{\sqrt{n}}\left\Vert \mathcal{P}_{T^{\star}}\left(\bm{H}_{\bm{L}}\right)\right\Vert _{\mathrm{F}}-\frac{\lambda}{2}\left\Vert \mathcal{P}_{T^{\star\perp}}\left(\bm{H}_{\bm{L}}\right)\right\Vert _{\ast}.
\end{align*}
Here, the penultimate line uses the fact $\bm{U}^{\star}\bm{V}^{\star\top}\in T^{\star}$
and the elementary inequalities $\vert\langle\bm{A},\bm{B}\rangle\vert\leq\Vert\bm{A}\Vert_{\mathrm{F}}\Vert\bm{B}\Vert_{\mathrm{F}}$
and $\vert\langle\bm{A},\bm{B}\rangle\vert\leq\Vert\bm{A}\Vert\Vert\bm{B}\Vert_{\ast}$,
whereas the last inequality relies on the properties of the dual certificate
$\bm{W}$, namely, (\ref{eq:dc-1}) and~(\ref{eq:dc-2}). Moving
on to $\theta_{2}$, one has 
\begin{align*}
\theta_{2} & =\left\langle \lambda\mathcal{P}_{\Omega_{\mathsf{obs}}\setminus\Omega^{\star}}\left(\bm{W}\right),\bm{H}_{\bm{L}}\right\rangle +\left\langle \lambda\mathcal{P}_{(\Omega_{\mathsf{obs}}\setminus\Omega^{\star})^{\mathrm{c}}}\left(\bm{W}\right),\bm{H}_{\bm{L}}\right\rangle \\
 & \overset{\text{}}{=}\left\langle \lambda\bm{W},\mathcal{P}_{\Omega_{\mathsf{obs}}\setminus\Omega^{\star}}\left(\bm{H}_{\bm{L}}\right)\right\rangle \overset{\text{(i)}}{\geq}-\left\Vert \lambda\bm{W}\right\Vert _{\infty}\left\Vert \mathcal{P}_{\Omega_{\mathsf{obs}}\setminus\Omega^{\star}}\left(\bm{H}_{\bm{L}}\right)\right\Vert _{1}\overset{\text{(ii)}}{\geq}-\frac{\tau}{2}\left\Vert \mathcal{P}_{\Omega_{\mathsf{obs}}\setminus\Omega^{\star}}\left(\bm{H}_{\bm{L}}\right)\right\Vert _{1}.
\end{align*}
Here, the second identity uses the assumption~(\ref{eq:dc-3}), the
first inequality (i) uses the elementary inequality $\vert\langle\bm{A},\bm{B}\rangle\vert\leq\Vert\bm{A}\Vert_{\infty}\Vert\bm{B}\Vert_{1}$,
and the last relation (ii) holds because of the assumption~(\ref{eq:dc-4}).
Substituting the above two bounds back into (\ref{eq:crude-proof-1})
gives 
\begin{equation}
\Delta\geq-\frac{\tau}{\sqrt{n}}\left\Vert \mathcal{P}_{T^{\star}}\left(\bm{H}_{\bm{L}}\right)\right\Vert _{\mathrm{F}}+\frac{\lambda}{2}\left\Vert \mathcal{P}_{T^{\star\perp}}\left(\bm{H}_{\bm{L}}\right)\right\Vert _{\ast}+\frac{\tau}{2}\left\Vert \mathcal{P}_{\Omega_{\mathsf{obs}}\setminus\Omega^{\star}}\left(\bm{H}_{\bm{L}}\right)\right\Vert _{1}.\label{eq:crude-proof-2}
\end{equation}
Continuing the lower bound, we have 
\begin{align*}
\left\Vert \mathcal{P}_{\Omega_{\mathsf{obs}}\setminus\Omega^{\star}}\left(\bm{H}_{\bm{L}}\right)\right\Vert _{1} & \overset{\text{(i)}}{\geq}\left\Vert \mathcal{P}_{\Omega_{\mathsf{obs}}\setminus\Omega^{\star}}\left(\bm{H}_{\bm{L}}\right)\right\Vert _{\mathrm{F}}=\left\Vert \mathcal{P}_{\Omega_{\mathsf{obs}}\setminus\Omega^{\star}}\mathcal{P}_{T^{\star}}\left(\bm{H}_{\bm{L}}\right)+\mathcal{P}_{\Omega_{\mathsf{obs}}\setminus\Omega^{\star}}\mathcal{P}_{T^{\star\perp}}\left(\bm{H}_{\bm{L}}\right)\right\Vert _{\mathrm{F}}\\
 & \overset{\text{(ii)}}{\geq}\left\Vert \mathcal{P}_{\Omega_{\mathsf{obs}}\setminus\Omega^{\star}}\mathcal{P}_{T^{\star}}\left(\bm{H}_{\bm{L}}\right)\right\Vert _{\mathrm{F}}-\left\Vert \mathcal{P}_{\Omega_{\mathsf{obs}}\setminus\Omega^{\star}}\mathcal{P}_{T^{\star\perp}}\left(\bm{H}_{\bm{L}}\right)\right\Vert _{\mathrm{F}}\\
 & \geq\left\Vert \mathcal{P}_{\Omega_{\mathsf{obs}}\setminus\Omega^{\star}}\mathcal{P}_{T^{\star}}\left(\bm{H}_{\bm{L}}\right)\right\Vert _{\mathrm{F}}-\left\Vert \mathcal{P}_{T^{\star\perp}}\left(\bm{H}_{\bm{L}}\right)\right\Vert _{\mathrm{F}},
\end{align*}
where (i) holds because $\Vert\bm{A}\Vert_{\mathrm{1}}\geq\Vert\bm{A}\Vert_{\mathrm{F}}$
for any matrix $\bm{A}$, and (ii) arises from the triangle inequality.
Putting the above relation and (\ref{eq:crude-proof-2}) together
results in 
\begin{align}
\Delta & \geq-\frac{\tau}{\sqrt{n}}\left\Vert \mathcal{P}_{T^{\star}}\left(\bm{H}_{\bm{L}}\right)\right\Vert _{\mathrm{F}}+\left(\frac{\lambda}{2}-\frac{\tau}{4}\right)\left\Vert \mathcal{P}_{T^{\star\perp}}\left(\bm{H}_{\bm{L}}\right)\right\Vert _{\ast}+\frac{\tau}{4}\left\Vert \mathcal{P}_{\Omega_{\mathsf{obs}}\setminus\Omega^{\star}}\mathcal{P}_{T^{\star}}\left(\bm{H}_{\bm{L}}\right)\right\Vert _{\mathrm{F}}+\frac{\tau}{4}\left\Vert \mathcal{P}_{\Omega_{\mathsf{obs}}\setminus\Omega^{\star}}\left(\bm{H}_{\bm{L}}\right)\right\Vert _{1}\nonumber \\
 & \geq\frac{\tau}{4}\left\Vert \mathcal{P}_{\Omega_{\mathsf{obs}}\setminus\Omega^{\star}}\mathcal{P}_{T^{\star}}\left(\bm{H}_{\bm{L}}\right)\right\Vert _{\mathrm{F}}-\frac{\tau}{\sqrt{n}}\left\Vert \mathcal{P}_{T^{\star}}\left(\bm{H}_{\bm{L}}\right)\right\Vert _{\mathrm{F}}+\frac{\lambda}{4}\left\Vert \mathcal{P}_{T^{\star\perp}}\left(\bm{H}_{\bm{L}}\right)\right\Vert _{\ast}+\frac{\tau}{4}\left\Vert \mathcal{P}_{\Omega_{\mathsf{obs}}\setminus\Omega^{\star}}\left(\bm{H}_{\bm{L}}\right)\right\Vert _{1},\label{eq:crude-proof-4}
\end{align}
where the last line holds since $\lambda=\tau\sqrt{np/\log n}\geq\tau$
(as long as $np\geq\log n$). Everything then boils down to lower
bounding $\|\mathcal{P}_{\Omega_{\mathsf{obs}}\setminus\Omega^{\star}}\mathcal{P}_{T^{\star}}(\bm{H}_{\bm{L}})\|_{\mathrm{F}}$.
To this end, one can use the assumption $\Vert\mathcal{P}_{T^{\star}}-p^{-1}(1-\rho_{\mathsf{s}})^{-1}\mathcal{P}_{T^{\star}}\mathcal{P}_{\Omega_{\mathsf{obs}}\setminus\Omega^{\star}}\mathcal{P}_{T^{\star}}\Vert\leq1/2$
and Fact~\ref{fact:near-isometry} to obtain 
\begin{align}
\left\Vert \mathcal{P}_{\Omega_{\mathsf{obs}}\setminus\Omega^{\star}}\mathcal{P}_{T^{\star}}\left(\bm{H}_{\bm{L}}\right)\right\Vert _{\mathrm{F}}^{2} & \geq\frac{1}{2}p\left(1-\rho_{\mathsf{s}}\right)\left\Vert \mathcal{P}_{T^{\star}}\left(\bm{H}_{\bm{L}}\right)\right\Vert _{\mathrm{F}}^{2}.\label{eq:crude-proof-3}
\end{align}
Take~(\ref{eq:crude-proof-4}) and~(\ref{eq:crude-proof-3}) collectively
to yield 
\begin{align*}
\Delta & \geq\left(\frac{\tau}{4}\sqrt{\frac{1}{2}p\left(1-\rho_{\mathsf{s}}\right)}-\frac{\tau}{\sqrt{n}}\right)\left\Vert \mathcal{P}_{T^{\star}}\left(\bm{H}_{\bm{L}}\right)\right\Vert _{\mathrm{F}}+\frac{\lambda}{4}\left\Vert \mathcal{P}_{T^{\star\perp}}\left(\bm{H}_{\bm{L}}\right)\right\Vert _{\ast}+\frac{\tau}{4}\left\Vert \mathcal{P}_{\Omega_{\mathsf{obs}}\setminus\Omega^{\star}}\left(\bm{H}_{\bm{L}}\right)\right\Vert _{1}\\
 & \geq\frac{\lambda}{4}\left\Vert \mathcal{P}_{T^{\star\perp}}\left(\bm{H}_{\bm{L}}\right)\right\Vert _{\ast}+\frac{\tau}{4}\left\Vert \mathcal{P}_{\Omega_{\mathsf{obs}}\setminus\Omega^{\star}}\left(\bm{H}_{\bm{L}}\right)\right\Vert _{1},
\end{align*}
where the last relation is guaranteed by $np\gg1$ and $\rho_{\mathsf{s}}\ll1$.
Recognizing that $\mathcal{P}_{\Omega_{\mathsf{obs}}\setminus\Omega^{\star}}(\bm{H}_{\bm{L}})=-\mathcal{P}_{\Omega_{\mathsf{obs}}\setminus\Omega^{\star}}(\bm{H}_{\bm{S}})$
finishes the proof.

\section{Equivalence between convex and nonconvex solutions (Proof of Theorem
\ref{thm:cvx-ncvx-equivalence})\label{sec:proof-Equivalence-cvx-ncvx}}

The goal of this section is to establish the intimate connection between
the convex and nonconvex solutions~(cf.~Theorem~\ref{thm:cvx-ncvx-equivalence}).
Before continuing, we remind the readers of the following notations: 
\begin{itemize}
\item $\bm{X}\bm{Y}^{\top}=\bm{U}\bm{\Sigma}\bm{V}^{\top}$: the rank-$r$
singular value decomposition of $\bm{X}\bm{Y}^{\top}$;
\item $T$: the tangent space of the set of rank-$r$ matrices at the estimate
$\bm{X}\bm{Y}^{\top}$.
\end{itemize}
In addition, we define 
\begin{equation}
\bm{\Delta}_{\bm{L}}\coloneqq\bm{L}_{\mathsf{cvx}}-\bm{X}\bm{Y}^{\top},\qquad\bm{\Delta}_{\bm{S}}\coloneqq\bm{S}_{\mathsf{cvx}}-\bm{S},\label{eq:defn-DeltaL-DeltaS}
\end{equation}
and denote the support of $\bm{S}$ by
\begin{equation}
\Omega\coloneqq\{(i,j)\,|\,S_{ij}\neq0\}.\label{eq:defn-Omega-S}
\end{equation}

\subsection{Preliminary facts}

We begin with two useful lemmas which demonstrate that the point $(\bm{X}\bm{Y}^{\top},\bm{S})$
described in Theorem~\ref{thm:cvx-ncvx-equivalence} satisfies approximate
optimality conditions w.r.t.~the convex program (\ref{eq:cvx}).

\begin{lemma}\label{lemma:R1}Instate the assumptions in Theorem~\ref{thm:cvx-ncvx-equivalence}.
The triple $(\bm{X},\bm{Y},\bm{S})$ as stated in Theorem \ref{thm:cvx-ncvx-equivalence}
satisfies 
\begin{equation}
\frac{1}{\lambda}\mathcal{P}_{\Omega_{\mathsf{obs}}}\left(\bm{X}\bm{Y}^{\top}+\bm{S}-\bm{M}\right)=-\bm{U}\bm{V}^{\top}+\bm{R}_{1}\label{eq:def-R1}
\end{equation}
for some matrix $\bm{R}_{1}\in\mathbb{R}^{n\times n}$ obeying 
\begin{equation}
\left\Vert \mathcal{P}_{T}\left(\bm{R}_{1}\right)\right\Vert _{\mathrm{F}}\lesssim\frac{\kappa p\left\Vert \nabla f\left(\bm{X},\bm{Y};\bm{S}\right)\right\Vert _{\mathrm{F}}}{\lambda\sqrt{\sigma_{\min}}}\lesssim\frac{1}{n^{19}}\qquad\text{and}\qquad\left\Vert \mathcal{P}_{T^{\perp}}\left(\bm{R}_{1}\right)\right\Vert \leq\frac{1}{2}.\label{eq:R1-conditions}
\end{equation}
\end{lemma}\begin{proof}The proof can be straightforwardly adapted
from \cite[Claim 2]{chen2019noisy} by replacing $\bm{E}$ therein
with $\bm{E}+\bm{S}^{\star}-\bm{S}$. We omit it for the sake of brevity.
\end{proof}

\begin{lemma}\label{lemma:R2}The point $(\bm{X}\bm{Y}^{\top},\bm{S})$
as stated in Theorem \ref{thm:cvx-ncvx-equivalence} obeys 
\begin{equation}
\frac{1}{\tau}\mathcal{P}_{\Omega_{\mathsf{obs}}}\left(\bm{X}\bm{Y}^{\top}+\bm{S}-\bm{M}\right)=-\mathsf{sign}\left(\bm{S}\right)+\bm{R}_{2}\label{eq:def-R2}
\end{equation}
for some matrix $\bm{R}_{2}\in\mathbb{R}^{n\times n}$, where $\bm{R}_{2}$
satisfies 
\begin{equation}
\mathcal{P}_{\Omega}\left(\bm{R}_{2}\right)=\bm{0}\qquad\text{and}\qquad\left\Vert \mathcal{P}_{\Omega^{\mathrm{c}}}(\bm{R}_{2})\right\Vert _{\infty}\leq1\label{eq:R2}
\end{equation}
with $\Omega$ defined in (\ref{eq:defn-Omega-S}). \end{lemma}\begin{proof}By
definition, one has $\bm{S}=\mathcal{P}_{\Omega_{\mathsf{obs}}}[\mathcal{S}_{\tau}(\bm{M}-\bm{X}\bm{Y}^{\top})]$.
Clearly, this is equivalent to saying that $\bm{S}$ is the unique
minimizer of the following convex program 
\begin{equation}
\bm{S}=\underset{\widehat{\bm{S}}\in\mathbb{R}^{n\times n}}{\arg\min}\:\,\frac{1}{2}\Bigl\Vert\mathcal{P}_{\Omega_{\mathsf{obs}}}\big(\bm{X}\bm{Y}^{\top}+\widehat{\bm{S}}-\bm{M}\big)\Bigr\Vert_{\mathrm{F}}^{2}+\frac{\lambda}{2}\left\Vert \bm{X}\right\Vert _{\mathrm{F}}^{2}+\frac{\lambda}{2}\left\Vert \bm{Y}\right\Vert _{\mathrm{F}}^{2}+\tau\big\|\widehat{\bm{S}}\big\|_{1}.\label{eq:S-cvx-problem}
\end{equation}
The claim of this lemma then follows from the optimality condition
of this convex program~(\ref{eq:S-cvx-problem}). \end{proof}

Additionally, in view of the crude error bound (\ref{eq:thm-cvx-ncvx-equivalence-cond-2})
and Condition~\ref{assumption:link-cvx-ncvx-noise}, the matrix $\bm{\Delta}_{\bm{L}}$
(cf.~(\ref{eq:defn-DeltaL-DeltaS})) obeys
\begin{align}
\left\Vert \bm{\Delta}_{\bm{L}}\right\Vert _{\mathrm{F}} & =\left\Vert \bm{L}_{\mathsf{cvx}}-\bm{X}\bm{Y}^{\top}\right\Vert _{\mathrm{F}}\leq\left\Vert \bm{L}_{\mathsf{cvx}}-\bm{L}^{\star}\right\Vert _{\mathrm{F}}+\left\Vert \bm{X}\bm{Y}^{\top}-\bm{L}^{\star}\right\Vert _{\mathrm{F}}\leq\left\Vert \bm{L}_{\mathsf{cvx}}-\bm{L}^{\star}\right\Vert _{\mathrm{F}}+n\left\Vert \bm{X}\bm{Y}^{\top}-\bm{L}^{\star}\right\Vert _{\infty}\nonumber \\
 & \lesssim\sigma n^{4}+n\tau\asymp\sigma n^{4},\label{eq:new-crude-bound}
\end{align}
where we use the the elementary inequality $\|\bm{A}\|_{\mathrm{F}}\leq n\|\bm{A}\|_{\infty}$
and the fact that $\tau\asymp\sigma\sqrt{\log n}$.

\subsection{Proof of Theorem \ref{thm:cvx-ncvx-equivalence}}

We now present the proof of Theorem~\ref{thm:cvx-ncvx-equivalence},
which consists of three main steps:
\begin{enumerate}
\item Showing that $(\bm{X}\bm{Y}^{\top},\bm{S})$ is not far from $(\bm{L}_{\mathsf{cvx}},\bm{S}_{\mathsf{cvx}})$
over $\Omega_{\mathsf{obs}}$, in the sense that $\mathcal{P}_{\Omega_{\mathsf{obs}}}(\bm{\Delta}_{\bm{L}}+\bm{\Delta}_{\bm{S}})\approx\bm{0}$;
\item Showing that $\bm{\Delta}_{\bm{L}}$ (resp.~$\bm{\Delta}_{\bm{S}}$)
is extremely small outside the tangent space $T$ (resp.~the support
$\Omega^{\star}$), and hence most of the energy of $\bm{\Delta}_{\bm{L}}$
(resp.~$\bm{\Delta}_{\bm{S}}$) --- if it is not vanishingly small
--- has to reside within $T$ (resp.~$\Omega^{\star}$);
\item Showing that $\bm{\Delta}_{\bm{S}}\approx\bm{0}$ and $\bm{\Delta}_{\bm{L}}\approx\bm{0}$,
with the assistance of the preceding two steps.
\end{enumerate}
In what follows, we shall detail each of these steps.

\subsubsection{Step 1: showing that $\mathcal{P}_{\Omega_{\mathsf{obs}}}(\bm{\Delta}_{\bm{L}}+\bm{\Delta}_{\bm{S}})\approx\bm{0}$}

Since $(\bm{L}_{\mathsf{cvx}},\bm{S}_{\mathsf{cvx}})$ is the minimizer
of the convex program (\ref{eq:cvx}), we have 
\begin{align*}
 & \tfrac{1}{2}\left\Vert \mathcal{P}_{\Omega_{\mathsf{obs}}}\left(\bm{L}_{\mathsf{cvx}}+\bm{S}_{\mathsf{cvx}}-\bm{M}\right)\right\Vert _{\mathrm{F}}^{2}+\lambda\left\Vert \bm{L}_{\mathsf{cvx}}\right\Vert _{\ast}+\tau\left\Vert \bm{S}_{\mathsf{cvx}}\right\Vert _{1}\\
 & \quad=\tfrac{1}{2}\left\Vert \mathcal{P}_{\Omega_{\mathsf{obs}}}\left(\bm{X}\bm{Y}^{\top}+\bm{\Delta}_{\bm{L}}+\bm{S}+\bm{\Delta}_{\bm{S}}-\bm{M}\right)\right\Vert _{\mathrm{F}}^{2}+\lambda\left\Vert \bm{X}\bm{Y}^{\top}+\bm{\Delta}_{\bm{L}}\right\Vert _{\ast}+\tau\left\Vert \bm{S}+\bm{\Delta}_{\bm{S}}\right\Vert _{1}\\
\text{} & \quad\leq\tfrac{1}{2}\left\Vert \mathcal{P}_{\Omega_{\mathsf{obs}}}\left(\bm{X}\bm{Y}^{\top}+\bm{S}-\bm{M}\right)\right\Vert _{\mathrm{F}}^{2}+\lambda\left\Vert \bm{X}\bm{Y}^{\top}\right\Vert _{\ast}+\tau\left\Vert \bm{S}\right\Vert _{1}.
\end{align*}
Here, the equality arises from the relations $\bm{L}_{\mathsf{cvx}}=\bm{X}\bm{Y}^{\top}+\bm{\Delta}_{\bm{L}}$
and $\bm{S}_{\mathsf{cvx}}=\bm{S}+\bm{\Delta}_{\bm{S}}$. Expanding
the squares and rearranging terms, we arrive at 
\begin{align}
\tfrac{1}{2}\left\Vert \mathcal{P}_{\Omega_{\mathsf{obs}}}\left(\bm{\Delta}_{\bm{L}}+\bm{\Delta}_{\bm{S}}\right)\right\Vert _{\mathrm{F}}^{2} & \leq-\left\langle \mathcal{P}_{\Omega_{\mathsf{obs}}}\left(\bm{X}\bm{Y}^{\top}+\bm{S}-\bm{M}\right),\bm{\Delta}_{\bm{L}}+\bm{\Delta}_{\bm{S}}\right\rangle +\lambda\left\Vert \bm{X}\bm{Y}^{\top}\right\Vert _{\ast}-\lambda\left\Vert \bm{X}\bm{Y}^{\top}+\bm{\Delta}_{\bm{L}}\right\Vert _{\ast}\nonumber \\
 & \quad\qquad+\tau\left\Vert \bm{S}\right\Vert _{1}-\tau\left\Vert \bm{S}+\bm{\Delta}_{\bm{S}}\right\Vert _{1}.\label{eq:DeltaL-DeltaS-UB1}
\end{align}
In view of the convexity of the nuclear norm $\|\cdot\|_{*}$ and
the $\ell_{1}$ norm $\|\cdot\|_{1}$, one has \begin{subequations}
\begin{align}
\left\Vert \bm{X}\bm{Y}^{\top}+\bm{\Delta}_{\bm{L}}\right\Vert _{\ast} & \geq\left\Vert \bm{X}\bm{Y}^{\top}\right\Vert _{\ast}+\big\langle\bm{U}\bm{V}^{\top}+\bm{W},\bm{\Delta}_{\bm{L}}\big\rangle\overset{(\text{i})}{=}\left\Vert \bm{X}\bm{Y}^{\top}\right\Vert _{\ast}+\big\langle\bm{U}\bm{V}^{\top},\bm{\Delta}_{\bm{L}}\big\rangle+\left\Vert \mathcal{P}_{T^{\perp}}(\bm{\Delta}_{\bm{L}})\right\Vert _{\ast};\label{eq:nuclear-convexity}\\
\left\Vert \bm{S}+\bm{\Delta}_{\bm{S}}\right\Vert _{1} & \geq\left\Vert \bm{S}\right\Vert _{1}+\big\langle\mathsf{sign}\left(\bm{S}\right)+\bm{G},\bm{\Delta}_{\bm{S}}\big\rangle\overset{(\text{ii})}{=}\left\Vert \bm{S}\right\Vert _{1}+\big\langle\mathsf{sign}\left(\bm{S}\right),\bm{\Delta}_{\bm{S}}\big\rangle+\left\Vert \mathcal{P}_{\Omega^{\mathrm{c}}}(\bm{\Delta}_{\bm{S}})\right\Vert _{1}.\label{eq:L1-convexity}
\end{align}
\end{subequations}Here, $\bm{U}\bm{V}^{\top}+\bm{W}$ is a sub-gradient
of $\|\cdot\|_{*}$ at $\bm{X}\bm{Y}^{\top}$. The identity (i) holds
by choosing $\bm{W}$ such that $\langle\bm{W},\bm{\Delta}_{\bm{L}}\rangle=\Vert\mathcal{P}_{T^{\perp}}(\bm{\Delta}_{\bm{L}})\Vert_{\ast}$.
Similarly, $\mathsf{sign}(\bm{S})+\bm{G}$ is a sub-gradient of $\|\cdot\|_{1}$
at $\bm{S}$ and one can choose $\bm{G}$ obeying $\langle\bm{G},\bm{\Delta}_{\bm{S}}\rangle=\Vert\mathcal{P}_{\Omega^{\mathrm{c}}}(\bm{\Delta}_{\bm{S}})\Vert_{1}$
to make (ii) valid. These taken together with (\ref{eq:DeltaL-DeltaS-UB1})
lead to 
\begin{align*}
\tfrac{1}{2}\left\Vert \mathcal{P}_{\Omega_{\mathsf{obs}}}\left(\bm{\Delta}_{\bm{L}}+\bm{\Delta}_{\bm{S}}\right)\right\Vert _{\mathrm{F}}^{2} & \leq-\big\langle\mathcal{P}_{\Omega_{\mathsf{obs}}}\left(\bm{X}\bm{Y}^{\top}+\bm{S}-\bm{M}\right),\bm{\Delta}_{\bm{L}}+\bm{\Delta}_{\bm{S}}\big\rangle-\lambda\big\langle\bm{U}\bm{V}^{\top},\bm{\Delta}_{\bm{L}}\big\rangle-\lambda\left\Vert \mathcal{P}_{T^{\perp}}\left(\bm{\Delta}_{\bm{L}}\right)\right\Vert _{\ast}\\
 & \quad-\tau\big\langle\mathsf{sign}\left(\bm{S}\right),\bm{\Delta}_{\bm{S}}\big\rangle-\tau\left\Vert \mathcal{P}_{\Omega^{\mathrm{c}}}\left(\bm{\Delta}_{\bm{S}}\right)\right\Vert _{1}.
\end{align*}
Recall the definitions of $\bm{R}_{1}$ and $\bm{R}_{2}$ from Lemmas
\ref{lemma:R1} and \ref{lemma:R2}. We can then simplify the above
inequality as 
\begin{equation}
\tfrac{1}{2}\left\Vert \mathcal{P}_{\Omega_{\mathsf{obs}}}\left(\bm{\Delta}_{\bm{L}}+\bm{\Delta}_{\bm{S}}\right)\right\Vert _{\mathrm{F}}^{2}\leq\underbrace{-\lambda\left\langle \bm{R}_{1},\bm{\Delta}_{\bm{L}}\right\rangle -\lambda\left\Vert \mathcal{P}_{T^{\perp}}(\bm{\Delta}_{\bm{L}})\right\Vert _{\ast}}_{\eqqcolon\theta_{1}}\underbrace{-\tau\left\langle \bm{R}_{2},\bm{\Delta}_{\bm{S}}\right\rangle -\tau\left\Vert \mathcal{P}_{\Omega^{\mathrm{c}}}(\bm{\Delta}_{\bm{S}})\right\Vert _{1}}_{\eqqcolon\theta_{2}}.\label{eq:cvx-ncvx-no1}
\end{equation}
In the sequel, we develop bounds on $\theta_{1}$ and $\theta_{2}$.
\begin{enumerate}
\item With regards to $\theta_{1}$, one can further decompose it into 
\begin{align}
\theta_{1} & =-\lambda\left\langle \bm{R}_{1},\mathcal{P}_{T}\left(\bm{\Delta}_{\bm{L}}\right)\right\rangle -\lambda\left\langle \bm{R}_{1},\mathcal{P}_{T^{\perp}}\left(\bm{\Delta}_{\bm{L}}\right)\right\rangle -\lambda\left\Vert \mathcal{P}_{T^{\perp}}\left(\bm{\Delta}_{\bm{L}}\right)\right\Vert _{\ast}\nonumber \\
 & \leq\lambda\left\Vert \mathcal{P}_{T}\left(\bm{R}_{1}\right)\right\Vert _{\mathrm{F}}\left\Vert \mathcal{P}_{T}\left(\bm{\Delta}_{\bm{L}}\right)\right\Vert _{\mathrm{F}}-\lambda\left(1-\left\Vert \mathcal{P}_{T^{\perp}}\left(\bm{R}_{1}\right)\right\Vert \right)\left\Vert \mathcal{P}_{T^{\perp}}\left(\bm{\Delta}_{\bm{L}}\right)\right\Vert _{\ast}\nonumber \\
 & \leq\lambda\left\Vert \mathcal{P}_{T}\left(\bm{R}_{1}\right)\right\Vert _{\mathrm{F}}\left\Vert \mathcal{P}_{T}\left(\bm{\Delta}_{\bm{L}}\right)\right\Vert _{\mathrm{F}}-\frac{\lambda}{2}\left\Vert \mathcal{P}_{T^{\perp}}\left(\bm{\Delta}_{\bm{L}}\right)\right\Vert _{\ast},\label{eq:Delta-L-1}
\end{align}
where the middle line arises from the elementary inequalities $|\langle\bm{A},\bm{B}\rangle|\leq\|\bm{A}\|_{\mathrm{F}}\|\bm{B}\|_{\mathrm{F}}$
and $|\langle\bm{A},\bm{B}\rangle|\leq\|\bm{A}\|\|\bm{B}\|_{\mathrm{*}}$,
and the last inequality holds since $\|\mathcal{P}_{T^{\perp}}(\bm{R}_{1})\|\leq1/2$
(see Lemma~\ref{lemma:R1}).
\item Similarly, one can decompose $\theta_{2}$ into 
\begin{align}
\theta_{2} & =-\tau\left\langle \bm{R}_{2},\mathcal{P}_{\Omega}\left(\bm{\Delta}_{\bm{S}}\right)\right\rangle -\tau\left\langle \bm{R}_{2},\mathcal{P}_{\Omega^{\mathrm{c}}}\left(\bm{\Delta}_{\bm{S}}\right)\right\rangle -\tau\left\Vert \mathcal{P}_{\Omega^{\mathrm{c}}}\left(\bm{\Delta}_{\bm{S}}\right)\right\Vert _{1}\nonumber \\
 & \leq\tau\left\langle \mathcal{P}_{\Omega}\left(\bm{R}_{2}\right),\mathcal{P}_{\Omega}\left(\bm{\Delta}_{\bm{S}}\right)\right\rangle -\tau\left(1-\left\Vert \mathcal{P}_{\Omega^{\mathrm{c}}}\left(\bm{R}_{2}\right)\right\Vert _{\infty}\right)\left\Vert \mathcal{P}_{\Omega^{\mathrm{c}}}\left(\bm{\Delta}_{\bm{S}}\right)\right\Vert _{1}\leq0.\label{eq:Delta-S-1}
\end{align}
Here, the first inequality comes from the facts that $|\langle\bm{A},\bm{B}\rangle|\leq\|\bm{A}\|_{\mathrm{\infty}}\|\bm{B}\|_{1}$
and $|\langle\bm{A},\bm{B}\rangle|\leq\|\bm{A}\|_{\mathrm{F}}\|\bm{B}\|_{\mathrm{F}}$,
and the second one utilizes the facts that $\mathcal{P}_{\Omega}(\bm{R}_{2})=\bm{0}$
and $\|\mathcal{P}_{\Omega^{\mathrm{c}}}(\bm{R}_{2})\|_{\infty}\le1$
(cf.~Lemma~\ref{lemma:R2}).
\end{enumerate}
Combining (\ref{eq:cvx-ncvx-no1}), (\ref{eq:Delta-L-1}) and (\ref{eq:Delta-S-1})
yields 
\begin{align}
\tfrac{1}{2}\left\Vert \mathcal{P}_{\Omega_{\mathsf{obs}}}\left(\bm{\Delta}_{\bm{L}}+\bm{\Delta}_{\bm{S}}\right)\right\Vert _{\mathrm{F}}^{2} & \leq\lambda\left\Vert \mathcal{P}_{T}(\bm{R}_{1})\right\Vert _{\mathrm{F}}\left\Vert \mathcal{P}_{T}(\bm{\Delta}_{\bm{L}})\right\Vert _{\mathrm{F}}-\frac{\lambda}{2}\left\Vert \mathcal{P}_{T^{\perp}}(\bm{\Delta}_{\bm{L}})\right\Vert _{\ast}\label{eq:UB10-Delta}\\
 & \lesssim\frac{\lambda}{n^{19}}\left\Vert \bm{\Delta}_{\bm{L}}\right\Vert _{\mathrm{F}}\lesssim\frac{\sigma\sqrt{np}}{n^{19}}\sigma n^{4}\lesssim\frac{\sigma^{2}}{n^{14.5}},\nonumber 
\end{align}
where we make use of the upper bound $\|\mathcal{P}_{T}(\bm{R}_{1})\|_{\mathrm{F}}\lesssim n^{-19}$
(cf.~Lemma~\ref{lemma:R1}), the choice $\lambda\asymp\sigma\sqrt{np}$
as well as the crude error bound $\|\bm{\Delta}_{\bm{L}}\|_{\mathrm{F}}\lesssim\sigma n^{4}$
(cf.~(\ref{eq:new-crude-bound})). Consequently, we have demonstrated
that $\mathcal{P}_{\Omega_{\mathsf{obs}}}(\bm{\Delta}_{\bm{L}}+\bm{\Delta}_{\bm{S}})\approx\bm{0}$
in the sense that 
\begin{equation}
\left\Vert \mathcal{P}_{\Omega_{\mathsf{obs}}}\left(\bm{\Delta}_{\bm{L}}+\bm{\Delta}_{\bm{S}}\right)\right\Vert _{\mathrm{F}}\lesssim\frac{\sigma}{n^{7.25}}\leq\frac{\sigma}{n^{7}}.\label{eq:Delta-L-Delta-S-sum-small}
\end{equation}

\subsubsection{Step 2: showing that $\mathcal{P}_{T^{\perp}}(\bm{\Delta}_{\bm{L}})\approx\bm{0}$
and $\mathcal{P}_{(\Omega^{\star})^{\mathrm{c}}}(\bm{\Delta}_{\bm{S}})\approx\bm{0}$}

We begin by demonstrating that $\mathcal{P}_{T^{\perp}}(\bm{\Delta}_{\bm{L}})\approx\bm{0}$.
From the inequality (\ref{eq:UB10-Delta}), we have 
\begin{align*}
\frac{1}{2}\left\Vert \mathcal{P}_{T^{\perp}}(\bm{\Delta}_{\bm{L}})\right\Vert _{\ast} & \leq\left\Vert \mathcal{P}_{T}(\bm{R}_{1})\right\Vert _{\mathrm{F}}\left\Vert \mathcal{P}_{T}(\bm{\Delta}_{\bm{L}})\right\Vert _{\mathrm{F}}-\frac{1}{2\lambda}\left\Vert \mathcal{P}_{\Omega_{\mathsf{obs}}}\left(\bm{\Delta}_{\bm{L}}+\bm{\Delta}_{\bm{S}}\right)\right\Vert _{\mathrm{F}}^{2}\\
 & \leq\left\Vert \mathcal{P}_{T}(\bm{R}_{1})\right\Vert _{\mathrm{F}}\left\Vert \mathcal{P}_{T}(\bm{\Delta}_{\bm{L}})\right\Vert _{\mathrm{F}}\lesssim\frac{1}{n^{19}}\left\Vert \bm{\Delta}_{\bm{L}}\right\Vert _{\mathrm{F}},
\end{align*}
where the last inequality again results from the estimate $\|\mathcal{P}_{T}(\bm{R}_{1})\|_{\mathrm{F}}\lesssim n^{-19}$
given in Lemma \ref{lemma:R1}. Invoking the condition $\|\bm{\Delta}_{\bm{L}}\|_{\mathrm{F}}\lesssim\sigma n^{4}$
(cf.~(\ref{eq:new-crude-bound})) yields 
\begin{equation}
\left\Vert \mathcal{P}_{T^{\perp}}(\bm{\Delta}_{\bm{L}})\right\Vert _{\mathrm{F}}\leq\left\Vert \mathcal{P}_{T^{\perp}}(\bm{\Delta}_{\bm{L}})\right\Vert _{\ast}\lesssim\frac{1}{n^{19}}\sigma n^{4}\lesssim\frac{\sigma}{n^{15}}\leq\frac{\sigma}{n^{14}},\label{eq:PTperp-DeltaL}
\end{equation}
which demonstrates that the energy of $\bm{\Delta}_{\bm{L}}$ outside
$T$ is extremely small.

We now move on to $\mathcal{P}_{(\Omega^{\star})^{\mathrm{c}}}(\bm{\Delta}_{\bm{S}})$.
This term obeys
\[
\mathcal{P}_{(\Omega^{\star})^{\mathrm{c}}}(\bm{\Delta}_{\bm{S}})=\mathcal{P}_{\Omega_{\mathsf{obs}}\setminus\Omega^{\star}}(\bm{\Delta}_{\bm{S}}),
\]
where the relation holds since $\bm{\Delta}_{\bm{S}}$ is supported
on $\Omega_{\mathsf{obs}}$. To facilitate the analysis of $\Omega_{\mathsf{obs}}\backslash\Omega^{\star}$,
we introduce another index subset 
\begin{equation}
\Omega_{1}\coloneqq\left\{ (i,j)\in\Omega_{\mathsf{obs}}\,:\,|(\bm{\Delta}_{\bm{S}})_{ij}|\leq\|\mathcal{P}_{\Omega_{\mathsf{obs}}}(\bm{\Delta}_{\bm{L}}+\bm{\Delta}_{\bm{S}})\|_{\infty}\right\} .\label{eq:defn-Omega-1-auxiliary}
\end{equation}
The usefulness of $\Omega_{1}$ can be seen through the following
claim, whose claim is postponed to the end of this section.

\begin{claim}\label{claim:Omega-subset}Under Condition \ref{assumption:link-cvx-ncvx-noise},
we have 
\[
\Omega_{\mathsf{obs}}\backslash\Omega^{\star}\subseteq\Omega_{1}.
\]
\end{claim}

An immediate consequence of Claim~\ref{claim:Omega-subset} is that
\begin{align}
\left\Vert \mathcal{P}_{(\Omega^{\star})^{\mathrm{c}}}\left(\bm{\Delta}_{\bm{S}}\right)\right\Vert _{\mathrm{F}} & =\left\Vert \mathcal{P}_{\Omega_{\mathsf{obs}}\setminus\Omega^{\star}}\left(\bm{\Delta}_{\bm{S}}\right)\right\Vert _{\mathrm{F}}\leq\left\Vert \mathcal{P}_{\Omega_{1}}\left(\bm{\Delta}_{\bm{S}}\right)\right\Vert _{\mathrm{F}}\nonumber \\
 & \leq n\left\Vert \mathcal{P}_{\Omega_{1}}\left(\bm{\Delta}_{\bm{S}}\right)\right\Vert _{\infty}\leq n\left\Vert \mathcal{P}_{\Omega_{\mathsf{obs}}}\left(\bm{\Delta}_{\bm{L}}+\bm{\Delta}_{\bm{S}}\right)\right\Vert _{\infty}\nonumber \\
 & \leq n\left\Vert \mathcal{P}_{\Omega_{\mathsf{obs}}}\left(\bm{\Delta}_{\bm{L}}+\bm{\Delta}_{\bm{S}}\right)\right\Vert _{\mathrm{F}}\leq\frac{\sigma}{n^{6}},\label{eq:PSc-DeltaS}
\end{align}
which justifies our assertion that the energy of $\bm{\Delta}_{\bm{S}}$
outside $\Omega^{\star}$ is extremely small. Here, the last inequality
arises from (\ref{eq:Delta-L-Delta-S-sum-small}).

\subsubsection{Step 3: controlling the size of $\bm{\Delta}_{\bm{S}}$ (and hence
that of $\bm{\Delta}_{\bm{L}}$)}

In view of (\ref{eq:Delta-L-Delta-S-sum-small}) and the triangle
inequality, we have 
\begin{align}
\frac{\sigma}{n^{7}} & \geq\left\Vert \mathcal{P}_{\Omega_{\mathsf{obs}}}\left(\bm{\Delta}_{\bm{L}}+\bm{\Delta}_{\bm{S}}\right)\right\Vert _{\mathrm{F}}\nonumber \\
 & \geq\left\Vert \mathcal{P}_{\Omega_{\mathsf{obs}}}\mathcal{P}_{T}\left(\bm{\Delta}_{\bm{L}}\right)\right\Vert _{\mathrm{F}}-\left\Vert \mathcal{P}_{\Omega_{\mathsf{obs}}}\mathcal{P}_{T^{\perp}}\left(\bm{\Delta}_{\bm{L}}\right)\right\Vert _{\mathrm{F}}-\left\Vert \mathcal{P}_{\Omega^{\star}}\left(\bm{\Delta}_{\bm{S}}\right)\right\Vert _{\mathrm{F}}-\left\Vert \mathcal{P}_{\Omega_{\mathsf{obs}}\backslash\Omega^{\star}}\left(\bm{\Delta}_{\bm{S}}\right)\right\Vert _{\mathrm{F}}\nonumber \\
 & \geq\left\Vert \mathcal{P}_{\Omega_{\mathsf{obs}}}\mathcal{P}_{T}\left(\bm{\Delta}_{\bm{L}}\right)\right\Vert _{\mathrm{F}}-\left\Vert \mathcal{P}_{\Omega^{\star}}\left(\bm{\Delta}_{\bm{S}}\right)\right\Vert _{\mathrm{F}}-\left\Vert \mathcal{P}_{T^{\perp}}\left(\bm{\Delta}_{\bm{L}}\right)\right\Vert _{\mathrm{F}}-\left\Vert \mathcal{P}_{(\Omega^{\star})^{\mathrm{c}}}\left(\bm{\Delta}_{\bm{S}}\right)\right\Vert _{\mathrm{F}}\nonumber \\
 & \geq\left\Vert \mathcal{P}_{\Omega_{\mathsf{obs}}}\mathcal{P}_{T}(\bm{\Delta}_{\bm{L}})\right\Vert _{\mathrm{F}}-\left\Vert \mathcal{P}_{\Omega^{\star}}\left(\bm{\Delta}_{\bm{S}}\right)\right\Vert _{\mathrm{F}}-\frac{\sigma}{n^{6}}-\frac{\sigma}{n^{14}},\label{eq:cvx-ncvx-no2}
\end{align}
where the last step follows from (\ref{eq:PTperp-DeltaL}) and (\ref{eq:PSc-DeltaS}).
By Condition~\ref{assumption:injectivity}, we have
\[
\left\Vert \mathcal{P}_{\Omega_{\mathsf{obs}}}\mathcal{P}_{T}\left(\bm{\Delta}_{\bm{L}}\right)\right\Vert _{\mathrm{F}}\geq\sqrt{\frac{c_{\mathrm{inj}}}{\kappa}p}\left\Vert \mathcal{P}_{T}\left(\bm{\Delta}_{\bm{L}}\right)\right\Vert _{\mathrm{F}}\quad\text{and}\quad\left\Vert \mathcal{P}_{\Omega^{\star}}\mathcal{P}_{T}\left(\bm{\Delta}_{\bm{L}}\right)\right\Vert _{\mathrm{F}}\leq\frac{1}{2}\sqrt{\frac{c_{\mathrm{inj}}}{\kappa}p}\left\Vert \mathcal{P}_{T}\left(\bm{\Delta}_{\bm{L}}\right)\right\Vert _{\mathrm{F}},
\]
given that $\mathcal{P}_{T}(\bm{\Delta}_{\bm{L}})\in T$. The latter
inequality combined with (\ref{eq:Delta-L-Delta-S-sum-small}) and
(\ref{eq:PTperp-DeltaL}) further gives 
\begin{align*}
\left\Vert \mathcal{P}_{\Omega^{\star}}\left(\bm{\Delta}_{\bm{S}}\right)\right\Vert _{\mathrm{F}} & \leq\left\Vert \mathcal{P}_{\Omega^{\star}}\left(\bm{\Delta}_{\bm{S}}+\bm{\Delta}_{\bm{L}}\right)\right\Vert _{\mathrm{F}}+\left\Vert \mathcal{P}_{\Omega^{\star}}\left(\bm{\Delta}_{\bm{L}}\right)\right\Vert _{\mathrm{F}}\\
 & \leq\left\Vert \mathcal{P}_{\Omega_{\mathsf{obs}}}\left(\bm{\Delta}_{\bm{S}}+\bm{\Delta}_{\bm{L}}\right)\right\Vert _{\mathrm{F}}+\left\Vert \mathcal{P}_{\Omega^{\star}}\mathcal{P}_{T}\left(\bm{\Delta}_{\bm{L}}\right)\right\Vert _{\mathrm{F}}+\left\Vert \mathcal{P}_{\Omega^{\star}}\mathcal{P}_{T^{\perp}}\left(\bm{\Delta}_{\bm{L}}\right)\right\Vert _{\mathrm{F}}\\
 & \leq\frac{\sigma}{n^{7}}+\frac{1}{2}\sqrt{\frac{c_{\mathrm{inj}}}{\kappa}p}\left\Vert \mathcal{P}_{T}\bm{\Delta}_{\bm{L}}\right\Vert _{\mathrm{F}}+\frac{\sigma}{n^{14}}.
\end{align*}
Substituting the above bounds into (\ref{eq:cvx-ncvx-no2}) gives
\begin{align*}
\frac{\sigma}{n^{7}} & \geq\sqrt{\frac{c_{\mathrm{inj}}}{\kappa}p}\left\Vert \mathcal{P}_{T}\left(\bm{\Delta}_{\bm{L}}\right)\right\Vert _{\mathrm{F}}-\frac{1}{2}\sqrt{\frac{c_{\mathrm{inj}}}{\kappa}p}\left\Vert \mathcal{P}_{T}\left(\bm{\Delta}_{\bm{L}}\right)\right\Vert _{\mathrm{F}}-\frac{\sigma}{n^{7}}-\frac{\sigma}{n^{6}}-\frac{2\sigma}{n^{14}}\\
 & \geq\frac{1}{2}\sqrt{\frac{c_{\mathrm{inj}}}{\kappa}p}\left\Vert \mathcal{P}_{T}\left(\bm{\Delta}_{\bm{L}}\right)\right\Vert _{\mathrm{F}}-\frac{2\sigma}{n^{6}},
\end{align*}
which further yields 
\[
\left\Vert \mathcal{P}_{T}\left(\bm{\Delta}_{\bm{L}}\right)\right\Vert _{\mathrm{F}}\lesssim\frac{\sigma}{n^{6}}\sqrt{\frac{\kappa}{p}}\leq\frac{\sigma}{n^{5}},
\]
provided that $n^{2}p\gg\kappa$. This combined with (\ref{eq:PTperp-DeltaL})
allows one to control the size of $\bm{\Delta}_{\bm{L}}$: 
\begin{align*}
\bigl\Vert\bm{\Delta}_{\bm{L}}\bigr\Vert_{\mathrm{F}} & \leq\left\Vert \mathcal{P}_{T}\left(\bm{\Delta}_{\bm{L}}\right)\right\Vert _{\mathrm{F}}+\left\Vert \mathcal{P}_{T^{\perp}}\left(\bm{\Delta}_{\bm{L}}\right)\right\Vert _{\mathrm{F}}\lesssim\frac{\sigma}{n^{5}}.
\end{align*}
In view of (\ref{eq:Delta-L-Delta-S-sum-small}) and the fact that
$\bm{\Delta}_{\bm{S}}$ is supported on $\Omega_{\mathsf{obs}}$,
we have
\begin{align*}
\left\Vert \bm{\Delta}_{\bm{S}}\right\Vert _{\mathrm{F}} & =\left\Vert \mathcal{P}_{\Omega_{\mathsf{obs}}}\left(\bm{\Delta}_{\bm{S}}\right)\right\Vert _{\mathrm{F}}\leq\left\Vert \mathcal{P}_{\Omega_{\mathsf{obs}}}\left(\bm{\Delta}_{\bm{L}}+\bm{\Delta}_{\bm{S}}\right)\right\Vert _{\mathrm{F}}+\left\Vert \mathcal{P}_{\Omega_{\mathsf{obs}}}\left(\bm{\Delta}_{\bm{L}}\right)\right\Vert _{\mathrm{F}}\\
 & \leq\left\Vert \mathcal{P}_{\Omega_{\mathsf{obs}}}\left(\bm{\Delta}_{\bm{L}}+\bm{\Delta}_{\bm{S}}\right)\right\Vert _{\mathrm{F}}+\left\Vert \bm{\Delta}_{\bm{L}}\right\Vert _{\mathrm{F}}\lesssim\frac{\sigma}{n^{5}},
\end{align*}
thus concluding the proof.

\subsubsection{Proof of Claim \ref{claim:Omega-subset}}

We first recall the facts that 
\[
\bm{S}=\mathcal{P}_{\Omega_{\mathsf{obs}}}\left[\mathcal{S}_{\tau}\left(\bm{M}-\bm{X}\bm{Y}^{\top}\right)\right]\qquad\text{and}\qquad\bm{S}+\bm{\Delta}_{\bm{S}}=\mathcal{P}_{\Omega_{\mathsf{obs}}}\left[\mathcal{S}_{\tau}\left(\bm{M}-\bm{X}\bm{Y}^{\top}-\bm{\Delta}_{\bm{L}}\right)\right],
\]
where the second identity follows since $(\bm{L}_{\mathsf{cvx}},\bm{S}_{\mathsf{cvx}})=(\bm{X}\bm{Y}^{\top}+\bm{\Delta}_{\bm{L}},\bm{S}+\bm{\Delta}_{\bm{S}})$
is the optimizer of the convex program (\ref{eq:cvx}). These allow
us to write 
\begin{align}
\bm{\Delta}_{\bm{S}} & =\mathcal{P}_{\Omega_{\mathsf{obs}}}\left[\mathcal{S}_{\tau}\big(\bm{M}-\bm{X}\bm{Y}^{\top}-\bm{\Delta}_{\bm{L}}\big)-\mathcal{S}_{\tau}\left(\bm{M}-\bm{X}\bm{Y}^{\top}\right)\right]\nonumber \\
 & =\mathcal{P}_{\Omega_{\mathsf{obs}}}\left[\mathcal{S}_{\tau}\left[\bm{M}-\bm{X}\bm{Y}^{\top}+\bm{\Delta}_{\bm{S}}-(\bm{\Delta}_{\bm{L}}+\bm{\Delta}_{\bm{S}})\right]-\mathcal{S}_{\tau}\left(\bm{M}-\bm{X}\bm{Y}^{\top}\right)\right].\label{eq:Delta-S-characterization}
\end{align}
This characterization of $\bm{\Delta}_{\bm{S}}$ turns out to be crucial
when establishing the inclusion $\Omega_{\mathsf{obs}}\backslash\Omega^{\star}\subseteq\Omega_{1}$.
Towards this end, we need to introduce another index subset 
\[
\Omega_{2}\coloneqq\left\{ (i,j)\in\Omega_{\mathsf{obs}}\,:\,\tau-\|\mathcal{P}_{\Omega_{\mathsf{obs}}}(\bm{\Delta}_{\bm{L}}+\bm{\Delta}_{\bm{S}})\|_{\infty}\leq\big|\big(\bm{M}-\bm{X}\bm{Y}^{\top}\big)_{ij}\big|\leq\tau\right\} .
\]
As it turns out, the sets $\Omega,\Omega_{1}$ and $\Omega_{2}$ obey
the following three conditions
\[
\Omega_{2}\cap\Omega=\varnothing,\qquad\Omega_{\mathsf{obs}}\backslash\Omega\subseteq\Omega_{1}\cup\Omega_{2},\qquad\text{and}\qquad\Omega\cup\Omega_{2}\subseteq\Omega^{\star},
\]
which immediately lead to
\[
\Omega_{\mathsf{obs}}\backslash\Omega^{\star}\,\overset{(\text{i})}{\subseteq}\,\Omega_{\mathsf{obs}}\backslash(\Omega\cup\Omega_{2})\,\overset{(\mathrm{ii})}{=}\,(\Omega_{\mathsf{obs}}\backslash\Omega)\backslash\Omega_{2}\,\overset{(\mathrm{iii})}{\subseteq}\,(\Omega_{1}\cup\Omega_{2})\backslash\Omega_{2}\,\subseteq\,\Omega_{1}.
\]
Here, (i) follows since $\Omega\cup\Omega_{2}\subseteq\Omega^{\star}$,
(ii) holds true since $\Omega_{2}\cap\Omega=\varnothing$, and (iii)
results from the condition $\Omega_{\mathsf{obs}}\backslash\Omega\subseteq\Omega_{1}\cup\Omega_{2}$.
It then boils down to proving each of the above three conditions.
\begin{enumerate}
\item The first one $\Omega_{2}\cap\Omega=\varnothing$ is straightforward
to establish. Note that for any $(i,j)\in\Omega_{2}$, one must have
$\big|\big(\bm{M}-\bm{X}\bm{Y}^{\top}\big)_{ij}\big|\leq\tau$ and
hence $\big[\mathcal{S}_{\tau}(\bm{M}-\bm{X}\bm{Y}^{\top})\big]_{ij}=0$,
which means that $(i,j)\notin\Omega$. This proves the relation $\Omega_{2}\cap\Omega=\varnothing$.
\item Moving on to the second one $\Omega_{\mathsf{obs}}\backslash\Omega\subseteq\Omega_{1}\cup\Omega_{2}$,
we prove this via contradiction. Suppose that this inclusion is false,
i.e.~there exits an index $(i,j)\in\Omega_{\mathsf{obs}}\backslash\Omega$
such that 
\begin{equation}
\vert(\bm{\Delta}_{\bm{S}})_{ij}\vert>\Vert\mathcal{P}_{\Omega_{\mathsf{obs}}}(\bm{\Delta}_{\bm{L}}+\bm{\Delta}_{\bm{S}})\Vert_{\infty}\quad\text{and}\quad\vert(\bm{M}-\bm{X}\bm{Y}^{\top})_{ij}\vert<\tau-\Vert\mathcal{P}_{\Omega_{\mathsf{obs}}}(\bm{\Delta}_{\bm{L}}+\bm{\Delta}_{\bm{S}})\Vert_{\infty}.\label{eq:M-XY-UB1}
\end{equation}
Here, we have taken into account the fact that 
\[
\vert(\bm{M}-\bm{X}\bm{Y}^{\top})_{ij}\vert\leq\tau,\qquad\text{for any }(i,j)\in\Omega_{\mathsf{obs}}\backslash\Omega.
\]
To reach contradiction, we find it convenient to state the following
simple fact. \begin{fact}\label{fact:a-b-S}Suppose that $|a|\leq\tau$
and that $\mathcal{S}_{\tau}(a+b)\neq0$. Then 
\[
\big|\mathcal{S}_{\tau}(a+b)\big|\leq|b|+|a|-\tau.
\]
\end{fact}\begin{proof}Given that $\mathcal{S}_{\tau}(a+b)\neq0$,
one necessarily has $|a+b|>\tau$. Without loss of generality, assume
that $a+b>0$, which gives 
\[
\mathcal{S}_{\tau}(a+b)=a+b-\tau>0.
\]
This together with the fact $\tau\geq|a|$ yields $|\mathcal{S}_{\tau}(a+b)|=a+b-\tau\leq|b|+|a|-\tau$.
\end{proof}

With this fact in mind, we can deduce that 
\begin{align}
\big|\big(\bm{\Delta}_{\bm{S}}\big)_{ij}\big| & =\left|\left\{ \mathcal{S}_{\tau}\left[\bm{M}-\bm{X}\bm{Y}^{\top}+\bm{\Delta}_{\bm{S}}-(\bm{\Delta}_{\bm{L}}+\bm{\Delta}_{\bm{S}})\right]-\mathcal{S}_{\tau}\left(\bm{M}-\bm{X}\bm{Y}^{\top}\right)\right\} _{ij}\right|\nonumber \\
 & \overset{(\mathrm{i})}{=}\left|\left\{ \mathcal{S}_{\tau}\left[\bm{M}-\bm{X}\bm{Y}^{\top}+\bm{\Delta}_{\bm{S}}-(\bm{\Delta}_{\bm{L}}+\bm{\Delta}_{\bm{S}})\right]\right\} _{ij}\right|\nonumber \\
 & \overset{(\mathrm{ii})}{\leq}\left|\left[\bm{\Delta}_{\bm{S}}-\left(\bm{\Delta}_{\bm{L}}+\bm{\Delta}_{\bm{S}}\right)\right]_{ij}\right|+\big|\left(\bm{M}-\bm{X}\bm{Y}^{\top}\right)_{ij}\big|-\tau\nonumber \\
 & \overset{(\mathrm{iii})}{<}\big|\big(\bm{\Delta}_{\bm{S}}\big)_{ij}\big|+\left\Vert \mathcal{P}_{\Omega_{\mathsf{obs}}}(\bm{\Delta}_{\bm{L}}+\bm{\Delta}_{\bm{S}})\right\Vert _{\infty}-\Vert\mathcal{P}_{\Omega_{\mathsf{obs}}}(\bm{\Delta}_{\bm{L}}+\bm{\Delta}_{\bm{S}})\Vert_{\infty}\nonumber \\
 & =\big|\big(\bm{\Delta}_{\bm{S}}\big)_{ij}\big|,\label{eq:Delta-small-Delta}
\end{align}
where (i) holds true since $\big[\mathcal{S}_{\tau}(\bm{M}-\bm{X}\bm{Y}^{\top})\big]_{ij}=0$
for any $(i,j)\in\Omega_{\mathsf{obs}}\backslash\Omega$, (ii) follows
from Fact \ref{fact:a-b-S} (by taking $a=(\bm{M}-\bm{X}\bm{Y}^{\top})_{ij}$
and $b=\left[\bm{\Delta}_{\bm{S}}-\left(\bm{\Delta}_{\bm{L}}+\bm{\Delta}_{\bm{S}}\right)\right]_{ij}$),
and (iii) is a consequence of (\ref{eq:M-XY-UB1}) as well as the
triangle inequality. The inequality (\ref{eq:Delta-small-Delta}),
however, is clearly impossible. This establishes that $\Omega_{\mathsf{obs}}\backslash\Omega\subseteq\Omega_{1}\cup\Omega_{2}$.
\item We are left with the last one $\Omega\cup\Omega_{2}\subseteq\Omega^{\star}$,
which is equivalent to saying $\Omega\subseteq\Omega^{\star}$ and
$\Omega_{2}\subseteq\Omega^{\star}$. First, for any $(i,j)\in\Omega$,
one has 
\begin{align*}
\big|S_{ij}\big|>0\quad & \Longrightarrow\quad(i,j)\in\Omega_{\mathsf{obs}}\quad\text{and}\quad\bigl|\left(\bm{L}^{\star}+\bm{S}^{\star}+\bm{E}-\bm{X}\bm{Y}^{\top}\right)_{ij}\bigr|>\tau\\
 & \Longrightarrow\quad(i,j)\in\Omega_{\mathsf{obs}}\quad\text{and}\quad\bigl|S_{ij}^{\star}\bigr|>\tau-\left\Vert \bm{L}^{\star}-\bm{X}\bm{Y}^{\top}\right\Vert _{\infty}-\left\Vert \bm{E}\right\Vert _{\infty}>0.
\end{align*}
Here, the last step comes from the triangle inequality and Condition~\ref{assumption:link-cvx-ncvx-noise}.
This reveals that $\Omega\subseteq\Omega^{\star}$. Similarly, for
any $(i,j)\in\Omega_{2}$ we have
\begin{align*}
 & \tau-\left\Vert \mathcal{P}_{\Omega_{\mathsf{obs}}}(\bm{\Delta}_{\bm{L}}+\bm{\Delta}_{\bm{S}})\right\Vert _{\infty}\leq\big|\big(\bm{M}-\bm{X}\bm{Y}^{\top}\big)_{ij}\big|\\
\Longleftrightarrow\quad\quad & \tau-\left\Vert \mathcal{P}_{\Omega_{\mathsf{obs}}}(\bm{\Delta}_{\bm{L}}+\bm{\Delta}_{\bm{S}})\right\Vert _{\infty}\leq\big|\big(\bm{S}^{\star}+\bm{L}^{\star}-\bm{X}\bm{Y}^{\top}+\bm{E}\big)_{ij}\big|\\
\Longrightarrow\quad\quad & \bigl|S_{ij}^{\star}\bigr|\geq\tau-\left\Vert \bm{L}^{\star}-\bm{X}\bm{Y}^{\top}\right\Vert _{\infty}-\left\Vert \bm{E}\right\Vert _{\infty}-\left\Vert \mathcal{P}_{\Omega_{\mathsf{obs}}}(\bm{\Delta}_{\bm{L}}+\bm{\Delta}_{\bm{S}})\right\Vert _{\mathrm{F}}\geq\frac{\tau}{2}-\frac{\sigma}{n^{7}}>0,
\end{align*}
where we have used Condition~\ref{assumption:link-cvx-ncvx-noise},
the bound~(\ref{eq:Delta-L-Delta-S-sum-small}), and the fact that
$\tau\gg\sigma$. This demonstrates that $\Omega_{2}\subseteq\Omega^{\star}$.
We have therefore justified that $\Omega\cup\Omega_{2}\subseteq\Omega^{\star}$.
\end{enumerate}

\section{Analysis of the nonconvex procedure (Proof of Theorem \ref{thm:ncvx-property})
\label{sec:Analysis-nonconvex}}

This section is devoted to establishing Theorem~\ref{thm:ncvx-property}.
For notational convenience, we introduce 
\begin{equation}
\bm{F}^{t}\coloneqq\left[\bm{X}^{t\top},\bm{Y}^{t\top}\right]^{\top}\in\mathbb{R}^{2n\times r}\qquad\text{and}\qquad\bm{F}^{\star}\coloneqq\left[\bm{X}^{\star\top},\bm{Y}^{\star\top}\right]^{\top}\in\mathbb{R}^{2n\times r}.\label{eq:defn-Ft-Fstar}
\end{equation}
These allow us to express succinctly the rotation matrix $\bm{H}^{t}$
defined in~(\ref{eq:defn-Ht}) as 
\begin{equation}
\bm{H}^{t}=\arg\min_{\bm{R}\in\mathcal{O}^{r\times r}}\left\Vert \bm{F}^{t}\bm{R}-\bm{F}^{\star}\right\Vert _{\mathrm{F}}.\label{eq:defn-Ht-alt}
\end{equation}
With the definitions of $\bm{F}^{t}$ and $\bm{H}^{t}$ in mind, it
suffices to justify that: for all $0\leq t\leq t_{0}=n^{47}$, the
following hypotheses\begin{subequations}\label{eq:induction-hypotheses}
\begin{align}
\left\Vert \bm{F}^{t}\bm{H}^{t}-\bm{F}^{\star}\right\Vert _{\mathrm{F}} & \leq C_{\mathrm{F}}\left(\frac{\sigma}{\sigma_{\min}}\sqrt{\frac{n}{p}}+\frac{\lambda}{p\sigma_{\min}}\right)\left\Vert \bm{X}^{\star}\right\Vert _{\mathrm{F}},\label{eq:induction-fro}\\
\left\Vert \bm{F}^{t}\bm{H}^{t}-\bm{F}^{\star}\right\Vert  & \leq C_{\mathrm{op}}\left(\frac{\sigma}{\sigma_{\min}}\sqrt{\frac{n}{p}}+\frac{\lambda}{p\sigma_{\min}}\right)\left\Vert \bm{X}^{\star}\right\Vert ,\label{eq:induction-op}\\
\left\Vert \bm{F}^{t}\bm{H}^{t}-\bm{F}^{\star}\right\Vert _{2,\infty} & \leq C_{\infty}\kappa\left(\frac{\sigma}{\sigma_{\min}}\sqrt{\frac{n\log n}{p}}+\frac{\lambda}{p\sigma_{\min}}\right)\left\Vert \bm{F}^{\star}\right\Vert _{2,\infty},\label{eq:induction-two-to-infty}\\
\left\Vert \bm{X}^{t\top}\bm{X}^{t}-\bm{Y}^{t\top}\bm{Y}^{t}\right\Vert _{\mathrm{F}} & \leq C_{\mathrm{B}}\kappa\eta\left(\frac{\sigma}{\sigma_{\min}}\sqrt{\frac{n}{p}}+\frac{\lambda}{p\sigma_{\min}}\right)\sqrt{r}\sigma_{\max}^{2},\label{eq:induction-balance}\\
\left\Vert \bm{S}^{t}-\bm{S}^{\star}\right\Vert  & \leq C_{\mathrm{S}}\sigma\sqrt{np}\label{eq:induction-S-op-norm}
\end{align}
\end{subequations}hold for some universal constants $C_{\mathrm{F}}$,
$C_{\mathrm{op}}$, $C_{\infty}$, $C_{\mathrm{B}}$, $C_{\mathrm{S}}>0$,
and, in addition, 
\begin{equation}
F\left(\bm{X}^{t},\bm{Y}^{t};\bm{S}^{t}\right)\leq F\left(\bm{X}^{t-1},\bm{Y}^{t-1};\bm{S}^{t-1}\right)-\frac{\eta}{2}\left\Vert \nabla f\left(\bm{X}^{t-1},\bm{Y}^{t-1};\bm{S}^{t-1}\right)\right\Vert _{\mathrm{F}}^{2}\label{eq:induction-descent}
\end{equation}
holds for all $1\leq t\leq t_{0}=n^{47}$.

Clearly, the bounds (\ref{eq:thm-ncvx-fro}), (\ref{eq:thm-ncvx-op}),
(\ref{eq:thm-ncvx-two-infty}), and (\ref{eq:thm-ncvx-S}) in Theorem~\ref{thm:ncvx-property}
follow immediately from (\ref{eq:induction-fro}), (\ref{eq:induction-op}),
(\ref{eq:induction-two-to-infty}), and (\ref{eq:induction-S-op-norm}),
respectively. It remains to justify the small gradient bound (\ref{eq:small-gradient})
on the basis of (\ref{eq:induction-hypotheses}) and (\ref{eq:induction-descent}),
which is exactly the content of the following lemma.

\begin{lemma}[\textbf{Small gradient}]\label{lemma:small-gradient}Set
$\lambda=C_{\lambda}\sigma\sqrt{np\log n}$ for some large constant
$C_{\lambda}>0$. Suppose that $n^{2}p\gg\kappa^{3}\mu rn\log^{2}n$
and that the noise satisfies $\frac{\sigma}{\sigma_{\min}}\sqrt{\frac{n}{p}}\ll\frac{1}{\sqrt{\kappa^{4}\mu r\log n}}$.
Take $\eta\asymp1/(n\kappa^{3}\sigma_{\max})$. If the iterates satisfy
(\ref{eq:induction-hypotheses}) for all $0\leq t\leq t_{0}$ and
(\ref{eq:induction-descent}) for all $1\leq t\leq t_{0}$, then with
probability at least $1-O(n^{-50})$, one has
\[
\min_{0\leq t<t_{0}}\left\Vert \nabla f\left(\bm{X}^{t},\bm{Y}^{t};\bm{S}^{t}\right)\right\Vert _{\mathrm{F}}\leq\frac{1}{n^{20}}\frac{\lambda}{p}\sqrt{\sigma_{\min}}.
\]
\end{lemma}\begin{proof}See Appendix \ref{subsec:Proof-of-Lemma-small-gradient}.\end{proof}

The remainder of this section is thus dedicated to showing that (\ref{eq:induction-hypotheses})
and (\ref{eq:induction-descent}) hold for $\{(\bm{F}^{t},\bm{S}^{t})\}_{0\leq t\leq t_{0}}$,
which we accomplish via mathematical induction. Throughout this section,
we let $\bm{X}_{l,\cdot}$ denote the $l$th row of a matrix $\bm{X}$.

\subsection{Leave-one-out analysis\label{subsec:Leave-one-out-analysis}}

The above hypotheses (\ref{eq:induction-hypotheses}) require, among
other things, sharp control of the $\ell_{2,\infty}$ estimation errors,
which calls for fine-grained statistical analyses. In order to decouple
complicated statistical dependency, we resort to the following leave-one-out
analysis framework that has been successfully applied to analyze other
nonconvex algorithms \cite{zhong2017near,ma2017implicit,chen2019nonconvex,chen2018gradient,chen2019noisy,cai2019nonconvex,li2020breaking,ding2018leave}.

\paragraph{Leave-one-out loss functions.}

For each $1\leq l\leq n$, we define the following auxiliary loss
functions 
\begin{align*}
 & F^{(l)}\left(\bm{X},\bm{Y},\bm{S}\right)\\
 & \quad\coloneqq\underbrace{\frac{1}{2p}\left\Vert \mathcal{P}_{(\Omega_{\mathsf{obs}})_{-l,\cdot}}\left(\bm{X}\bm{Y}^{\top}+\bm{S}-\bm{M}\right)\right\Vert _{\mathrm{F}}^{2}+\frac{1}{2}\left\Vert \mathcal{P}_{l,\cdot}\left(\bm{X}\bm{Y}^{\top}-\bm{L}^{\star}\right)\right\Vert _{\mathrm{F}}^{2}+\frac{\lambda}{2p}\left\Vert \bm{X}\right\Vert _{\mathrm{F}}^{2}+\frac{\lambda}{2p}\left\Vert \bm{Y}\right\Vert _{\mathrm{F}}^{2}}_{\eqqcolon f^{\left(l\right)}\left(\bm{X},\bm{Y};\bm{S}\right)}+\frac{\tau}{p}\left\Vert \bm{S}\right\Vert _{1}.
\end{align*}
Here, $\mathcal{P}_{(\Omega_{\mathsf{obs}})_{-l,\cdot}}(\cdot)$ (resp.~$\mathcal{P}_{l,\cdot}(\cdot)$)
denotes orthogonal projection onto the space of matrices supported
on the index set $\{(i,j)\in\Omega_{\mathsf{obs}}\mid i\neq l\}$
(resp.~$\{(i,j)\mid i=l\}$), namely, for any matrix $\bm{B}\in\mathbb{R}^{n\times n}$
one has 
\[
\left[\mathcal{P}_{(\Omega_{\mathsf{obs}})_{-l,\cdot}}\left(\bm{B}\right)\right]_{ij}=\begin{cases}
B_{ij}, & \text{if }(i,j)\in\Omega_{\mathsf{obs}}\text{ and }i\neq l,\\
0, & \text{otherwise}
\end{cases}\quad\text{and}\quad\left[\mathcal{P}_{l,\cdot}\left(\bm{B}\right)\right]_{ij}=\begin{cases}
B_{ij}, & \text{if }i=l,\\
0, & \text{otherwise}.
\end{cases}
\]
The above auxiliary loss function is obtained by dropping the randomness
coming from the $l$th row of $\bm{M}$, which, as we shall see shortly,
facilitates analysis in establishing the incoherence properties (\ref{eq:induction-two-to-infty}).
Similarly, we define for each $n+1\leq l\leq2n$ that 
\begin{align*}
 & F^{(l)}\left(\bm{X},\bm{Y},\bm{S}\right)\\
 & \quad\coloneqq\underbrace{\frac{1}{2p}\left\Vert \mathcal{P}_{(\Omega_{\mathsf{obs}})_{\cdot,-(l-n)}}\left(\bm{X}\bm{Y}^{\top}+\bm{S}-\bm{M}\right)\right\Vert _{\mathrm{F}}^{2}+\frac{1}{2}\left\Vert \mathcal{P}_{\cdot,(l-n)}\left(\bm{X}\bm{Y}^{\top}-\bm{L}^{\star}\right)\right\Vert _{\mathrm{F}}^{2}+\frac{\lambda}{2p}\left\Vert \bm{X}\right\Vert _{\mathrm{F}}^{2}+\frac{\lambda}{2p}\left\Vert \bm{Y}\right\Vert _{\mathrm{F}}^{2}}_{\eqqcolon f^{\left(l\right)}\left(\bm{X},\bm{Y};\bm{S}\right)}+\tau\left\Vert \bm{S}\right\Vert _{1},
\end{align*}
where the projection operators $\mathcal{P}_{(\Omega_{\mathsf{obs}})_{\cdot,-(l-n)}}(\cdot)$
and $\mathcal{P}_{\cdot,(l-n)}(\cdot)$ are defined such that for
any matrix $\bm{B}\in\mathbb{R}^{n\times n}$, 
\[
\left[\mathcal{P}_{(\Omega_{\mathsf{obs}})_{\cdot,-(l-n)}}\left(\bm{B}\right)\right]_{ij}=\begin{cases}
B_{ij}, & \text{if }(i,j)\in\Omega_{\mathsf{obs}}\text{ and }j\neq l-n,\\
0, & \text{otherwise}
\end{cases}\quad\text{and}\quad\left[\mathcal{P}_{\cdot,(l-n)}\left(\bm{B}\right)\right]_{ij}=\begin{cases}
B_{ij}, & \text{if }j=l-n,\\
0, & \text{otherwise}.
\end{cases}
\]
Again, this auxiliary loss function is produced in a way that is independent
from the $(l-n)$-th column of $\bm{M}$. In the above notation, $f^{\left(l\right)}\left(\bm{X},\bm{Y};\bm{S}\right)$
is a function of $\bm{X}$ and $\bm{Y}$ with $\bm{S}$ frozen.

\paragraph{Leave-one-out auxiliary sequences.}

For each $1\leq l\leq2n$, we construct a sequence of leave-one-out
iterates $\{\bm{F}^{t,(l)},\bm{S}^{t,(l)}\}_{t\geq0}$ via Algorithm
\ref{alg:gd-mc-LOO}.

\begin{algorithm}
\caption{Construction of the $l$th leave-one-out sequences.}

\label{alg:gd-mc-LOO}\begin{algorithmic}

\STATE \textbf{{Initialization}}: $\bm{X}^{0,(l)}=\bm{X}^{\star}$,
$\bm{Y}^{0,(l)}=\bm{Y}^{\star}$, $\bm{S}^{0,(l)}=\bm{S}^{\star}$,
$\bm{F}^{0,(l)}\coloneqq\left[\begin{array}{c}
\bm{X}^{0,(l)}\\
\bm{Y}^{0,(l)}
\end{array}\right]$, and the step size $\eta>0$.

\STATE \textbf{{Gradient updates}}: \textbf{for }$t=0,1,\ldots,t_{0}-1$
\textbf{do}

\STATE \vspace{-1em}
 \begin{subequations} \label{subeq:GD-rules-LOO} 
\begin{align}
\bm{F}^{t+1,(l)} & \coloneqq\left[\begin{array}{c}
\bm{X}^{t+1,(l)}\\
\bm{Y}^{t+1,(l)}
\end{array}\right]=\left[\begin{array}{c}
\bm{X}^{t,(l)}-\eta\nabla_{\bm{X}}f^{(l)}(\bm{X}^{t,(l)},\bm{Y}^{t,(l)};\bm{S}^{t,(l)})\\
\bm{Y}^{t,(l)}-\eta\nabla_{\bm{Y}}f^{(l)}(\bm{X}^{t,(l)},\bm{Y}^{t,(l)};\bm{S}^{t,(l)})
\end{array}\right];\label{eq:defn-F-t-l}\\
\bm{S}^{t+1,(l)} & \coloneqq\begin{cases}
\mathcal{S}_{\tau}\left[\mathcal{P}_{(\Omega_{\mathsf{obs}})_{-l,\cdot}}\left(\bm{M}-\bm{X}^{t+1,(l)}\bm{Y}^{t+1,(l)\top}\right)\right]+\mathcal{P}_{l,\cdot}\left(\bm{S}^{\star}\right), & \text{if }1\leq l\leq n,\\
\mathcal{S}_{\tau}\left[\mathcal{P}_{(\Omega_{\mathsf{obs}})_{\cdot,-(l-n)}}\left(\bm{M}-\bm{X}^{t+1,(l)}\bm{Y}^{t+1,(l)\top}\right)\right]+\mathcal{P}_{\cdot,(l-n)}\left(\bm{S}^{\star}\right), & \text{if }n+1\leq l\leq2n.
\end{cases}\label{eq:defn-S-t-l}
\end{align}
\end{subequations}

\end{algorithmic} 
\end{algorithm}

\paragraph{Properties of leave-one-out sequences.}

There are several features of the leave-one-out sequences that prove
useful for our statistical analysis: (1) for the $l$th leave-one-out
sequence, one can exploit the statistical independence to control
the estimation error of $\bm{F}^{t,(l)}$ in the $l$th row; (2) the
leave-one-out sequences and the original sequence $(\bm{F}^{t},\bm{S}^{t})$
are exceedingly close (since we have only discarded a small amount
of information). These properties taken collectively allow us to control
the estimation error of $\bm{F}^{t}$ in each row. To formalize these
features, we make an additional set of induction hypotheses\begin{subequations}\label{eq:induction-hypotheses-LOO}
\begin{align}
\big\|\bm{F}^{t}\bm{H}^{t}-\bm{F}^{t,(l)}\bm{R}^{t,(l)}\big\|_{\mathrm{F}} & \leq C_{1}\left(\frac{\sigma}{\sigma_{\min}}\sqrt{\frac{n\log n}{p}}+\frac{\lambda}{p\sigma_{\min}}\right)\left\Vert \bm{F}^{\star}\right\Vert _{2,\infty},\label{eq:induction-loo-perturbation}\\
\max_{1\leq l\leq2n}\big\|\big(\bm{F}^{t,(l)}\bm{H}^{t,(l)}-\bm{F}^{\star}\big)_{l,\cdot}\big\|_{2} & \leq C_{2}\kappa\left(\frac{\sigma}{\sigma_{\min}}\sqrt{\frac{n\log n}{p}}+\frac{\lambda}{p\sigma_{\min}}\right)\left\Vert \bm{F}^{\star}\right\Vert _{2,\infty},\label{eq:induction-loo-close}\\
\max_{1\leq l\leq n}\big\|\mathcal{P}_{-l,\cdot}\big(\bm{S}^{t}-\bm{S}^{t,(l)}\big)\big\|_{\mathrm{F}} & \leq C_{3}\frac{\sigma}{\sigma_{\min}}\sqrt{n\log n}\left\Vert \bm{F}^{\star}\right\Vert \left\Vert \bm{F}^{\star}\right\Vert _{2,\infty},\label{eq:induction-S-loo}\\
\max_{n<l\leq2n}\big\|\mathcal{P}_{\cdot,-(l-n)}\big(\bm{S}^{t}-\bm{S}^{t,(l)}\big)\big\|_{\mathrm{F}} & \leq C_{3}\frac{\sigma}{\sigma_{\min}}\sqrt{n\log n}\left\Vert \bm{F}^{\star}\right\Vert \left\Vert \bm{F}^{\star}\right\Vert _{2,\infty}.\label{eq:induction-S-loo'}
\end{align}
\end{subequations}Here, the rotation matrices $\bm{H}^{t,(l)}$ and
$\bm{R}^{t,(l)}$ are defined respectively by 
\[
\bm{H}^{t,(l)}\coloneqq\arg\min_{\bm{R}\in\mathcal{O}^{r\times r}}\big\|\bm{F}^{t,(l)}\bm{R}-\bm{F}^{\star}\big\|_{\mathrm{F}},\qquad\text{and}\qquad\bm{R}^{t,(l)}\coloneqq\arg\min_{\bm{R}\in\mathcal{O}^{r\times r}}\big\|\bm{F}^{t,(l)}\bm{R}-\bm{F}^{t}\big\|_{\mathrm{F}}.
\]

\subsection{Key lemmas for establishing the induction hypotheses}

This subsection establishes the induction hypotheses made in Appendix
\ref{subsec:Leave-one-out-analysis}, namely~(\ref{eq:induction-hypotheses}),
(\ref{eq:induction-descent}) and (\ref{eq:induction-hypotheses-LOO}).
Before continuing, we find it convenient to introduce another function
of $\bm{X}$ and $\bm{Y}$ (with $\bm{S}$ frozen) as follows 
\begin{equation}
f_{\mathsf{aug}}\left(\bm{X},\bm{Y};\bm{S}\right)\coloneqq\frac{1}{2p}\left\Vert \mathcal{P}_{\Omega_{\mathsf{obs}}}\left(\bm{X}\bm{Y}^{\top}+\bm{S}-\bm{M}\right)\right\Vert _{\mathrm{F}}^{2}+\frac{\lambda}{2p}\left\Vert \bm{X}\right\Vert _{\mathrm{F}}^{2}+\frac{\lambda}{2p}\left\Vert \bm{Y}\right\Vert _{\mathrm{F}}^{2}+\frac{1}{8}\left\Vert \bm{X}^{\top}\bm{X}-\bm{Y}^{\top}\bm{Y}\right\Vert _{\mathrm{F}}^{2}.\label{eq:defn-faug}
\end{equation}
The difference between $f_{\mathsf{aug}}$ and $f$ lies in the following
balancing term
\[
f_{\mathsf{diff}}\left(\bm{X},\bm{Y}\right)\coloneqq-\frac{1}{8}\left\Vert \bm{X}^{\top}\bm{X}-\bm{Y}^{\top}\bm{Y}\right\Vert _{\mathrm{F}}^{2},
\]
that is, $f=f_{\mathsf{aug}}+f_{\mathsf{diff}}$.

The following four lemmas, which are inherited from \cite{chen2019noisy}
with little modification, are concerned with local strong convexity
as well as the hypotheses (\ref{eq:induction-fro}), (\ref{eq:induction-op}),
(\ref{eq:induction-balance}), (\ref{eq:induction-loo-close}) and
(\ref{eq:induction-two-to-infty}).

\begin{lemma}[\textbf{Restricted strong convexity}]\label{lemma:local-cvx}Set
$\lambda=C_{\lambda}\sigma\sqrt{np}$ for some large enough constant $C_{\lambda}>0$. Suppose that the
sample size obeys $n^{2}p\gg\kappa\mu rn\log n$ and that the noise
satisfies $\frac{\sigma}{\sigma_{\min}}\sqrt{\frac{n\log n}{p}}\ll1$.
Let the function $f_{\mathsf{aug}}$ be defined in (\ref{eq:defn-faug}).
Then with probability at least $1-O(n^{-100})$,
\begin{align*}
\mathsf{vec}\left(\bm{\Delta}\right)^{\top}\nabla^{2}f_{\mathsf{aug}}\left(\bm{X},\bm{Y};\bm{S}\right)\mathsf{vec}\left(\bm{\Delta}\right) & \geq\tfrac{1}{10}\sigma_{\min}\left\Vert \bm{\Delta}\right\Vert _{\mathrm{F}}^{2}\\
\max\left\{ \left\Vert \nabla^{2}f_{\mathsf{aug}}\left(\bm{X},\bm{Y};\bm{S}\right)\right\Vert ,\left\Vert \nabla^{2}f\left(\bm{X},\bm{Y}\right)\right\Vert \right\}  & \leq10\sigma_{\max}
\end{align*}
hold uniformly over all $\bm{X},\bm{Y}\in\mathbb{R}^{n\times r},\bm{S}\in\mathbb{R}^{n\times n}$
obeying 
\[
\left\Vert \left[\begin{array}{c}
\bm{X}-\bm{X}^{\star}\\
\bm{Y}-\bm{Y}^{\star}
\end{array}\right]\right\Vert _{2,\infty}\leq\frac{1}{1000\kappa\sqrt{n}}\left\Vert \bm{X}^{\star}\right\Vert ,\qquad\left\Vert \bm{S}-\bm{S}^{\star}\right\Vert \leq C_{\mathrm{S}}\sigma\sqrt{np}
\]
and all $\bm{\Delta}=\left[\begin{array}{c}
\bm{\Delta}_{\bm{X}}\\
\bm{\Delta}_{\bm{Y}}
\end{array}\right]\in\mathbb{R}^{2n\times r}$ lying in the set 
\[
\left\{ \left.\left[\begin{array}{c}
\bm{X}_{1}\\
\bm{Y}_{1}
\end{array}\right]\hat{\bm{H}}-\left[\begin{array}{c}
\bm{X}_{2}\\
\bm{Y}_{2}
\end{array}\right]\,\right|\,\,\left\Vert \left[\begin{array}{c}
\bm{X}_{2}-\bm{X}^{\star}\\
\bm{Y}_{2}-\bm{Y}^{\star}
\end{array}\right]\right\Vert \leq\frac{1}{500\kappa}\left\Vert \bm{X}^{\star}\right\Vert ,\hat{\bm{H}}\coloneqq\arg\min_{\bm{R}\in\mathcal{O}^{r\times r}}\left\Vert \left[\begin{array}{c}
\bm{X}_{1}\\
\bm{Y}_{1}
\end{array}\right]\bm{R}-\left[\begin{array}{c}
\bm{X}_{2}\\
\bm{Y}_{2}
\end{array}\right]\right\Vert _{\mathrm{F}}\right\} .
\]
\end{lemma}

\begin{lemma}[\textbf{Frobenius norm error w.r.t.~$\bm{F}$}]\label{lemma:fro}Set
$\lambda=C_{\lambda}\sigma\sqrt{np}$ for some large enough constant $C_{\lambda}>0$. Suppose that the
sample size obeys $n^{2}p\gg\kappa\mu rn\log^{2}n$ and that the noise
satisfies $\frac{\sigma}{\sigma_{\min}}\sqrt{\frac{n}{p}}\ll\frac{1}{\sqrt{\kappa^{4}\mu r\log n}}$.
If the iterates satisfy (\ref{eq:induction-hypotheses}) in the $t$th
iteration, then with probability at least $1-O(n^{-100})$ , 
\begin{align*}
\left\Vert \bm{F}^{t+1}\bm{H}^{t+1}-\bm{F}^{\star}\right\Vert _{\mathrm{F}} & \leq C_{\mathrm{F}}\left(\frac{\sigma}{\sigma_{\min}}\sqrt{\frac{n}{p}}+\frac{\lambda}{p\sigma_{\min}}\right)\left\Vert \bm{X}^{\star}\right\Vert _{\mathrm{F}}
\end{align*}
holds as long as $0<\eta\ll1/(\kappa^{5/2}\sigma_{\max})$.\end{lemma}

\begin{lemma}[\textbf{Spectral norm error w.r.t.~$\bm{F}$}]\label{lemma:spectral}Set
$\lambda=C_{\lambda}\sigma\sqrt{np}$
for some large enough constant $C_{\lambda}>0$. Suppose that the
sample size obeys $n^{2}p\gg\kappa^{4}\mu^{2}r^{2}n\log^{2}n$ and
that the noise satisfies $\frac{\sigma}{\sigma_{\min}}\sqrt{\frac{n}{p}}\ll\frac{1}{\sqrt{\kappa^{4}\log n}}$.
If the iterates satisfy (\ref{eq:induction-hypotheses}) in the $t$th
iteration, then with probability at least $1-O(n^{-100})$, one has
\begin{align*}
\left\Vert \bm{F}^{t+1}\bm{H}^{t+1}-\bm{F}^{\star}\right\Vert  & \leq C_{\mathrm{op}}\left(\frac{\sigma}{\sigma_{\min}}\sqrt{\frac{n}{p}}+\frac{\lambda}{p\sigma_{\min}}\right)\left\Vert \bm{X}^{\star}\right\Vert 
\end{align*}
holds as long as $0<\eta\ll1/(\kappa^{3}\sigma_{\max}\sqrt{r})$ and
$C_{\mathrm{op}}\gg1$.\end{lemma}

\begin{lemma}[\textbf{Approximate balancedness}]\label{lemma:balancing}Set
$\lambda=C_{\lambda}\sigma\sqrt{np}$
for some large enough constant $C_{\lambda}>0$. Suppose that the
sample size obeys $n^{2}p\gg\kappa^{2}\mu^{2}r^{2}n\log n$ and that
the noise satisfies $\frac{\sigma}{\sigma_{\min}}\sqrt{\frac{n}{p}}\ll\frac{1}{\sqrt{\kappa^{2}\log n}}$.
If the iterates satisfy (\ref{eq:induction-hypotheses}) in the $t$th
iteration, then with probability at least $1-O(n^{-100})$, 
\begin{align*}
\left\Vert \bm{X}^{t+1\top}\bm{X}^{t+1}-\bm{Y}^{t+1\top}\bm{Y}^{t+1}\right\Vert _{\mathrm{F}} & \leq C_{\mathrm{B}}\kappa\eta\left(\frac{\sigma}{\sigma_{\min}}\sqrt{\frac{n}{p}}+\frac{\lambda}{p\sigma_{\min}}\right)\sqrt{r}\sigma_{\max}^{2}\\
\max_{1\leq l\leq2n}\big\|\bm{X}^{t+1,(l)\top}\bm{X}^{t+1,(l)}-\bm{Y}^{t+1,(l)\top}\bm{Y}^{t+1,(l)}\big\|_{\mathrm{F}} & \leq C_{\mathrm{B}}\kappa\eta\left(\frac{\sigma}{\sigma_{\min}}\sqrt{\frac{n}{p}}+\frac{\lambda}{p\sigma_{\min}}\right)\sqrt{r}\sigma_{\max}^{2}
\end{align*}
hold for some sufficiently large constant $C_{\mathrm{B}}\gg C_{\mathrm{op}}^{2}$,
provided that $0<\eta<1/\sigma_{\min}$. \end{lemma}

\begin{lemma}[\textbf{$\ell_{2,\infty}$ norm error of leave-one-out
sequences}]\label{lemma:loo-l-contraction}Set $\lambda=C_{\lambda}\sigma\sqrt{np}$ for some large enough constant
$C_{\lambda}>0$. Suppose that the sample size obeys $n^{2}p\gg\kappa^{4}\mu^{2}r^{2}n\log^{3}n$
and that the noise satisfies $\frac{\sigma}{\sigma_{\min}}\sqrt{\frac{n}{p}}\ll\frac{1}{\sqrt{\kappa^{2}\log n}}$.
If the iterates satisfy (\ref{eq:induction-hypotheses}) in the $t$th
iteration, then with probability at least $1-O(n^{-100})$, 
\[
\max_{1\leq l\leq2n}\big\|\big(\bm{F}^{t+1,(l)}\bm{H}^{t+1,(l)}-\bm{F}^{\star}\big)_{l,\cdot}\big\|_{2}\leq C_{2}\kappa\left(\frac{\sigma}{\sigma_{\min}}\sqrt{\frac{n\log n}{p}}+\frac{\lambda}{p\sigma_{\min}}\right)\left\Vert \bm{F}^{\star}\right\Vert _{2,\infty}
\]
holds, provided that $0<\eta\ll1/(\kappa^{2}\sqrt{r}\sigma_{\max})$,
$C_{\mathrm{op}}\gg1$ and $C_{2}\gg C_{\mathrm{op}}$. \end{lemma}

\begin{lemma}[\textbf{$\ell_{2,\infty}$ norm error of the true
sequence}]\label{lemma:2-infty-contraction}Set $\lambda=C_{\lambda}\sigma\sqrt{np}$ for some large enough constant
$C_{\lambda}>0$. Suppose that $n\geq\mu r$ and that the noise satisfies
$\frac{\sigma}{\sigma_{\min}}\sqrt{\frac{n}{p}}\ll\frac{1}{\sqrt{\kappa^{2}\log n}}$.
If the iterates satisfy (\ref{eq:induction-hypotheses}) and (\ref{eq:induction-hypotheses-LOO})
in the $t$th iteration, then with probability at least $1-O(n^{-99})$,
one has
\begin{align*}
\big\|\bm{F}^{t+1}\bm{H}^{t+1}-\bm{F}^{\star}\big\|_{\mathrm{2,\infty}} & \leq C_{\infty}\kappa\left(\frac{\sigma}{\sigma_{\min}}\sqrt{\frac{n\log n}{p}}+\frac{\lambda}{p\sigma_{\min}}\right)\left\Vert \bm{F}^{\star}\right\Vert _{2,\infty},
\end{align*}
provided that $C_{\infty}\geq5C_{1}+C_{2}$. \end{lemma}

\begin{proof}[Proof of Lemmas \ref{lemma:local-cvx}, \ref{lemma:fro},
\ref{lemma:spectral}, \ref{lemma:balancing}, \ref{lemma:loo-l-contraction}
and \ref{lemma:2-infty-contraction}]As it turns out, Lemmas \ref{lemma:local-cvx},
\ref{lemma:fro}, \ref{lemma:spectral} and \ref{lemma:balancing}
follow immediately from the proofs of \cite[Lemmas 17, 10, 11, 15]{chen2019noisy}
respectively. More specifically, the proofs can be accomplished by
replacing $\bm{E}$ in the proofs therein with $\tilde{\bm{E}}\coloneqq\bm{E}+\bm{S}^{\star}-\bm{S}^{t}$.
To see this, we remark that the only property of the perturbation
matrix $\bm{E}$ utilized in the proofs therein is that $\Vert\mathcal{P}_{\Omega_{\mathsf{obs}}}(\bm{E})\Vert\lesssim\sigma\sqrt{np}$
with probability at least $1-O(n^{-10})$; under our hypotheses, the
new matrix $\tilde{\bm{E}}$ clearly satisfies this property since
\[
\|\mathcal{P}_{\Omega_{\mathsf{obs}}}\bigl(\tilde{\bm{E}}\bigr)\|\leq\left\Vert \mathcal{P}_{\Omega_{\mathsf{obs}}}\left(\bm{E}\right)\right\Vert +\left\Vert \bm{S}^{\star}-\bm{S}^{t}\right\Vert \lesssim\sigma\sqrt{np}.
\]

Regarding Lemma \ref{lemma:loo-l-contraction}, we note that $\bm{S}_{l,\cdot}^{t,(l)}\equiv\bm{S}_{l,\cdot}^{\star}$
by construction. Therefore, the update rule regarding the $l$th row
of $\{\bm{X}_{l,\cdot}^{t,(l)}\}_{t\geq0}$ and $\{\bm{Y}_{l,\cdot}^{t,(l)}\}_{t\geq0}$
is exactly the same as that in the leave-one-out sequence introduced
in \cite{chen2019noisy}. Thus, Lemma \ref{lemma:loo-l-contraction}
follows immediately from the proof of \cite[Lemma 13]{chen2019noisy}.

Finally, the proof of Lemma \ref{lemma:2-infty-contraction} is exactly
the same as the proof of \cite[Lemma 14]{chen2019noisy}.\end{proof}

Next, we justify the hypotheses (\ref{eq:induction-S-op-norm}), (\ref{eq:induction-loo-perturbation})
and (\ref{eq:induction-S-loo}) in the following three lemmas, which
require more careful analysis of the properties about $\{\bm{S}^{t}\}$.

\begin{lemma}[\textbf{Spectral norm error w.r.t.~$\bm{S}$}]\label{lemma:S-induction}Set $\tau=C_{\tau}\sigma\sqrt{\log n}$
for some large enough constant $C_{\tau}>0$. Suppose that the
sample size obeys $n^{2}p\gg\kappa^{4}\mu rn\log n$, the noise satisfies $\frac{\sigma}{\sigma_{\min}}\sqrt{\frac{n}{p}}\ll1/\sqrt{\kappa^2\log n}$, the outlier
fraction satisfies $\rho_{\mathsf{s}}\leq\rho_{\mathsf{aug}}\ll1/\sqrt{\kappa^{5}\mu r\log^2 n}$
and $n^{2}p\rho_{\mathsf{aug}}\gg\mu nr\log^2 n$. If the iterates satisfy (\ref{eq:induction-op}) and (\ref{eq:induction-two-to-infty})
in the $(t+1)$-th iteration, then with probability at least $1-O(n^{-100})$,
\[
\left\Vert \bm{S}^{t+1}-\bm{S}^{\star}\right\Vert \leq C_{\mathrm{S}}\sigma\sqrt{np}
\]
holds for some constant $C_{\mathrm{S}}>0$ that does not rely on
the choice of other constants.\end{lemma}\begin{proof}See Appendix
\ref{subsec:Proof-of-Lemma-S-induction}.\end{proof}

\begin{lemma}[\textbf{Leave-one-out perturbation w.r.t.~$\bm{F}$}]\label{lemma:LOO-F-perturbation}Set
$\lambda=C_{\lambda}\sigma\sqrt{np}$
for some large enough constant $C_{\lambda}>0$. Suppose that the
sample size obeys $n^{2}p\gg\kappa^{4}\mu^{2}r^{2}n\log^{4}n$, the noise satisfies $\frac{\sigma}{\sigma_{\min}}\sqrt{\frac{n}{p}}\ll1/\sqrt{\kappa^{4}\mu r\log n}$, the outlier fraction satisfies $\rho_{\mathsf{s}}\leq\rho_{\mathsf{aug}}\ll1/(\kappa^{3}\mu r\log n)$ 
and $n^{2}p\rho_{\mathsf{aug}}\gg\mu rn\log n$. If the iterates satisfy
(\ref{eq:induction-hypotheses}) and (\ref{eq:induction-hypotheses-LOO})
in the $t$th iteration, then with probability at least $1-O(n^{-100})$,
\[
\max_{1\leq l\leq2n}\big\|\bm{F}^{t+1}\bm{H}^{t+1}-\bm{F}^{t+1,(l)}\bm{R}^{t+1,(l)}\big\|_{\mathrm{F}}\leq C_{1}\left(\frac{\sigma}{\sigma_{\min}}\sqrt{\frac{n\log n}{p}}+\frac{\lambda}{p\sigma_{\min}}\right)\left\Vert \bm{F}^{\star}\right\Vert _{2,\infty}
\]
holds for some constant $C_{1}>0$, provided that $\eta\ll1/(n\kappa^{2}\sigma_{\max})$
and $C_{1}\gg C_{3}$.\end{lemma}\begin{proof}See Appendix \ref{subsec:Proof-of-Lemma-LOO-F-perturbation}.\end{proof}

\begin{lemma}[\textbf{Leave-one-out perturbation w.r.t.~$\bm{S}$}]\label{lemma:LOO-S-perturbation}Set $\tau=C_{\tau}\sigma\sqrt{\log n}$
for some large enough constant $C_{\tau}>0$. Suppose that the
sample size satisfies $n^{2}p\gg\kappa^{4}\mu^{2}r^{2}n\log n$, the noise obeys $\frac{\sigma}{\sigma_{\min}}\sqrt{\frac{n}{p}}\ll1/\sqrt{\kappa^2\log n}$ and the outlier
fraction satisfies $\rho_{\mathsf{s}}\leq\rho_{\mathsf{aug}}\ll1/\kappa$.
If the iterates satisfy (\ref{eq:induction-op}), (\ref{eq:induction-two-to-infty})
and (\ref{eq:induction-loo-perturbation}) in the $(t+1)$-th iteration,
then with probability at least $1-O(n^{-100})$, 
\begin{align*}
\max_{1\leq l\leq n}\big\|\mathcal{P}_{-l,\cdot}\big(\bm{S}^{t+1}-\bm{S}^{t+1,(l)}\big)\big\|_{\mathrm{F}} & \leq C_{3}\frac{\sigma}{\sigma_{\min}}\sqrt{n\log n}\left\Vert \bm{F}^{\star}\right\Vert \left\Vert \bm{F}^{\star}\right\Vert _{2,\infty},\\
\max_{n<l\leq2n}\big\|\mathcal{P}_{\cdot,-(l-n)}\big(\bm{S}^{t+1}-\bm{S}^{t+1,(l)}\big)\big\|_{\mathrm{F}} & \leq C_{3}\frac{\sigma}{\sigma_{\min}}\sqrt{n\log n}\left\Vert \bm{F}^{\star}\right\Vert \left\Vert \bm{F}^{\star}\right\Vert _{2,\infty}
\end{align*}
hold for some constant $C_3$ that does not rely on the choice of other constants.\end{lemma}\begin{proof}See Appendix
\ref{subsec:Proof-of-Lemma-LOO-S-perturbation}.\end{proof}

Finally, it remains to justify (\ref{eq:induction-descent}), which
is a straightforward consequence from standard gradient descent theory
and implies the existence of a point with nearly zero gradient.

\begin{lemma}[\textbf{Monotonicity of the function values}]\label{lemma:function-value-decreasing}Set
$\lambda=C_{\lambda}\sigma\sqrt{np}$
for some large enough constant $C_{\lambda}>0$. Suppose that the
noise satisfies $\frac{\sigma}{\sigma_{\min}}\sqrt{\frac{n}{p}}\ll1$.
If the iterates satisfy (\ref{eq:induction-hypotheses}) in the $t$th
iteration, then with probability at least $1-O(n^{-100})$, 
\[
F\left(\bm{X}^{t+1},\bm{Y}^{t+1};\bm{S}^{t+1}\right)\leq F\left(\bm{X}^{t},\bm{Y}^{t};\bm{S}^{t}\right)-\frac{\eta}{2}\left\Vert \nabla f\left(\bm{X}^{t},\bm{Y}^{t};\bm{S}^{t}\right)\right\Vert _{\mathrm{F}}^{2}
\]
holds as long as $\eta\ll1/(\kappa n\sigma_{\max})$. \end{lemma}\begin{proof}See
Appendix~\ref{sec:Proof-of-Lemma-funcation-value}. \end{proof}

\subsection{Proof of Lemma \ref{lemma:small-gradient}\label{subsec:Proof-of-Lemma-small-gradient}}

Summing (\ref{eq:induction-descent}) over $t=1,\ldots,t_{0}$ gives
\begin{align*}
F\left(\bm{X}^{t_{0}},\bm{Y}^{t_{0}},\bm{S}^{t_{0}}\right) & \leq F\left(\bm{X}^{0},\bm{Y}^{0},\bm{S}^{0}\right)-\frac{\eta}{2}\sum_{t=0}^{t_{0}-1}\left\Vert \nabla f\left(\bm{X}^{t},\bm{Y}^{t};\bm{S}^{t}\right)\right\Vert _{\mathrm{F}}^{2},
\end{align*}
which further implies 
\begin{align}
\min_{0\leq t<t_{0}}\left\Vert \nabla f\left(\bm{X}^{t},\bm{Y}^{t};\bm{S}^{t}\right)\right\Vert _{\mathrm{F}}^{2} & \leq\frac{1}{t_{0}}\sum_{t=0}^{t_{0}-1}\left\Vert \nabla f\left(\bm{X}^{t},\bm{Y}^{t};\bm{S}^{t}\right)\right\Vert _{\mathrm{F}}^{2}\nonumber \\
 & \leq\frac{2}{\eta t_{0}}\left[F\left(\bm{X}^{\star},\bm{Y}^{\star},\bm{S}^{\star}\right)-F\left(\bm{X}^{t_{0}},\bm{Y}^{t_{0}},\bm{S}^{t_{0}}\right)\right].\label{eq:min-grad-f-UB}
\end{align}
Here, the last inequality results from our choice $(\bm{X}^{0},\bm{Y}^{0},\bm{S}^{0})=(\bm{X}^{\star},\bm{Y}^{\star},\bm{S}^{\star})$.
Therefore, it suffices to control $F\left(\bm{X}^{\star},\bm{Y}^{\star},\bm{S}^{\star}\right)-F\left(\bm{X}^{t_{0}},\bm{Y}^{t_{0}},\bm{S}^{t_{0}}\right)$.

We first decompose the difference into 
\begin{align*}
 & F\left(\bm{X}^{\star},\bm{Y}^{\star},\bm{S}^{\star}\right)-F\left(\bm{X}^{t_{0}},\bm{Y}^{t_{0}},\bm{S}^{t_{0}}\right)=f\left(\bm{X}^{\star},\bm{Y}^{\star};\bm{S}^{\star}\right)+\frac{\tau}{p}\left\Vert \bm{S}^{\star}\right\Vert _{1}-f\left(\bm{X}^{t_{0}},\bm{Y}^{t_{0}};\bm{S}^{t_{0}}\right)-\frac{\tau}{p}\left\Vert \bm{S}^{t_{0}}\right\Vert _{1}\\
 & \quad=\underbrace{\vphantom{\frac{\tau}{p}}f\left(\bm{X}^{\star},\bm{Y}^{\star};\bm{S}^{\star}\right)-f\left(\bm{X}^{t_{0}},\bm{Y}^{t_{0}};\bm{S}^{\star}\right)}_{\eqqcolon \Delta_{1}}+\underbrace{\vphantom{\frac{\tau}{p}}f\left(\bm{X}^{t_{0}},\bm{Y}^{t_{0}};\bm{S}^{\star}\right)-f\left(\bm{X}^{t_{0}},\bm{Y}^{t_{0}};\bm{S}^{t_{0}}\right)}_{\eqqcolon \Delta_{2}}+\underbrace{\frac{\tau}{p}\left\Vert \bm{S}^{\star}\right\Vert _{1}-\frac{\tau}{p}\left\Vert \bm{S}^{t_{0}}\right\Vert _{1}}_{\eqqcolon \Delta_{3}}.
\end{align*}
In what follows, we shall bound $\Delta_{1},\Delta_{2}$ and $\Delta_{3}$
separately.
\begin{enumerate}
\item In view of the proof of \cite[Lemma 9]{chen2019noisy}, we have 
\[
\left|\Delta_{1}\right|\lesssim r\kappa^{2}\left(\frac{\lambda}{p}\right)^{2},
\]
provided that $\frac{\sigma}{\sigma_{\min}}\sqrt{\frac{n}{p}}\ll\frac{1}{\sqrt{\kappa^{4}\mu r\log n}}$.
\item When it comes to $\Delta_{2}$, we deduce that 
\begin{align}
\left|\Delta_{2}\right| & =\left|\frac{1}{2}\left\Vert \mathcal{P}_{\Omega_{\mathsf{obs}}}\left(\bm{X}^{t_{0}}\bm{Y}^{t_{0}\top}+\bm{S}^{\star}-\bm{M}\right)\right\Vert _{\mathrm{F}}^{2}+\frac{\lambda}{2p}\|\bm{F}^{t_{0}}\|_{\mathrm{F}}^{2}-\frac{1}{2}\left\Vert \mathcal{P}_{\Omega_{\mathsf{obs}}}\left(\bm{X}^{t_{0}}\bm{Y}^{t_{0}\top}+\bm{S}^{t_{0}}-\bm{M}\right)\right\Vert _{\mathrm{F}}^{2}-\frac{\lambda}{2p}\|\bm{F}^{t_{0}}\|_{\mathrm{F}}^{2}\right|\nonumber \\
 & =\left|\left\langle \mathcal{P}_{\Omega_{\mathsf{obs}}}\left(\bm{X}^{t_{0}}\bm{Y}^{t_{0}\top}+\bm{S}^{t_{0}}-\bm{M}\right),\bm{S}^{\star}-\bm{S}^{t_{0}}\right\rangle +\frac{1}{2}\left\Vert \bm{S}^{\star}-\bm{S}^{t_{0}}\right\Vert _{\mathrm{F}}^{2}\right|\nonumber \\
 & \leq\left\Vert \mathcal{P}_{\Omega_{\mathsf{obs}}}\left(\bm{X}^{t_{0}}\bm{Y}^{t_{0}\top}+\bm{S}^{t_{0}}-\bm{M}\right)\right\Vert _{\mathrm{F}}\left\Vert \bm{S}^{\star}-\bm{S}^{t_{0}}\right\Vert _{\mathrm{F}}+\frac{1}{2}\left\Vert \bm{S}^{\star}-\bm{S}^{t_{0}}\right\Vert _{\mathrm{F}}^{2},\label{eq:Delta_2}
\end{align}
where the last step arises from the elementary inequality $\langle\bm{A},\bm{B}\rangle\leq\|\bm{A}\|_{\mathrm{F}}\|\bm{B}\|_{\mathrm{F}}$
and the triangle inequality. It is straightforward to derive from
(\ref{eq:induction-S-op-norm}) that 
\[
\Vert\bm{S}^{\star}-\bm{S}^{t_{0}}\Vert_{\mathrm{F}}\leq\sqrt{n}\Vert\bm{S}^{\star}-\bm{S}^{t_{0}}\Vert\lesssim\sigma n\sqrt{p}.
\]
Moving on to the first term in (\ref{eq:Delta_2}), one has by the
triangle inequality 
\begin{align*}
\left\Vert \mathcal{P}_{\Omega_{\mathsf{obs}}}\left(\bm{X}^{t_{0}}\bm{Y}^{t_{0}\top}+\bm{S}^{t_{0}}-\bm{M}\right)\right\Vert _{\mathrm{F}} & \leq\left\Vert \mathcal{P}_{\Omega_{\mathsf{obs}}}\left(\bm{X}^{t_{0}}\bm{Y}^{t_{0}\top}-\bm{X}^{\star}\bm{Y}^{\star\top}\right)\right\Vert _{\mathrm{F}}+\left\Vert \mathcal{P}_{\Omega_{\mathsf{obs}}}\left(\bm{S}^{t_{0}}-\bm{S}^{\star}\right)\right\Vert _{\mathrm{F}}+\left\Vert \mathcal{P}_{\Omega_{\mathsf{obs}}}\left(\bm{E}\right)\right\Vert _{\mathrm{F}}\\
 & \lesssim\left\Vert \mathcal{P}_{\Omega_{\mathsf{obs}}}\left(\bm{X}^{t_{0}}\bm{Y}^{t_{0}\top}-\bm{X}^{\star}\bm{Y}^{\star\top}\right)\right\Vert _{\mathrm{F}}+\sigma n\sqrt{p},
\end{align*}
where the last bound follows from Lemma~\ref{lemma:noise-bound}
and the bound above $\|\mathcal{P}_{\Omega_{\mathsf{obs}}}(\bm{S}^{t_{0}}-\bm{S}^{\star})\|_{\mathrm{F}}\leq\|\bm{S}^{t_{0}}-\bm{S}^{\star}\|_{\mathrm{F}}\lesssim\sigma n\sqrt{p}$.
We can further decompose $\|\mathcal{P}_{\Omega_{\mathsf{obs}}}(\bm{X}^{t_{0}}\bm{Y}^{t_{0}\top}-\bm{X}^{\star}\bm{Y}^{\star\top})\|_{\mathrm{F}}$
into 
\begin{align}
\left\Vert \mathcal{P}_{\Omega_{\mathsf{obs}}}\left(\bm{X}^{t_{0}}\bm{Y}^{t_{0}\top}-\bm{X}^{\star}\bm{Y}^{\star\top}\right)\right\Vert _{\mathrm{F}} & \leq\left\Vert \mathcal{P}_{\Omega_{\mathsf{obs}}}\left[\left(\bm{X}^{t_{0}}\bm{H}^{t_{0}}-\bm{X}^{\star}\right)\bm{Y}^{\star\top}\right]\right\Vert _{\mathrm{F}}+\left\Vert \mathcal{P}_{\Omega_{\mathsf{obs}}}\left[\bm{X}^{t_{0}}\bm{H}^{t_{0}}\left(\bm{Y}^{t_{0}}\bm{H}^{t_{0}}-\bm{Y}^{\star}\right)^{\top}\right]\right\Vert _{\mathrm{F}}\nonumber \\
 & \overset{(\text{i})}{\lesssim}\sqrt{\kappa p}\left\Vert \left(\bm{X}^{t_{0}}\bm{H}^{t_{0}}-\bm{X}^{\star}\right)\bm{Y}^{\star\top}\right\Vert _{\mathrm{F}}+\sqrt{\kappa p}\left\Vert \bm{X}^{t_{0}}\bm{H}^{t_{0}}\left(\bm{Y}^{t_{0}}\bm{H}^{t_{0}}-\bm{Y}^{\star}\right)^{\top}\right\Vert _{\mathrm{F}}\nonumber \\
 & \lesssim\sqrt{\kappa p}\left\Vert \bm{X}^{t_{0}}\bm{H}^{t_{0}}-\bm{X}^{\star}\right\Vert _{\mathrm{F}}\left\Vert \bm{Y}^{\star}\right\Vert +\sqrt{\kappa p}\left\Vert \bm{X}^{t_{0}}\bm{H}^{t_{0}}\right\Vert \left\Vert \bm{Y}^{t_{0}}\bm{H}^{t_{0}}-\bm{Y}^{\star}\right\Vert _{\mathrm{F}}\nonumber \\
 & \overset{(\text{ii})}{\lesssim}\sqrt{\kappa p}\left(\frac{\sigma}{\sigma_{\min}}\sqrt{\frac{n}{p}}+\frac{\lambda}{p\sigma_{\min}}\right)\left\Vert \bm{X}^{\star}\right\Vert _{\mathrm{F}}\left\Vert \bm{X}^{\star}\right\Vert \lesssim\kappa^{3/2}\frac{\lambda}{\sqrt{p}}\sqrt{r}.\label{eq:P_Omega_tangent}
\end{align}
Here, the relation (i) utilizes Lemma \ref{lemma:sampling-operator-norm-in-T},
and the facts that $(\bm{X}^{t_{0}}\bm{H}^{t_{0}}-\bm{X}^{\star})\bm{Y}^{\star\top}\in T^{\star}$
and that $\bm{X}^{t_{0}}\bm{H}^{t_{0}}(\bm{Y}^{t_{0}}\bm{H}^{t_{0}}-\bm{Y}^{\star})^{\top}\in T^{t_{0}}$,
where $T^{t_{0}}$ denotes the tangent space at $\bm{X}^{t_{0}}\bm{Y}^{t_{0}\top}$.
In addition, the last line (ii) holds because of the hypothesis~(\ref{eq:induction-fro})
and the simple fact $\|\bm{X}^{t_{0}}\bm{H}^{t_{0}}\|\leq2\|\bm{X}^{\star}\|$, which is an immediate consequence of the hypothesis (\ref{eq:induction-op}) provided that $\frac{\sigma}{\sigma_{\min}}\sqrt{\frac{n}{p}}\ll1$.
Collecting the bounds together, we arrive at 
\[
\left|\Delta_{2}\right|\lesssim\left(\kappa^{3/2}\frac{\lambda}{\sqrt{p}}\sqrt{r}+\sigma n\sqrt{p}\right)\cdot\sigma n\sqrt{p}+\sigma^{2}n^{2}p+\frac{\lambda}{p}\cdot\kappa\frac{\lambda}{p}r\lesssim\sigma^{2}n^{2}p,
\]
with the proviso that $np\gg\kappa^{3}r$.
\item In the end, we have the following upper bound on $\Delta_{3}$: 
\[
\left|\Delta_{3}\right|\leq\frac{\tau}{p}\left\Vert \bm{S}^{t_{0}}-\bm{S}^{\star}\right\Vert _{1}\leq\frac{\tau}{p}n\left\Vert \bm{S}^{t_{0}}-\bm{S}^{\star}\right\Vert _{\mathrm{F}}\lesssim\frac{1}{p}\frac{\lambda}{\sqrt{np/\log n}}\sigma n^{2}\sqrt{p}\asymp\frac{\lambda}{p}\sigma n^{3/2}\sqrt{\log n},
\]
where we have made use of the elementary fact that $\|\bm{A}\|_{1}\leq n\|\bm{A}\|_{\mathrm{F}}$
for all $\bm{A}\in\mathbb{R}^{n\times n}$. 
\end{enumerate}
Putting the above bounds together, one can reach 
\[
\left|F\left(\bm{X}^{\star},\bm{Y}^{\star},\bm{S}^{\star}\right)-F\left(\bm{X}^{t_{0}},\bm{Y}^{t_{0}},\bm{S}^{t_{0}}\right)\right|\lesssim r\kappa^{2}\left(\frac{\lambda}{p}\right)^{2}+\sigma^{2}n^{2}p+\frac{\lambda}{p}\sigma n^{3/2}\sqrt{\log n}\lesssim n\left(\frac{\lambda}{p}\right)^{2}\sqrt{\log n}
\]
as long as $n\gg\kappa^{2}r$ and $\lambda\asymp\sigma\sqrt{np}$.
Substitution into (\ref{eq:min-grad-f-UB}) allows us to conclude
that 
\[
\min_{0\leq t\leq t_{0}}\left\Vert \nabla f\left(\bm{X}^{t},\bm{Y}^{t};\bm{S}^{t}\right)\right\Vert _{\mathrm{F}}\lesssim\sqrt{\frac{1}{\eta t_{0}}n\left(\frac{\lambda}{p}\right)^{2}\sqrt{\log n}}\leq\frac{1}{n^{20}}\frac{\lambda}{p}\sqrt{\sigma_{\min}},
\]
provided that $\eta\asymp1/(n\kappa^{3}\sigma_{\max})$, $t_{0}\geq n^{47}$
and $n\geq\kappa$.

\subsection{Proof of Lemma \ref{lemma:S-induction}\label{subsec:Proof-of-Lemma-S-induction}}

In view of the definitions $\Omega^{\star}=\{(i,j):S_{ij}^{\star}\neq0\}\subseteq\Omega_{\mathsf{aug}}\subseteq\Omega_{\mathsf{obs}}$
and $\bm{S}^{t+1}=\mathcal{S}_{\tau}[\mathcal{P}_{\Omega_{\mathsf{obs}}}(\bm{L}^{\star}+\bm{S}^{\star}+\bm{E}-\bm{X}^{t+1}\bm{Y}^{t+1\top})]$,
we have the decomposition 
\begin{align}
 & \bm{S}^{t+1}-\bm{S}^{\star}=\mathcal{P}_{\Omega_{\mathsf{aug}}}\left(\bm{S}^{t+1}\right)-\mathcal{P}_{\Omega_{\mathsf{aug}}}\left(\bm{S}^{\star}\right)+\mathcal{P}_{\Omega_{\mathsf{aug}}^{\mathrm{c}}}\left(\bm{S}^{t+1}\right)\nonumber \\
 & \quad=\underbrace{\mathcal{S}_{\tau}\left[\mathcal{P}_{\Omega_{\mathsf{aug}}}\left(\bm{X}^{\star}\bm{Y}^{\star\top}+\bm{S}^{\star}+\bm{E}-\bm{X}^{t+1}\bm{Y}^{t+1\top}\right)\right]-\mathcal{P}_{\Omega_{\mathsf{aug}}}\left(\bm{S}^{\star}\right)}_{\eqqcolon \bm{A}^{t+1}}+\underbrace{\mathcal{S}_{\tau}\left[\mathcal{P}_{\Omega_{\mathsf{obs}}\backslash\Omega_{\mathsf{aug}}}\left(\bm{X}^{\star}\bm{Y}^{\star\top}+\bm{E}-\bm{X}^{t+1}\bm{Y}^{t+1\top}\right)\right]}_{\eqqcolon \bm{B}^{t+1}}.\label{eq:decomposition-St-Sstar}
\end{align}
We shall control $\|\bm{A}^{t+1}\|$ and $\|\bm{B}^{t+1}\|$ separately.
\begin{enumerate}
\item We begin by controlling the size of $\bm{A}^{t+1}$, which can be
further decomposed into 
\begin{align}
\bm{A}^{t+1} & =\mathcal{P}_{\Omega_{\mathsf{aug}}}\left(\bm{E}\right)+\underbrace{\mathcal{S}_{\tau}\left[\mathcal{P}_{\Omega_{\mathsf{aug}}}\left(\bm{S}^{\star}+\bm{E}\right)\right]-\mathcal{P}_{\Omega_{\mathsf{aug}}}\left(\bm{S}^{\star}+\bm{E}\right)}_{\eqqcolon \bm{A}_{1}}\nonumber \\
 & \quad+\underbrace{\mathcal{S}_{\tau}\left[\mathcal{P}_{\Omega_{\mathsf{aug}}}\left(\bm{X}^{\star}\bm{Y}^{\star\top}-\bm{X}^{t+1}\bm{Y}^{t+1\top}+\bm{S}^{\star}+\bm{E}\right)\right]-\mathcal{S}_{\tau}\left[\mathcal{P}_{\Omega_{\mathsf{aug}}}\left(\bm{S}^{\star}+\bm{E}\right)\right]}_{\eqqcolon \bm{A}_{2}^{t+1}}.\label{eq:defn-A_t+1}
\end{align}
First of all, we know that $\Vert\mathcal{P}_{\Omega_{\mathsf{aug}}}(\bm{E})\Vert\lesssim\sigma\sqrt{np\rho_{\mathsf{aug}}}\leq\sigma\sqrt{np}$,
as long as $n^{2}p\rho_{\mathsf{aug}}\gg n\log^{2}n$. This arises
from standard concentration results for the spectral norm of sub-Gaussian
random matrices (cf.~Lemma~\ref{lemma:noise-bound}). Regarding~$\bm{A}_{1}$,
we know from the definition of $\mathcal{S}_{\tau}(\cdot)$ that $\Vert\bm{A}_{1}\Vert_{\infty}\leq\tau$.
More precisely, we have 
\[
\left(\bm{A}_{1}\right)_{ij}=\begin{cases}
-\tau & \text{if}\ S_{ij}^{\star}+E_{ij}\geq\tau,\\
-S_{ij}^{\star}-E_{ij} & \text{if}\ -\tau<S_{ij}^{\star}+E_{ij}<\tau,\\
\tau & \text{if}\ S_{ij}^{\star}+E_{ij}\leq-\tau.
\end{cases}
\]
Recall from Assumption~\ref{assumption:random-sign} that $\bm{S}^{\star}$
has random signs on its support $\Omega^{\star}\subseteq\Omega_{\mathsf{aug}}$
and $E_{ij}$ is symmetric around zero. It then follows from standard
concentration results for the spectral norm of matrices with i.i.d.~entries
that 
\[
\left\Vert \bm{A}_{1}\right\Vert \lesssim\tau\sqrt{np\rho_{\mathsf{aug}}}=C_\tau\sigma\sqrt{np\rho_{\mathsf{aug}}\log n},
\]
provided that $n^{2}p\rho_{\mathsf{aug}}\gg n\log^{2}n$. Moving on to $\bm{A}_{2}^{t+1}$, since it is supported
on $\Omega_{\mathsf{aug}}$, we can further decompose $\|\bm{A}_{2}^{t+1}\|$
into 
\begin{align}
\left\Vert \bm{A}_{2}^{t+1}\right\Vert =\left\Vert \mathcal{P}_{\Omega_{\mathsf{aug}}}\left(\bm{A}_{2}^{t+1}\right)\right\Vert  & \leq p\rho_{\mathsf{aug}}\left\Vert \bm{A}_{2}^{t+1}\right\Vert +\left\Vert \mathcal{P}_{\Omega_{\mathsf{aug}}}\left(\bm{A}_{2}^{t+1}\right)-p\rho_{\mathsf{aug}}\bm{A}_{2}^{t+1}\right\Vert .\label{eq:A-2-control-1}
\end{align}
Invoking Lemma \ref{lemma:spectral-projection} with $\bm{A}=\bm{A}_{2}^{t+1}$,
$\bm{B}=\bm{I}_{n}$ and $\rho_{0}=p\rho_{\mathsf{aug}}$, we have
\begin{equation}
\left\Vert \mathcal{P}_{\Omega_{\mathsf{aug}}}\left(\bm{A}_{2}^{t+1}\right)-p\rho_{\mathsf{aug}}\bm{A}_{2}^{t+1}\right\Vert \leq C\sqrt{np\rho_{\mathsf{aug}}}\left\Vert \bm{A}_{2}^{t+1}\right\Vert _{2,\infty},\label{eq:A-2-control-2}
\end{equation}
with the proviso that $n^{2}p\rho_{\mathsf{aug}}\gg n\log n$. Combine
the above bounds to reach 
\begin{align*}
\left\Vert \bm{A}_{2}^{t+1}\right\Vert  & \leq p\rho_{\mathsf{aug}}\left\Vert \bm{A}_{2}^{t+1}\right\Vert +C\sqrt{np\rho_{\mathsf{aug}}}\left\Vert \bm{A}_{2}^{t+1}\right\Vert _{2,\infty}\leq\frac{1}{2}\left\Vert \bm{A}_{2}^{t+1}\right\Vert +C\sqrt{np\rho_{\mathsf{aug}}}\left\Vert \bm{A}_{2}^{t+1}\right\Vert _{2,\infty},
\end{align*}
as soon as $\rho_{\mathsf{s}}\leq\rho_{\mathsf{aug}}\leq1/2$. We
are then in need of an upper bound on $\|\bm{A}_{2}^{t+1}\|_{2,\infty}$,
which is supplied in the following fact.

\begin{fact}\label{fact:A_2_infty} Suppose that $n^{2}p\rho_{\mathsf{aug}}\gg\mu rn\log n$ and $\frac{\sigma}{\sigma_{\min}}\sqrt{\frac{n\log n}{p}}\ll1/\kappa$. Then with probability exceeding $1-O(n^{-100})$,
one has 
\[
\left\Vert \bm{A}_{2}^{t+1}\right\Vert _{2,\infty}\leq\sqrt{40\kappa p\rho_{\mathsf{aug}}}\left(C_{\infty}\kappa+2C_{\mathrm{op}}\right)\left(\frac{\sigma}{\sigma_{\min}}\sqrt{\frac{n\log n}{p}}+\frac{\lambda}{p\sigma_{\min}}\right)\left\Vert \bm{F}^{\star}\right\Vert _{2,\infty}\left\Vert \bm{X}^{\star}\right\Vert.
\]
\end{fact}With the help of Fact~\ref{fact:A_2_infty}, we can continue
the upper bound as follows
\[
\left\Vert \bm{A}_{2}^{t+1}\right\Vert \leq2C\sqrt{np\rho_{\mathsf{aug}}}\left\Vert \bm{A}_{2}^{t+1}\right\Vert _{2,\infty}\lesssim \left(C_{\infty}+C_{\mathrm{op}}\right)\sqrt{ np}\kappa^{3/2}\rho_{\mathsf{aug}}\frac{\sigma}{\sigma_{\min}}\sqrt{n\log n}\left\Vert \bm{F}^{\star}\right\Vert _{2,\infty}\left\Vert \bm{X}^{\star}\right\Vert .
\]
All in all, we obtain the following bound on $\bm{A}^{t+1}$: 
\begin{align*}
\|\bm{A}^{t+1}\| & \leq\|\mathcal{P}_{\Omega_{\mathsf{aug}}}(\bm{E})\|+\|\bm{A}_{1}\|+\left\Vert \bm{A}_{2}^{t+1}\right\Vert \\
 & \lesssim\sigma\sqrt{np}+C_\tau\sigma\sqrt{np\rho_{\mathsf{aug}}\log n}+\left(C_{\infty}+C_{\mathrm{op}}\right)\sqrt{np}\kappa^{3/2}\rho_{\mathsf{aug}}\frac{\sigma}{\sigma_{\min}}\sqrt{n\log n}\left\Vert \bm{F}^{\star}\right\Vert _{2,\infty}\left\Vert \bm{X}^{\star}\right\Vert \\
 & \leq C_{S}\sigma\sqrt{np},
\end{align*}
with the proviso that $\rho_{\mathsf{s}}\leq\rho_{\mathsf{aug}}\ll1/\sqrt{\kappa^{5}\mu r\log^2 n}.$
Here, the last line uses the incoherence assumption $\|\bm{F}^{\star}\|_{2,\infty}\leq\sqrt{\mu r/n}\Vert\bm{X}^\star\Vert$ (cf.~(\ref{eq:incoherence-X})).
\item When it comes to $\bm{B}^{t+1}$, we first note that 
\begin{align}
 & \left\Vert \bm{X}^{\star}\bm{Y}^{\star\top}-\bm{X}^{t+1}\bm{Y}^{t+1\top}\right\Vert _{\infty}=\left\Vert \left(\bm{X}^{\star}-\bm{X}^{t+1}\bm{H}^{t+1}\right)\bm{Y}^{\star\top}+\bm{X}^{t+1}\bm{H}^{t+1}\left(\bm{Y}^{\star}-\bm{Y}^{t+1}\bm{H}^{t+1}\right)^{\top}\right\Vert _{\infty}\nonumber \\
 & \quad\leq\left\Vert \bm{X}^{\star}-\bm{X}^{t+1}\bm{H}^{t+1}\right\Vert _{2,\infty}\left\Vert \bm{Y}^{\star}\right\Vert _{2,\infty}+\left\Vert \bm{X}^{t+1}\bm{H}^{t+1}\right\Vert _{2,\infty}\left\Vert \bm{Y}^{\star}-\bm{Y}^{t+1}\bm{H}^{t+1}\right\Vert _{2,\infty}\nonumber \\
 & \quad\leq3C_{\infty}\kappa\left(\frac{\sigma}{\sigma_{\min}}\sqrt{\frac{n\log n}{p}}+\frac{\lambda}{p\sigma_{\min}}\right)\left\Vert \bm{F}^{\star}\right\Vert _{2,\infty}^{2}\leq3C_{\infty}\kappa\left(\frac{\sigma}{\sigma_{\min}}\sqrt{\frac{n\log n}{p}}+\frac{\lambda}{p\sigma_{\min}}\right)\frac{\mu r}{n}\sigma_{\max}.\label{eq:X-star-Y-star-X-t-Y-t-infty-norm}
\end{align}
Here, we have plugged in~(\ref{eq:induction-two-to-infty}) for the
$(t+1)$-th iteration and its immediate consequence $\|\bm{X}^{t+1}\bm{H}^{t+1}\|_{2,\infty}\leq\|\bm{F}^{t+1}\|_{2,\infty}\leq2\|\bm{F}^{\star}\|_{2,\infty}$,
as long as $\frac{\sigma}{\sigma_{\min}}\sqrt{\frac{n\log n}{p}}\ll1/\kappa$.
As a result, for all $(i,j)$ we have 
\begin{align*}
\left|\left(\bm{M}-\bm{X}^{t+1}\bm{Y}^{t+1\top}\right)_{ij}\right| & =\left|\left(\bm{X}^{\star}\bm{Y}^{\star\top}+\bm{E}-\bm{X}^{t+1}\bm{Y}^{t+1\top}\right)_{ij}\right|\leq\big|E_{ij}\big|+\left\Vert \bm{X}^{\star}\bm{Y}^{\star\top}-\bm{X}^{t+1}\bm{Y}^{t+1\top}\right\Vert _{\infty}\\
 & \overset{\text{(i)}}{\leq}\big|E_{ij}\big|+3C_{\infty}\kappa\left(\frac{\sigma}{\sigma_{\min}}\sqrt{\frac{n\log n}{p}}+\frac{\lambda}{p\sigma_{\min}}\right)\frac{\mu r}{n}\sigma_{\max}\\
 & \overset{\text{(ii)}}{\leq}C_{\lambda}\sigma\sqrt{\log n}=\tau.
\end{align*}
Here, the inequality (i) comes from~(\ref{eq:X-star-Y-star-X-t-Y-t-infty-norm}),
and the last line (ii) relies on the property of sub-Gaussian random
variables (namely, $|E_{ij}|\leq\tau/2$ with probability exceeding $1-O(n^{-102})$) and
the sample size condition $n^{2}p\gg\kappa^{4}\mu^{2}r^{2}n\log n$.
An immediate consequence is that with probability at least $1-O(n^{-100})$,
\begin{equation}
\bm{B}^{t+1}=\mathcal{S}_{\tau}\left[\mathcal{P}_{\Omega_{\mathsf{obs}}\backslash\Omega_{\mathsf{aug}}}\left(\bm{X}^{\star}\bm{Y}^{\star\top}+\bm{E}-\bm{X}^{t+1}\bm{Y}^{t+1\top}\right)\right]=\bm{0}.\label{eq:shrink-to-zero-1}
\end{equation}
\end{enumerate}
Substituting the above two bounds into (\ref{eq:decomposition-St-Sstar}),
we conclude that $\left\Vert \bm{S}^{t+1}-\bm{S}^{\star}\right\Vert \leq C_{S}\sigma\sqrt{np}$
as claimed.

\begin{proof}[Proof of Fact~\ref{fact:A_2_infty}]In view of the
definition of $\bm{A}^{t+1}$ in~(\ref{eq:defn-A_t+1}), we have
\begin{align*}
\left\Vert \bm{A}_{2}^{t+1}\right\Vert _{2,\infty} & \leq\left\Vert \mathcal{P}_{\Omega_{\mathsf{aug}}}\left(\bm{X}^{t+1}\bm{Y}^{t+1\top}-\bm{X}^{\star}\bm{Y}^{\star\top}\right)\right\Vert _{2,\infty},
\end{align*}
where we use the non-expansiveness of the proximal operator $\mathcal{S}_{\tau}(\cdot)$.
Apply a similar argument as in bounding (\ref{eq:P_Omega_tangent})
to obtain 
\begin{align*}
 & \left\Vert \mathcal{P}_{\Omega_{\mathsf{aug}}}\left(\bm{X}^{t+1}\bm{Y}^{t+1\top}-\bm{X}^{\star}\bm{Y}^{\star\top}\right)\right\Vert _{2,\infty}\\
 & \quad\leq\left\Vert \mathcal{P}_{\Omega_{\mathsf{aug}}}\left[\left(\bm{X}^{t+1}\bm{H}^{t+1}-\bm{X}^{\star}\right)\bm{Y}^{\star\top}\right]\right\Vert _{2,\infty}+\left\Vert \mathcal{P}_{\Omega_{\mathsf{aug}}}\left[\bm{X}^{t+1}\bm{H}^{t+1}\left(\bm{Y}^{t+1}\bm{H}^{t+1}-\bm{Y}^{\star}\right)^{\top}\right]\right\Vert _{2,\infty}\\
 & \quad\leq\sqrt{40\kappa p\rho_{\mathsf{aug}}}\left\Vert \left(\bm{X}^{t+1}\bm{H}^{t+1}-\bm{X}^{\star}\right)\bm{Y}^{\star\top}\right\Vert _{2,\infty}+\sqrt{40\kappa p\rho_{\mathsf{aug}}}\left\Vert \bm{X}^{t+1}\bm{H}^{t+1}\left(\bm{Y}^{t+1}\bm{H}^{t+1}-\bm{Y}^{\star}\right)^{\top}\right\Vert _{2,\infty}\\
 & \quad\leq\sqrt{40\kappa p\rho_{\mathsf{aug}}}\left(C_{\infty}\kappa+2C_{\mathrm{op}}\right)\left(\frac{\sigma}{\sigma_{\min}}\sqrt{\frac{n\log n}{p}}+\frac{\lambda}{p\sigma_{\min}}\right)\left\Vert \bm{F}^{\star}\right\Vert _{2,\infty}\left\Vert \bm{X}^{\star}\right\Vert ,
\end{align*}
as long as $n^{2}p\rho_{\mathsf{aug}}\gg\mu rn\log n$. Here, the
last line uses the induction hypotheses~(\ref{eq:induction-op})
and~(\ref{eq:induction-two-to-infty}) for the $(t+1)$-th iteration
and their immediate consequence~$\|\bm{X}^{t+1}\bm{H}^{t+1}\|_{2,\infty}\leq2\|\bm{F}^{\star}\|_{2,\infty}$, as long as $\frac{\sigma}{\sigma_{\min}}\sqrt{\frac{n\log n}{p}}\ll1/\kappa$. Taking the preceding two bounds together concludes the proof. \end{proof}

\subsection{Proof of Lemma \ref{lemma:LOO-F-perturbation}\label{subsec:Proof-of-Lemma-LOO-F-perturbation}}

Without loss of generality, we only consider the case when $1\leq l\leq n$.
The case with $n+1\leq l\leq2n$ can be derived similarly with very
minor modification, and hence we omit it for the sake of brevity.

To begin with, since $(\bm{H}^{t+1},\bm{R}^{t+1,(l)})$ is the choice
of the rotation matrix that best aligns $\bm{F}^{t+1}$ and $\bm{F}^{t+1,(l)}$,
we have 
\[
\big\|\bm{F}^{t+1}\bm{H}^{t+1}-\bm{F}^{t+1,(l)}\bm{R}^{t+1,(l)}\big\|_{\mathrm{F}}\leq\big\|\bm{F}^{t+1}\bm{H}^{t}-\bm{F}^{t+1,(l)}\bm{R}^{t,(l)}\big\|_{\mathrm{F}}.
\]
In view of the gradient update rule, one has 
\begin{align*}
 & \bm{F}^{t+1}\bm{H}^{t}-\bm{F}^{t+1,(l)}\bm{R}^{t,(l)}\\
 & \quad=\left[\bm{F}^{t}-\eta\nabla f\left(\bm{F}^{t};\bm{S}^{t}\right)\right]\bm{H}^{t}-\left[\bm{F}^{t,(l)}-\eta\nabla f^{(l)}\big(\bm{F}^{t,(l)};\bm{S}^{t,(l)}\big)\right]\bm{R}^{t,(l)}\\
 & \quad=\bm{F}^{t}\bm{H}^{t}-\eta\nabla f\left(\bm{F}^{t}\bm{H}^{t};\bm{S}^{t}\right)-\left[\bm{F}^{t,(l)}\bm{R}^{t,(l)}-\eta\nabla f^{(l)}(\bm{F}^{t,(l)}\bm{R}^{t,(l)};\bm{S}^{t,(l)})\right]\\
 & \quad=\underbrace{\bm{F}^{t}\bm{H}^{t}-\bm{F}^{t,(l)}\bm{R}^{t,(l)}-\eta\left[\nabla f_{\mathsf{aug}}\big(\bm{F}^{t}\bm{H}^{t};\bm{S}^{t}\big)-\nabla f_{\mathsf{aug}}\big(\bm{F}^{t,(l)}\bm{R}^{t,(l)};\bm{S}^{t}\big)\right]}_{\eqqcolon \bm{C}_{1}}-\underbrace{\eta\left[\nabla f_{\mathsf{diff}}\left(\bm{F}^{t}\bm{H}^{t}\right)-\nabla f_{\mathsf{diff}}\big(\bm{F}^{t,(l)}\bm{R}^{t,(l)}\big)\right]}_{\eqqcolon \bm{C}_{2}}\\
 & \quad\quad+\underbrace{\eta\left[\nabla f^{(l)}\big(\bm{F}^{t,(l)}\bm{R}^{t,(l)};\bm{S}^{t,(l)}\big)-\nabla f\big(\bm{F}^{t,(l)}\bm{R}^{t,(l)};\bm{S}^{t,(l)}\big)\right]}_{\eqqcolon \bm{C}_{3}}+\underbrace{\eta\left[\nabla f\big(\bm{F}^{t,(l)}\bm{R}^{t,(l)};\bm{S}^{t,(l)}\big)-\nabla f\big(\bm{F}^{t,(l)}\bm{R}^{t,(l)};\bm{S}^{t}\big)\right]}_{\eqqcolon \bm{C}_{4}}.
\end{align*}
Here, the second identity relies on the facts that $\nabla f(\bm{F};\bm{S})\bm{R}=\nabla f(\bm{F}\bm{R};\bm{S})$
and $\nabla f^{(l)}(\bm{F};\bm{S})\bm{R}=\nabla f^{(l)}(\bm{F}\bm{R};\bm{S})$
for any orthonormal matrix $\bm{R}\in\mathcal{O}^{r\times r}$. We
shall then control $\bm{C}_{1}$, $\bm{C}_{2}$, $\bm{C}_{3}$ and
$\bm{C}_{4}$ separately.

Employing the same strategy used to bound $\bm{A}_{1}$ and $\bm{A}_{2}$
in the proof of \cite[Lemma 12]{chen2019noisy}, we can demonstrate
that 
\begin{align*}
\left\Vert \bm{C}_{1}\right\Vert _{\mathrm{F}} & \leq\left(1-\frac{\sigma_{\min}}{20}\eta\right)\big\|\bm{F}^{t}\bm{H}^{t}-\bm{F}^{t,(l)}\bm{R}^{t,(l)}\big\|_{\mathrm{F}}\quad\text{and}\quad\left\Vert \bm{C}_{2}\right\Vert _{\mathrm{F}}\leq\eta\left(\sigma\sqrt{\frac{n}{p}}+\frac{\lambda}{p}\right)\left\Vert \bm{F}^{\star}\right\Vert _{2,\infty},
\end{align*}
provided that $\frac{\sigma}{\sigma_{\min}}\sqrt{\frac{n}{p}}\ll1/\sqrt{\kappa^{4}\mu r\log n}$
and $\eta\ll1/(n\kappa^{2}\sigma_{\max})$. With regards to $\bm{C}_{3}$,
it is seen from the definitions of $\nabla f$ and $\nabla f^{(l)}$
that 
\[
\bm{C}_{3}=\eta\left[\begin{array}{c}
\left[\mathcal{P}_{l,\cdot}\left(\bm{X}^{t,(l)}\bm{Y}^{t,(l)}-\bm{L}^{\star}\right)-p^{-1}\mathcal{P}_{(\Omega_{\mathsf{obs}})_{l,\cdot}}\left(\bm{X}^{t,(l)}\bm{Y}^{t,(l)}-\bm{L}^{\star}\right)\right]\bm{Y}^{t,(l)}\bm{R}^{t,(l)}+p^{-1}\mathcal{P}_{(\Omega_{\mathsf{obs}})_{l,\cdot}}\left(\bm{E}\right)\bm{Y}^{t,(l)}\bm{R}^{t,(l)}\\
\left[\mathcal{P}_{l,\cdot}\left(\bm{X}^{t,(l)}\bm{Y}^{t,(l)}-\bm{L}^{\star}\right)-p^{-1}\mathcal{P}_{(\Omega_{\mathsf{obs}})_{l,\cdot}}\left(\bm{X}^{t,(l)}\bm{Y}^{t,(l)}-\bm{L}^{\star}\right)\right]^{\top}\bm{X}^{t,(l)}\bm{R}^{t,(l)}+p^{-1}\mathcal{P}_{(\Omega_{\mathsf{obs}})_{l,\cdot}}\left(\bm{E}\right)^{\top}\bm{X}^{t,(l)}\bm{R}^{t,(l)}
\end{array}\right],
\]
which has the same form as $\bm{A}_{3}$ in the proof of \cite[Lemma 12]{chen2019noisy}.
It thus follows from \cite[Claim 5, 6 and 7]{chen2019noisy} that
\[
\left\Vert \bm{C}_{3}\right\Vert _{\mathrm{F}}\lesssim\eta\sigma\sqrt{\frac{n\log n}{p}}\left\Vert \bm{F}^{\star}\right\Vert _{2,\infty}+\eta\sqrt{\frac{\mu^{2}r^{2}\log n}{np}}\left\Vert \bm{F}^{t,(l)}\bm{R}^{t,(l)}-\bm{F}^{\star}\right\Vert _{2,\infty}\sigma_{\max},
\]
provided that $\frac{\sigma}{\sigma_{\min}}\sqrt{\frac{n}{p}}\ll\frac{1}{\sqrt{\kappa^{2}\log n}}$
and that $n^{2}p\gg n\log^{3}n$.

We are then left with controlling the term $\bm{C}_{4}$. Towards
this, we invoke the definition of $f$ to decompose 
\begin{align*}
\bm{C}_{4} & =\eta\left[\begin{array}{c}
p^{-1}\mathcal{P}_{\Omega_{\mathsf{obs}}}\left(\bm{S}^{t,(l)}-\bm{S}^{t}\right)\bm{Y}^{t,(l)}\bm{R}^{t,(l)}\\
p^{-1}\mathcal{P}_{\Omega_{\mathsf{obs}}}\left(\bm{S}^{t,(l)}-\bm{S}^{t}\right)^{\top}\bm{X}^{t,(l)}\bm{R}^{t,(l)}
\end{array}\right]\\
 & =\underbrace{\frac{\eta}{p}\left[\begin{array}{c}
\mathcal{P}_{-l,\cdot}\big(\bm{S}^{t,(l)}-\bm{S}^{t}\big)\bm{Y}^{t,(l)}\bm{R}^{t,(l)}\\
\big[\mathcal{P}_{-l,\cdot}\big(\bm{S}^{t,(l)}-\bm{S}^{t}\big)\big]^{\top}\bm{X}^{t,(l)}\bm{R}^{t,(l)}
\end{array}\right]}_{\eqqcolon \bm{D}_{1}}+\underbrace{\frac{\eta}{p}\left[\begin{array}{c}
\mathcal{P}_{l,\cdot}\left(\bm{S}^{t,(l)}-\bm{S}^{t}\right)\bm{Y}^{t,(l)}\bm{R}^{t,(l)}\\
\big[\mathcal{P}_{l,\cdot}\big(\bm{S}^{t,(l)}-\bm{S}^{t}\big)\big]^{\top}\bm{X}^{t,(l)}\bm{R}^{t,(l)}
\end{array}\right]}_{\eqqcolon \bm{D}_{2}}.
\end{align*}
Here, we have used the fact that both $\bm{S}^{t,(l)}$ and $\bm{S}^{t}$
are supported on $\Omega^{\star}\subseteq\Omega_{\mathsf{obs}}$.
Regarding the first matrix $\bm{D}_{1}$, we have the following fact.

\begin{fact}\label{Fact-D1}Suppose that the sample size obeys $n^2p\gg\kappa^4\mu^2r^2n\log n$, the noise satisfies $\frac{\sigma}{\sigma_{\min}}\sqrt{\frac{n\log n}{p}}\ll1/\kappa$, the outlier fraction satisfies $\rho_{\mathsf{s}}\leq\rho_{\mathsf{aug}}\ll1/\kappa^{3}$
and $n^{2}p\rho_{\mathsf{aug}}\gg\mu rn\log n$ hold. Then with probability
at least $1-O(n^-100)$, we have 
\[
\bigl\Vert\bm{D}_{1}\bigr\Vert_{\mathrm{F}}\lesssim\eta\sigma\sqrt{\frac{n\log n}{p}}\bigl\Vert\bm{F}^{\star}\bigr\Vert_{2,\infty}.
\]
\end{fact}

With regards to $\bm{D}_{2}$, recall that $\bm{S}_{l,\cdot}^{t,(l)}=\bm{S}_{l,\cdot}^{\star}$.
Using the decomposition (\ref{eq:decomposition-St-Sstar}) in the
proof of Lemma \ref{lemma:S-induction}, and recalling that $\bm{B}^{t+1}=\bm{0}$
from the proof of Lemma \ref{lemma:S-induction}, we obtain 
\begin{equation}
\mathcal{P}_{l,\cdot}\big(\bm{S}^{t,(l)}-\bm{S}^{t}\big)\bm{Y}^{t,(l)}\bm{R}^{t,(l)}=\mathcal{P}_{l,\cdot}\big(\bm{A}_{1}+\bm{E}\big)\bm{Y}^{t,(l)}\bm{R}^{t,(l)}+\mathcal{P}_{l,\cdot}\big(\bm{A}_{2}^{t}\big)\bm{Y}^{t,(l)}\bm{R}^{t,(l)},\label{eq:Pl-decompose-11}
\end{equation}
where 
\begin{align*}
\bm{A}_{1} & =\mathcal{S}_{\tau}\left[\mathcal{P}_{\Omega_{\mathsf{aug}}}\left(\bm{S}^{\star}+\bm{E}\right)\right]-\mathcal{P}_{\Omega_{\mathsf{aug}}}\left(\bm{S}^{\star}+\bm{E}\right);\\
\bm{A}_{2}^{t} & \coloneqq\mathcal{S}_{\tau}\left[\mathcal{P}_{\Omega_{\mathsf{aug}}}\left(\bm{X}^{\star}\bm{Y}^{\star\top}-\bm{X}^{t}\bm{Y}^{t\top}+\bm{S}^{\star}+\bm{E}\right)\right]-\mathcal{S}_{\tau}\left[\mathcal{P}_{\Omega_{\mathsf{aug}}}\left(\bm{S}^{\star}+\bm{E}\right)\right].
\end{align*}
For the first term $\mathcal{P}_{l,\cdot}(\bm{A}_{1}+\bm{E})\bm{Y}^{t,(l)}\bm{R}^{t,(l)}$,
the independence between $\bm{Y}^{t,(l)}\bm{R}^{t,(l)}$ and the $l$-th
row of $\bm{A}_{1}+\bm{E}$ allows us to obtain the following bound.

\begin{fact}\label{Fact-D2-1}Suppose that $\rho_{\mathsf{s}}\leq\rho_{\mathsf{aug}}\ll1/\log n$
and that $n^{2}p\gg n\log^{4}n$. Then with probability
at least $1-O(n^-100)$, we have 
\[
\left\Vert \mathcal{P}_{l,\cdot}\left(\bm{A}_{1}\right)\bm{Y}^{t,(l)}\bm{R}^{t,(l)}\right\Vert _{\mathrm{F}}\lesssim\sigma\sqrt{np\log n}\left\Vert \bm{Y}^{\star}\right\Vert _{2,\infty}.
\]
\end{fact} The term involving $\bm{A}_{2}^{t}$ is controlled in
the following claim, which relies heavily on the small scale of the
entries in $\bm{A}_{2}^{t}$.

\begin{fact}\label{Fact:D2-2}Suppose that $n\gg\kappa\mu r$, $\frac{\sigma}{\sigma_{\min}}\sqrt{\frac{n\log n}{p}}\ll1/\kappa$,  $\rho_{\mathsf{s}}\leq\rho_{\mathsf{aug}}\ll1/(\kappa\mu r)$
and that $n^2 p\rho_{\mathsf{aug}}\gg n\log n$. Then with probability
at least $1-O(n^-100)$, we have 
\[
\left\Vert \mathcal{P}_{l,\cdot}\left(\bm{A}_{2}^{t}\right)\bm{Y}^{t,(l)}\bm{R}^{t,(l)}\right\Vert _{\mathrm{F}}\lesssim\sigma\sqrt{np\log n}\left\Vert \bm{Y}^{\star}\right\Vert _{2,\infty}.
\]
\end{fact}

Combining the two bounds in Facts \ref{Fact-D2-1} and \ref{Fact:D2-2}
gives 
\[
\left\Vert \mathcal{P}_{l,\cdot}\big(\bm{S}^{t,(l)}-\bm{S}^{t}\big)\bm{Y}^{t,(l)}\bm{R}^{t,(l)}\right\Vert _{\mathrm{F}}\lesssim\sigma\sqrt{np\log n}\left\Vert \bm{Y}^{\star}\right\Vert _{2,\infty}.
\]
The same bound applies to $\Vert\mathcal{P}_{l,\cdot}(\bm{S}^{t,(l)}-\bm{S}^{t})^{\top}\bm{X}^{t,(l)}\bm{R}^{t,(l)}\Vert_{\mathrm{F}}$
via the same technique. As a result, we have 
\[
\Vert\bm{D}_{2}\Vert_{\mathrm{F}}\lesssim\eta\sigma\sqrt{\frac{n\log n}{p}}\Vert\bm{F}^{\star}\Vert_{2,\infty}.
\]

Putting the above bounds together yields 
\begin{align*}
 & \big\|\bm{F}^{t+1}\bm{H}^{t+1}-\bm{F}^{t+1,(l)}\bm{R}^{t+1,(l)}\big\|_{\mathrm{F}}\leq\left\Vert \bm{C}_{1}\right\Vert _{\mathrm{F}}+\left\Vert \bm{C}_{2}\right\Vert _{\mathrm{F}}+\left\Vert \bm{C}_{3}\right\Vert _{\mathrm{F}}+\left\Vert \bm{D}_{1}\right\Vert _{\mathrm{F}}+\left\Vert \bm{D}_{2}\right\Vert _{\mathrm{F}}\\
 & \quad\leq\left(1-\frac{\sigma_{\min}}{20}\eta\right)\left\Vert \bm{F}^{t}\bm{H}^{t}-\bm{F}^{t,(l)}\bm{R}^{t,(l)}\right\Vert _{\mathrm{F}}+\eta\left(\sigma\sqrt{\frac{n}{p}}+\frac{\lambda}{p}\right)\left\Vert \bm{F}^{\star}\right\Vert _{2,\infty}\\
 & \quad\quad+\tilde{C}\left(\eta\sigma\sqrt{\frac{n\log n}{p}}\left\Vert \bm{F}^{\star}\right\Vert _{2,\infty}+\eta\sqrt{\frac{\mu^{2}r^{2}\log n}{np}}\left\Vert \bm{F}^{t,(l)}\bm{R}^{t,(l)}-\bm{F}^{\star}\right\Vert _{2,\infty}\sigma_{\max}\right)+\tilde{C}\eta\sigma\sqrt{\frac{n\log n}{p}}\bigl\Vert\bm{F}^{\star}\bigr\Vert_{2,\infty}\\
 & \quad\overset{\text{(i)}}{\leq}\left(1-\frac{\sigma_{\min}}{20}\eta\right)C_{1}\left(\frac{\sigma}{\sigma_{\min}}\sqrt{\frac{n\log n}{p}}+\frac{\lambda}{p\sigma_{\min}}\right)\left\Vert \bm{F}^{\star}\right\Vert _{2,\infty}+\eta\left(\sigma\sqrt{\frac{n}{p}}+\frac{\lambda}{p}\right)\left\Vert \bm{F}^{\star}\right\Vert _{2,\infty}+\tilde{C}\frac{\lambda}{p}\bigl\Vert\bm{F}^{\star}\bigr\Vert_{2,\infty}\\
 & \quad\quad+\tilde{C}\eta\sqrt{\frac{\mu^{2}r^{2}\log n}{np}}\left(C_{\infty}\kappa+C_{1}\right)\left(\frac{\sigma}{\sigma_{\min}}\sqrt{\frac{n\log n}{p}}+\frac{\lambda}{p\sigma_{\min}}\right)\left\Vert \bm{F}^{\star}\right\Vert _{2,\infty}\sigma_{\max}+\tilde{C}\eta\sigma\sqrt{\frac{n\log n}{p}}\bigl\Vert\bm{F}^{\star}\bigr\Vert_{2,\infty}\\
 & \quad\overset{\text{(ii)}}{\leq}C_{1}\left(\frac{\sigma}{\sigma_{\min}}\sqrt{\frac{n\log n}{p}}+\frac{\lambda}{p\sigma_{\min}}\right)\left\Vert \bm{F}^{\star}\right\Vert _{2,\infty},
\end{align*}
where (i) invokes (\ref{eq:induction-loo-perturbation}) and its immediate
consequence that 
\begin{align}
\left\Vert \bm{F}^{t,(l)}\bm{R}^{t,(l)}-\bm{F}^{\star}\right\Vert _{2,\infty} & \leq\left\Vert \bm{F}^{t}\bm{H}^{t}-\bm{F}^{t,(l)}\bm{R}^{t,(l)}\right\Vert _{\mathrm{F}}+\left\Vert \bm{F}^{t}\bm{H}^{t}-\bm{F}^{\star}\right\Vert _{2,\infty}\\
 & \le\left(C_{\infty}\kappa+C_{1}\right)\left(\frac{\sigma}{\sigma_{\min}}\sqrt{\frac{n\log n}{p}}+\frac{\lambda}{p\sigma_{\min}}\right)\left\Vert \bm{F}^{\star}\right\Vert _{2,\infty}.\label{eq:F-t-l-F-star-close}
\end{align}
The last line (ii) holds as long as $n^{2}p\gg\kappa^{4}\mu^{2}r^{2}n\log n$
and $C_{1}$ is large enough.

\begin{proof}[Proof of Fact~\ref{Fact-D1}]First notice that $\bm{S}^{t}$
is supported on $\Omega_{\mathsf{aug}}$, which is a consequence of
(\ref{eq:decomposition-St-Sstar}) and (\ref{eq:shrink-to-zero-1}) as long as $\frac{\sigma}{\sigma_{\min}}\sqrt{\frac{n\log n}{p}}\ll1/\kappa$ and $n^2p\gg\kappa^4\mu^2r^2n\log n$.
By replacing $\bm{X}^{t+1}$ (resp.~$\bm{Y}^{t+1})$ with $\bm{X}^{t+1,(l)}$ (resp.~$\bm{Y}^{t+1,(l)})$) and invoking (\ref{eq:F-t-l-F-star-close}) instead of (\ref{eq:induction-two-to-infty}), the same arguments yield the fact that $\bm{S}^{t,(l)}$ is also supported
on $\Omega_{\mathsf{aug}}$. Define $\omega_{ij}\coloneqq\ind_{(i,j)\in\Omega_{\mathsf{aug}}}$.
The Frobenius norm of the upper block of $\bm{D}_{1}$ can be bounded
by 
\begin{align*}
 & \left\Vert \mathcal{P}_{-l,\cdot}\big(\bm{S}^{t,(l)}-\bm{S}^{t}\big)\bm{Y}^{t,(l)}\bm{R}^{t,(l)}\right\Vert _{\mathrm{F}}^{2}=\sum_{i:i\neq l}\sum_{j=1}^{r}\left[\sum_{k=1}^{n}\left(\bm{S}^{t,(l)}-\bm{S}^{t}\right)_{ik}Y_{kj}^{t,(l)}\right]^{2}\\
 & \quad=\sum_{i:i\neq l}\sum_{j=1}^{r}\left[\sum_{k=1}^{n}\omega_{ik}\left(\bm{S}^{t,(l)}-\bm{S}^{t}\right)_{ik}Y_{kj}^{t,(l)}\right]^{2}\leq\sum_{i:i\neq l}\sum_{j=1}^{r}\left[\sum_{k=1}^{n}\omega_{ik}\left(\bm{S}^{t,(l)}-\bm{S}^{t}\right)_{ik}^{2}\right]\left[\sum_{k=1}^{n}\omega_{ik}\bigl(Y_{kj}^{t,(l)}\bigr)^{2}\right],
\end{align*}
where we use the Cauchy-Schwarz inequality in the last step. Converting
to the matrix notation, we obtain
\[
\sum_{k=1}^{n}\omega_{ik}\bigl(Y_{kj}^{t,(l)}\bigr)^{2}=\left\Vert \mathcal{P}_{\Omega_{\mathsf{aug}}}\left(\bm{e}_{i}\bm{e}_{j}^{\top}\bm{Y}^{t,(l)\top}\right)\right\Vert _{\mathrm{F}}^{2}.
\]
Applying a similar argument as in bounding (\ref{eq:P_Omega_tangent}),
one can obtain from Lemma \ref{lemma:sampling-operator-norm-in-T}
that 
\[
\sum_{j=1}^{r}\left\Vert \mathcal{P}_{\Omega_{\mathsf{aug}}}\left(\bm{e}_{i}\bm{e}_{j}^{\top}\bm{Y}^{t,(l)\top}\right)\right\Vert _{\mathrm{F}}^{2}\lesssim\kappa p\rho_{\mathsf{aug}}\sum_{j=1}^{r}\bigl\Vert\bm{Y}_{\cdot,j}^{t,(l)}\bigr\Vert_{\mathrm{F}}^{2}=\kappa p\rho_{\mathsf{aug}}\|\bm{Y}^{t,(l)}\|_{\mathrm{F}}^{2},
\]
provided that $n^{2}p\rho_{\mathsf{aug}}\gg\mu rn\log n$. This allows
us to reach 
\begin{align*}
\left\Vert \mathcal{P}_{-l,\cdot}\big(\bm{S}^{t,(l)}-\bm{S}^{t}\big)\bm{Y}^{t,(l)}\bm{R}^{t,(l)}\right\Vert _{\mathrm{F}} & \lesssim\sqrt{\sum_{i:i\neq l}\left[\sum_{k=1}^{n}\omega_{ik}\left(\bm{S}^{t,(l)}-\bm{S}^{t}\right)_{ik}^{2}\right]}\cdot\sqrt{\kappa p\rho_{\mathsf{aug}}}\bigl\Vert\bm{Y}^{t,(l)}\bigr\Vert_{\mathrm{F}}\\
 & =\sqrt{\kappa p\rho_{\mathsf{aug}}}\bigl\Vert\mathcal{P}_{-l,\cdot}\left(\bm{S}^{t,(l)}-\bm{S}^{t}\right)\bigr\Vert_{\mathrm{F}}\bigl\Vert\bm{Y}^{t,(l)}\bigr\Vert_{\mathrm{F}}\\
 & \leq C_{3}\sqrt{\kappa p\rho_{\mathsf{aug}}}\frac{\sigma}{\sigma_{\min}}\sqrt{n\log n}\bigl\Vert\bm{F}^{\star}\bigr\Vert^{2}\bigl\Vert\bm{F}^{\star}\bigr\Vert_{2,\infty}\\
 & \lesssim\sigma\sqrt{np\log n}\bigl\Vert\bm{F}^{\star}\bigr\Vert_{2,\infty}.
\end{align*}
Here, the penultimate step comes from the hypothesis (\ref{eq:induction-S-loo}),
whereas the last step holds as long as $\rho_{\mathsf{s}}\leq\rho_{\mathsf{aug}}\ll1/\kappa^{3}$.
The Frobenius norm of the lower block of $\bm{D}_{1}$ admits the
same bound. As a result, we obtain $\bigl\Vert\bm{D}_{1}\bigr\Vert_{\mathrm{F}}\lesssim\eta\sigma\sqrt{\frac{n\log n}{p}}\bigl\Vert\bm{F}^{\star}\bigr\Vert_{2,\infty}$
as claimed. \end{proof}

\begin{proof}[Proof of Fact~\ref{Fact-D2-1}] Regarding the first
term on the right-hand side of (\ref{eq:Pl-decompose-11}), we can
write 
\[
\big\|\mathcal{P}_{l,\cdot}\big(\bm{A}_{1}+\bm{E}\big)\bm{Y}^{t,(l)}\bm{R}^{t,(l)}\big\|_{\mathrm{F}}=\Big\|\sum\nolimits _{j=1}^{n}\big(\bm{A}_{1}+\bm{E}\big)_{lj}\bm{Y}_{j,\cdot}^{t,(l)}\Big\|_{2}=\Big\|\sum\nolimits _{j=1}^{n}\underbrace{\omega_{lj}\left[\mathcal{S}_{\tau}(S_{lj}^{\star}+E_{ij})-S_{lj}^{\star}\right]\bm{Y}_{j,\cdot}^{t,(l)}}_{\eqqcolon \bm{u}_{j}}\Big\|_{2},
\]
where $\omega_{lj}\coloneqq\ind\{(l,j)\in\Omega_{\mathsf{aug}}\}$ is a Bernoulli
random variable with mean $p\rho_{\mathsf{aug}}$. Since $\bm{Y}^{t,(l)}$
is independent of $\{\omega_{lj}\}_{1\leq j\leq n}$ and $\bm{S}_{l,\cdot}^{\star}$,
the vectors $\{\bm{u}_{j}\}_{j=1}^{n}$ are statistically independent
conditional on $\bm{Y}^{t,(l)}$. We can thus apply the matrix Bernstein
inequality to control this term. Specifically, conditional on $\bm{Y}^{t,(l)}$,
we have 
\[
\left\Vert \|\bm{u}_{j}\|_{2}\right\Vert _{\psi_{1}}\leq\big\|\bm{Y}^{t,(l)}\big\|_{2,\infty}\left\Vert \omega_{lj}\left[\mathcal{S}_{\tau}(S_{lj}^{\star}+E_{ij})-S_{lj}^{\star}\right]\right\Vert _{\psi_{1}}\overset{\text{(i)}}{\lesssim}\tau\left\Vert \bm{Y}^{t,(l)}\right\Vert _{2,\infty},
\]
\[
V\coloneqq\left\Vert \mathbb{E}\left[\sum\nolimits _{j=1}^{n}\omega_{lj}^{2}\left(\mathcal{S}_{\tau}(S_{lj}^{\star}+E_{ij})-S_{lj}^{\star}\right)^{2}\bm{Y}_{j,\cdot}^{t,(l)}\bm{Y}_{j,\cdot}^{t,(l)\top}\right]\right\Vert \overset{(\text{ii})}{\lesssim}p\rho_{\mathsf{aug}}\tau^{2}\big\|\bm{Y}^{t,(l)}\big\|_{\mathrm{F}}^{2},
\]
where $\|\cdot\|_{\psi_{1}}$ denotes the sub-exponential norm \cite{vershynin2016high}.
Here, the relation (i) holds since
\begin{align*}
\left\Vert \omega_{lj}\left[\mathcal{S}_{\tau}(S_{lj}^{\star}+E_{ij})-S_{lj}^{\star}\right]\right\Vert _{\psi_{1}} & \leq\left\Vert \mathcal{S}_{\tau}(S_{lj}^{\star}+E_{ij})-S_{lj}^{\star}\right\Vert _{\psi_{1}}\leq\left\Vert \mathcal{S}_{\tau}(S_{lj}^{\star}+E_{ij})-\left(S_{lj}^{\star}+E_{ij}\right)\right\Vert _{\psi_{1}}+\left\Vert E_{ij}\right\Vert _{\psi_{1}}\\
 & \leq\left|\mathcal{S}_{\tau}(S_{lj}^{\star}+E_{ij})-\left(S_{lj}^{\star}+E_{ij}\right)\right|+\left\Vert E_{ij}\right\Vert _{\psi_{2}}\leq2\tau,
\end{align*}
where we have used the fact that $|\mathcal{S}_{\tau}(x)-x|\leq\tau$
and $\|E_{ij}\|_{\psi_{1}}\leq\|E_{ij}\|_{\psi_{2}}\leq\sigma\leq\tau$.
In addition, the second inequality (ii) comes from the identity $\mathbb{E}[\omega_{lj}^{2}]=p\rho_{\mathsf{aug}}$
and the fact that 
\[
\mathbb{E}\left[\left(\mathcal{S}_{\tau}(S_{lj}^{\star}+E_{ij})-S_{lj}^{\star}\right)^{2}\right]\leq2\mathbb{E}\left[\left(\mathcal{S}_{\tau}(S_{lj}^{\star}+E_{ij})-S_{lj}^{\star}-E_{ij}\right)^{2}\right]+2\mathbb{E}\left[E_{ij}^{2}\right]\lesssim\tau^{2}.
\]
With the aid of the above bounds, we can invoke the matrix Bernstein
inequality \cite[Proposition 2]{MR2906869} to reach 
\begin{align*}
\left\Vert \sum\nolimits _{j=1}^{n}\bm{u}_{j}\right\Vert _{2} & \lesssim\sqrt{V\log n}+\left\Vert \|\bm{u}_{j}\|_{2}\right\Vert _{\psi_{1}}\log^{2}n\\
 & \lesssim\sqrt{p\rho_{\mathsf{aug}}\tau^{2}\left\Vert \bm{Y}^{t,(l)}\right\Vert _{\mathrm{F}}^{2}\log n}+\tau\big\|\bm{Y}^{t,(l)}\big\|_{2,\infty}\log^{2}n\\
 & \lesssim\left(\tau\sqrt{np\rho_{\mathsf{aug}}\log n}+\tau\log^{2}n\right)\big\|\bm{Y}^{t,(l)}\big\|_{2,\infty}
\end{align*}
with probability at least $1-O(n^{-10})$. Here, the last inequality
arises from $\Vert\bm{Y}^{t,(l)}\Vert_{\mathrm{F}}^{2}\leq n\Vert\bm{Y}^{t,(l)}\Vert_{2,\infty}^{2}$.
Consequently, we conclude that, with high probability, 
\[
\left\Vert \mathcal{P}_{l,\cdot}\left(\bm{A}_{1}+\bm{E}\right)\bm{Y}^{t,(l)}\bm{R}^{t,(l)}\right\Vert _{\mathrm{F}}\lesssim\left(\tau\sqrt{np\rho_{\mathsf{aug}}\log n}+\tau\log^{2}n\right)\big\|\bm{Y}^{t,(l)}\big\|_{2,\infty}\lesssim\sigma\sqrt{np\log n}\left\Vert \bm{Y}^{\star}\right\Vert _{2,\infty},
\]
with the proviso that $\rho_{\mathsf{s}}\leq\rho_{\mathsf{aug}}\ll1/\log n$
and $n^{2}p\gg n\log^{4}n$.\end{proof}

\begin{proof}[Proof of Fact~\ref{Fact:D2-2}]Regarding the second
term on the right-hand side of (\ref{eq:Pl-decompose-11}), we have
\begin{align*}
\left\Vert \mathcal{P}_{l,\cdot}\left(\bm{A}_{2}^{t}\right)\bm{Y}^{t,(l)}\bm{R}^{t,(l)}\right\Vert _{\mathrm{F}} & =\Big\|\sum\nolimits _{j=1}^{n}\left(\bm{A}_{2}^{t}\right)_{lj}\bm{Y}_{j,\cdot}^{t,(l)}\Big\|_{2}\\
 & \overset{\text{(i)}}{\leq}2np\rho_{\mathsf{aug}}\left\Vert \bm{A}_{2}^{t}\right\Vert _{\infty}\big\|\bm{Y}^{t,(l)}\big\|_{2,\infty}\\
 & \overset{\text{(ii)}}{\leq}12np\rho_{\mathsf{aug}}C_{\infty}\left(\frac{\sigma}{\sigma_{\min}}\sqrt{\frac{n\log n}{p}}+\frac{\lambda}{p\sigma_{\min}}\right)\frac{\mu r}{n}\sigma_{\max}\left\Vert \bm{Y}^{\star}\right\Vert _{2,\infty}\\
 & \overset{\text{(iii)}}{\lesssim}\sigma\sqrt{np\log n}\left\Vert \bm{Y}^{\star}\right\Vert _{2,\infty}.
\end{align*}
Here, the first upper bound (i) arises from the fact that $\{j\mid(\bm{A}_{2}^{t})_{lj}\neq0\}\subseteq\{j\mid(l,j)\in\Omega_{\mathsf{aug}}\}$,
whose cardinality is upper bounded by $2np\rho_{\mathsf{aug}}$ with
high probability as long as $np\rho_{\mathsf{aug}}\gg\log n$. The
second inequality (ii) comes from the simple fact that $\Vert\bm{Y}^{t,(l)}\Vert_{2,\infty}\leq2\Vert\bm{Y}^{\star}\Vert_{2,\infty}$
as well as the bound 
\begin{align*}
\left\Vert \bm{A}_{2}^{t}\right\Vert _{\infty} & =\left\Vert \mathcal{S}_{\tau}\left[\mathcal{P}_{\Omega_{\mathsf{aug}}}\left(\bm{X}^{\star}\bm{Y}^{\star\top}-\bm{X}^{t}\bm{Y}^{t\top}+\bm{S}^{\star}+\bm{E}\right)\right]-\mathcal{S}_{\tau}\left[\mathcal{P}_{\Omega_{\mathsf{aug}}}\left(\bm{S}^{\star}+\bm{E}\right)\right]\right\Vert _{\infty}\\
 & \leq\left\Vert \mathcal{P}_{\Omega_{\mathsf{aug}}}\left(\bm{X}^{\star}\bm{Y}^{\star\top}-\bm{X}^{t}\bm{Y}^{t\top}+\bm{S}^{\star}+\bm{E}\right)-\mathcal{P}_{\Omega_{\mathsf{aug}}}\left(\bm{S}^{\star}+\bm{E}\right)\right\Vert _{\infty}\\
 & \leq\left\Vert \bm{X}^{\star}\bm{Y}^{\star\top}-\bm{X}^{t}\bm{Y}^{t\top}\right\Vert _{\infty}\\
 & \leq3C_{\infty}\left(\frac{\sigma}{\sigma_{\min}}\sqrt{\frac{n\log n}{p}}+\frac{\lambda}{p\sigma_{\min}}\right)\frac{\mu r}{n}\sigma_{\max},
\end{align*}
where we use the non-expansiveness of $\mathcal{S}_{\tau}(\cdot)$
and the established bound (\ref{eq:X-star-Y-star-X-t-Y-t-infty-norm}), which holds as long as $\frac{\sigma}{\sigma_{\min}}\sqrt{\frac{n\log n}{p}}\ll1/\kappa$.
Last but not least, the relation (iii) holds as long as $\rho_{\mathsf{s}}\leq\rho_{\mathsf{aug}}\ll1/(\kappa\mu r)$
and $n\gg\kappa\mu r$.\end{proof}

\subsection{Proof of Lemma \ref{lemma:LOO-S-perturbation}\label{subsec:Proof-of-Lemma-LOO-S-perturbation}}

Without loss of generality, we assume $1\leq l\leq n$. Following
the definitions of $\bm{S}^{t+1,(l)}$ and $\bm{S}^{t+1}$, we have
\begin{align}
\big\|\mathcal{P}_{-l,\cdot}\big(\bm{S}^{t+1,(l)}-\bm{S}^{t+1}\big)\big\|_{\mathrm{F}} & =\left\Vert \mathcal{P}_{-l,\cdot}\left[\mathcal{S}_{\tau}\big(\bm{M}-\bm{X}^{t+1,(l)}\bm{Y}^{t+1,(l)\top}\big)-\mathcal{S}_{\tau}\big(\bm{M}-\bm{X}^{t+1}\bm{Y}^{t+1\top}\big)\right]\right\Vert _{\mathrm{F}}\nonumber \\
 & \leq\left\Vert \mathcal{P}_{\Omega_{\mathsf{aug}}}\left(\bm{\Delta}\right)\right\Vert _{\mathrm{F}}+\left\Vert \mathcal{P}_{\Omega_{\mathsf{aug}}^{\mathrm{c}}}\left(\bm{\Delta}\right)\right\Vert _{\mathrm{F}},\label{eq:Pdelta-Pcdelta}
\end{align}
where we denote $\bm{\Delta}\coloneqq\mathcal{S}_{\tau}(\bm{M}-\bm{X}^{t+1,(l)}\bm{Y}^{t+1,(l)\top})-\mathcal{S}_{\tau}(\bm{M}-\bm{X}^{t+1}\bm{Y}^{t+1\top})$.
Recall from Appendix~\ref{sec:An-equivalent-probabilistic} that
each $(i,j)$ is included in $\Omega_{\mathsf{aug}}$ independently
with probability $p\rho_{\mathsf{aug}}$, where $1\geq\rho_{\mathsf{aug}}\geq\rho_{\mathsf{s}}$.
\begin{enumerate}
\item For the first term $\Vert\mathcal{P}_{\Omega_{\mathsf{aug}}}\left(\bm{\Delta}\right)\Vert_{\mathrm{F}}$,
the non-expansiveness of the proximal operator $\mathcal{S}_{\tau}(\cdot)$
yields 
\begin{align*}
\left\Vert \mathcal{P}_{\Omega_{\mathsf{aug}}}\left(\bm{\Delta}\right)\right\Vert _{\mathrm{F}} & \leq\left\Vert \mathcal{P}_{\Omega_{\mathsf{aug}}}\big(\bm{X}^{t+1,(l)}\bm{Y}^{t+1,(l)\top}-\bm{X}^{t+1}\bm{Y}^{t+1\top}\big)\right\Vert _{\mathrm{F}}.
\end{align*}
Apply Lemma~\ref{lemma:sampling-operator-norm-in-T} and a similar
argument in bounding~(\ref{eq:P_Omega_tangent}) to obtain 
\begin{align*}
\left\Vert \mathcal{P}_{\Omega_{\mathsf{aug}}}\left(\bm{\Delta}\right)\right\Vert _{\mathrm{F}} & \leq\left\Vert \mathcal{P}_{\Omega_{\mathsf{aug}}}\left[\bm{X}^{t+1}\bm{H}^{t+1}\left(\bm{Y}^{t+1,(l)}\bm{R}^{t+1,(l)}-\bm{Y}^{t+1}\bm{H}^{t+1}\right)^{\top}\right]\right\Vert _{\mathrm{F}}\\
 & \quad+\left\Vert \mathcal{P}_{\Omega_{\mathsf{aug}}}\left[\left(\bm{X}^{t+1,(l)}\bm{R}^{t+1,(l)}-\bm{X}^{t+1}\bm{H}^{t+1}\right)\bm{R}^{t+1,(l)\top}\bm{Y}^{t+1,(l)\top}\right]\right\Vert _{\mathrm{F}}\\
 & \lesssim\sqrt{\kappa p\rho_{\mathsf{aug}}}\left\Vert \bm{X}^{t+1}\bm{H}^{t+1}\right\Vert \left\Vert \bm{Y}^{t+1,(l)}\bm{R}^{t+1,(l)}-\bm{Y}^{t+1}\bm{H}^{t+1}\right\Vert _{\mathrm{F}}\\
 & \quad+\sqrt{\kappa p\rho_{\mathsf{aug}}}\left\Vert \bm{Y}^{t+1,(l)}\bm{H}^{t+1,(l)}\right\Vert \left\Vert \bm{X}^{t+1,(l)}\bm{R}^{t+1,(l)}-\bm{X}^{t+1}\bm{H}^{t+1}\right\Vert _{\mathrm{F}},
\end{align*}
with the proviso that $n^{2}p\rho_{\mathsf{aug}}\gg\mu rn\log n$.
In view of (\ref{eq:induction-loo-perturbation}) and the simple facts
$\|\bm{X}^{t+1}\bm{H}^{t+1}\|\leq2\|\bm{X}^{\star}\|,\|\bm{Y}^{t+1,(l)}\bm{H}^{t+1,(l)}\|\leq2\|\bm{X}^{\star}\|$,
one has 
\begin{align*}
\left\Vert \mathcal{P}_{\Omega_{\mathsf{aug}}}\left(\bm{\Delta}\right)\right\Vert _{\mathrm{F}} & \lesssim\sqrt{\kappa p\rho_{\mathsf{aug}}}\left\Vert \bm{X}^{\star}\right\Vert \left(\frac{\sigma}{\sigma_{\min}}\sqrt{\frac{n\log n}{p}}+\frac{\lambda}{p\sigma_{\min}}\right)\left\Vert \bm{F}^{\star}\right\Vert _{2,\infty}\\
 & \leq C_{3}\frac{\sigma}{\sigma_{\min}}\sqrt{n\log n}\left\Vert \bm{F}^{\star}\right\Vert _{2,\infty}\left\Vert \bm{F}^{\star}\right\Vert ,
\end{align*}
provided that $\rho_{\mathsf{aug}}\ll1/\kappa$.
\item Regarding the second term $\Vert\mathcal{P}_{\Omega_{\mathsf{aug}}^{\mathrm{c}}}\left(\bm{\Delta}\right)\Vert_{\mathrm{F}}$,
we first recall from (\ref{eq:shrink-to-zero-1}) that 
\[
\mathcal{S}_{\tau}\left[\mathcal{P}_{\Omega_{\mathsf{aug}}^{\mathrm{c}}}\left(\bm{M}-\bm{X}^{t+1}\bm{Y}^{t+1\top}\right)\right]=\bm{0}.
\]
By replacing $\bm{X}^{t+1}$ (resp.~$\bm{Y}^{t+1})$ with $\bm{X}^{t+1,(l)}$ (resp.~$\bm{Y}^{t+1,(l)})$) and invoking (\ref{eq:F-t-l-F-star-close}) instead of (\ref{eq:induction-two-to-infty}), the same arguments that we used to prove (\ref{eq:shrink-to-zero-1}) also allow us to demonstrate
\[
\mathcal{S}_{\tau}\left[\mathcal{P}_{\Omega_{\mathsf{aug}}^{\mathrm{c}}}\left(\bm{M}-\bm{X}^{t+1,(l)}\bm{Y}^{t+1,(l)\top}\right)\right]=\bm{0}
\]
provided that $n^{2}p\gg\kappa^{4}\mu^{2}r^{2}n\log n$ and $\frac{\sigma}{\sigma_{\min}}\sqrt{\frac{n\log n}{p}}\ll1/\kappa$. Consequently,
we have $\mathcal{P}_{\Omega_{\mathsf{aug}}^{\mathrm{c}}}\left(\bm{\Delta}\right)=\bm{0}$.
\end{enumerate}
Substituting the above two bounds into (\ref{eq:Pdelta-Pcdelta}),
we conclude that 
\[
\left\Vert \mathcal{P}_{-l,\cdot}\big(\bm{S}^{t+1,(l)}-\bm{S}^{t+1}\big)\right\Vert _{\mathrm{F}}\leq\left\Vert \mathcal{P}_{\Omega_{\mathsf{aug}}}\left(\bm{\Delta}\right)\right\Vert _{\mathrm{F}}\leq C_{3}\frac{\sigma}{\sigma_{\min}}\sqrt{n\log n}\left\Vert \bm{F}^{\star}\right\Vert _{2,\infty}\left\Vert \bm{F}^{\star}\right\Vert .
\]

\subsection{Proof of Lemma~\ref{lemma:function-value-decreasing} \label{sec:Proof-of-Lemma-funcation-value}}

Following \cite[Lemma 18]{chen2019noisy}, we already know that 
\begin{equation}
f\left(\bm{X}^{t+1},\bm{Y}^{t+1};\bm{S}^{t}\right)\leq f\left(\bm{X}^{t},\bm{Y}^{t};\bm{S}^{t}\right)-\frac{\eta}{2}\left\Vert \nabla f\left(\bm{X}^{t},\bm{Y}^{t};\bm{S}^{t}\right)\right\Vert _{\mathrm{F}}^{2}.\label{eq:monotone-f}
\end{equation}
As a result, one has 
\begin{align*}
F\left(\bm{X}^{t+1},\bm{Y}^{t+1},\bm{S}^{t+1}\right) & \overset{(\mathrm{i})}{\leq}F\left(\bm{X}^{t+1},\bm{Y}^{t+1},\bm{S}^{t}\right)=f\left(\bm{X}^{t+1},\bm{Y}^{t+1};\bm{S}^{t}\right)+\tau\left\Vert \bm{S}^{t}\right\Vert _{1}\\
 & \overset{(\mathrm{ii})}{\leq}f\left(\bm{X}^{t},\bm{Y}^{t};\bm{S}^{t}\right)-\frac{\eta}{2}\left\Vert \nabla f\left(\bm{X}^{t},\bm{Y}^{t};\bm{S}^{t}\right)\right\Vert _{\mathrm{F}}^{2}+\tau\left\Vert \bm{S}^{t}\right\Vert _{1}\\
 & =F\left(\bm{X}^{t},\bm{Y}^{t},\bm{S}^{t}\right)-\frac{\eta}{2}\left\Vert \nabla f\left(\bm{X}^{t},\bm{Y}^{t};\bm{S}^{t}\right)\right\Vert _{\mathrm{F}}^{2},
\end{align*}
where (i) follows since, by construction, $\bm{S}^{t+1}$ is the minimizer
of $F(\bm{X}^{t+1},\bm{Y}^{t+1},\bm{S})$ for any given $(\bm{X}^{t+1},\bm{Y}^{t+1})$,
and (ii) arises from (\ref{eq:monotone-f}).

\bibliographystyle{alpha}
\bibliography{bibfileNonconvex}

\end{document}